\PassOptionsToPackage{x11names,table,dvipsnames}{xcolor}

\documentclass[acmtog]{acmart}

\AtBeginDocument{%
  \providecommand\BibTeX{{%
    \normalfont B\kern-0.5em{\scshape i\kern-0.25em b}\kern-0.8em\TeX}}}

\acmSubmissionID{562}

\usepackage{float}
\usepackage{lscape}
\usepackage{multirow}
\usepackage{enumitem}
\usepackage{bm}
\usepackage{mathtools}
\usepackage{amsfonts}
\usepackage{color,colortbl}
\usepackage{xcolor}
\usepackage{siunitx} 
\usepackage{url}
\usepackage{listings}
\usepackage{xspace}
\usepackage{setspace}
\usepackage{nicefrac}
\usepackage[utf8]{inputenc} 
\usepackage[T1]{fontenc}    

\definecolor{LavenderBlue}{rgb}{0.7020, 0.8039, 0.8902}

\newcommand{\algoNameFull}{Neural Progressive Meshes\xspace}
\newcommand{\mpage}[2]
{
\begin{minipage}{#1\linewidth}\centering
#2
\end{minipage}
}

\usepackage[ruled]{algorithm2e}

\SetAlFnt{\small}
\SetAlCapFnt{\small}
\SetAlCapNameFnt{\small}
\SetAlCapHSkip{0pt}

\acmJournal{TOG}

\author{Yun-Chun Chen}
\affiliation{
  \institution{University of Toronto}
  \country{Canada}
}
\email{ycchen@cs.toronto.edu}

\author{Vladimir G. Kim}
\affiliation{
  \institution{Adobe Research}
  \country{USA}
}
\email{vokim@adobe.com}

\author{Noam Aigerman}
\affiliation{
  \institution{Adobe Research}
  \country{USA}
}
\email{aigerman@adobe.com}

\author{Alec Jacobson}
\affiliation{
  \institution{Adobe Research, University of Toronto}
  \country{Canada}
}
\email{jacobson@cs.toronto.edu}

\copyrightyear{2023} 
\acmYear{2023} 
\setcopyright{acmlicensed}\acmConference[SIGGRAPH '23 Conference Proceedings]{Special Interest Group on Computer Graphics and Interactive Techniques Conference Conference Proceedings}{August 6--10, 2023}{Los Angeles, CA, USA}
\acmBooktitle{Special Interest Group on Computer Graphics and Interactive Techniques Conference Conference Proceedings (SIGGRAPH '23 Conference Proceedings), August 6--10, 2023, Los Angeles, CA, USA}
\acmPrice{15.00}
\acmDOI{10.1145/3588432.3591531}
\acmISBN{979-8-4007-0159-7/23/08}

\begin{document}

\title{Neural Progressive Meshes}

\begin{teaserfigure}
  \centering
  \includegraphics[width=\linewidth]{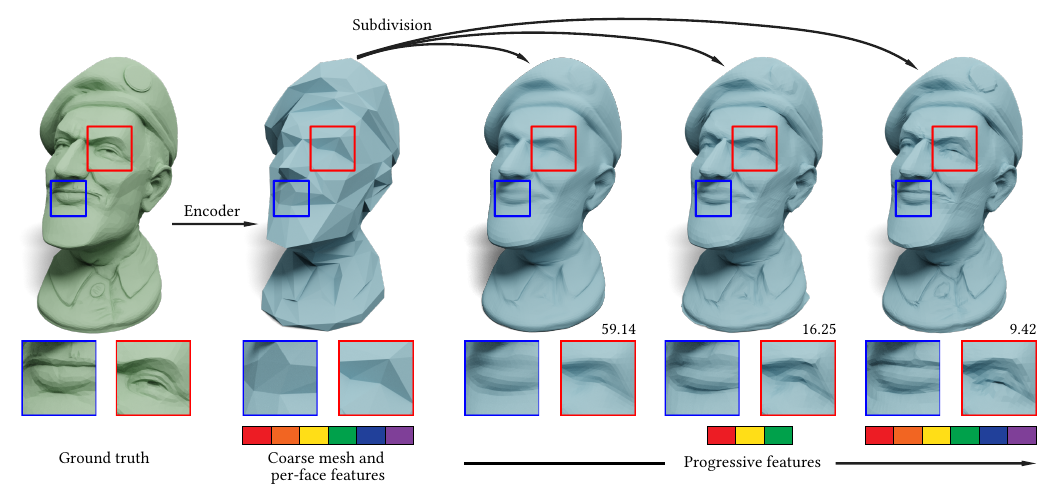}
  \caption{
  \textbf{\algoNameFull.} 
  We present a framework that learns a progressive compressed representation of meshes for transmission purposes.
  Given a high-resolution mesh in the database, the server trains a network that derives a compressed representation that can be transmitted to the client. 
  The client reconstructs the low-resolution mesh using a pre-trained subdivision decoder. 
  The reconstruction quality can be further improved with additional data transmitted progressively from the server to the client. 
  The numbers shown in the corner of each example are the compression ratios. 
  }
  \label{fig:teaser}
\end{teaserfigure}

\begin{abstract}
The recent proliferation of 3D content that can be consumed on hand-held devices necessitates efficient tools for transmitting large geometric data, e.g., 3D meshes, over the Internet. 
Detailed high-resolution assets can pose a challenge to storage as well as transmission bandwidth, and level-of-detail techniques are often used to transmit an asset using an appropriate bandwidth budget. 
It is especially desirable for these methods to transmit data progressively, improving the quality of the geometry with more data. 
Our key insight is that the geometric details of 3D meshes often exhibit similar local patterns even across different shapes, and thus can be effectively represented with a shared learned generative space. 
We learn this space using a subdivision-based encoder-decoder architecture trained in advance on a large collection of surfaces. 
We further observe that additional residual features can be transmitted progressively between intermediate levels of subdivision that enable the client to control the tradeoff between bandwidth cost and quality of reconstruction, providing a \emph{neural progressive mesh representation}. 
We evaluate our method on a diverse set of complex 3D shapes and demonstrate that it outperforms baselines in terms of compression ratio and reconstruction quality. 
\end{abstract}

\maketitle

\section{Introduction}

We propose a framework for learning a progressive compressed representation of meshes. 
Given a high-resolution mesh, our goal is to derive a compressed representation that can be transmitted to a client using a small bandwidth budget. 
The progressive nature of the compression entails the client can immediately reconstruct a meaningful mesh with lower reconstruction quality, and the server can further progressively transmit additional data to improve the reconstruction quality while the asset is being used. 

The need for such progressive compression techniques is consistently rising, following the rise in the need to transmit detailed 3D meshes from a centralized server repository to a client. 
Our method is especially suitable for virtual and augmented reality applications on mobile devices, which require selective transmissions of 3D content based on its visibility and available bandwidth. 
It additionally enables the client to stop decompression at a desired resolution.

Mesh decimation and level-of-detail (LoD) techniques are commonly used to reduce the size of a 3D asset either for rendering or transmission efficiency purposes~\cite{QSlim,lescoat2020spectral}. 
In addition to geometry simplification, remeshing can be used to optimize the size of a 3D asset~\cite{szymczak2002piecewise,surazhsky2003explicit}. 
Mesh simplification is a greedy process, and thus learning-based techniques have been proposed to devise a more efficient simplification approach~\cite{neural_mesh_simplification}. 
Most decimation techniques provide a single asset with a desired triangle budget and do not provide a way to progressively improve the quality of the asset. 
Progressive representations have been proposed to address this use case~\cite{progressive_meshes}, enabling incremental transmission of data to gradually improve the quality of the asset on the client side. 
All of these methods, however, inevitably lose details as they reduce the polygon count, approximating complex geometric details with planes. 
Surface subdivision~\cite{Loop,Butterfly} techniques could be used on the client side to increase the resolution of the transmitted low-resolution mesh, and the coarse mesh can be optimized specifically for a particular subdivision scheme~\cite{Subdivfit}. 
These subdivision schemes, however, use simple hand-crafted filters, and thus subdivided geometry lacks any intricate original details. 
In this work, we propose to learn the space of geometric details by encoding them in progressive per-face features, which could be used to guide a neural subdivision process, enabling it to reconstruct complex geometry such as the eye or the lips of a character as shown in Figure~\ref{fig:teaser}.

We follow recent advances in subdivision-based learning techniques for mesh analysis~\cite{subdivnet,meshcnn} and upsampling~\cite{neural_subdivision,Hertz2020deep}. 
At inference time, given an input mesh, the server first uses TetWild~\cite{hu2018tetrahedral} to preprocess it and then uses a subdivision-based encoder adapted from SubdivNet~\cite{subdivnet} to map geometric details of the original mesh to high-dimensional per-face features of a sequence of decimated meshes. 
The mesh at the lowest (coarsest) level of resolution can then be transmitted to the client, which uses a subdivision-based decoder adapted from Neural Subdivision~\cite{neural_subdivision} to reconstruct a high-resolution mesh. 
The per-face features can be additionally transmitted to the client, and our subdivision decoder is trained to use them to further improve the quality of the reconstruction. 
We train the encoder and the decoder jointly on a large and heterogeneous collection of shapes using a reconstruction loss. 
To allow progressive refinement, we also introduce a sparsity loss on the per-face features, favoring the irrelevant features to be zero. 

We evaluate our network on the Thingi10K~\cite{thingi10k} dataset, split into the training, validation, and test sets.
We create a benchmark, evaluating the reconstruction quality given a prescribed transmission limit. 
We demonstrate that our method outperforms various baselines that use mesh decimation, subdivision, or progressive representations. 

\section{Related Work}

\paragraph{Mesh compression and simplification.} 
Lossless mesh compression techniques often use entropy encoding for geometry, connectivity, and surface attributes~\cite{touma1998triangle,szymczak2001edgebreaker,deering1995geometry,taubin1998geometric,rossignac1999edgebreaker,alliez2001valence}. 
These methods perfectly preserve the original details and thus are limited in their ability to reduce size. 
Geometry simplification techniques aim to reduce the polygon count while retaining the geometric features of the original mesh as much as possible~\cite{QSlim,lescoat2020spectral,szymczak2002piecewise,surazhsky2003explicit}. 
Neural mesh simplification techniques~\cite{neural_mesh_simplification} have also been proposed to address the greedy nature of classical techniques. 
All of these methods reduce the polygon count at the expense of losing original details. 
Any of these methods could be used to produce the initial coarse mesh in our framework. 
In this paper, we opted for QSlim~\cite{QSlim} due to its simplicity and since it preserves the manifoldness and the watertightness of the input. 

Progressive mesh representations have also been proposed~\cite{progressive_meshes} to transmit details incrementally, but this technique does not aim to compress data, and thus can only reconstruct the original high-fidelity shape by transmitting all geometric information. 

\paragraph{Surface upsampling.} 
Mesh subdivision is a common tool to refine coarse meshes~\cite{Loop,catmull1978recursively,Butterfly}. 
However, these methods employ hard-coded priors and usually recover (piecewise) smooth shapes, losing non-trivial details. 
One can optimize the coarse shape with respect to a particular subdivision scheme to maximize the reconstruction quality~\cite{Subdivfit}, but it has limited capabilities since the scheme itself stays fixed. 

Neural networks can be used to significantly expand the space of geometric details created with a subdivision scheme~\cite{neural_subdivision,Hertz2020deep}. 
We build upon the Neural Subdivision~\cite{neural_subdivision} framework. 
Unlike Neural Subdivision, which encodes the local geometric details into the weights of the neural network, our method takes advantage of a subdivision-based encoder that encodes the local geometric details into per-face features. 
We introduce skip connections~\cite{u-net} between the same level in the encoder and the decoder to improve the reconstruction quality and allow for progressive learnable features as additional input. 

\citet{Morreale22} also propose to represent a surface via two neural networks: one is a generic multi-layer perception (MLP) network that reconstructs coarse surfaces and the other is a detailization convolutional architecture. 
Unlike our method, this technique requires training two separate neural networks for each shape instance and thus requires transmitting the entire neural network for each asset, preventing it from effectively compressing details shared across different shapes and increasing computational costs.

\begin{figure*}[!t]
  \centering
  \includegraphics[width=\linewidth]{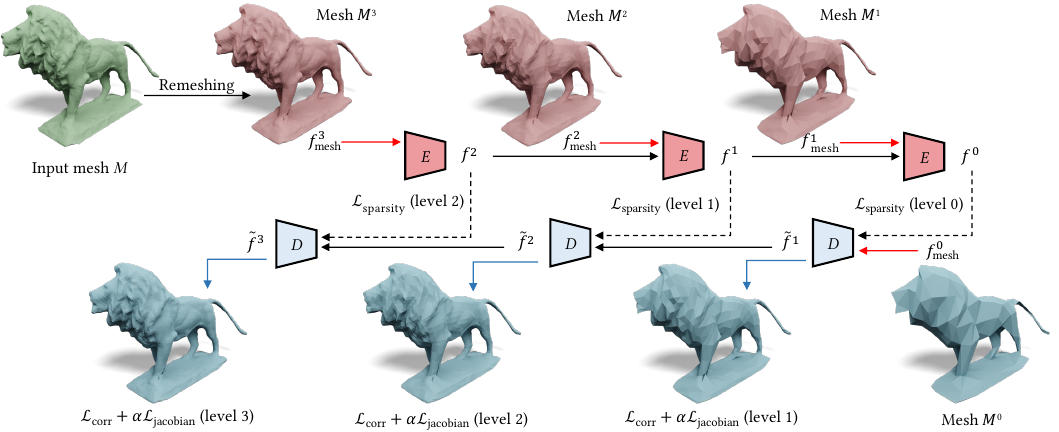}
  \caption{
  \textbf{Overview of \algoNameFull.}
  Our network consists of an encoder $E$ and a decoder $D$. 
  Given an input mesh $M$ to be transmitted, we first apply a remeshing scheme to obtain a sequence of LoD meshes $M^L \dots M^0$. 
  The encoder learns per-face features that encode the local geometric details of each LoD mesh. 
  The decoder learns to reconstruct a high-resolution mesh based on the transmitted coarsest mesh $M^0$. 
  The reconstruction quality can be iteratively improved with features being progressively transmitted from the encoder to the decoder.
  }
  \label{fig:network}
\end{figure*}

\paragraph{Geometric deep learning.}
A large number of neural shape representations have recently been developed. Unsupervised representation learning techniques often follow an encoder-decoder architecture~\cite{bengio2009learning,baldi2012autoencoders}, where the encoder network maps shapes to a high-dimensional feature code and the decoder reconstructs the original data. 
Existing shape encoders include 2D CNNs over projections of shapes~\cite{su2015multi,mccormac2017semanticfusion}, 3D CNNs over voxel grids~\cite{tran2015learning}, point-based architectures~\cite{pointnet,pointnet++}, surface-based techniques~\cite{fields_conv}, mesh-based approaches that learn directly over the discrete mesh structure~\cite{meshcnn}. 
SubdivNet~\cite{subdivnet} is a mesh-based approach that operates on meshes with Loop~\cite{Loop} subdivision sequence connectivity.
The goal is to learn per-face features for dense prediction tasks on mesh surfaces. 
We build upon SubdivNet and develop a subdivision-based encoder.
Our goal is different from SubdivNet in that we aim to encode the local geometric details into per-face features and use them to guide the subdivision process in the decoder.

One can also directly optimize for shape codes with respect to some reconstruction objective without using an encoder~\cite{deepsdf}. 
Existing decoders generate shapes by folding 2D atlases into 3D surfaces~\cite{AtlasNet}, deforming template meshes~\cite{3D-CODED}, or predicting occupancies~\cite{occNet} or SDFs~\cite{deepsdf,implicit_fields}. 

While one can view an encoder-decoder architecture as an extreme version of neural compression, where the entire shape is compressed to a single feature vector. 
Existing decoders usually do not perform well at reconstructing high-resolution mesh details in comparison to surface upsampling techniques. 
By transmitting an optimized low-resolution mesh, we can also guarantee the preservation of the original topology and the watertightness of the input. 

\section{Method}

\subsection{Overview}

Given a triangle mesh $M = (V, F)$ with vertex positions $V$ and faces $F$, our goal is to encode it into a data stream $d_{0 \dots t}$ which could be progressively transmitted to the client, where the client should be able to reconstruct a low-resolution shape from initial transmission $d_0$ and then iteratively improve it with subsequent transmissions $d_{1 \dots t}$ for some bandwidth and time budget $t$. 

We first obtain a sequence of LoD meshes $M^0 \dots M^L$, where $M^i=(V^i, F^i)$ and $M^0$ is the coarsest mesh with a fixed number of faces. The triangulation of each subsequent mesh $F^1\dots F^L$ is defined by a simple subdivision rule iteratively applied to $F^0$. We also preserve correspondences during the simplification step and use this mapping to define vertex positions at each level of subdivision $V^0\dots V^L$. 
When using our progressive representation, the server first transmits the coarsest mesh $d_0=M^0$ to the client. The client uses the same subdivision scheme to reconstruct the sequence of meshes $\tilde{M}^i=(\tilde{V}^i, F^i)$, where $i=1 \dots L$ (note that we apply the same subdivision rule on the client side, and thus the triangulation is exactly the same). We assume that the mesh at the highest level of subdivision $M^L$ is sufficiently close to the input for all practical purposes, and thus reduce our problem to designing a method that can efficiently compress vertex coordinates $V^L$ on the server side and enable reconstructing the coordinates $\tilde{V}^L$ on the client side.

To take advantage of the shared structures in local mesh geometry, we use an encoder-decoder approach, where we first encode geometric details as per-face features at each level $i$: $f^i$, and then the decoder uses the features to reconstruct the mesh at the highest level of detail. Our encoder $E$ directly leverages LoD meshes by learning filters that map features and vertex coordinates from higher-resolution to lower-resolution level: $E: [V^i, f^i] \rightarrow [V^{i-1}, f^{i-1}]$. Our decoder $D$, in a similar manner, maps from lower-resolution to higher-resolution level: $D: [V^i, f^i] \rightarrow [V^{i+1}, f^{i+1}]$. To speed up training and improve the quality of learned features, we connect corresponding faces at the same level of detail with skip connections, akin to U-Net architectures for images~\cite{u-net}. We train our network so that the decoder can still reconstruct a plausible shape via the learned subdivision process even before all features $f^i$ are transmitted. Specifically, to facilitate compression, we introduce a feature sparsity loss, which aims to set per-face features to 0 if they do not aid in reconstruction. In addition, we also have classical reconstruction losses based on vertex coordinates and their differential properties. See Figure~\ref{fig:network} for an illustration of our network architecture and training losses.

At inference time, after the initial coarse mesh $M^0$ is transmitted to the client, the decoder reconstructs the high-resolution mesh by running the learned subdivision process with per-face features $f^i$ set to 0. Our subsequent transmissions $d_{1 \dots t}$ simply assign non-zero features to some selected faces, enabling us to progressively improve the quality of the reconstructed shape. Due to the sparsity loss, we can simply sort the features by magnitude.

\subsection{\algoNameFull}

In this subsection, we discuss our neural progressive representation, including preprocessing and the encoder and decoder architectures.

\paragraph{LoD preprocessing.} 
To derive our LoD $M^0 \dots M^L$ representation used in the encoder, we first decimate the input mesh $M$ via QSlim~\cite{QSlim} to obtain a coarse mesh $M^0$ with $|F^0|=400$ faces. 
The target number of faces for simplification is picked to yield sufficiently coarse meshes to facilitate compression but also retain enough topological details for subdivision. To subdivide the coarse mesh into higher-resolution meshes, we first split each edge at the midpoint, subdividing each triangle into 4. This gives us the triangulations $F^1\dots F^L$, where $|F^i| = 4|F^{i-1}|$. To get vertex coordinates at the subdivision levels, we use successive self-parameterization~\cite{neural_subdivision}, which allows us to map each point on each mesh $M^i$ to the original mesh $M$, and we use that coordinate for all vertices: $V^1 \dots V^L$. We set $L=3$ for all experiments, which is selected to give us enough triangle budget to reconstruct shapes in our dataset. We sometimes refer to this step as remeshing, as $M^L$ could be viewed as a remeshed version of $M$, which has similar geometry but a different triangulation.

\paragraph{Encoder.} 
Our encoder $E$ operates on the sequence of LoD meshes: $M^L, M^{L-1}\dots M^0$, where all triangles in a high-resolution mesh can be grouped into groups of four and mapped to a single triangle on the next level of resolution based on the LoD scheme. We define convolution and pooling operators based on this mapping following SubdivNet~\cite{subdivnet}. 
The input per-face features at the highest level in the encoder are 13-dimensional (i.e., $f_\text{mesh}^L \in \mathbb{R}^{13}$), composed of a 7-dimensional shape feature (face area, three interior angles, and the inner product between the face normal and the vertex normals) and a 6-dimensional pose feature (face center coordinate and face normal). 
Unlike SubdivNet, where the input per-face features at the subsequent levels are just the output per-face features from the previous level, in our encoder, the input per-face features at the subsequent levels are a concatenation of the output per-face features from the previous level (i.e., $f^i$) and the 13-dimensional per-face features computed based on mesh $M^i$ (i.e., $f_\text{mesh}^i$). 
This design allows us to encode the local geometric details of each LoD mesh into the feature encoding process. 
Our encoder maps the input per-face features at level $i$ to learned per-face features at the subsequent level (i.e., level $i-1$): $f^{i-1} \in \mathbb{R}^8,~\forall~1 \leq i \leq L$. 

\begin{figure}[!t]
  \centering
  \includegraphics[width=\linewidth]{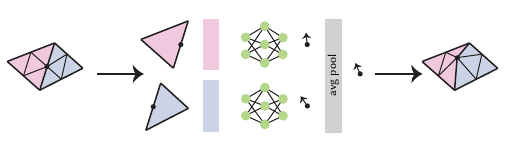}
  \caption{
  \textbf{Vertex position prediction.} 
  The decoder takes as input the features of the two adjacent faces and predicts the displacement for the midpoint. 
  }
  \label{fig:fine-mesh-vertex-update}
\end{figure}

\paragraph{Decoder.}
Our decoder $D$ operates on the mesh transmitted at the coarsest level $M^0$ and can optionally leverage learned per-face features $f^i$. We first compute 13-dimensional shape and pose features (as described in the previous paragraph) of the coarse mesh $M^0$ to derive the per-face features $f_\text{mesh}^0$. We then concatenate features $f_\text{mesh}^0$ with learned per-face features $f^0$. If features $f^0$ are not transmitted, we simply set $f^0=0$ for all faces. We treat the concatenated features as the input to the decoder. 
Instead of using half-flap representations as Neural Subdivision does, we adapt the Neural Subdivision architecture and develop a subdivision-based decoder that uses the features of the two adjacent triangles to predict vertex positions at the next level of subdivision $\tilde{V}^i, i=1 \dots L$, as shown in Figure~\ref{fig:fine-mesh-vertex-update}. 
Our decoder maps the input per-face features at level $i$ and optionally the learned per-face features transmitted from the same level in the encoder to per-face features at the next subdivision level in the decoder (i.e., level $i+1$): $f^{i+1} \in \mathbb{R}^8,~\forall~0 \leq i \leq L-1$. 

\begin{figure*}[t]
  \centering
  \mpage{0.19}{\includegraphics[width=\linewidth]{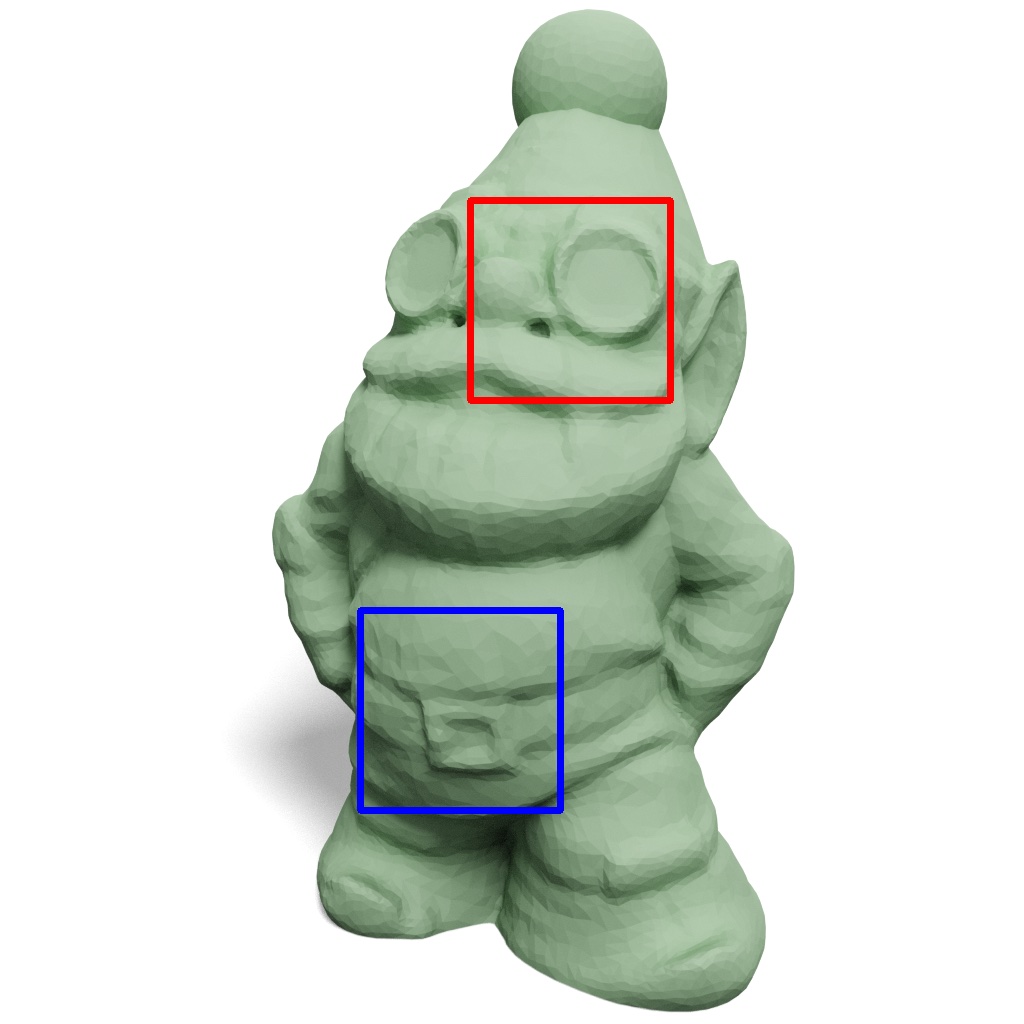}} \hfill
  \mpage{0.19}{\includegraphics[width=\linewidth]{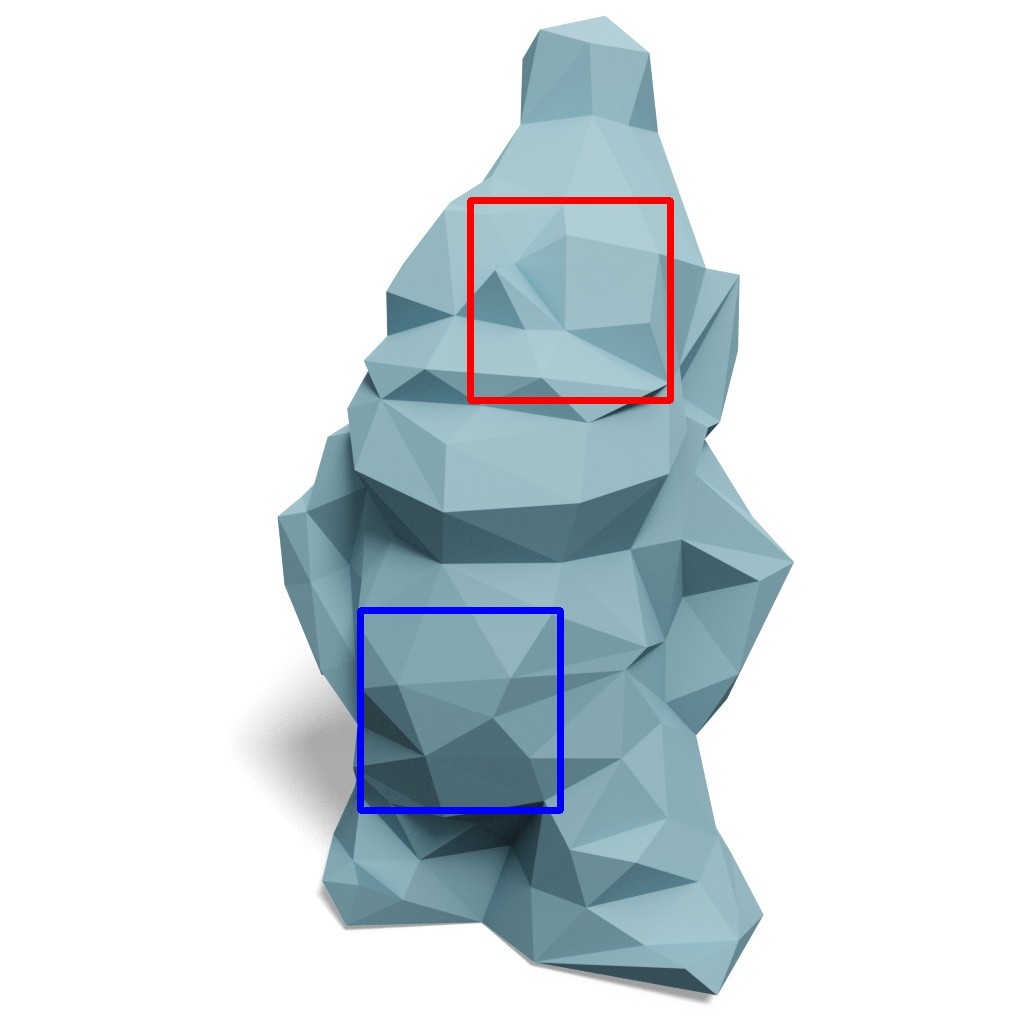}} \hfill
  \mpage{0.19}{\includegraphics[width=\linewidth]{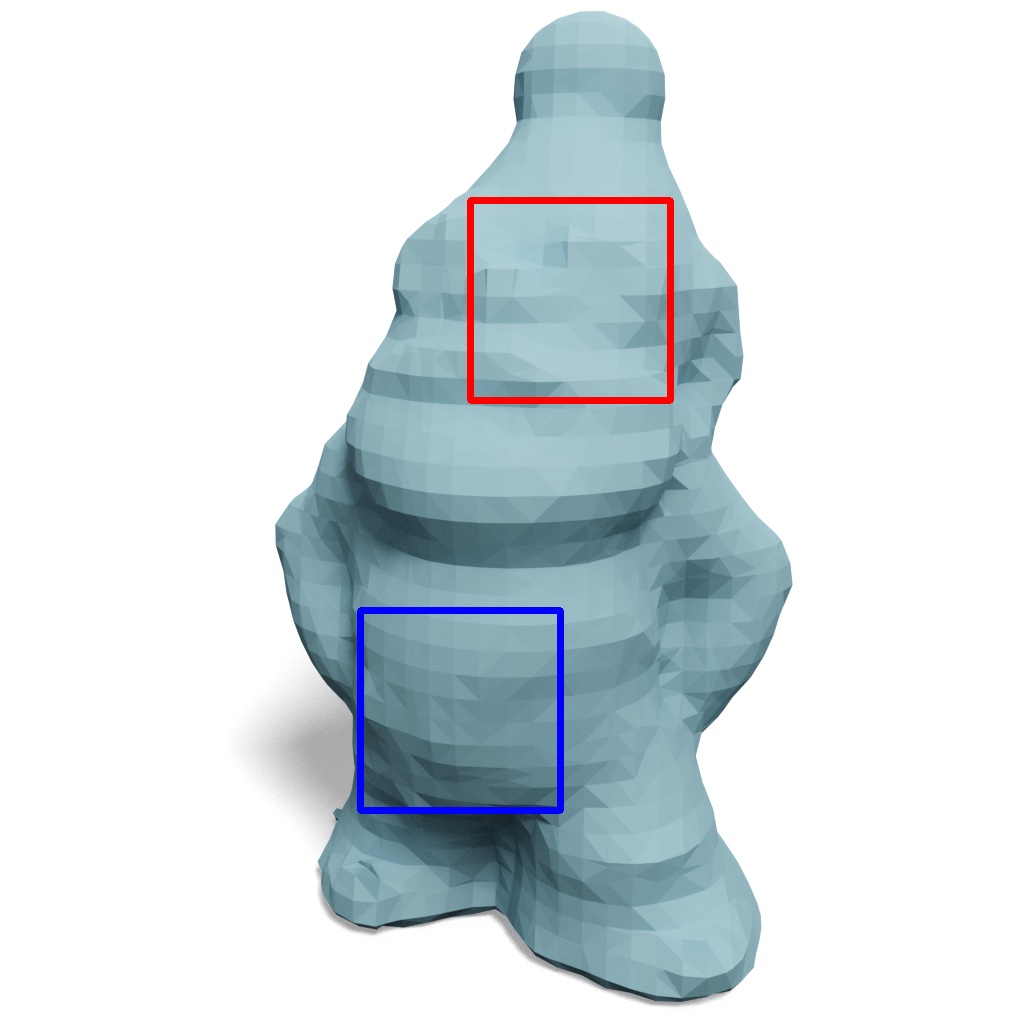}} \hfill
  \mpage{0.19}{\includegraphics[width=\linewidth]{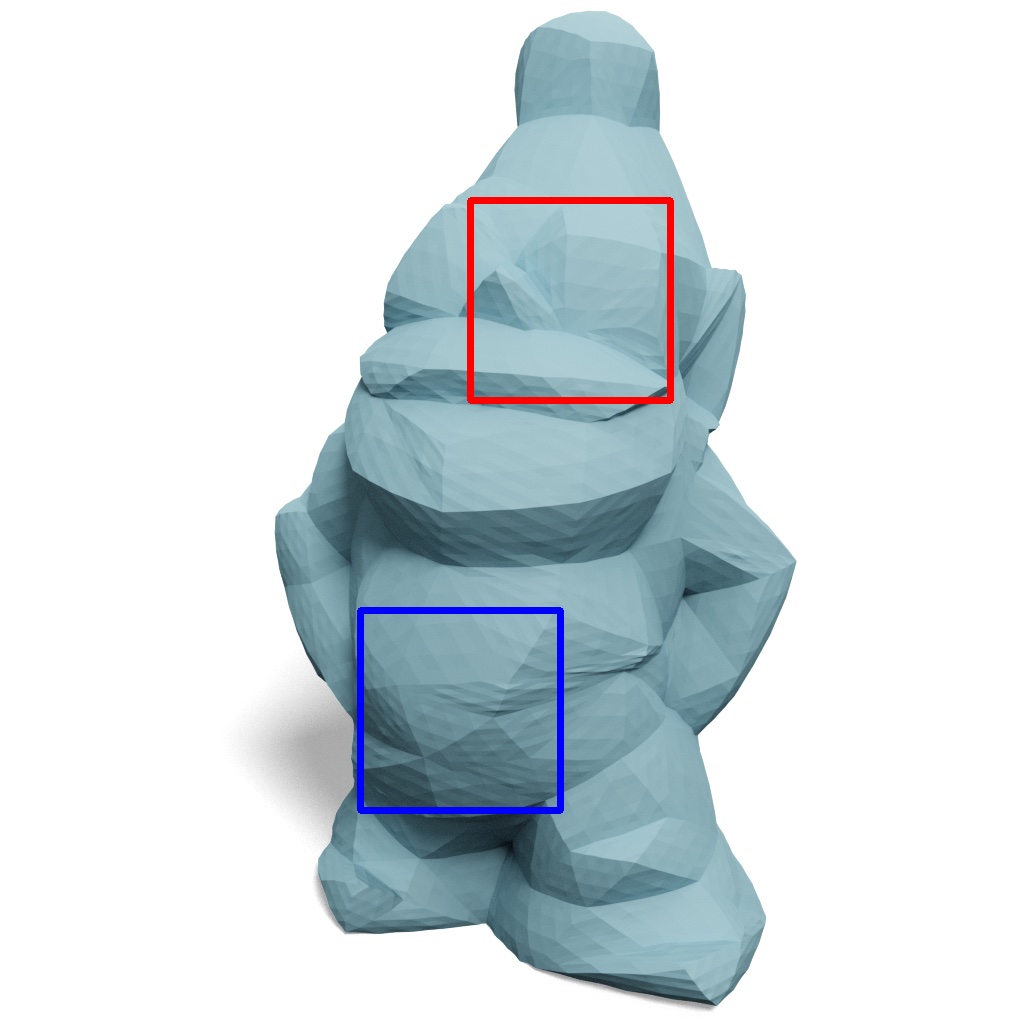}} \hfill
  \mpage{0.19}{\includegraphics[width=\linewidth]{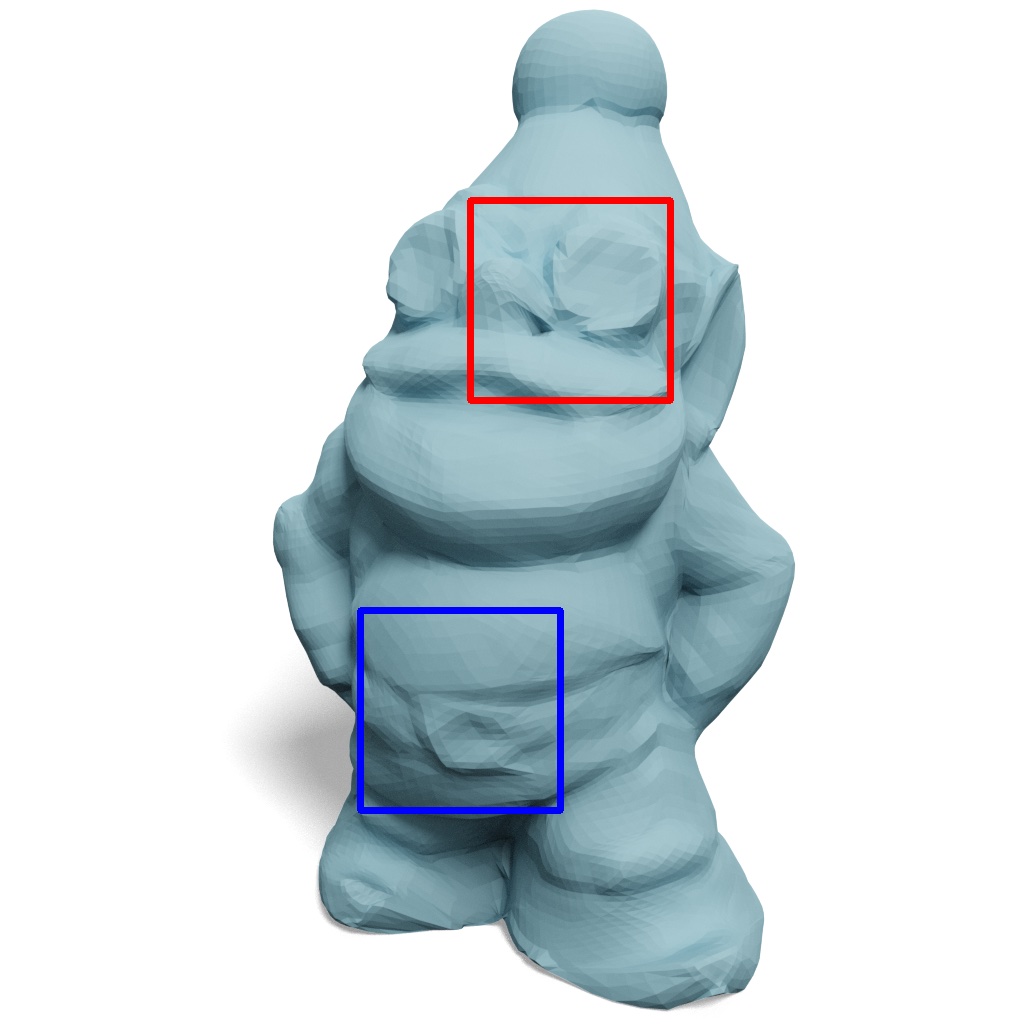}} \\
  \vspace{1.0mm}
  \mpage{0.19}{\includegraphics[width=0.475\linewidth]{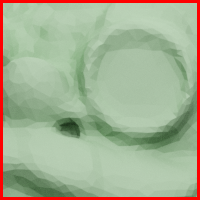} \hfill \includegraphics[width=0.475\linewidth]{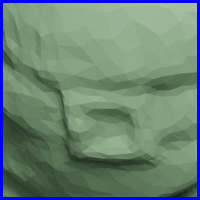}} \hfill
  \mpage{0.19}{\includegraphics[width=0.475\linewidth]{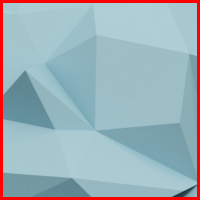} \hfill \includegraphics[width=0.475\linewidth]{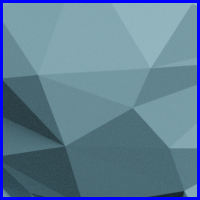}} \hfill
  \mpage{0.19}{\includegraphics[width=0.475\linewidth]{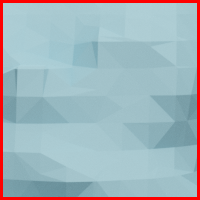} \hfill \includegraphics[width=0.475\linewidth]{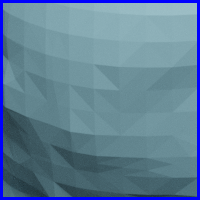}} \hfill
  \mpage{0.19}{\includegraphics[width=0.475\linewidth]{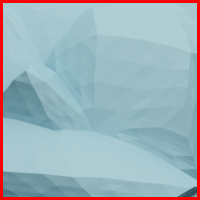} \hfill \includegraphics[width=0.475\linewidth]{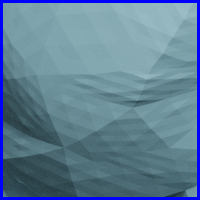}} \hfill
  \mpage{0.19}{\includegraphics[width=0.475\linewidth]{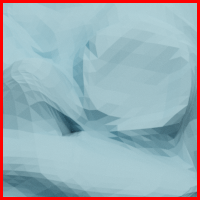} \hfill \includegraphics[width=0.475\linewidth]{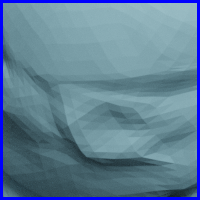}} \\
  \vspace{1.0mm}
  \mpage{0.19}{$d_\text{pm}$ ($\times 10^{-4}$) / $d_\text{normal}$} \hfill
  \mpage{0.19}{44.94 / 17.42$^\circ$} \hfill
  \mpage{0.19}{28.56 / 15.78$^\circ$} \hfill
  \mpage{0.19}{17.09 / 13.69$^\circ$} \hfill
  \mpage{0.19}{5.32 /  8.62$^\circ$} \\
  \vspace{1.0mm}
  \mpage{0.19}{Ground truth} \hfill
  \mpage{0.19}{QSlim} \hfill
  \mpage{0.19}{SubdivFit} \hfill
  \mpage{0.19}{Neural Subdivision} \hfill
  \mpage{0.19}{Ours} \\
  \caption{
  \textbf{Visual comparisons with decimation and subdivision methods on Thingi10K.} 
  The compression ratio ($CR$) is 61.39 for all methods.
  }
  \label{fig:exp-thingi10k-results}
\end{figure*}

\subsection{Network Training}
We train our encoder-decoder network end-to-end using reconstruction and sparsity losses. The former favors higher quality of reconstruction and the latter favors sparser features and thus compression of the signal. 

\paragraph{Reconstruction losses.}
Our reconstruction loss is composed of two terms. First, the $\ell_2$ distance between vertex positions predicted by the decoder and true LoD positions:
\begin{equation}
  \mathcal{L}_\text{corr} = \sum_{i=1}^{L}\frac{1}{|V^i|} \|\tilde{V}^i - V^i\|_2.
  \label{eq:corr-loss}
\end{equation} 
The second term is a loss in the gradient domain, measuring the similarity of Jacobians, which helps match differential properties of true and predicted LoD surfaces, such as normals and curvature:
\begin{equation}
  \mathcal{L}_\text{jacobian} =
  \sum_{i=1}^L
  \frac{1}{|F^i|}\sum_{j=1}^{|F^i|}\|J_j^i-I\|_2,
  \label{eq:jacobian-loss}
\end{equation}
where $J^i_j$ is the Jacobian of the deformation that maps the $j^\text{th}$ triangle of the true LoD mesh $M^i$ to its predicted counterpart $\tilde{M}^i$, and $I$ is the identity matrix. 

\paragraph{Sparsity loss.}
To avoid transmitting features that encode redundant information in regions whose geometry could be inferred by the decoder without any aid, we introduce a sparsity loss:
\begin{equation}
  \mathcal{L}_\text{sparsity} =
  \sum_{i=0}^{L-1} \frac{1}{|F^i|} \|f^i\|_1.
\end{equation}
After network training, we sort the features based on the magnitude and transmit them progressively from the encoder to the decoder. 

\paragraph{Total loss.}
We now define total training loss as a sum of weighted terms with $\alpha=1$ and $\beta=0.1$ for all experiments:
\begin{equation}
  \mathcal{L} = \mathcal{L}_\text{corr} + \alpha \mathcal{L}_\text{jacobian} + \beta \mathcal{L}_\text{sparsity}.
\end{equation}

\begin{figure*}[t]
  \centering
  \mpage{0.19}{\includegraphics[width=\linewidth]{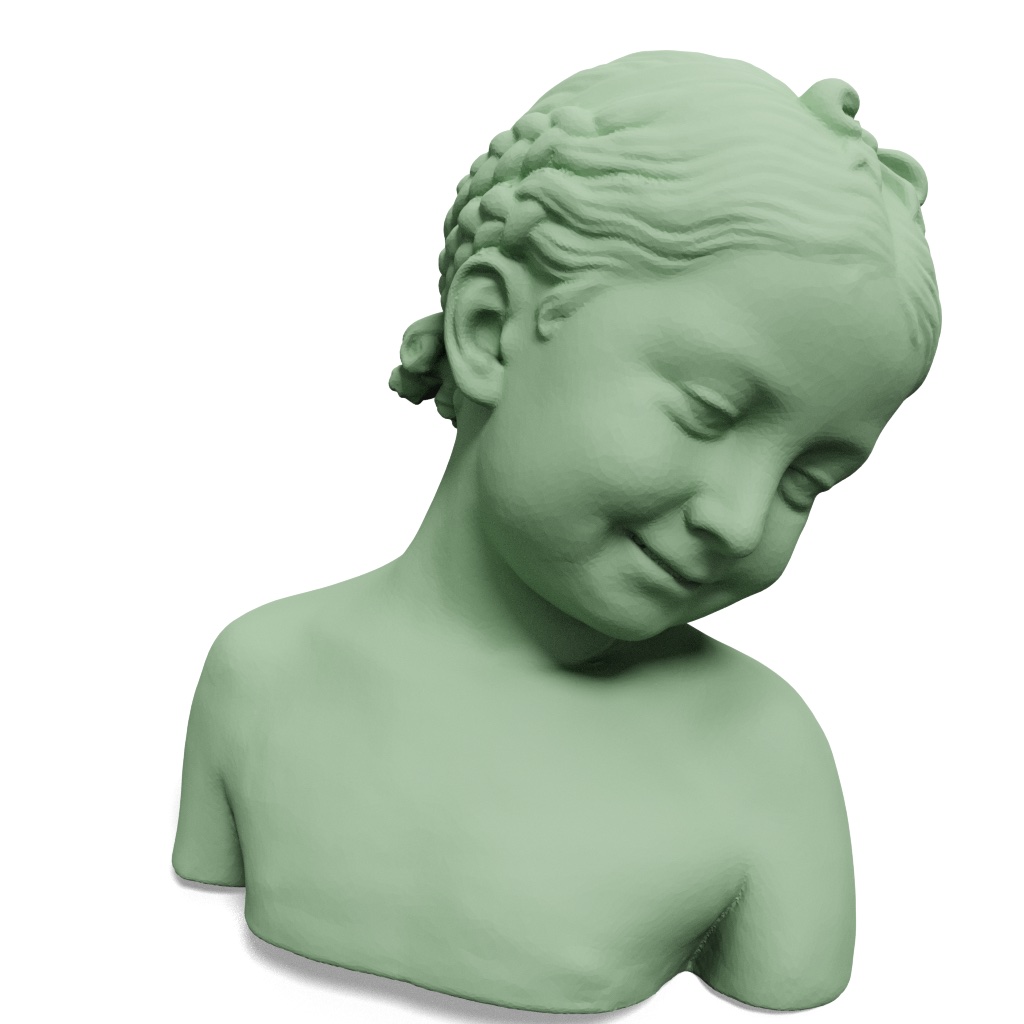}} \hfill
  \mpage{0.19}{\includegraphics[width=\linewidth]{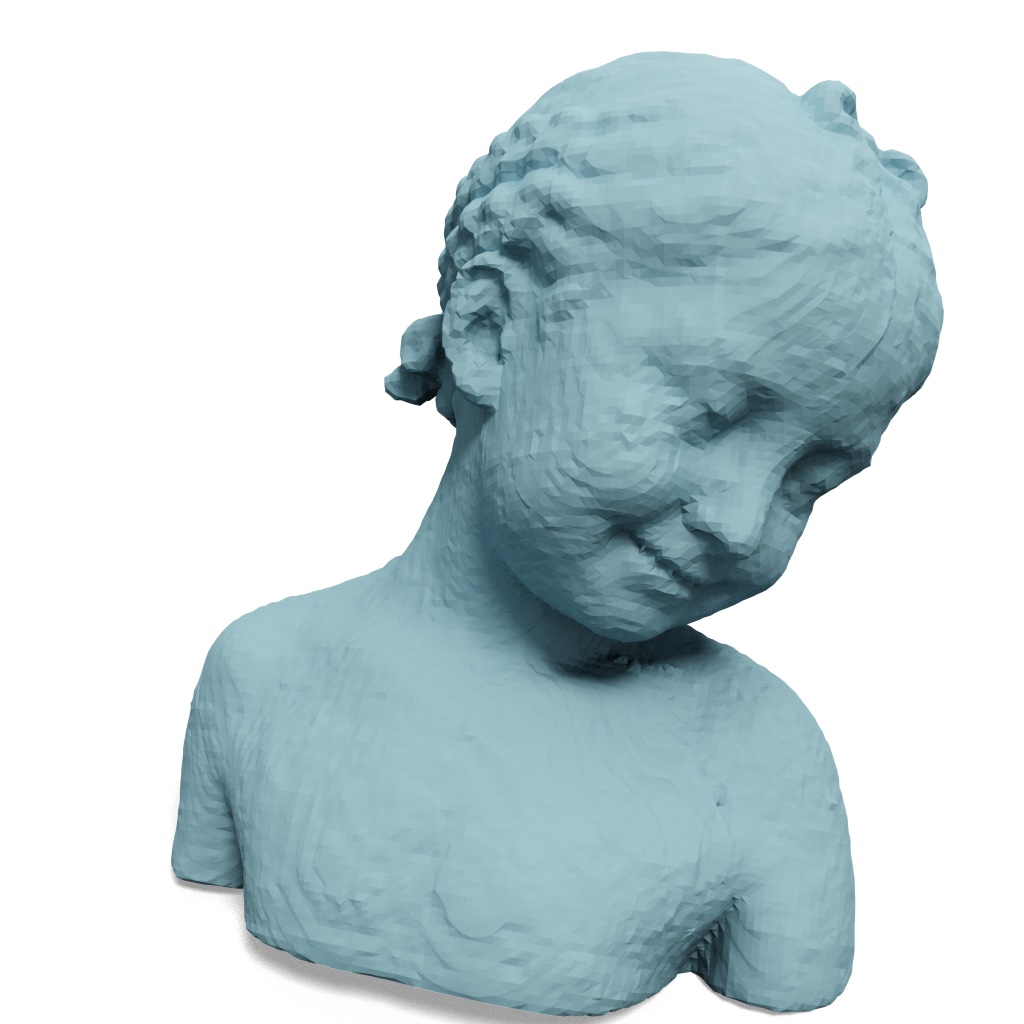}} \hfill
  \mpage{0.19}{\includegraphics[width=\linewidth]{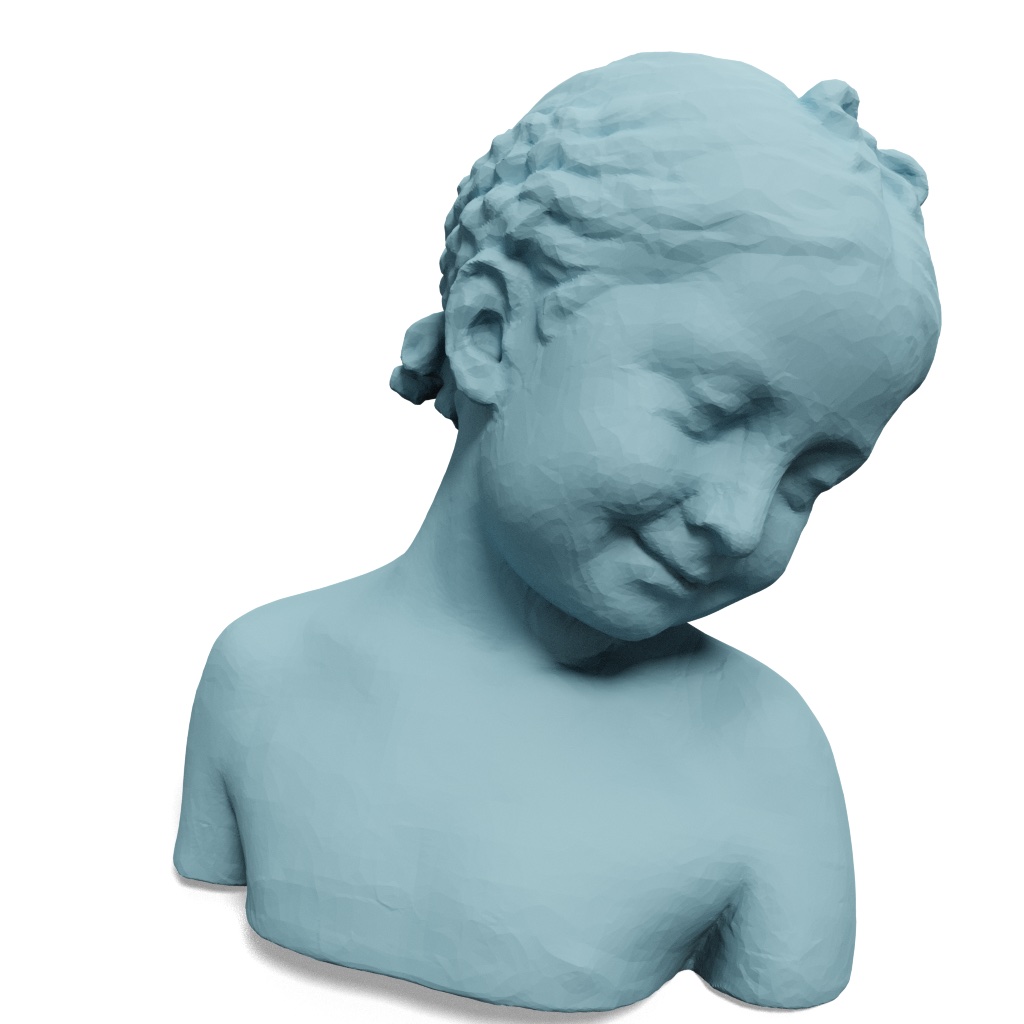}} \hfill
  \mpage{0.19}{\includegraphics[width=\linewidth]{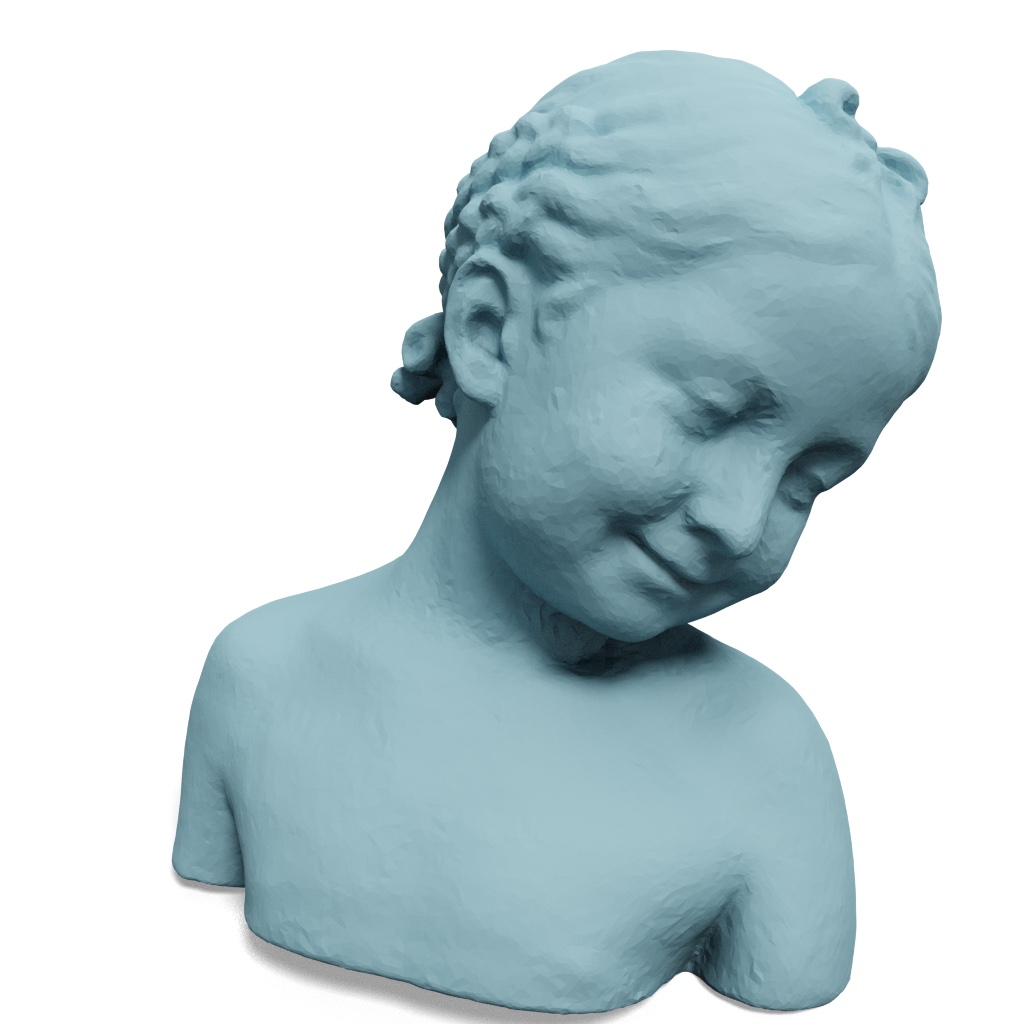}} \hfill
  \mpage{0.19}{\includegraphics[width=\linewidth]{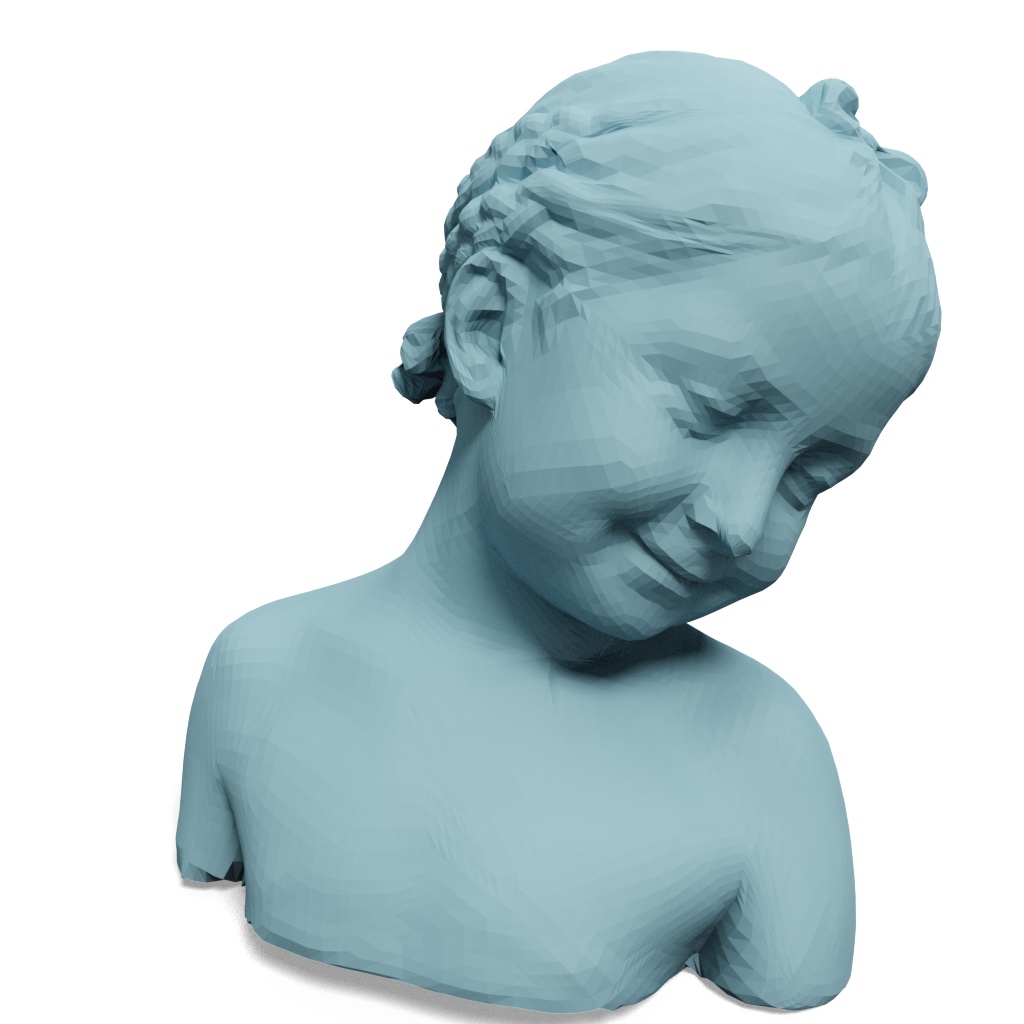}} \\
  \vspace{1.0mm}
  \mpage{0.19}{$CR$ / $d_\text{pm}$ ($\times 10^{-4}$) / $d_\text{normal}$} \hfill
  \mpage{0.19}{1.15 / 2.80 / 10.15$^\circ$} \hfill
  \mpage{0.19}{1.15 / 2.91 / 8.79$^\circ$} \hfill
  \mpage{0.19}{1.15 / 2.52 / 4.73$^\circ$} \hfill
  \mpage{0.19}{39.32 / 2.74 / 8.81$^\circ$} \\
  \vspace{1.0mm}
  \mpage{0.19}{Ground truth} \hfill
  \mpage{0.19}{ACORN} \hfill
  \mpage{0.19}{NGLOD} \hfill
  \mpage{0.19}{NCS} \hfill
  \mpage{0.19}{Ours} \\
  \caption{
  \textbf{Visual comparisons with neural overfitting methods.} 
  Our result is comparable to neural overfitting methods in $d_\text{pm}$ and $d_\text{normal}$ while having a much higher compression ratio.
  }
  \label{fig:exp-others-results}
\end{figure*}

\paragraph{Input preconditions.}
Our approach is not constrained to watertight, non-self-intersecting, near-delaunay triangulated meshes, and can be trained with such data by adding a preprocessing step. 
Given an input mesh, we use TetWild~\cite{hu2018tetrahedral} to preprocess it.

\section{Experiments}

\subsection{Experimental Setup}

\paragraph{Dataset.}
We evaluate our method on Thingi10K~\cite{thingi10k}, a dataset of diverse models with interesting geometric and topological features. We start with the preprocessed watertight meshes provided by \citet{hu2018tetrahedral} and filter out models that have more than 1 connected component or are not edge manifold, leaving us with 6,418 meshes. 
We sample 1,000 meshes and split them into training (80\%), validation (10\%), and test (10\%) sets for experiments. 

\bgroup
\def\arraystretch{1.2}
\begin{table}[t]
  \begin{center}
  \caption{
  \textbf{Comparisons with baseline methods on Thingi10K.} 
  All methods have the same compression ratio ($CR=61.39$). 
  Our method performs the best on both metrics that measure the quality of reconstruction.
  }
  \label{table:exp-sota-results}
  {
  \addtolength{\tabcolsep}{8.7pt}
  \begin{tabular}{lrr}
    \toprule
    \rowcolor{LavenderBlue}
    Method & $d_\text{pm}$ ($\times 10^{-4}$) $\downarrow$ & $d_\text{normal}$ $\downarrow$ \\  
    \midrule
    QSlim & 30.11 & 13.21$^\circ$ \\
    \rowcolor[HTML]{EFEFEF}
    Loop & 68.47 & 14.98$^\circ$ \\
    Butterfly & 40.55 & 16.99$^\circ$ \\
    \rowcolor[HTML]{EFEFEF}
    SubdivFit & 27.33 & 15.41$^\circ$ \\
    Neural Subdivision & 19.03 & 11.21$^\circ$ \\
    \rowcolor[HTML]{EFEFEF}
    Ours & 4.12 & 7.19$^\circ$ \\
    \bottomrule
  \end{tabular}
  }
  \end{center}
\end{table}
\egroup

\paragraph{Evaluation metrics.}
We adopt the mean point-to-mesh distance $d_\text{pm}$ and the average normal error $d_\text{normal}$ to measure the similarity of the reconstructed mesh to the ground truth. The mean point-to-mesh distance first uniformly samples 1 million points on the surface of the final subdivided mesh $\tilde{M}^L$. Then it computes the average distance between each sampled point and the ground-truth mesh $M$. The normal error computes the average angle (in degrees) between the normal of the sampled point on mesh $\tilde{M}^L$ and the normal of the projected point on the ground-truth mesh $M$.

We use the compression ratio ($CR$) to evaluate the effectiveness of different methods in transmitting LoDs. Since we did not want our metric to be skewed by various additional potential post-processes: generic compression algorithms, changing floating-point resolution, and topology-specific compression, we decided not to account for the transmission of topological data. However, in practice, this puts our method at a disadvantage, since we only need to transmit the topology for the coarse mesh, where all other triangulations are defined by subdivision. For all methods, we measure $CR$ as:
$$
CR = \frac{3|V|}{3|V^0|+\sum_{f\in\mathcal{T}} \text{dim}(f) },
$$
where $\text{dim}(f)$ is the size of all transmitted features $f\in\mathcal{T}$.

\paragraph{Implementation details.}
Each level in our encoder is composed of a mesh convolution~\cite{subdivnet}, a batch normalization~\cite{BatchNorm}, a ReLU~\cite{ReLU}, and an average pooling~\cite{subdivnet}.
Each level in our decoder is composed of two modules, one for predicting vertex displacements and the other for feature learning. The vertex displacement prediction module is a single fully connected layer. 
The feature learning module consists of a bilinear upsampling~\cite{subdivnet}, a mesh convolution~\cite{subdivnet}, a batch normalization, and a ReLU. 

We train our network using the ADAM~\cite{ADAM} optimizer in PyTorch~\cite{PyTorch}. The initial learning rate is set to $1 \times 10^{-3}$, and the learning rate decay is set to $1 \times 10^{-6}$. We scale each mesh in the dataset to fit a unit cube. We then employ two data augmentation strategies. First, we randomly decimate each mesh to 10 different coarse meshes. Second, we randomly rotate the meshes at all levels of subdivision by a random rotation (in $90^\circ$ increments) around each of the three axes. 

\subsection{Mesh Compression}
We evaluate our method and several baselines for the mesh compression task using compression and reconstruction metrics. Since all methods will have a tradeoff between the compression ratio and the reconstruction quality, for all comparisons presented in this subsection we set the parameters of each method to reach the same compression ratio as ours, and only compare the reconstruction quality. We show quantitative results in Table~\ref{table:exp-sota-results} and qualitative results in Figures~\ref{fig:exp-thingi10k-results} and ~\ref{app-fig:exp-thingi10k-results} for our method and baselines. 
See Section 3 in the supplemental material for more results of our method. 
We next detail the choice of our baselines.

We first compare to a mesh decimation approach, QSlim~\cite{QSlim}, which reduces the mesh size by greedily collapsing edges with the lowest quadric error. This method is expected to lose details since it does not have an upsampling step. One can run a classical subdivision scheme (e.g., Loop~\cite{Loop} or Butterfly~\cite{Butterfly}). However, as observed in Table~\ref{table:exp-sota-results}, it only worsens the performance, since these methods are not aware of the priors used by QSlim. One can specifically optimize the simplified mesh based on the subdivision method, e.g., SubdivFit~\cite{Subdivfit}, which improves the reconstruction accuracy. We find the learnable subdivision method (Neural Subdivision~\cite{neural_subdivision}) further improves the accuracy. Note that our method yields the highest accuracy at the same level of compression since it learns surface-specific features across a collection of shapes and can adaptively transmit features only in regions that need the most details. 

\subsection{Comparison to Mesh Compression by Neural Overfitting}
A few recent techniques have been proposed for compressing data by overfitting neural representations to high-resolution meshes, e.g., NCS~\cite{Morreale22}, ACORN~\cite{ACORN}, and NGLOD~\cite{NGLOD}. These methods require transmitting all network weights for each mesh, do not learn a space of local details, and have been overfitted to very high-resolution meshes. Since these methods would not be very effective at our lower-resolution meshes, we run our method on their data and show results in 
Figure~\ref{fig:exp-others-results}. 
The compression ratio for these methods is defined as the ratio between the ground truth mesh file size and the network file size. 
Even though our method was trained on much lower resolution data, it achieves a comparable quality of reconstruction (measured by $d_\text{pm}$ and $d_\text{normal}$) compared to neural overfitting baselines while offering a much higher compression ratio.

\subsection{Progressive Meshes}
We now demonstrate progressive transmission, a key feature of our method. Note that all mesh compression techniques we discussed so far can only transmit a single shape with a particular bandwidth budget. To upgrade the resolution of the shape, the entire mesh needs to be re-transmitted at a higher resolution, potentially transmitting redundant information multiple times. 

To evaluate the effectiveness of progressively transmitting features, we conduct an analysis by varying the number of features transmitted from the encoder to the decoder and look at the quality of the resulting reconstructions (see Table~\ref{table:exp-ablation-prog-feat}, Figures~\ref{fig:exp-progressive-featature} and ~\ref{app-fig:prog-feat}). 
See Section 2 in the supplemental material for more visual results. 
Note the gradual improvement in the quality of the reconstructed local details as more features are being transmitted. 

\begin{figure*}[t]
  \centering
  \mpage{0.235}{\includegraphics[width=\linewidth]{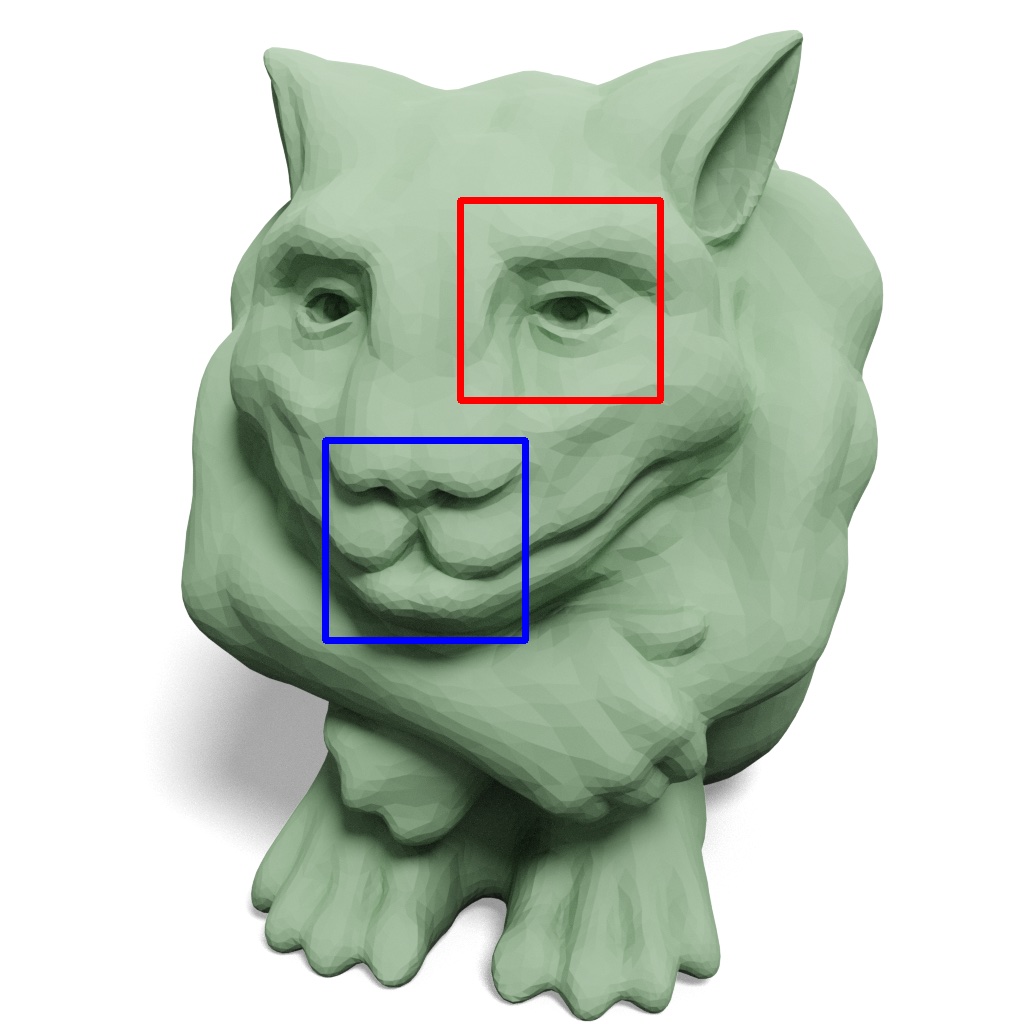}} \hfill
  \mpage{0.235}{\includegraphics[width=\linewidth]{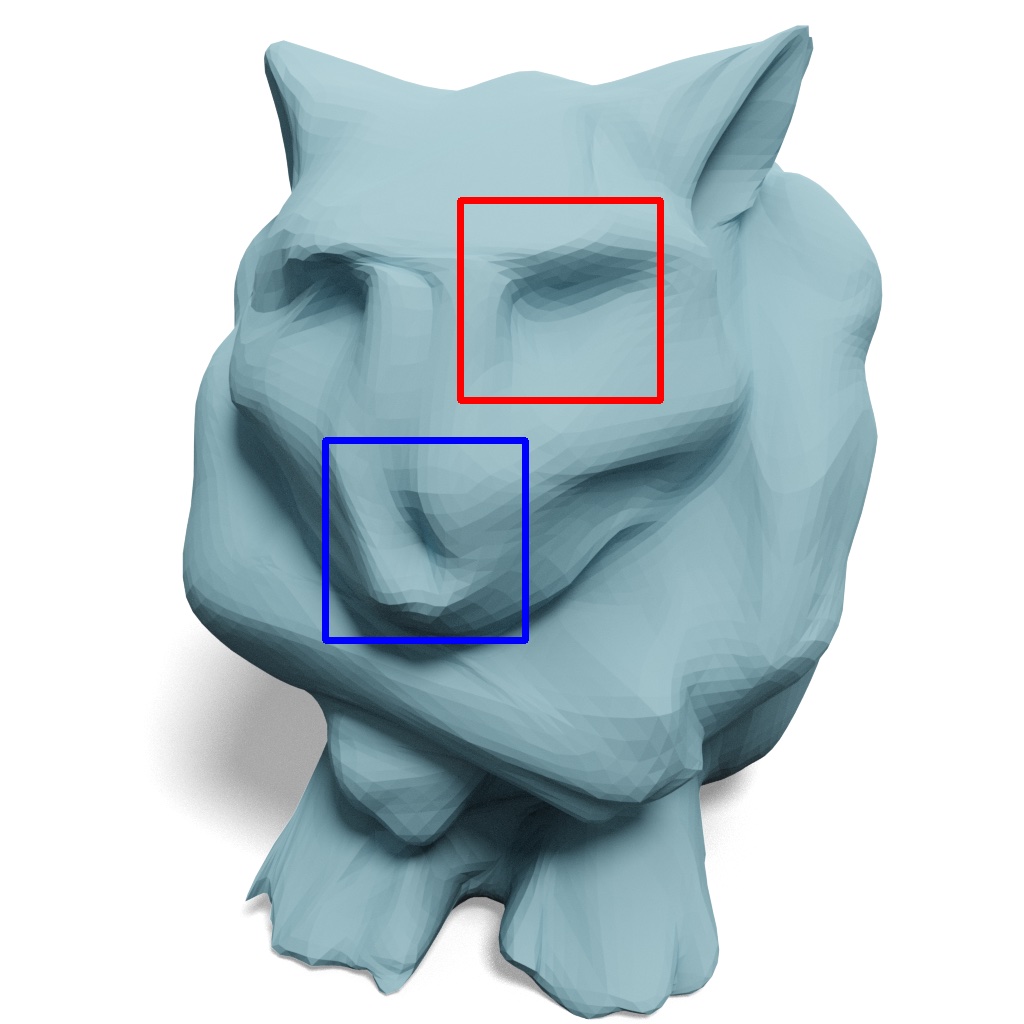}} \hfill
  \mpage{0.235}{\includegraphics[width=\linewidth]{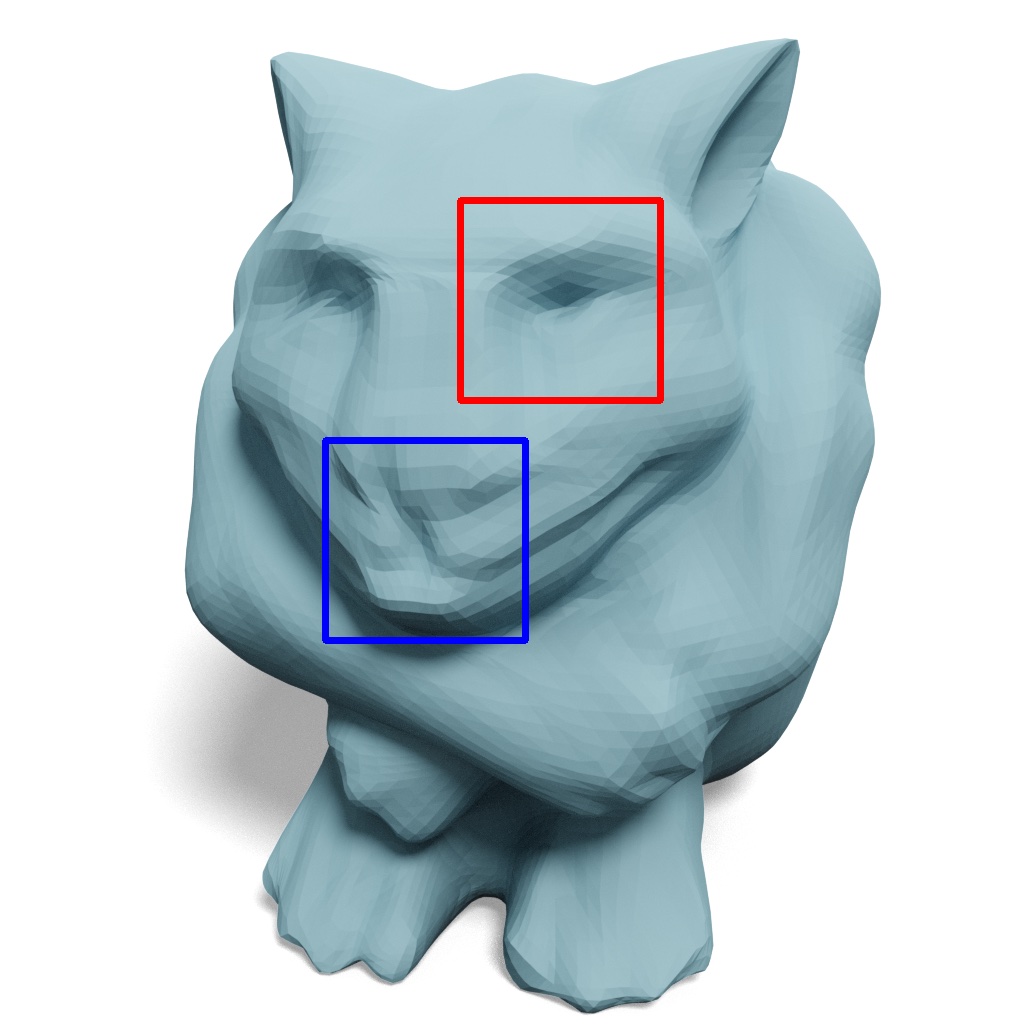}} \hfill
  \mpage{0.235}{\includegraphics[width=\linewidth]{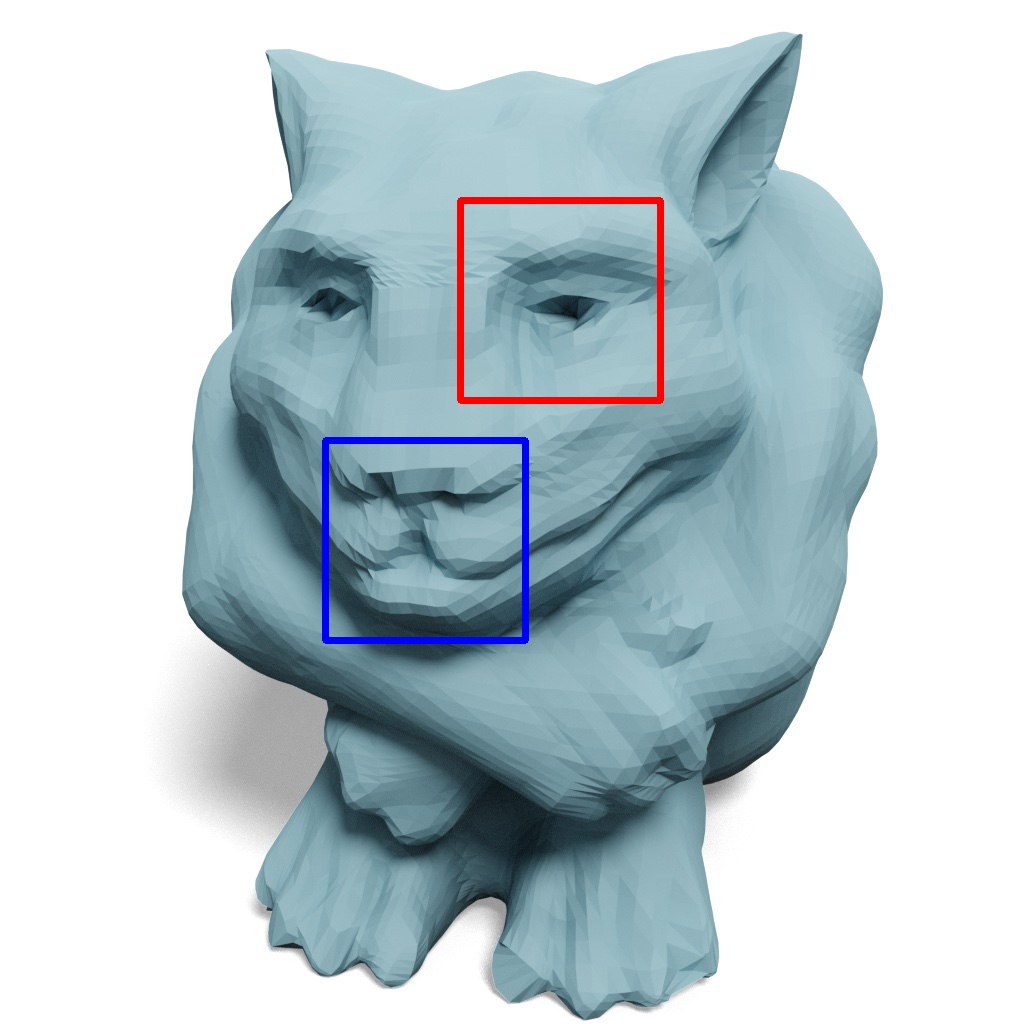}} \\
  \vspace{1.0mm}
  \mpage{0.235}{\includegraphics[width=0.475\linewidth]{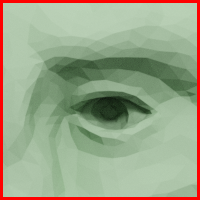} \hfill \includegraphics[width=0.475\linewidth]{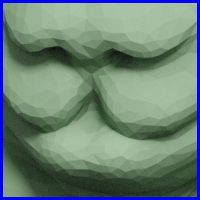}} \hfill
  \mpage{0.235}{\includegraphics[width=0.475\linewidth]{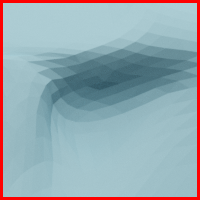} \hfill \includegraphics[width=0.475\linewidth]{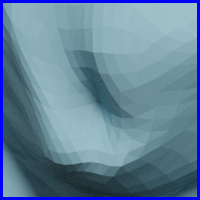}} \hfill
  \mpage{0.235}{\includegraphics[width=0.475\linewidth]{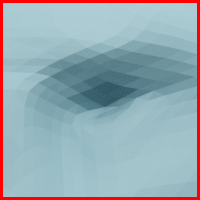} \hfill \includegraphics[width=0.475\linewidth]{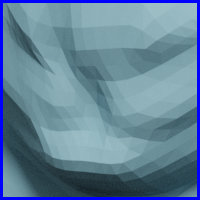}} \hfill
  \mpage{0.235}{\includegraphics[width=0.475\linewidth]{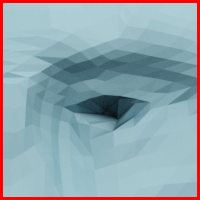} \hfill \includegraphics[width=0.475\linewidth]{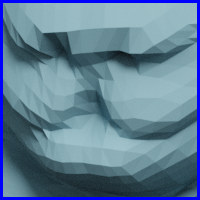}} \\
  \vspace{1.0mm}
  \mpage{0.235}{$CR$ / $d_\text{pm}$ ($\times 10^{-4}$) / $d_\text{normal}$} \hfill
  \mpage{0.235}{64.73 / 14.87 / 10.33$^\circ$} \hfill
  \mpage{0.235}{17.78 / 11.48 / 6.94$^\circ$} \hfill
  \mpage{0.235}{10.31 /  5.37 / 5.31$^\circ$} \\
  \vspace{1.0mm}
  \mpage{0.235}{Ground truth} \hfill
  \mpage{0.235}{Ours w/o features} \hfill
  \mpage{0.235}{Ours + 40 features} \hfill
  \mpage{0.235}{Ours + 400 features} \\
  \caption{
  \textbf{Progressive features.} 
  Our results can be progressively improved if additional features are transmitted.
  }
  \label{fig:exp-progressive-featature}
\end{figure*}

\begin{figure}[t]
  \includegraphics[width=\linewidth]{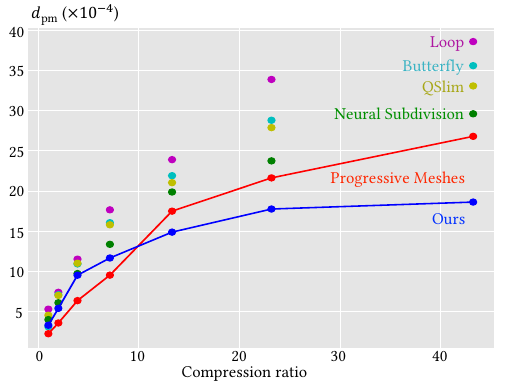}
  \caption{
  \textbf{Compression ratio vs. point-to-mesh distance curve.} 
  }
  \label{fig:exp-comp-rate-vs-error}
\end{figure}

We further compare our method to baselines for variable compression ratio in Figure~\ref{fig:exp-comp-rate-vs-error}, where the x-axis shows CR and the y-axis shows point-to-mesh distance. As noted, previously discussed mesh simplification baselines (QSlim) and different subdivision schemes applied to QSlim (Loop, Butterfly, Neural Subdivision) do not allow transmitting incremental data to progressively improve the subdivision quality. Thus, these methods are not directly comparable, and we render them as scatter points for context. It is worth noting, nevertheless, that for any compression ratio, our method provides a higher reconstruction quality than these alternatives. 

\bgroup
\def\arraystretch{1.2}
\begin{table}[t]
  \begin{center}
  \caption{
  \textbf{Ablation study on progressive features.}
  }
  \label{table:exp-ablation-prog-feat}
  {
  \addtolength{\tabcolsep}{8.7pt}
  \begin{tabular}{lrr}
    \toprule
    \rowcolor{LavenderBlue}
    Method & $d_\text{pm}$ ($\times 10^{-4}$) $\downarrow$ & $d_\text{normal}$ $\downarrow$ \\  
    \midrule
    Ours w/o features      & 14.56 & 12.36$^\circ$ \\
    \rowcolor[HTML]{EFEFEF}
    Ours + 40 features  & 6.81 & 9.20$^\circ$ \\
    Ours + 400 features & 4.12 & 7.19$^\circ$ \\
    \bottomrule
  \end{tabular}
  }
  \end{center}
\end{table}
\egroup

Progressive Meshes~\cite{progressive_meshes} is the pioneering method that inspired our work and allowed us to use incremental features to progressively recover the details of a mesh. Since Progressive Meshes is a lossless method, it outperforms our approach for a smaller $CR$. However, we observe that the gap widens for $CR>10$, suggesting that our method is especially effective when an asset needs to be significantly reduced in size. See Figure~\ref{app-fig:exp-prog-meshes} for a visual comparison.

\subsection{Levels of Detail}
\begin{figure*}[t]
  \centering
  \mpage{0.19}{\includegraphics[width=\linewidth]{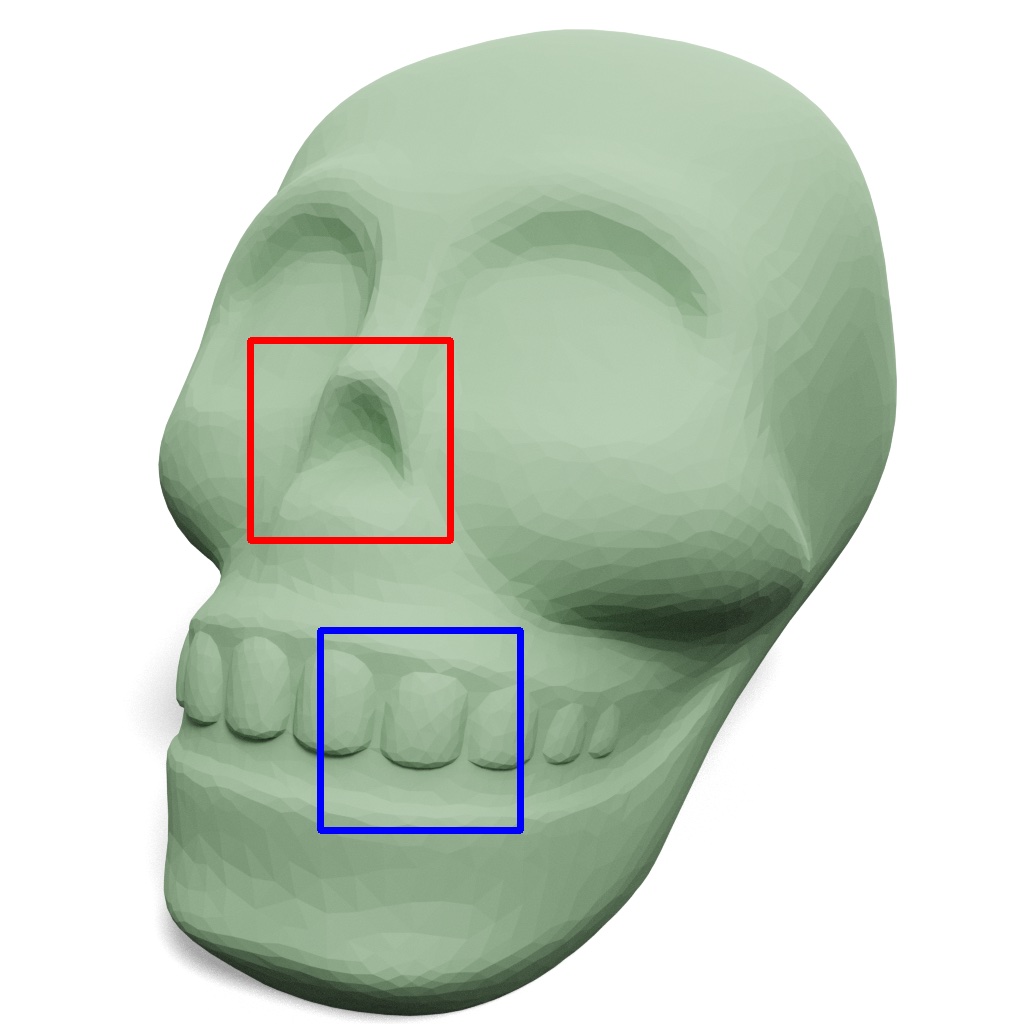}} \hfill
  \mpage{0.19}{\includegraphics[width=\linewidth]{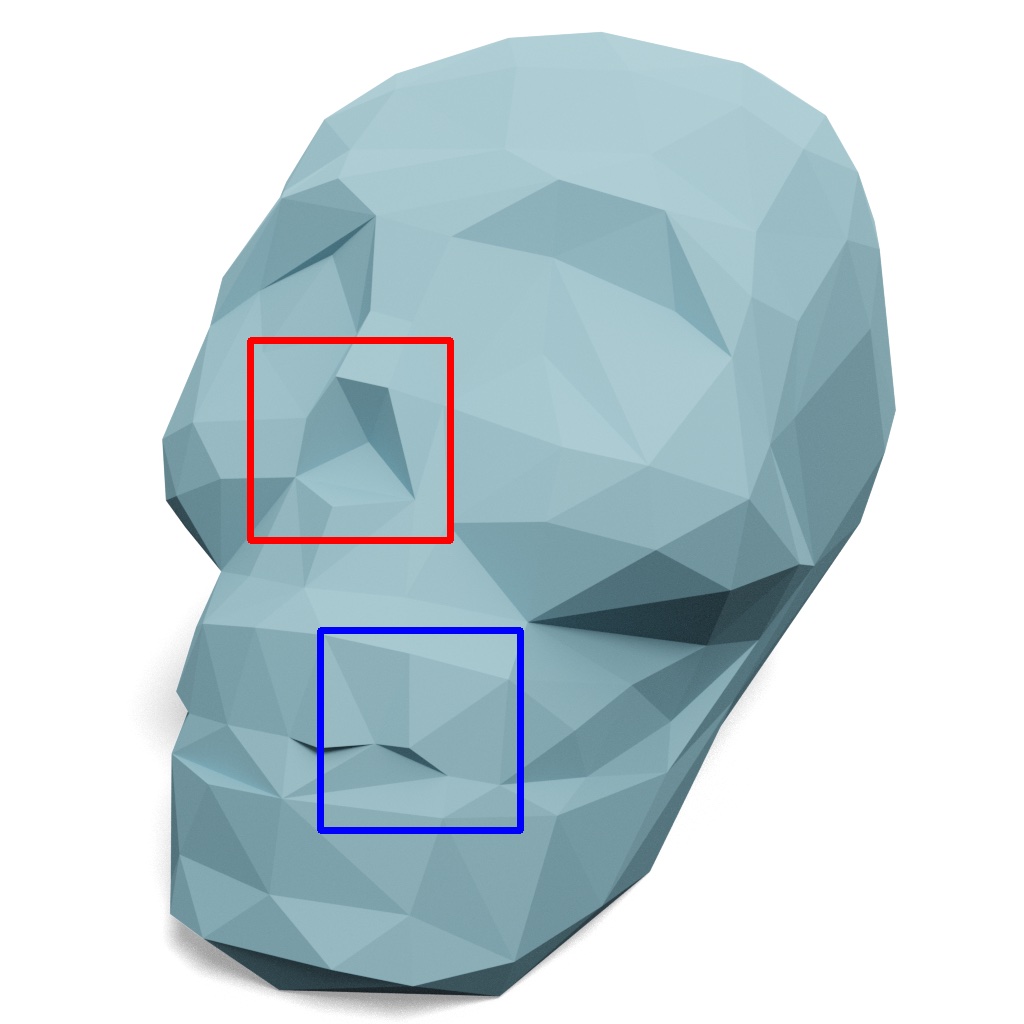}} \hfill
  \mpage{0.19}{\includegraphics[width=\linewidth]{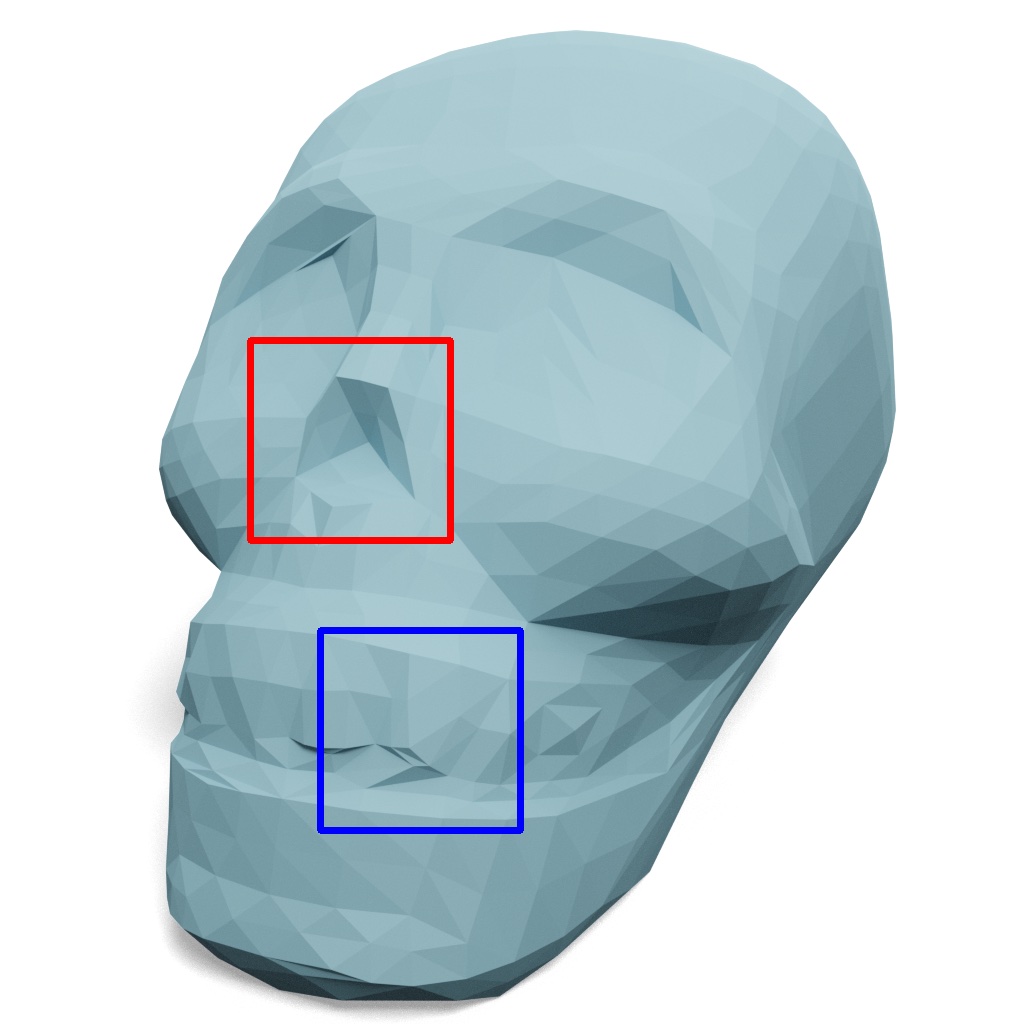}} \hfill
  \mpage{0.19}{\includegraphics[width=\linewidth]{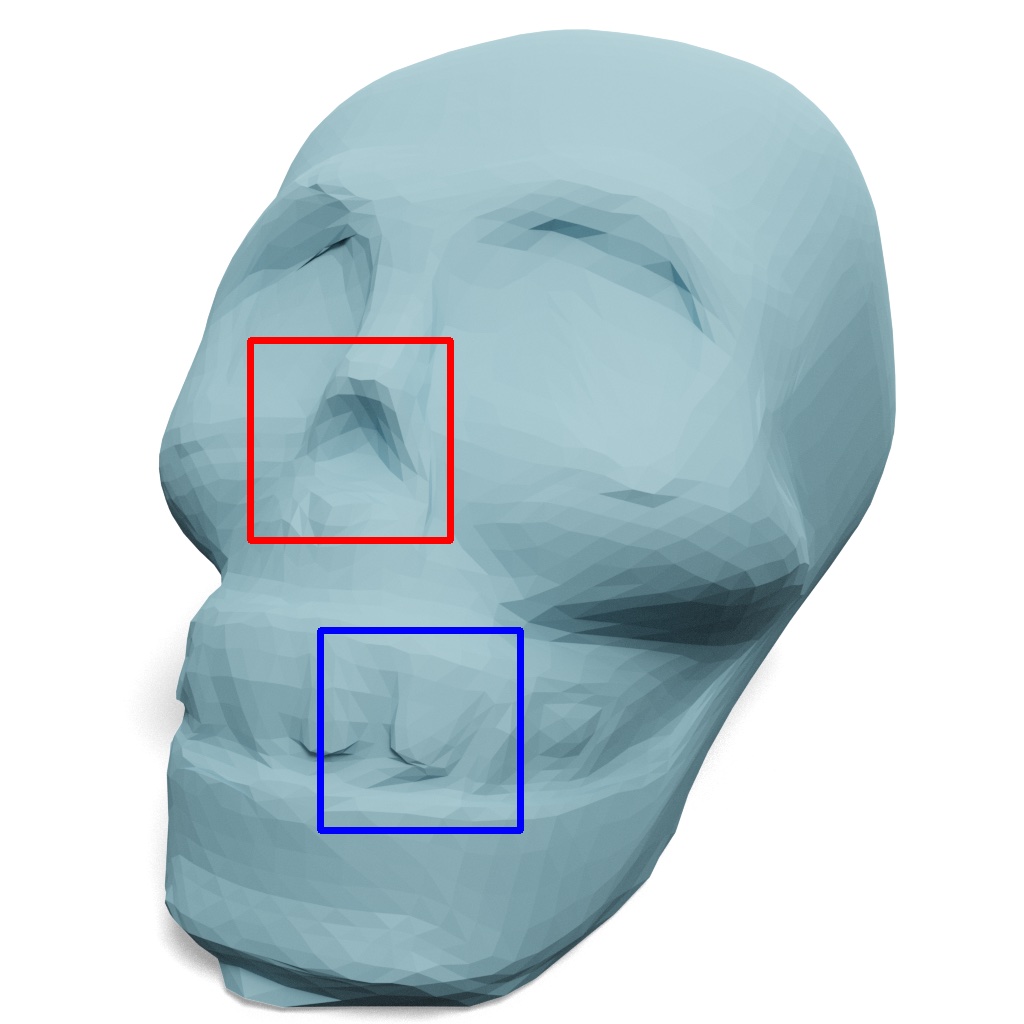}} \hfill
  \mpage{0.19}{\includegraphics[width=\linewidth]{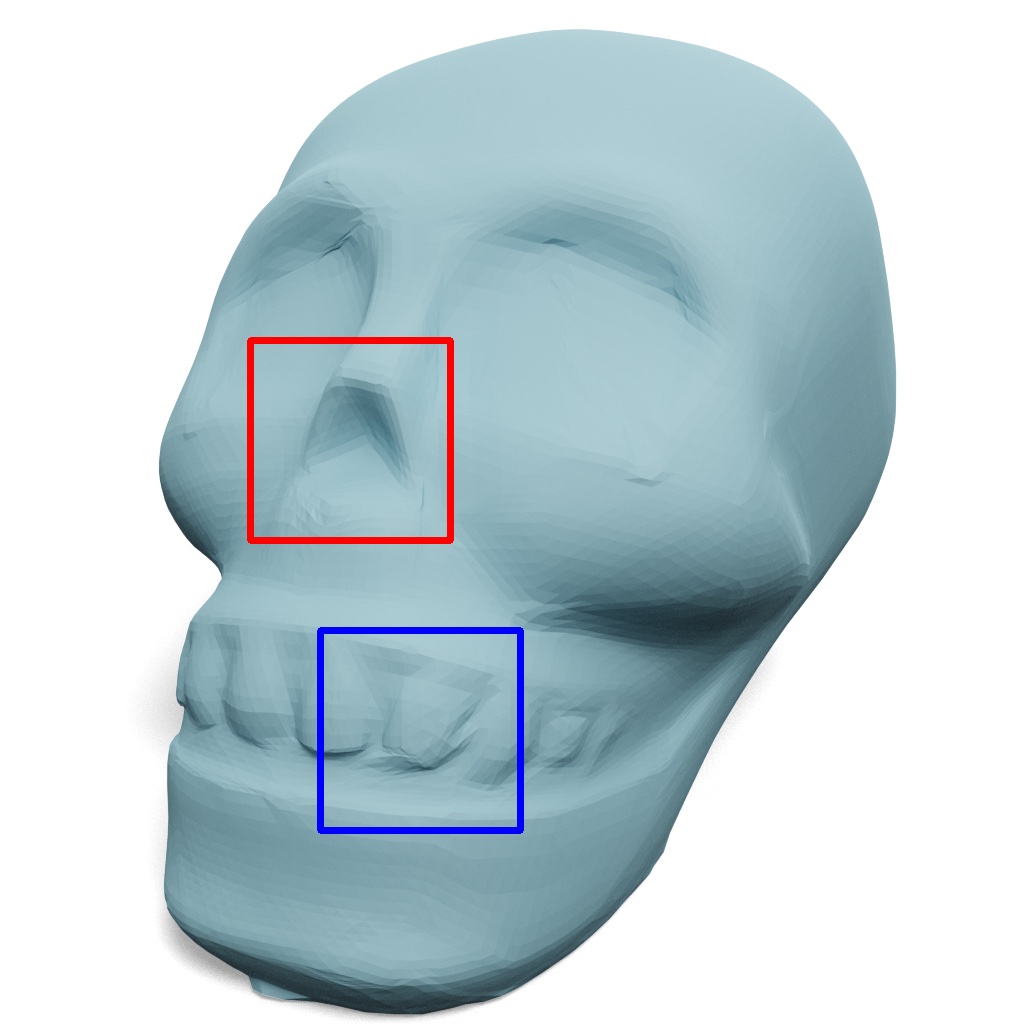}} \\
  \vspace{1.0mm}
  \mpage{0.19}{\includegraphics[width=0.475\linewidth]{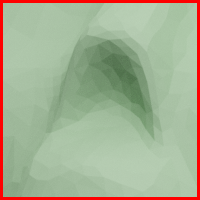} \hfill \includegraphics[width=0.475\linewidth]{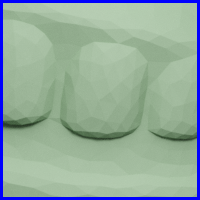}} \hfill
  \mpage{0.19}{\includegraphics[width=0.475\linewidth]{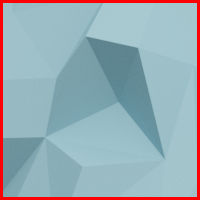} \hfill \includegraphics[width=0.475\linewidth]{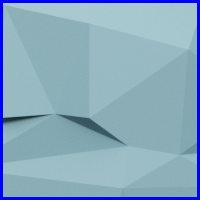}} \hfill
  \mpage{0.19}{\includegraphics[width=0.475\linewidth]{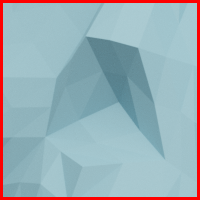} \hfill \includegraphics[width=0.475\linewidth]{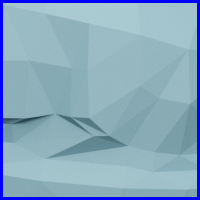}} \hfill
  \mpage{0.19}{\includegraphics[width=0.475\linewidth]{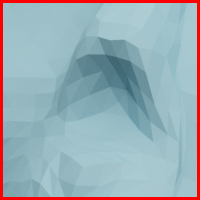} \hfill \includegraphics[width=0.475\linewidth]{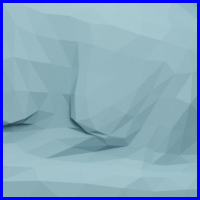}} \hfill
  \mpage{0.19}{\includegraphics[width=0.475\linewidth]{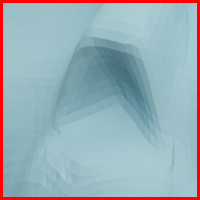} \hfill \includegraphics[width=0.475\linewidth]{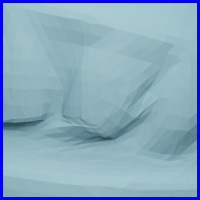}} \\
  \vspace{1.0mm}
  \mpage{0.19}{$CR$ / $d_\text{pm}$ ($\times 10^{-4}$) / $d_\text{normal}$} \hfill
  \mpage{0.19}{22.99 / 24.85 / 6.60$^\circ$} \hfill
  \mpage{0.19}{4.40 /  8.60 / 4.84$^\circ$} \hfill
  \mpage{0.19}{4.00 /  6.15 / 4.64$^\circ$} \hfill
  \mpage{0.19}{3.66 /  1.82 / 2.94$^\circ$} \\
  \vspace{1.0mm}
  \mpage{0.19}{Ground truth} \hfill
  \mpage{0.19}{Coarse mesh} \hfill
  \mpage{0.19}{Ours (level 1)} \hfill
  \mpage{0.19}{Ours (level 2)} \hfill
  \mpage{0.19}{Ours (level 3)} \\
  \caption{
  \textbf{Levels of detail.}
  Our decoder is able to subdivide a coarse mesh to different LoD meshes, providing the user with the flexibility to determine the resolution of the subdivided mesh.
  }
  \label{fig:exp-lod}
\end{figure*}

Our method design provides flexibility to the user on the client side to determine the resolution of the subdivided mesh. This choice does not affect the compression ratio but can be used to optimize the use of computational resources (e.g., displaying lower-level of subdivision for far-away assets).  
We conduct an ablation study that evaluates the quality of the subdivided mesh at each subdivision level (see Table~\ref{table:exp-ablation-lod} and Figure~\ref{fig:exp-lod}). As expected, the quality increases with each subdivision level, but returns are diminishing, even though the number of triangles increases exponentially by a factor of 4. 

\subsection{Limitations and Failure Cases}
Our method fails when applied to a shape with complex topological details and intricate thin features. 
See Section 4 in the supplemental material for an example.
Although not suitable for lossless compression, our method provides a superior lossy compression for a wide section of the size-accuracy spectrum.

\subsection{Runtime} 
We train and test the network on an Intel(R) Core(TM) i7-12700K CPU machine with one NVIDIA A40 GPU. The network training takes around 2 days. At test time, the server needs 4.84s to remesh a 100,000-face mesh to 400 faces (CPU-only), and the encoder forward pass takes 4.02s to predict per-face features on the GPU. It takes 4.58s for the client to subdivide a 400-face mesh to 25,600 faces (i.e., 3 subdivision levels) on the GPU. 
Our method can also run on the CPU. In the CPU-only case, the encoder forward pass takes 7.42s to predict per-face features, and the decoder forward pass takes 6.93s to subdivide a 400-face mesh to 25,600 faces.

\section{Conclusions}
We propose \algoNameFull, a novel representation that allows us to learn the space of surface details and efficiently compress them into per-face features. These features can be transmitted progressively, enabling our method to iteratively improve the quality of the reconstructed mesh as more data is transmitted. We demonstrate that our method is especially effective when only a small fraction of the original shape can be transmitted and outperforms other compression techniques, mesh simplification and subdivision approaches, and progressive mesh representations. 

\bgroup
\def\arraystretch{1.2}
\begin{table}[t]
  \begin{center}
  \caption{
  \textbf{Ablation study on levels of detail.}
  }
  \label{table:exp-ablation-lod}
  \resizebox{\linewidth}{!} 
  {
  \addtolength{\tabcolsep}{8.7pt}
  \begin{tabular}{lrrr}
    \toprule
    \rowcolor{LavenderBlue}
    Method & \# triangles & $d_\text{pm}$ ($\times 10^{-4}$) $\downarrow$ & $d_\text{normal}$ $\downarrow$ \\
    \midrule
    Ours level 1 & 1,600 & 12.47 & 11.23$^\circ$ \\
    \rowcolor[HTML]{EFEFEF}
    Ours level 2 & 6,400 & 4.36 & 8.66$^\circ$ \\
    Ours level 3 & 25,600 & 4.12 & 7.19$^\circ$ \\
    \bottomrule
  \end{tabular}
  }
  \end{center}
\end{table}
\egroup

\begin{acks}
This project is funded in part by NSERC Discovery (RGPIN–2022–04680), the Ontario Early Research Award program, the Canada Research Chairs Program, a Sloan Research Fellowship, the DSI Catalyst Grant program and gifts by Adobe Systems. 
We thank Hsueh-Ti Derek Liu for help with the Neural Subdivision code and Silvia Sellán, Abhishek Madan and Selena Ling for help with making figures.
\end{acks}

\appendix

\section{Feature Importance based on Reconstruction Losses}

\begin{figure*}[!t]
  \centering
  \mpage{0.31}{\includegraphics[width=\linewidth]{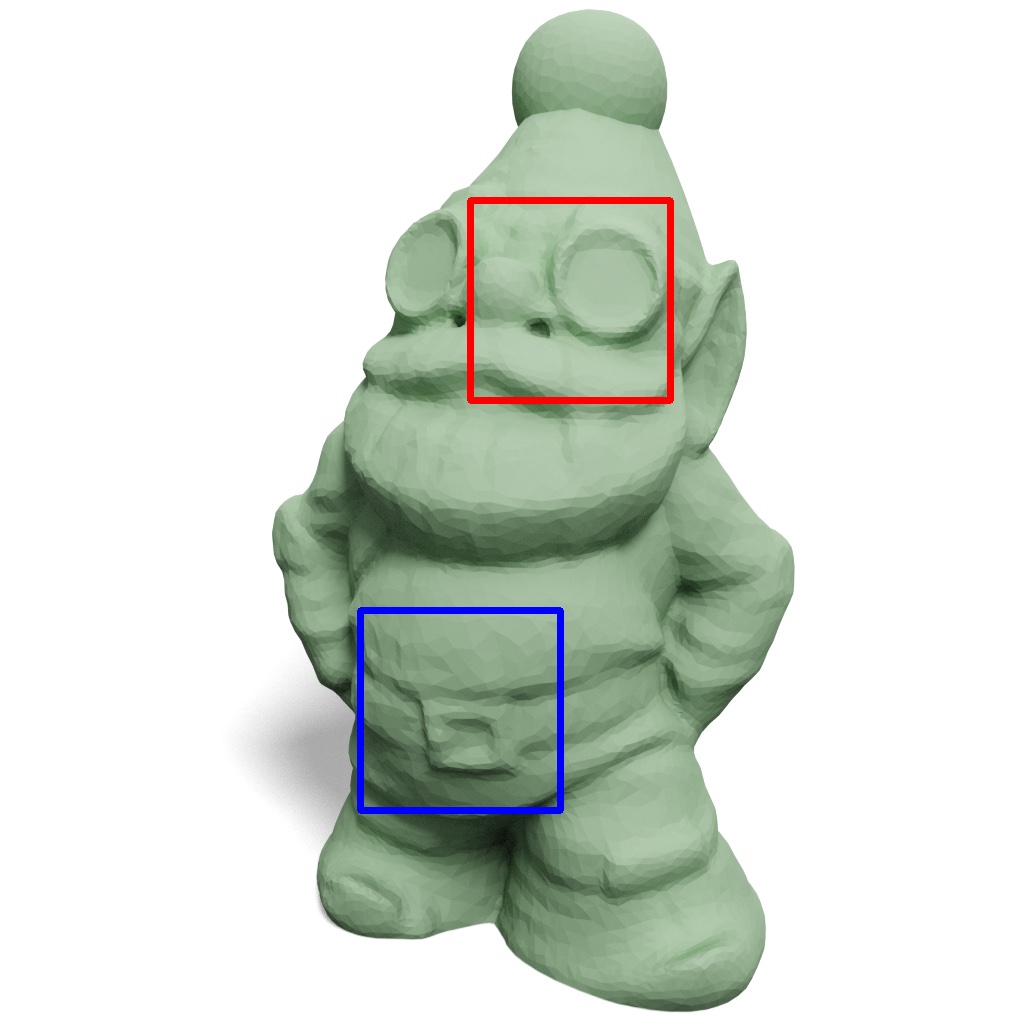}} \hfill
  \mpage{0.31}{\includegraphics[width=\linewidth]{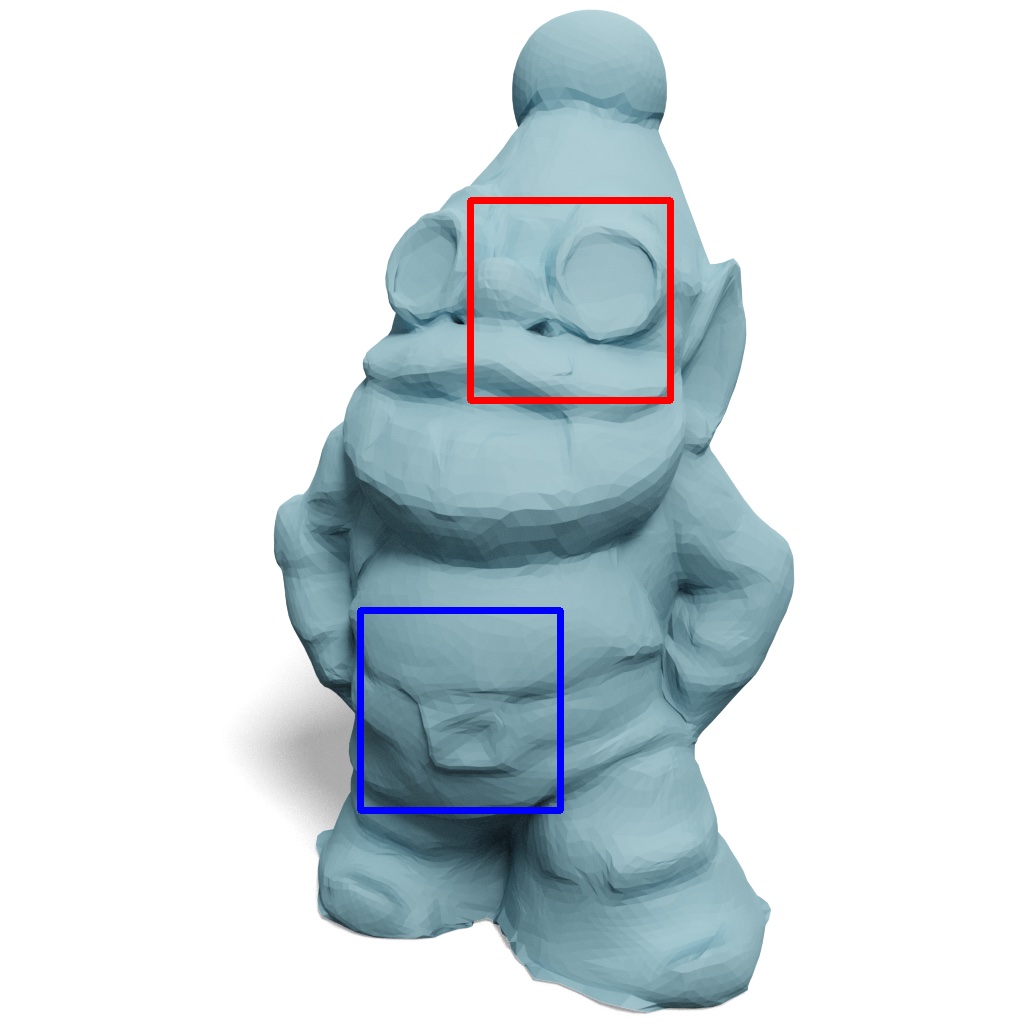}} \hfill
  \mpage{0.31}{\includegraphics[width=\linewidth]{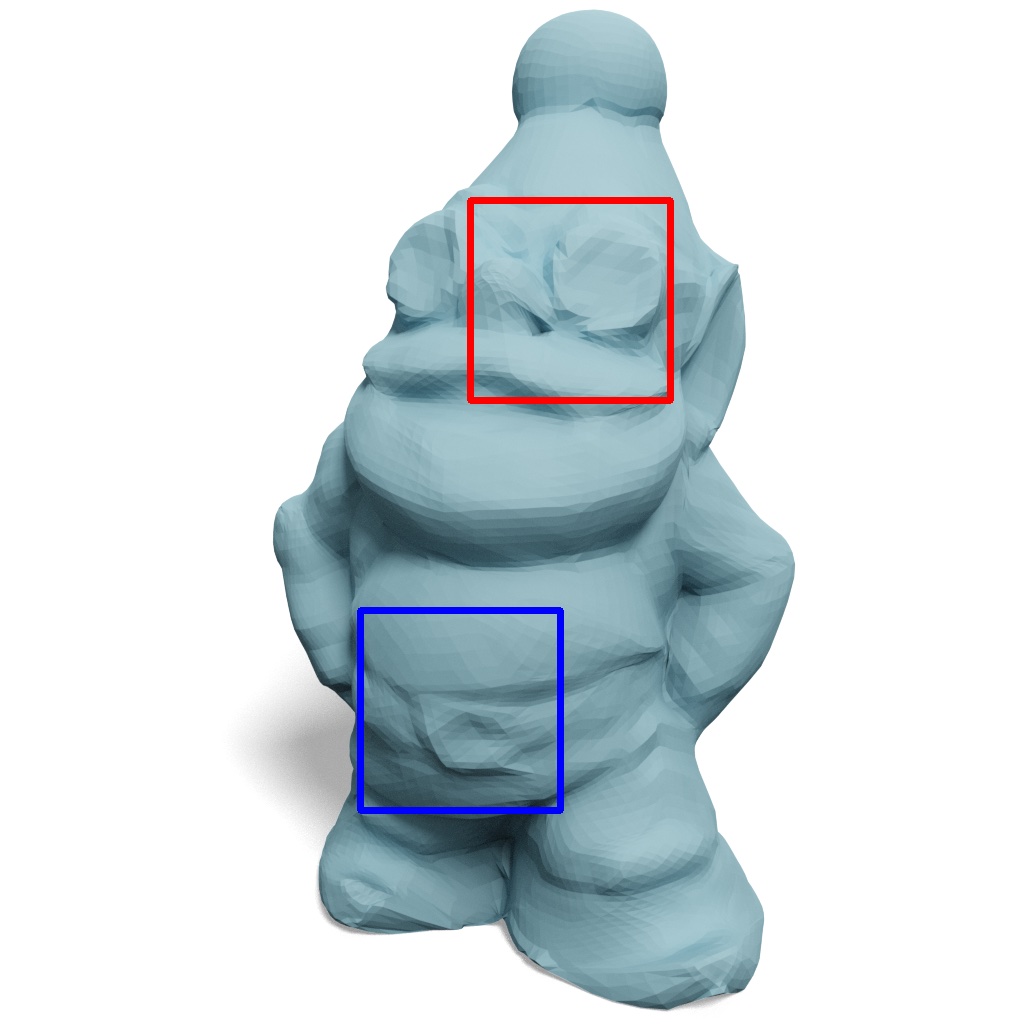}} \\
  \vspace{1.0mm}
  \mpage{0.31}{\includegraphics[width=0.475\linewidth]{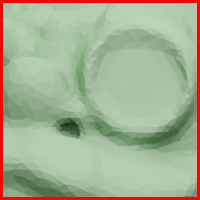} \hfill \includegraphics[width=0.475\linewidth]{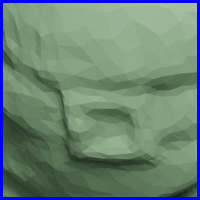}} \hfill
  \mpage{0.31}{\includegraphics[width=0.475\linewidth]{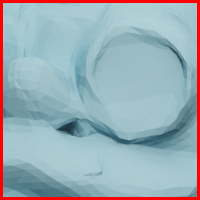} \hfill \includegraphics[width=0.475\linewidth]{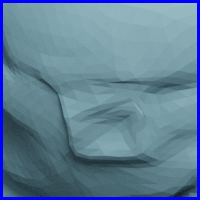}} \hfill
  \mpage{0.31}{\includegraphics[width=0.475\linewidth]{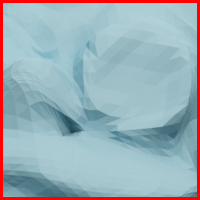} \hfill \includegraphics[width=0.475\linewidth]{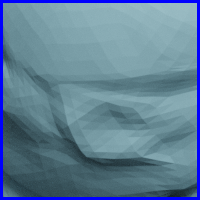}} \\
  \vspace{1.0mm}
  \mpage{0.31}{$CR$ / $d_\text{pm}$ ($\times 10^{-4}$) / $d_\text{normal}$} \hfill
  \mpage{0.31}{61.39 / 5.24 / 8.55$^\circ$} \hfill
  \mpage{0.31}{61.39 / 5.32 / 8.62$^\circ$} \\
  \vspace{1.0mm}
  \mpage{0.31}{Ground truth} \hfill
  \mpage{0.31}{Ours (reconstruction losses)} \hfill
  \mpage{0.31}{Ours (feature magnitude)} \\
  \caption{
  \textbf{Feature importance.} 
  Using the reconstruction losses to determine feature importance achieves better quantitative and qualitative results than using the feature magnitude.
  }
  \label{supp-fig:exp-rec}
\end{figure*}

Figure~\ref{supp-fig:exp-rec} shows an example.
Using the reconstruction losses to determine feature importance, the quantitative and qualitative results slightly improve over the method that uses the magnitude of the feature to determine the importance of the feature.

\section{Comparison to Progressive Meshes}

\begin{figure*}[!t]
  \centering
  \mpage{0.22}{\includegraphics[width=\linewidth]{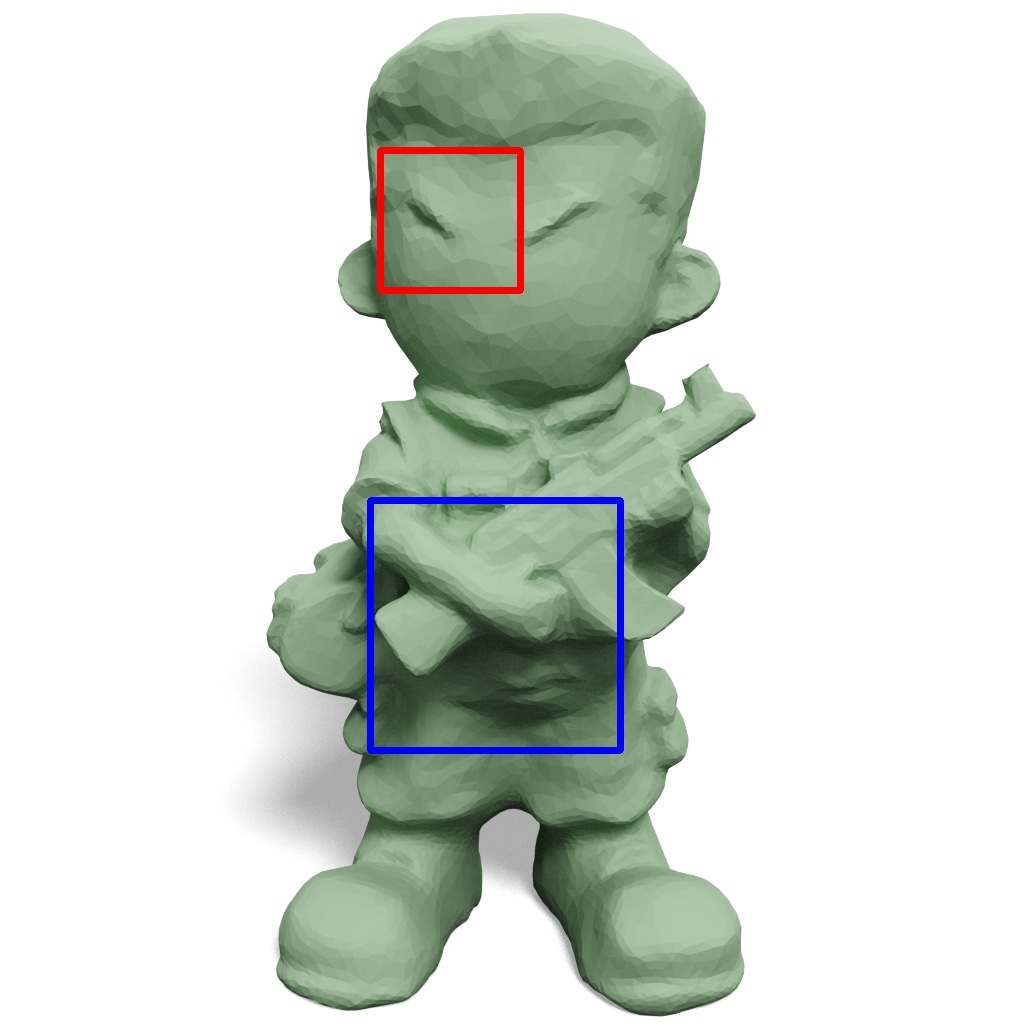}} \hfill
  \mpage{0.22}{\includegraphics[width=\linewidth]{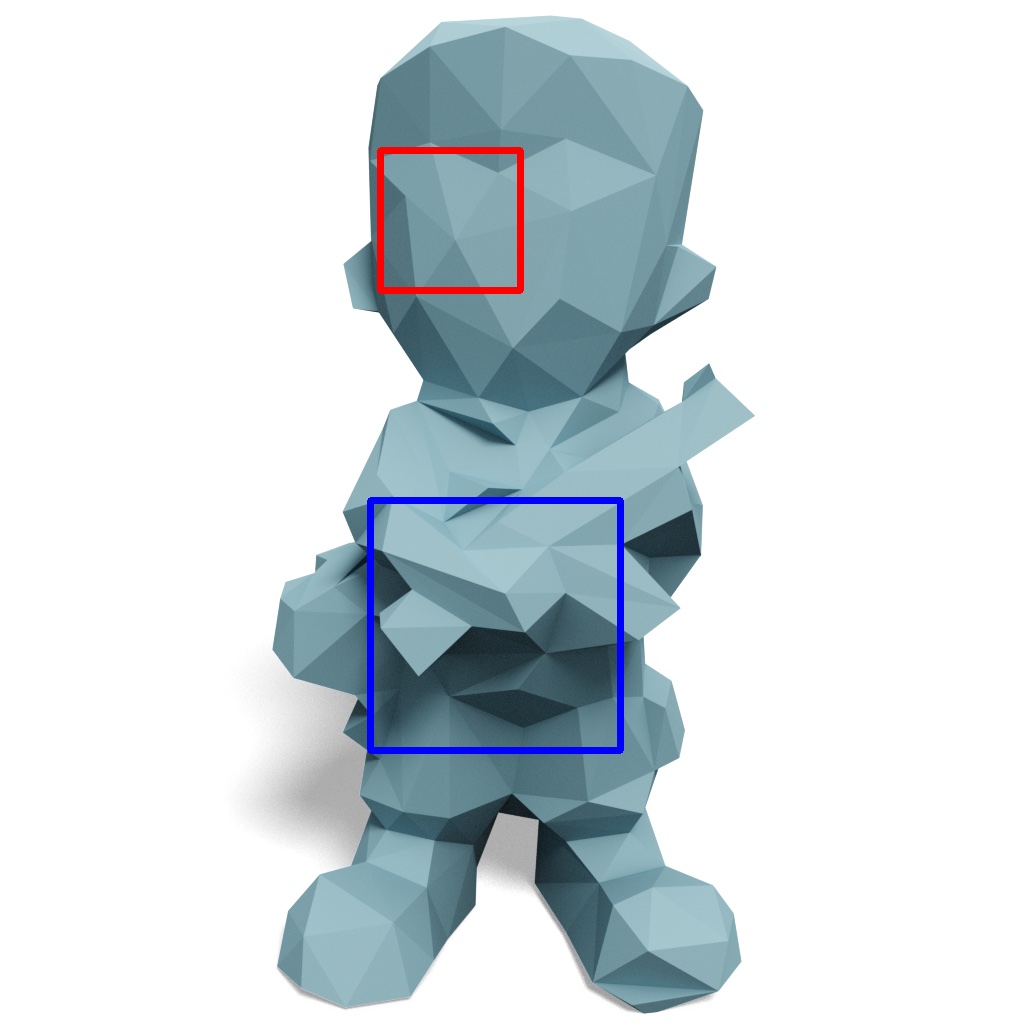}} \hfill
  \mpage{0.22}{\includegraphics[width=\linewidth]{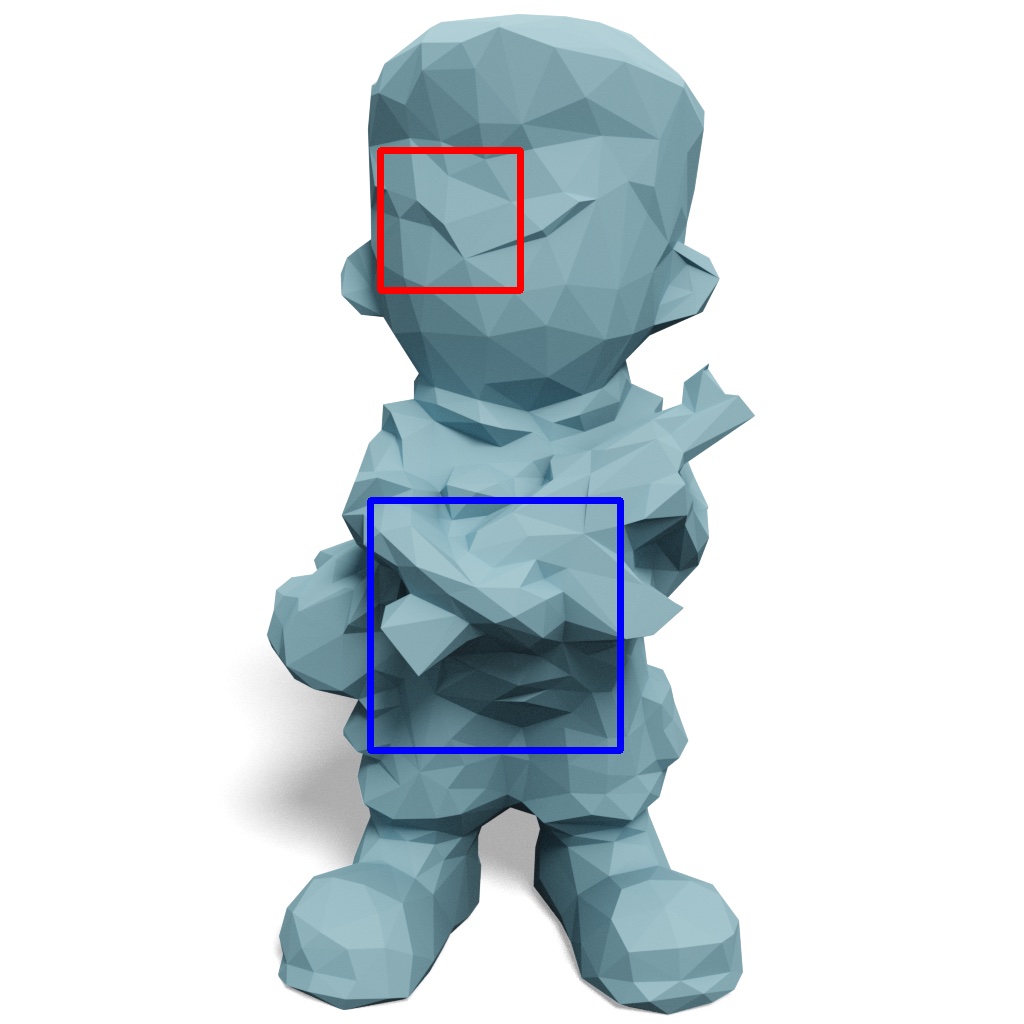}} \hfill
  \mpage{0.22}{\includegraphics[width=\linewidth]{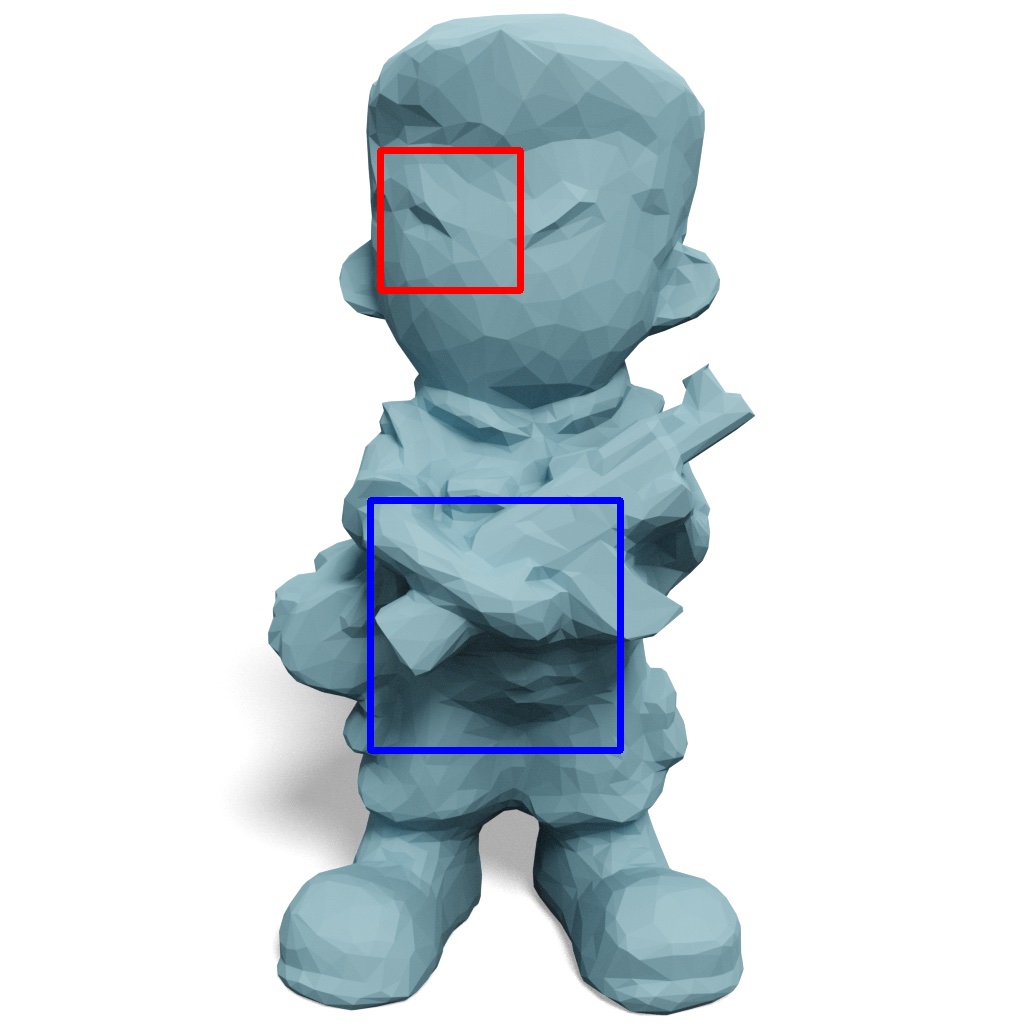}} \\
  \vspace{1.0mm}
  \mpage{0.22}{\includegraphics[width=0.475\linewidth]{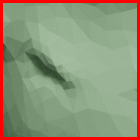} \hfill \includegraphics[width=0.475\linewidth]{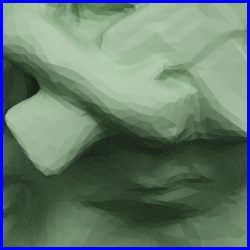}} \hfill
  \mpage{0.22}{\includegraphics[width=0.475\linewidth]{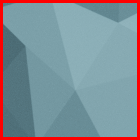} \hfill \includegraphics[width=0.475\linewidth]{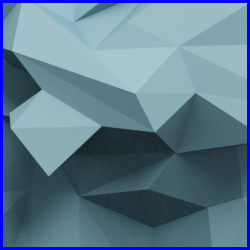}} \hfill
  \mpage{0.22}{\includegraphics[width=0.475\linewidth]{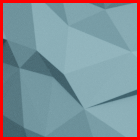} \hfill \includegraphics[width=0.475\linewidth]{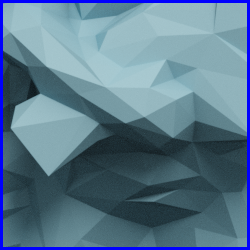}} \hfill
  \mpage{0.22}{\includegraphics[width=0.475\linewidth]{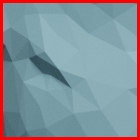} \hfill \includegraphics[width=0.475\linewidth]{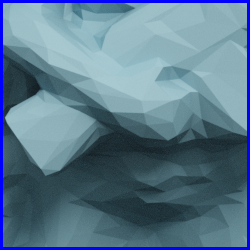}} \\
  \vspace{1.0mm}
  \mpage{0.22}{$CR$ / $d_\text{pm}$ ($\times 10^{-4}$) / $d_\text{normal}$} \hfill
  \mpage{0.22}{63.37 / 34.57 / 16.89$^\circ$} \hfill
  \mpage{0.22}{41.47 / 12.38 / 10.39$^\circ$} \hfill
  \mpage{0.22}{10.09 /  4.02 / 6.60$^\circ$} \\
  \vspace{1.0mm}
  \mpage{0.22}{Ground truth} \hfill
  \mpage{0.22}{Progressive Meshes} \hfill
  \mpage{0.22}{Progressive Meshes} \hfill
  \mpage{0.22}{Progressive Meshes} \\
  \vspace{1.0mm}
  \mpage{0.22}{\includegraphics[width=\linewidth]{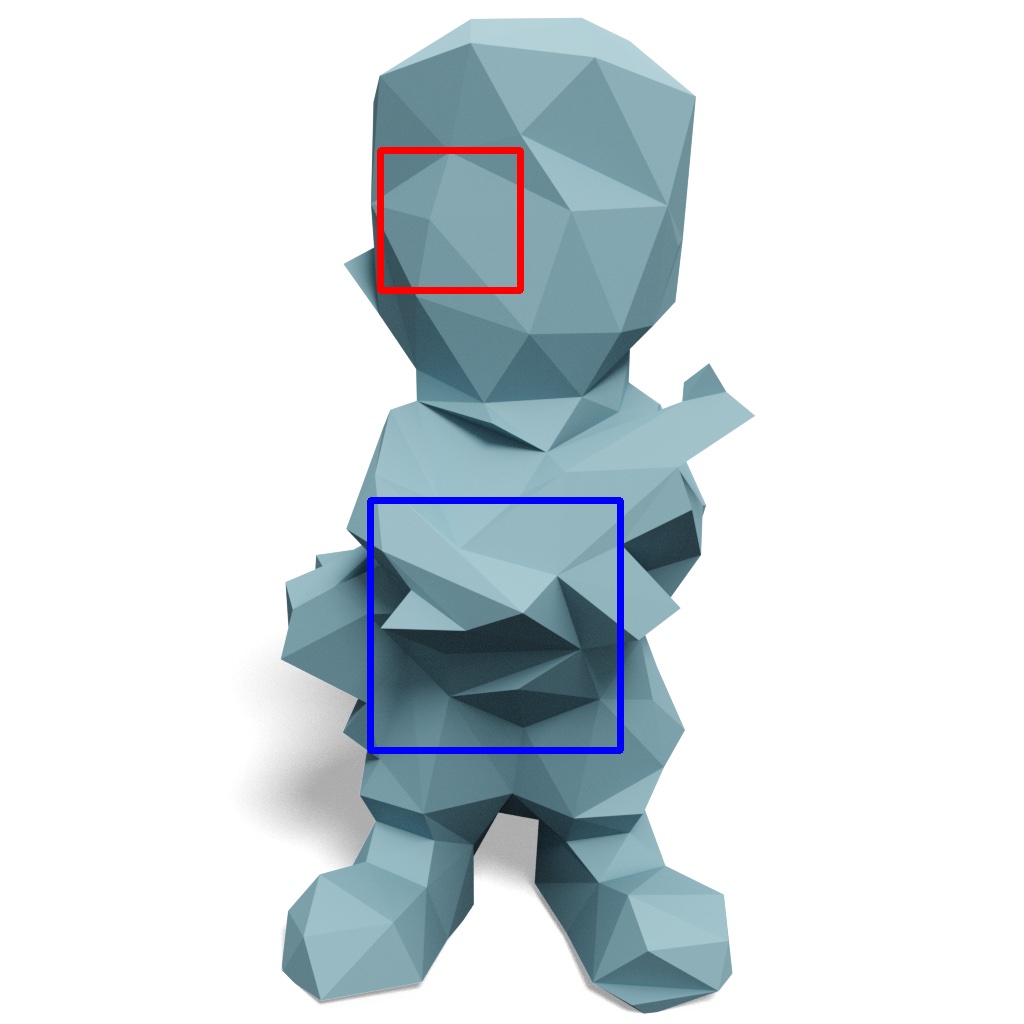}} \hfill
  \mpage{0.22}{\includegraphics[width=\linewidth]{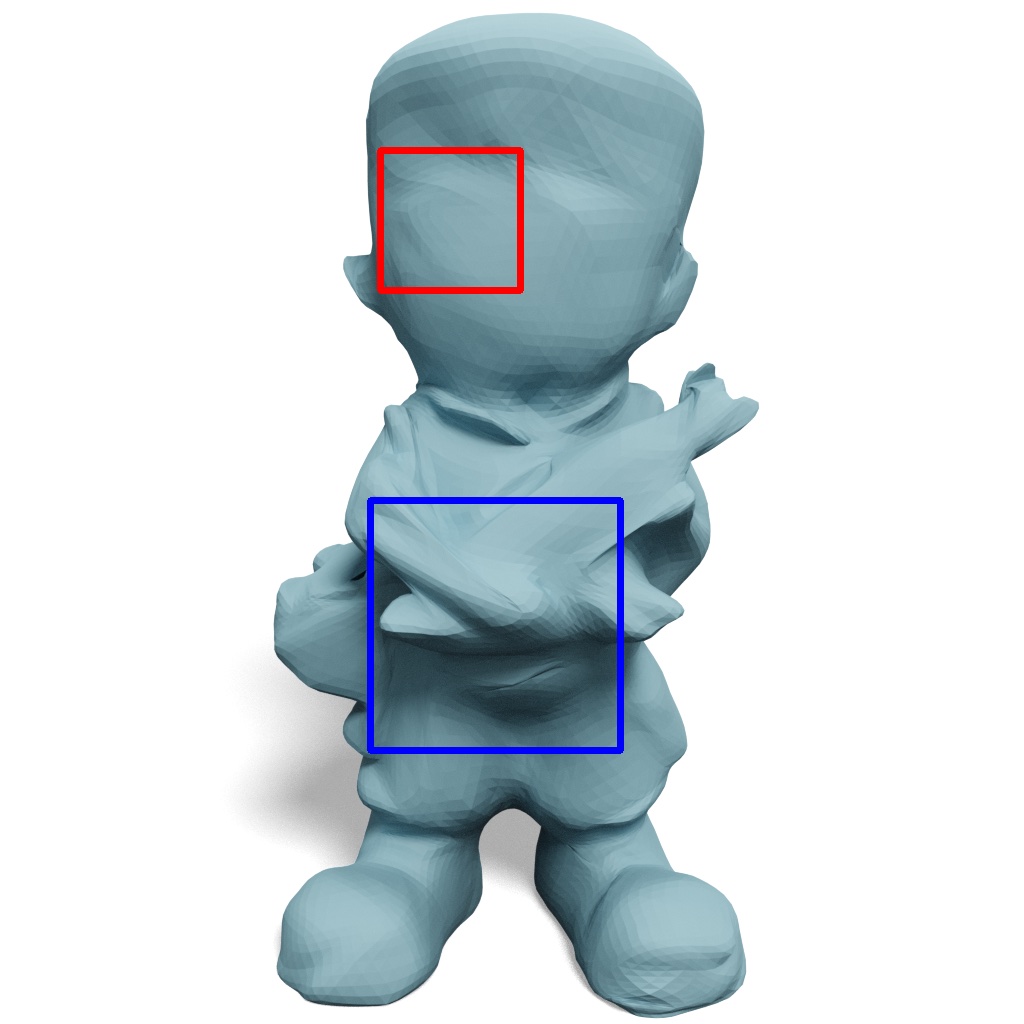}} \hfill
  \mpage{0.22}{\includegraphics[width=\linewidth]{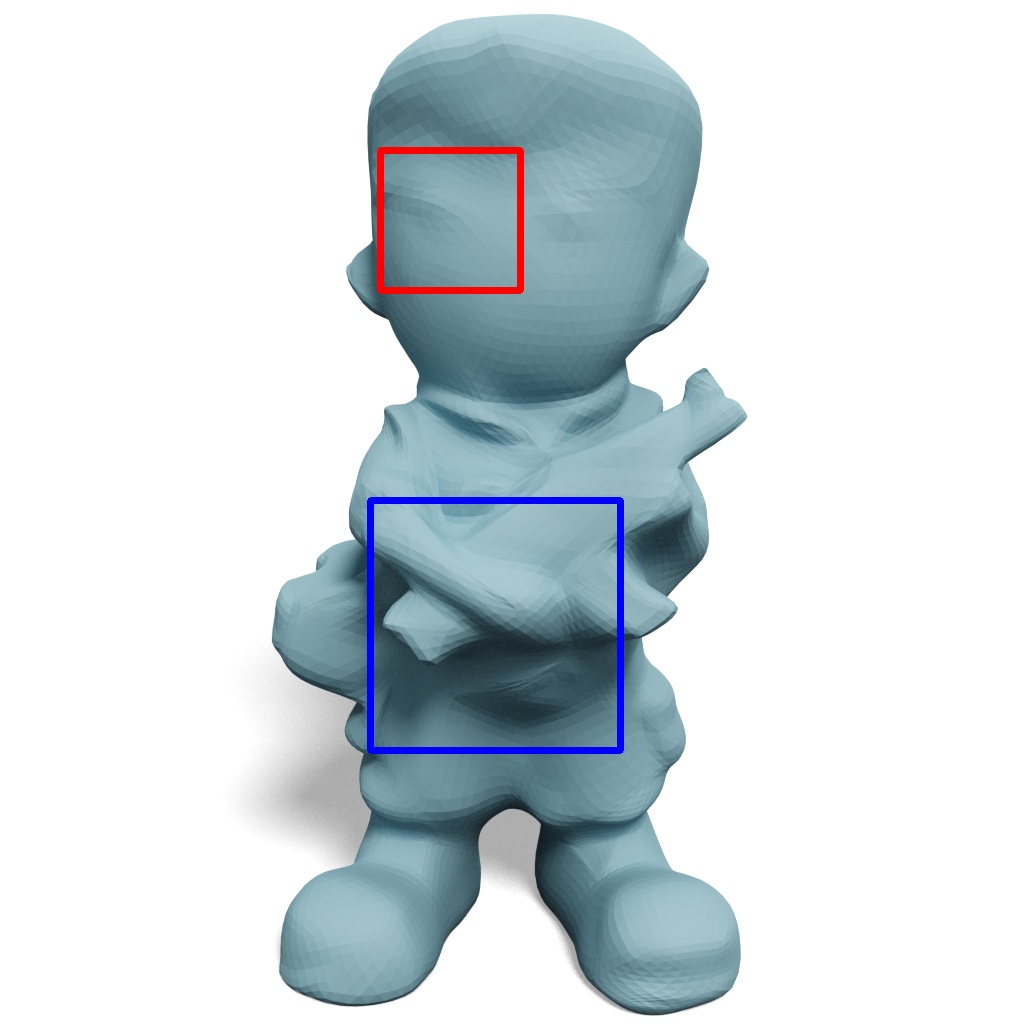}} \hfill
  \mpage{0.22}{\includegraphics[width=\linewidth]{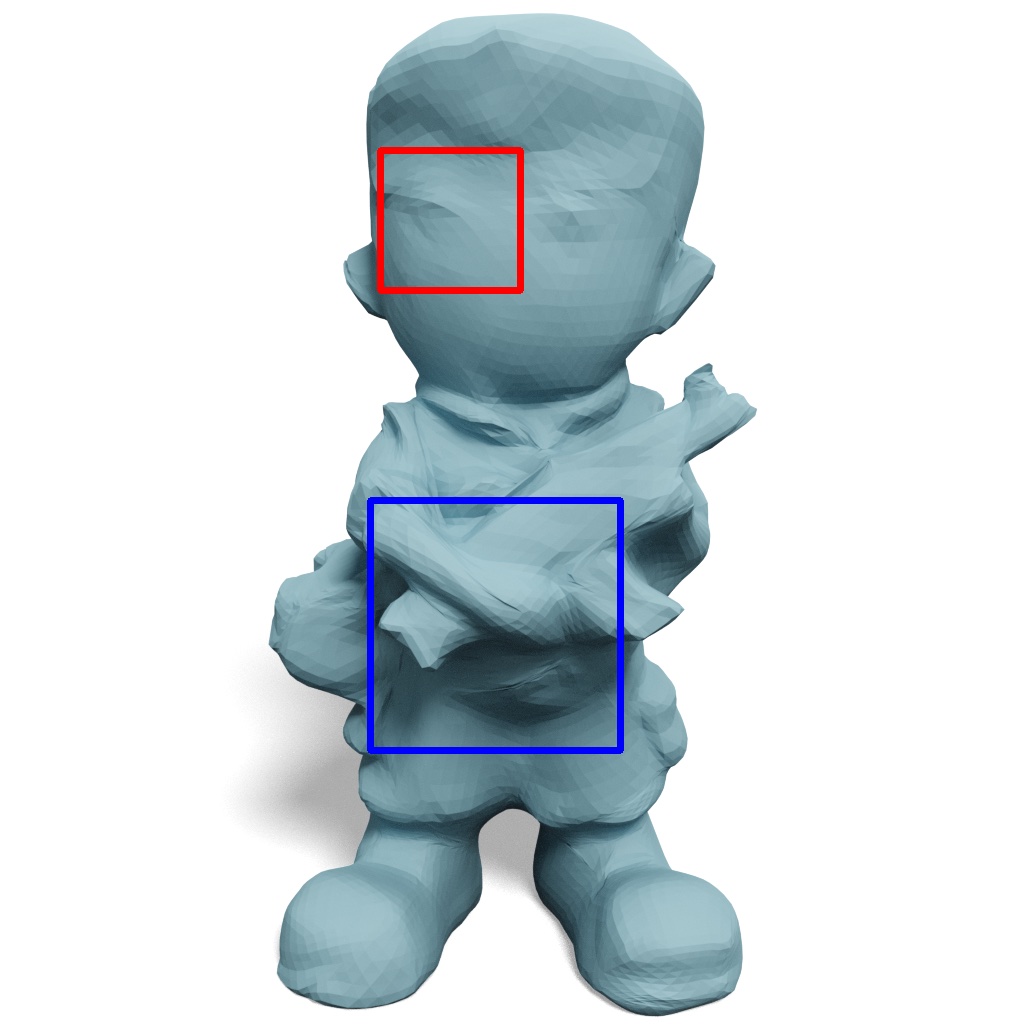}} \\
  \vspace{1.0mm}
  \mpage{0.22}{\includegraphics[width=0.475\linewidth]{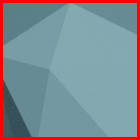} \hfill \includegraphics[width=0.475\linewidth]{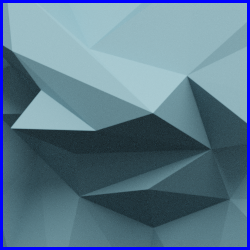}} \hfill
  \mpage{0.22}{\includegraphics[width=0.475\linewidth]{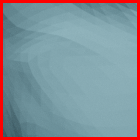} \hfill \includegraphics[width=0.475\linewidth]{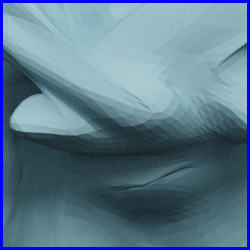}} \hfill
  \mpage{0.22}{\includegraphics[width=0.475\linewidth]{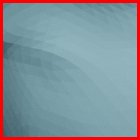} \hfill \includegraphics[width=0.475\linewidth]{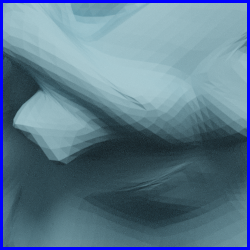}} \hfill
  \mpage{0.22}{\includegraphics[width=0.475\linewidth]{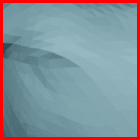} \hfill \includegraphics[width=0.475\linewidth]{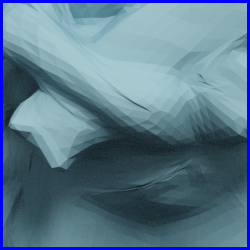}} \\
  \vspace{1.0mm}
  \mpage{0.22}{63.37 / 51.61 / 20.34$^\circ$} \hfill
  \mpage{0.22}{63.37 / 14.50 / 13.69$^\circ$} \hfill
  \mpage{0.22}{41.47 / 10.07 / 9.35$^\circ$} \hfill
  \mpage{0.22}{10.09 /  5.51 / 9.11$^\circ$} \\
  \vspace{1.0mm}
  \mpage{0.22}{QSlim} \hfill
  \mpage{0.22}{Ours w/o features} \hfill
  \mpage{0.22}{Ours + 40 features} \hfill
  \mpage{0.22}{Ours + 400 features} \\
  \caption{
  \textbf{Comparison to Progressive Meshes.} 
  }
  \label{supp-fig:exp-prog-meshes-1}
\end{figure*}

\begin{figure*}[!t]
  \centering
  \mpage{0.22}{\includegraphics[width=\linewidth]{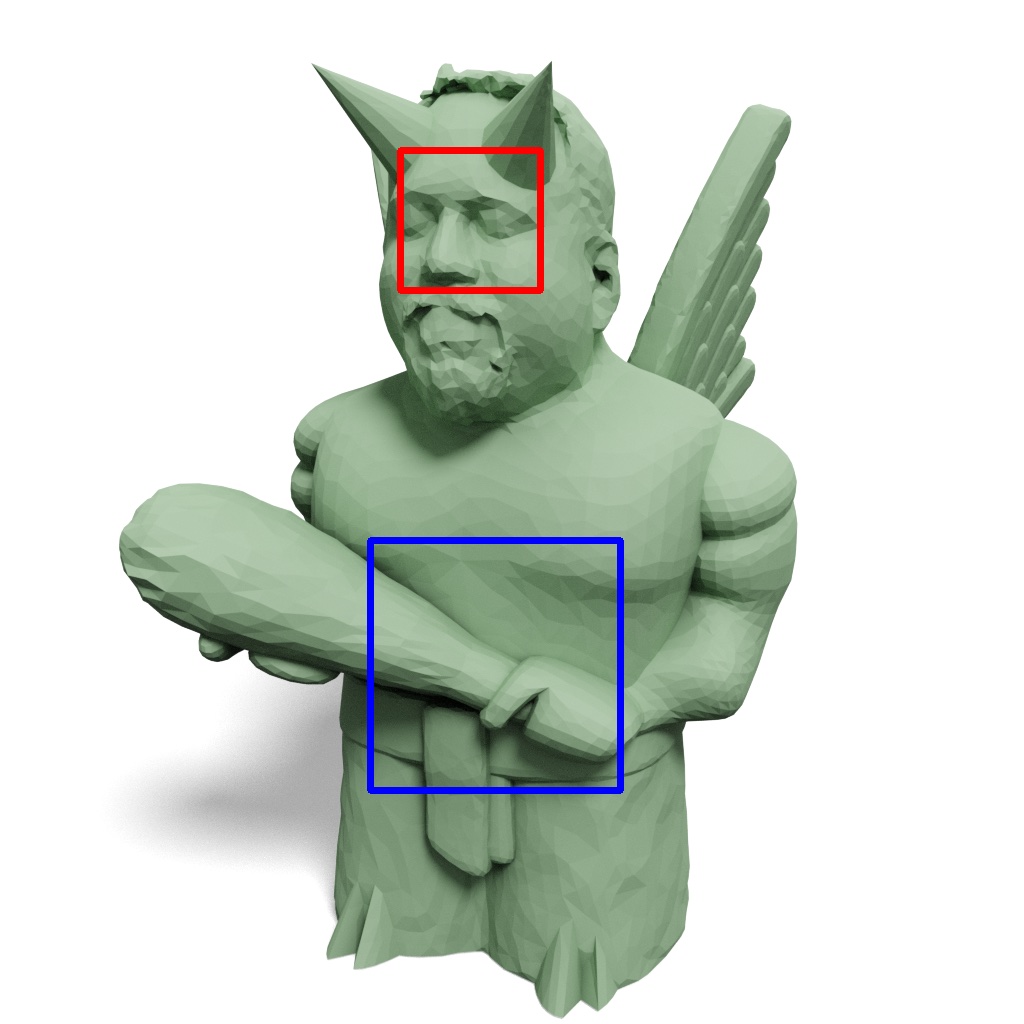}} \hfill
  \mpage{0.22}{\includegraphics[width=\linewidth]{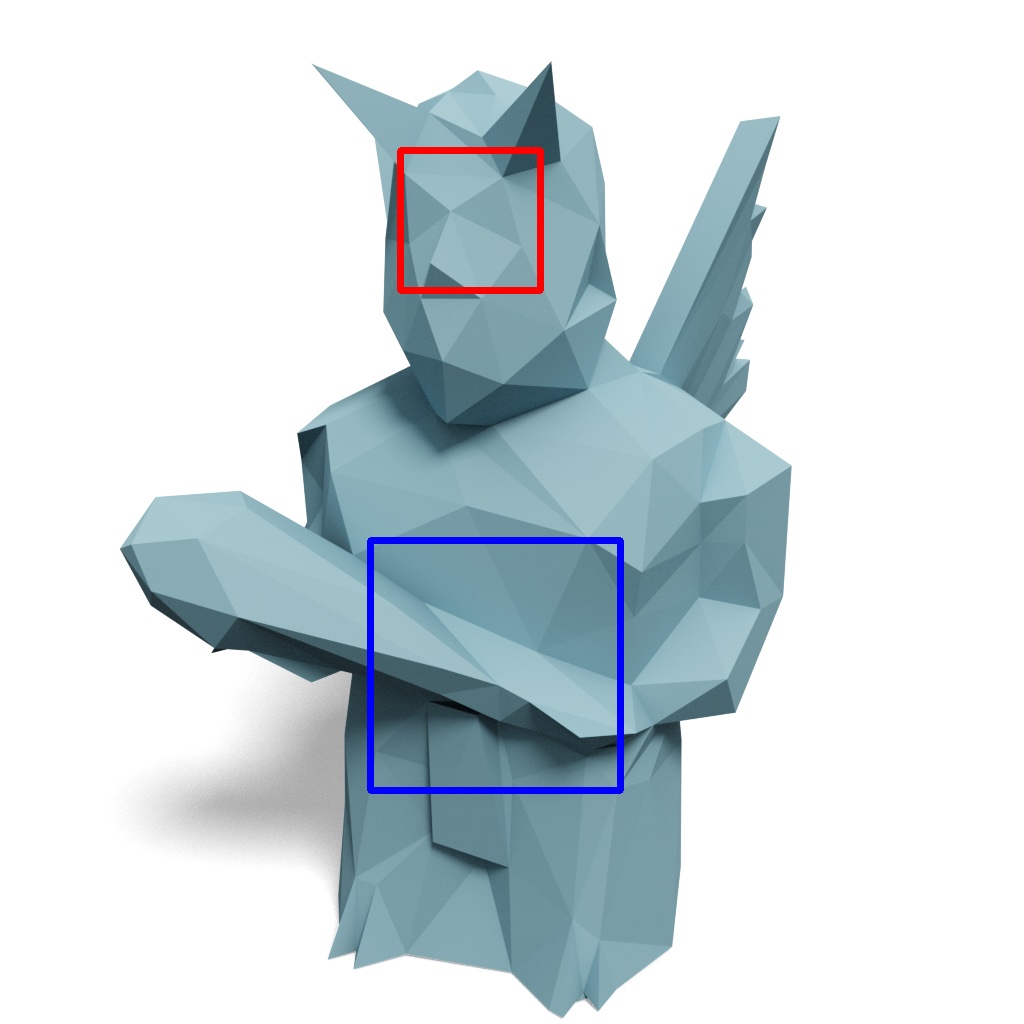}} \hfill
  \mpage{0.22}{\includegraphics[width=\linewidth]{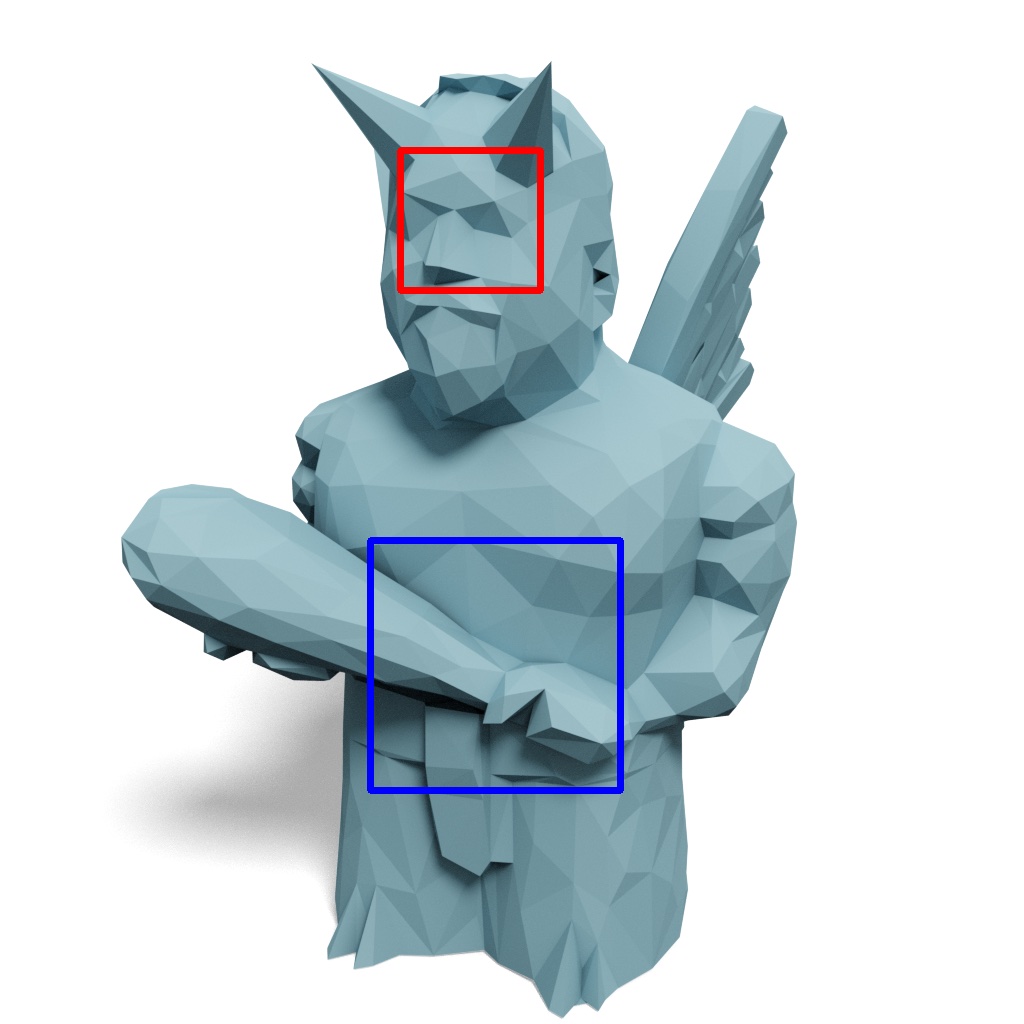}} \hfill
  \mpage{0.22}{\includegraphics[width=\linewidth]{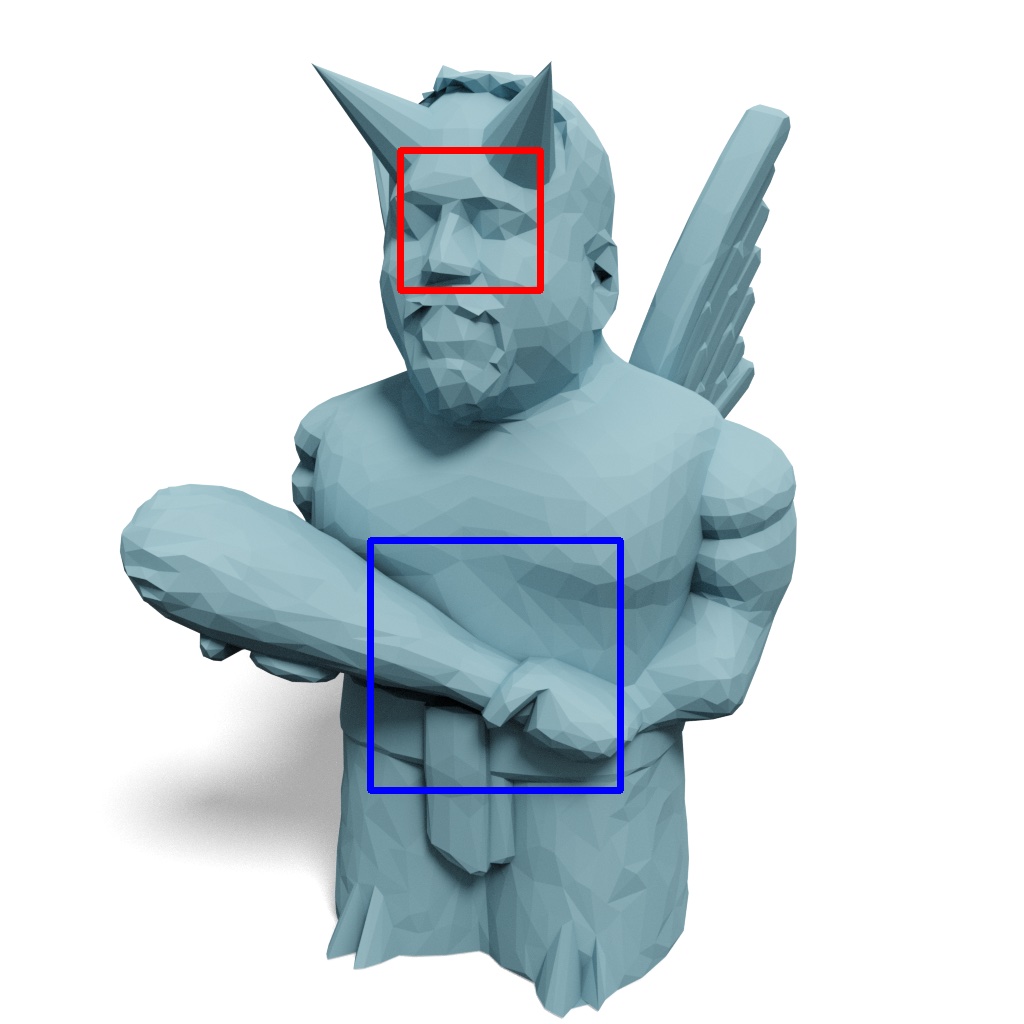}} \\
  \vspace{1.0mm}
  \mpage{0.22}{\includegraphics[width=0.475\linewidth]{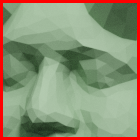} \hfill \includegraphics[width=0.475\linewidth]{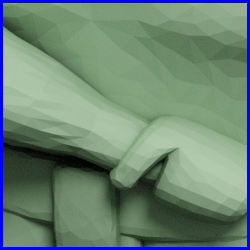}} \hfill
  \mpage{0.22}{\includegraphics[width=0.475\linewidth]{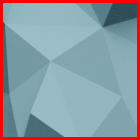} \hfill \includegraphics[width=0.475\linewidth]{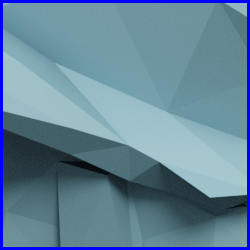}} \hfill
  \mpage{0.22}{\includegraphics[width=0.475\linewidth]{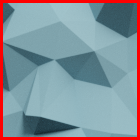} \hfill \includegraphics[width=0.475\linewidth]{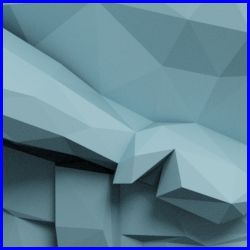}} \hfill
  \mpage{0.22}{\includegraphics[width=0.475\linewidth]{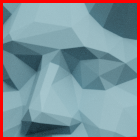} \hfill \includegraphics[width=0.475\linewidth]{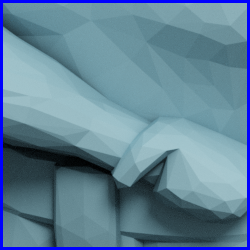}} \\
  \vspace{1.0mm}
  \mpage{0.22}{$CR$ / $d_\text{pm}$ ($\times 10^{-4}$) / $d_\text{normal}$} \hfill
  \mpage{0.22}{66.74 / 37.09 / 18.82$^\circ$} \hfill
  \mpage{0.22}{43.52 / 14.70 / 11.37$^\circ$} \hfill
  \mpage{0.22}{9.54 /  3.82 /  6.13$^\circ$} \\
  \vspace{1.0mm}
  \mpage{0.22}{Ground truth} \hfill
  \mpage{0.22}{Progressive Meshes} \hfill
  \mpage{0.22}{Progressive Meshes} \hfill
  \mpage{0.22}{Progressive Meshes} \\
  \vspace{1.0mm}
  \mpage{0.22}{\includegraphics[width=\linewidth]{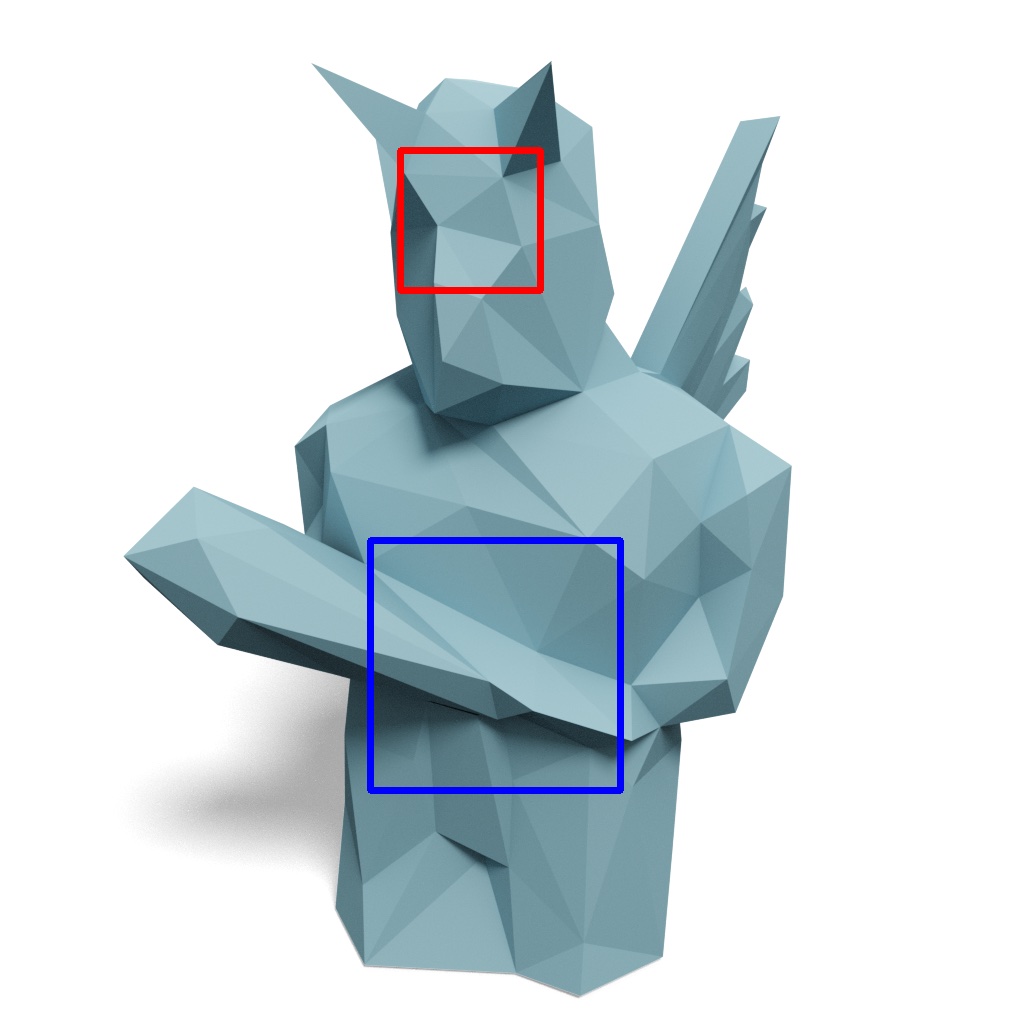}} \hfill
  \mpage{0.22}{\includegraphics[width=\linewidth]{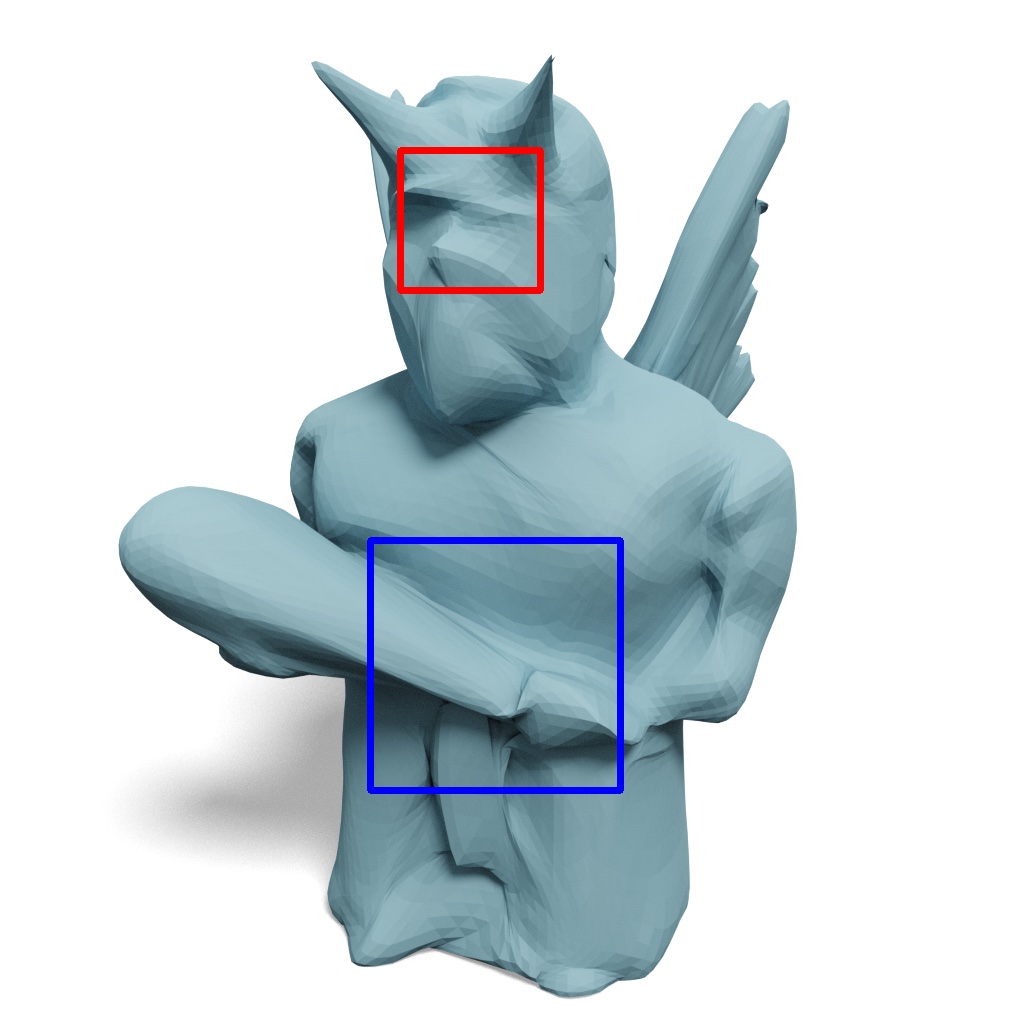}} \hfill
  \mpage{0.22}{\includegraphics[width=\linewidth]{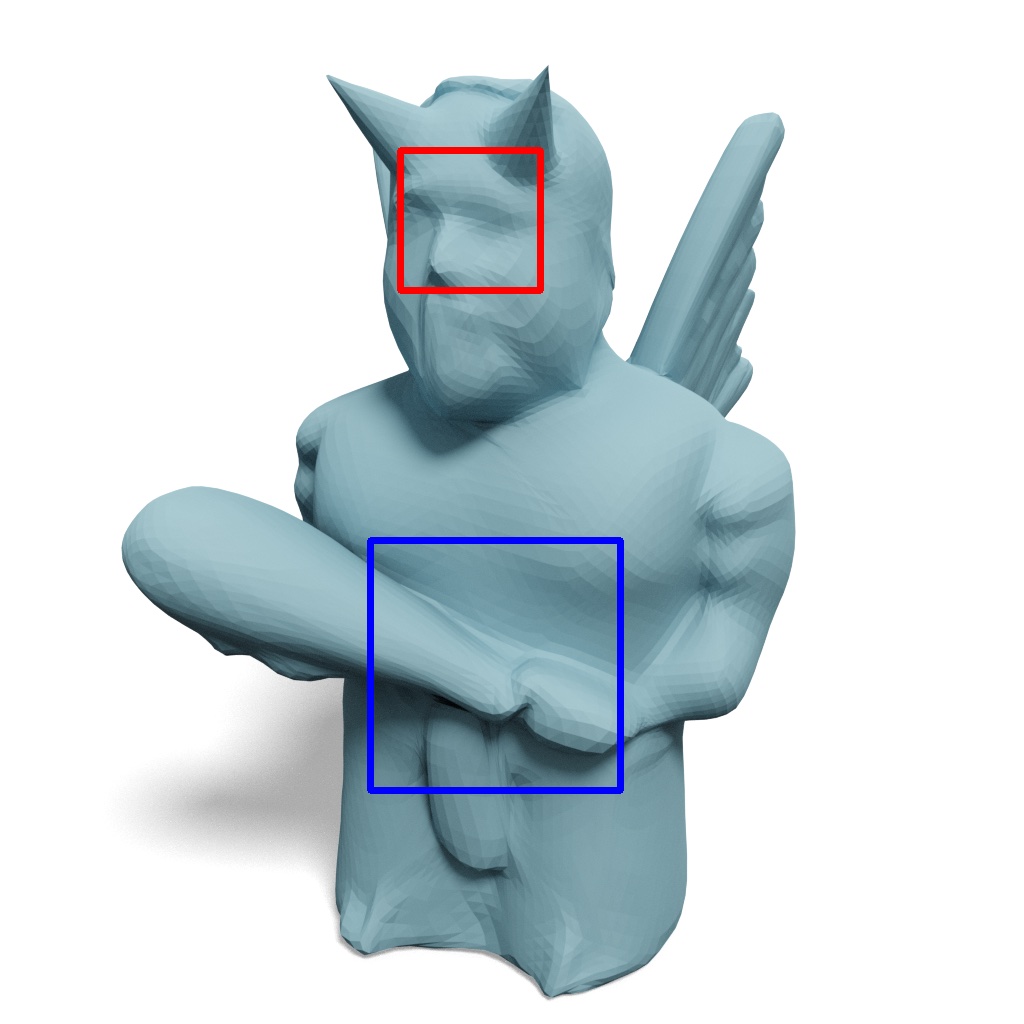}} \hfill
  \mpage{0.22}{\includegraphics[width=\linewidth]{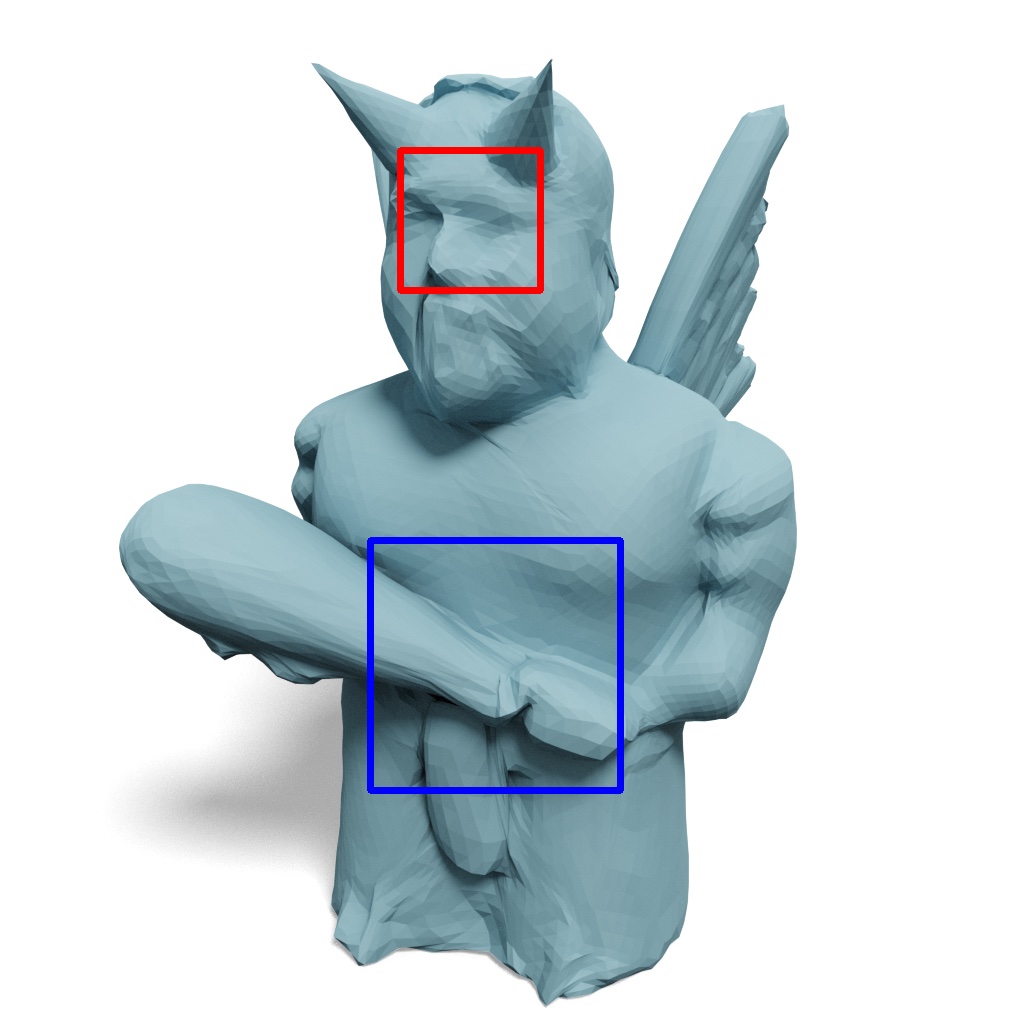}} \\
  \vspace{1.0mm}
  \mpage{0.22}{\includegraphics[width=0.475\linewidth]{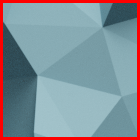} \hfill \includegraphics[width=0.475\linewidth]{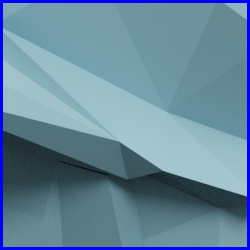}} \hfill
  \mpage{0.22}{\includegraphics[width=0.475\linewidth]{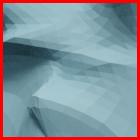} \hfill \includegraphics[width=0.475\linewidth]{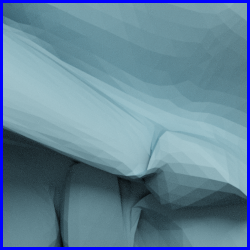}} \hfill
  \mpage{0.22}{\includegraphics[width=0.475\linewidth]{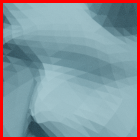} \hfill \includegraphics[width=0.475\linewidth]{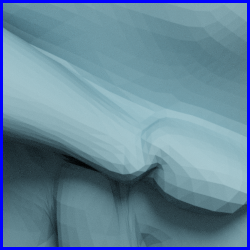}} \hfill
  \mpage{0.22}{\includegraphics[width=0.475\linewidth]{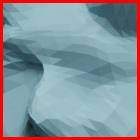} \hfill \includegraphics[width=0.475\linewidth]{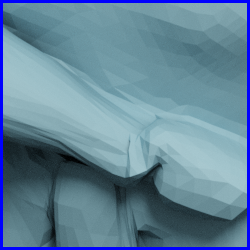}} \\
  \vspace{1.0mm}
  \mpage{0.22}{66.74 / 49.60 / 22.24$^\circ$} \hfill
  \mpage{0.22}{66.74 / 17.65 / 17.05$^\circ$} \hfill
  \mpage{0.22}{43.52 / 10.91 / 12.16$^\circ$} \hfill
  \mpage{0.22}{9.54 /  6.57 / 10.92$^\circ$} \\
  \vspace{1.0mm}
  \mpage{0.22}{QSlim} \hfill
  \mpage{0.22}{Ours w/o features} \hfill
  \mpage{0.22}{Ours + 40 features} \hfill
  \mpage{0.22}{Ours + 400 features} \\
  \caption{
  \textbf{Comparison to Progressive Meshes.} 
  }
  \label{supp-fig:exp-prog-meshes-2}
\end{figure*}

Figures~\ref{supp-fig:exp-prog-meshes-1} and~\ref{supp-fig:exp-prog-meshes-2} show two visual comparisons.

\section{Our Results}

\begin{figure*}[t]
  \centering
  \mpage{0.235}{\includegraphics[width=\linewidth]{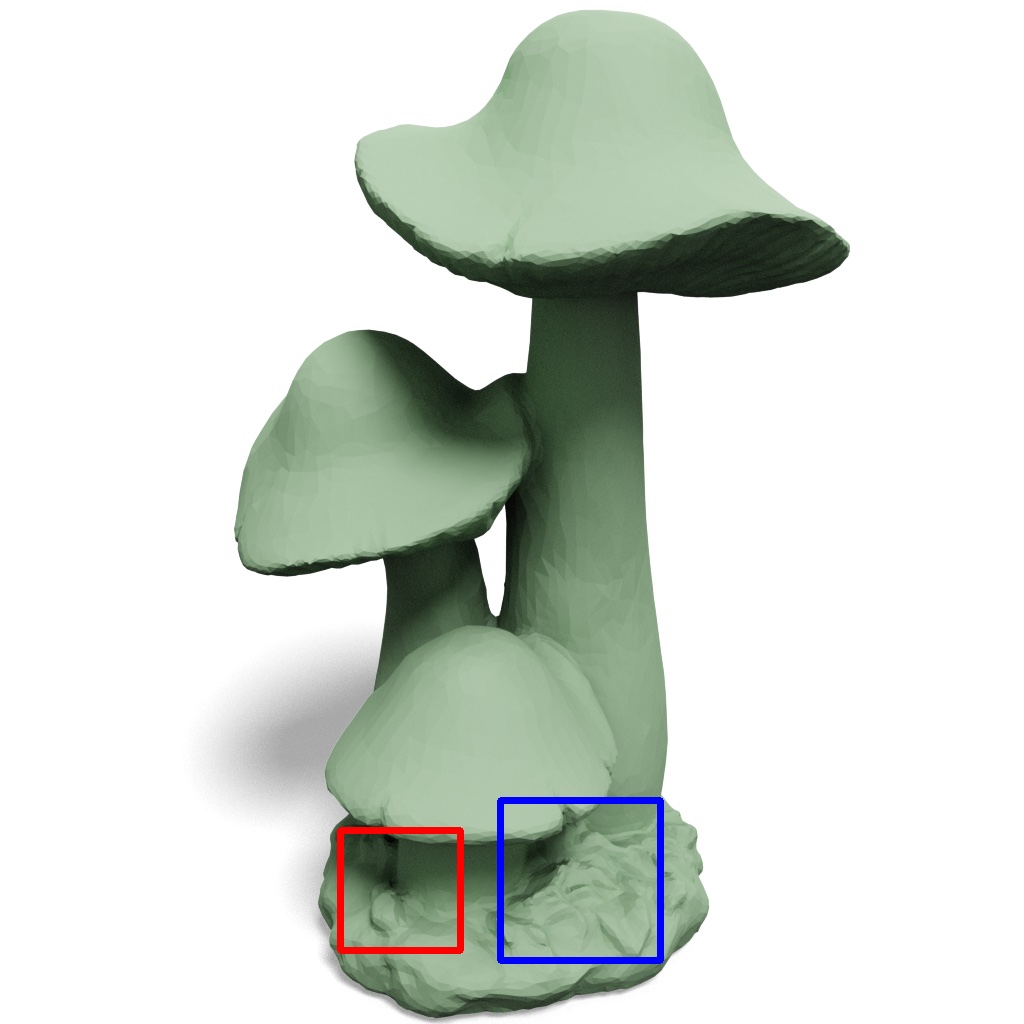}} \hfill
  \mpage{0.235}{\includegraphics[width=\linewidth]{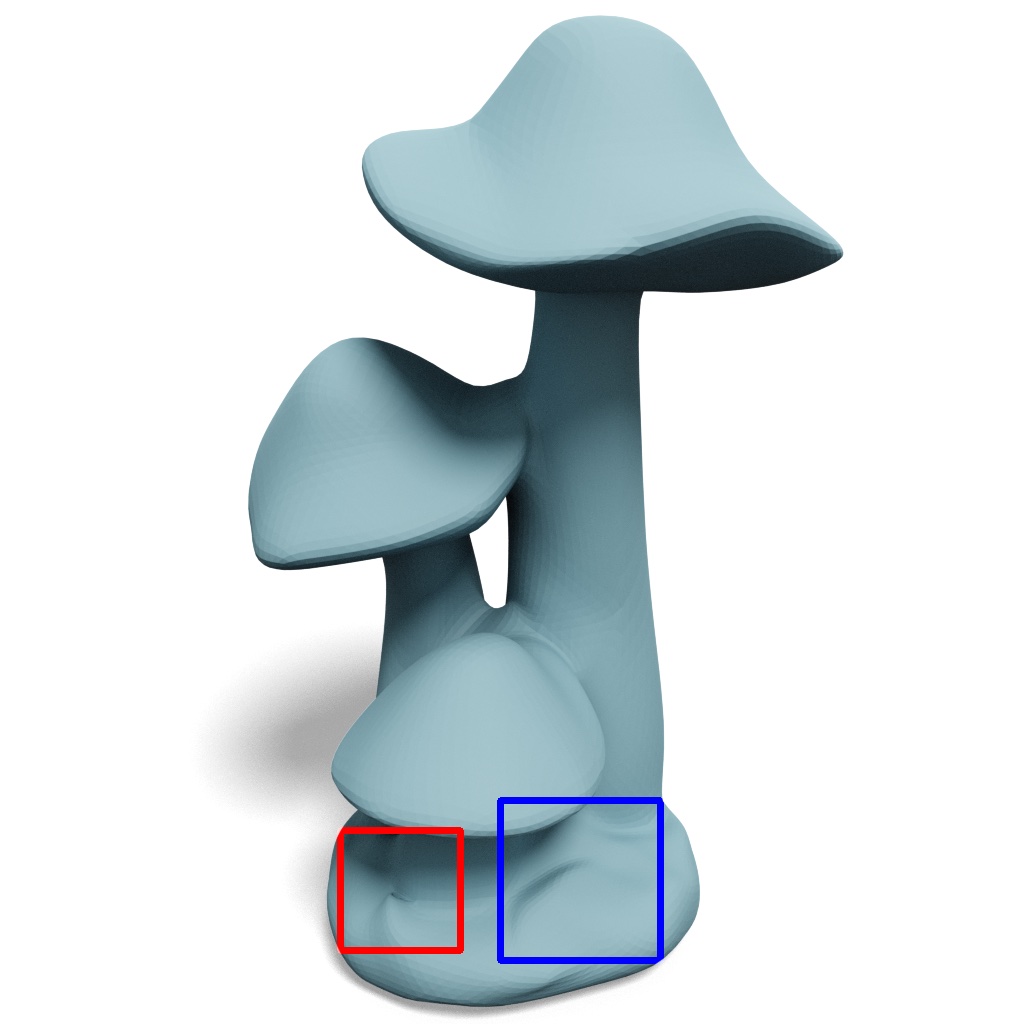}} \hfill
  \mpage{0.235}{\includegraphics[width=\linewidth]{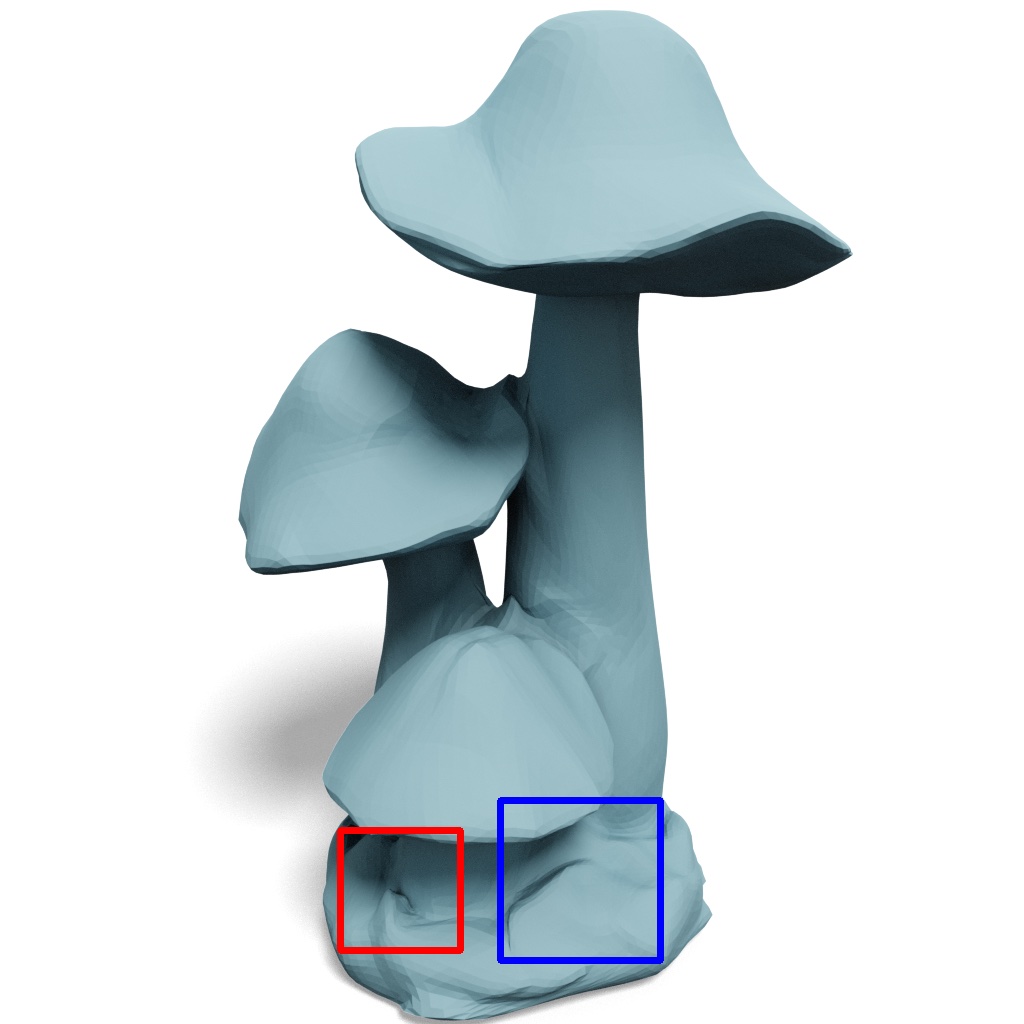}} \hfill
  \mpage{0.235}{\includegraphics[width=\linewidth]{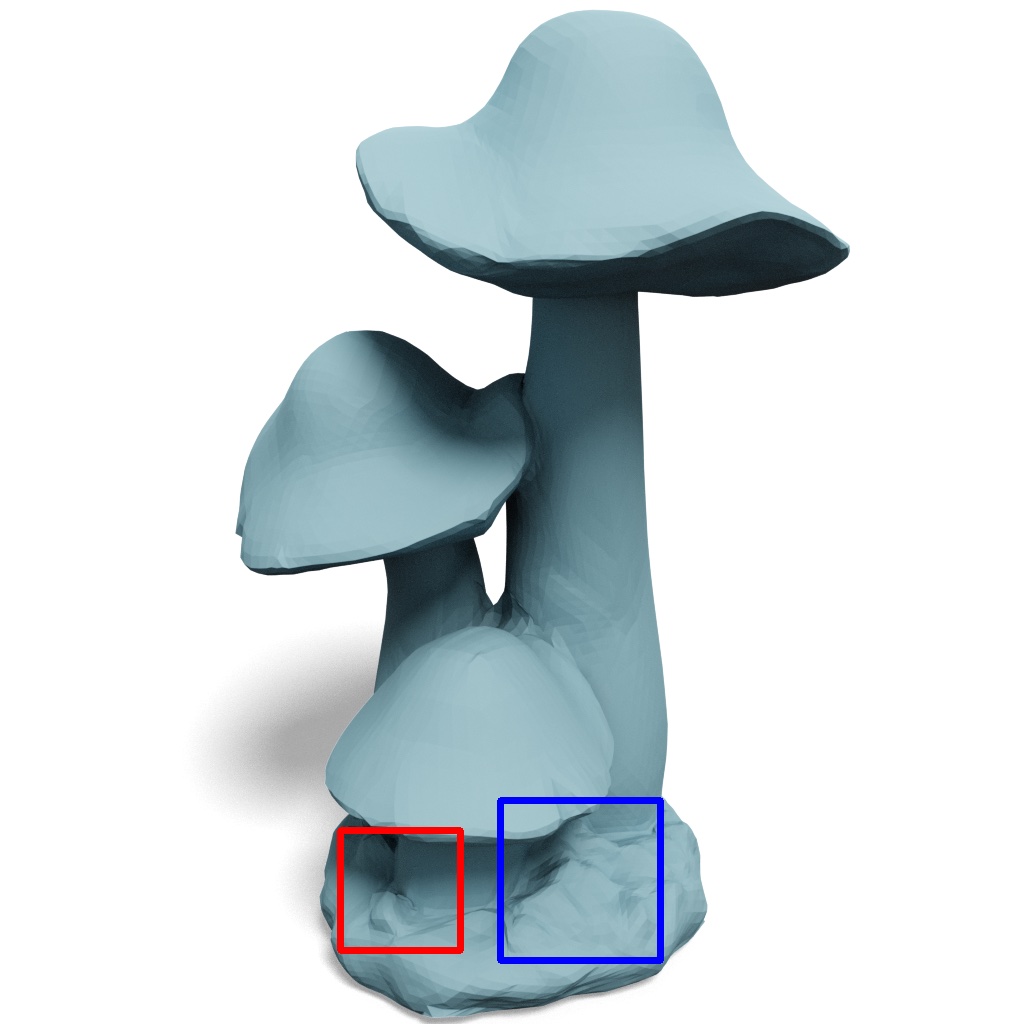}} \\
  \vspace{1.0mm}
  \mpage{0.235}{\includegraphics[width=0.475\linewidth]{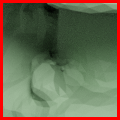} \hfill \includegraphics[width=0.475\linewidth]{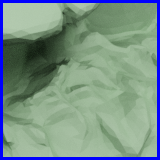}} \hfill
  \mpage{0.235}{\includegraphics[width=0.475\linewidth]{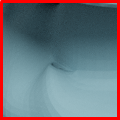} \hfill \includegraphics[width=0.475\linewidth]{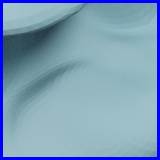}} \hfill
  \mpage{0.235}{\includegraphics[width=0.475\linewidth]{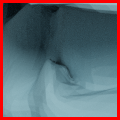} \hfill \includegraphics[width=0.475\linewidth]{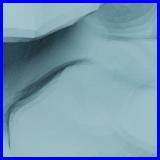}} \hfill
  \mpage{0.235}{\includegraphics[width=0.475\linewidth]{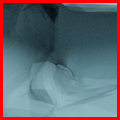} \hfill \includegraphics[width=0.475\linewidth]{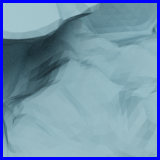}} \\
  \vspace{1.0mm}
  \mpage{0.235}{$CR$ / $d_\text{pm}$ ($\times 10^{-4}$) / $d_\text{normal}$} \hfill
  \mpage{0.235}{69.48 / 25.11 / 11.74$^\circ$} \hfill
  \mpage{0.235}{45.31 / 12.15 / 11.30$^\circ$} \hfill
  \mpage{0.235}{10.96 /  4.53 /  8.26$^\circ$} \\
  \vspace{1.0mm}
  \mpage{0.235}{\includegraphics[width=\linewidth]{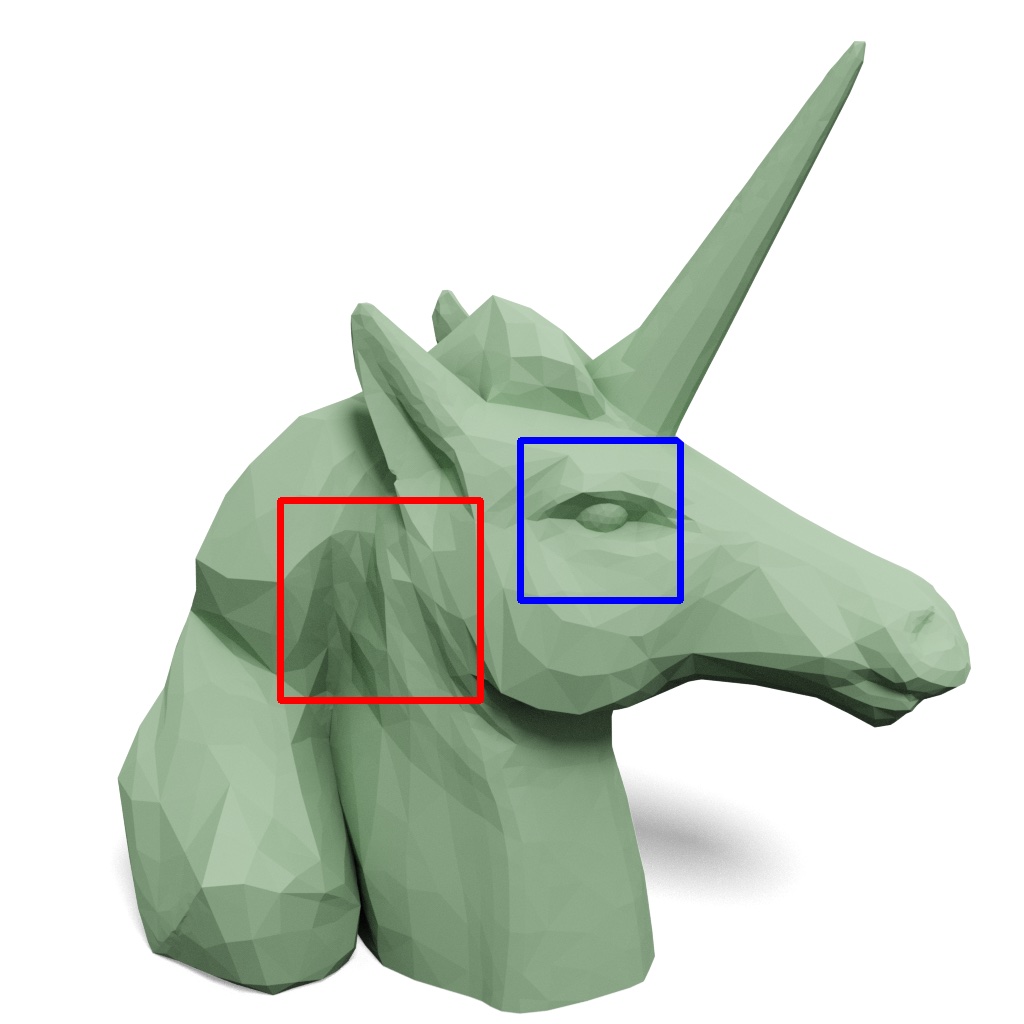}} \hfill
  \mpage{0.235}{\includegraphics[width=\linewidth]{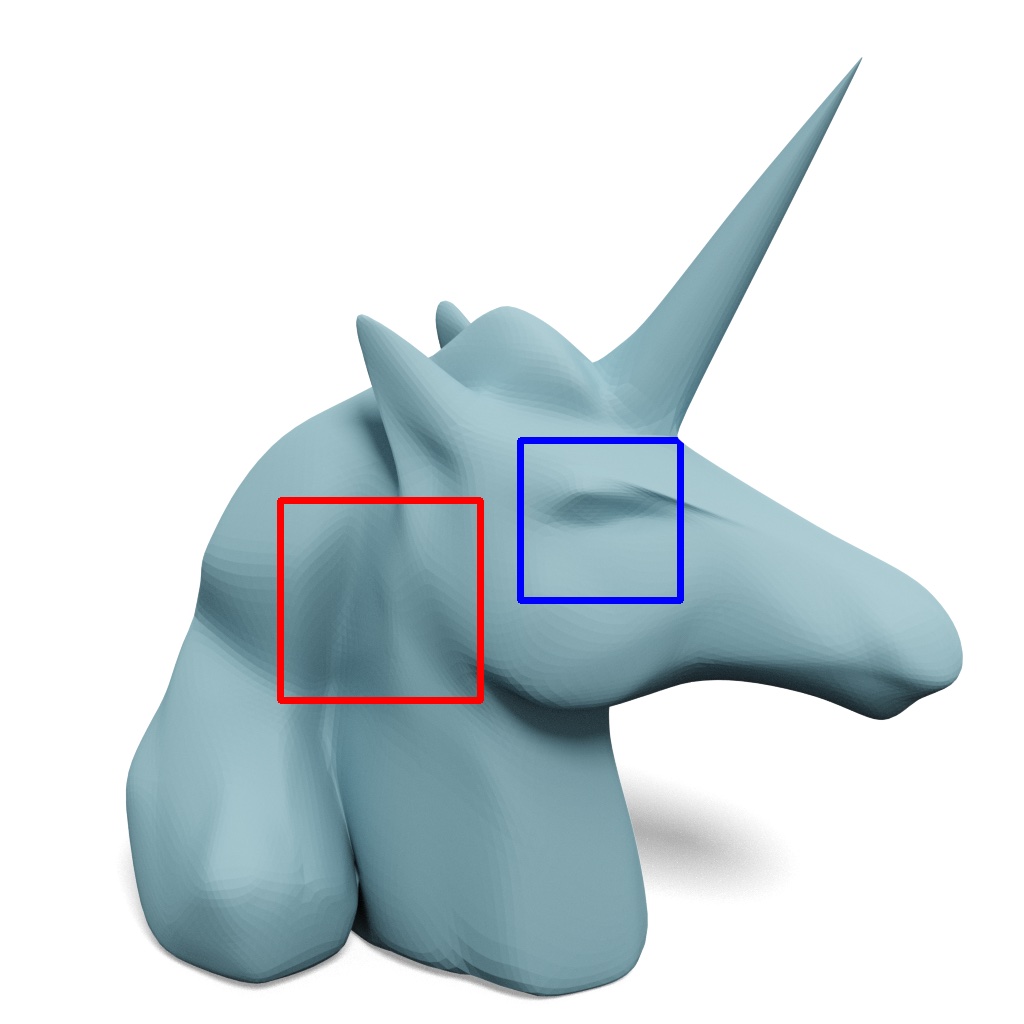}} \hfill
  \mpage{0.235}{\includegraphics[width=\linewidth]{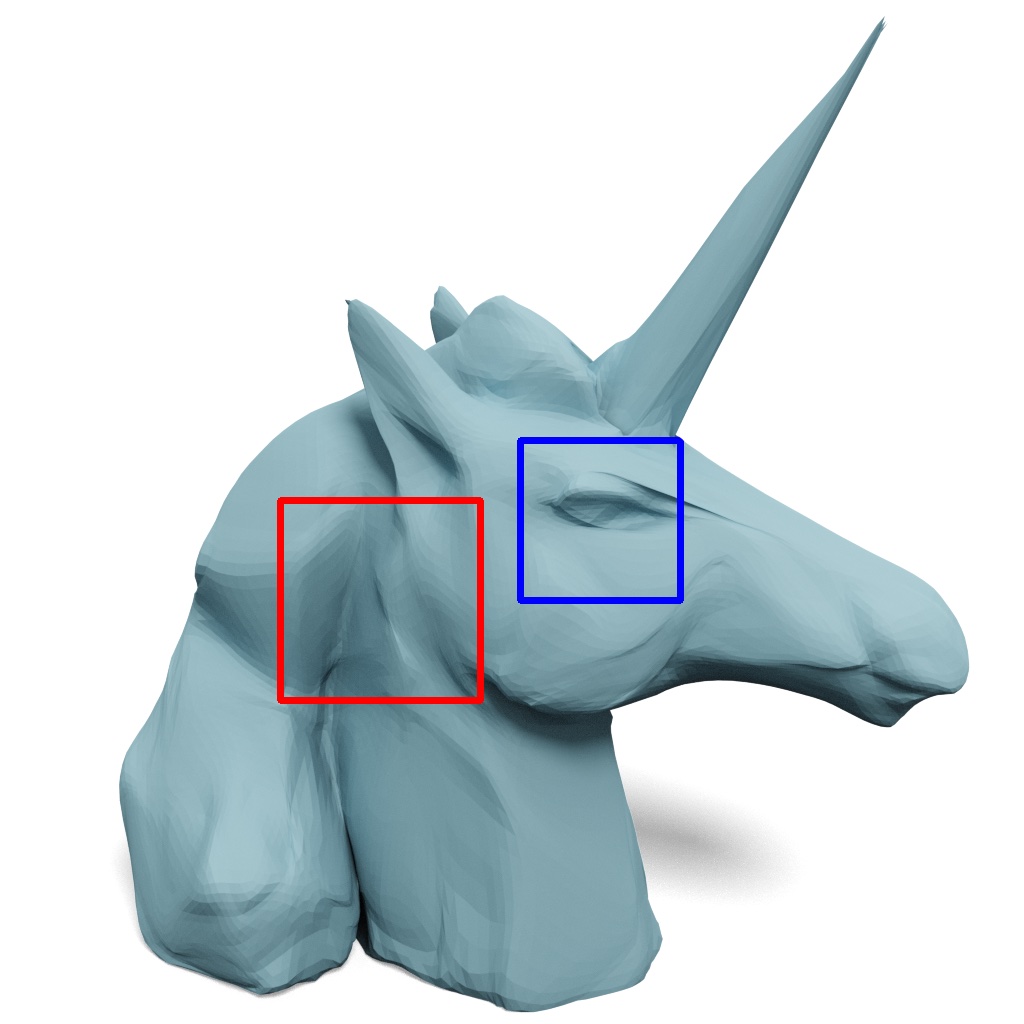}} \hfill
  \mpage{0.235}{\includegraphics[width=\linewidth]{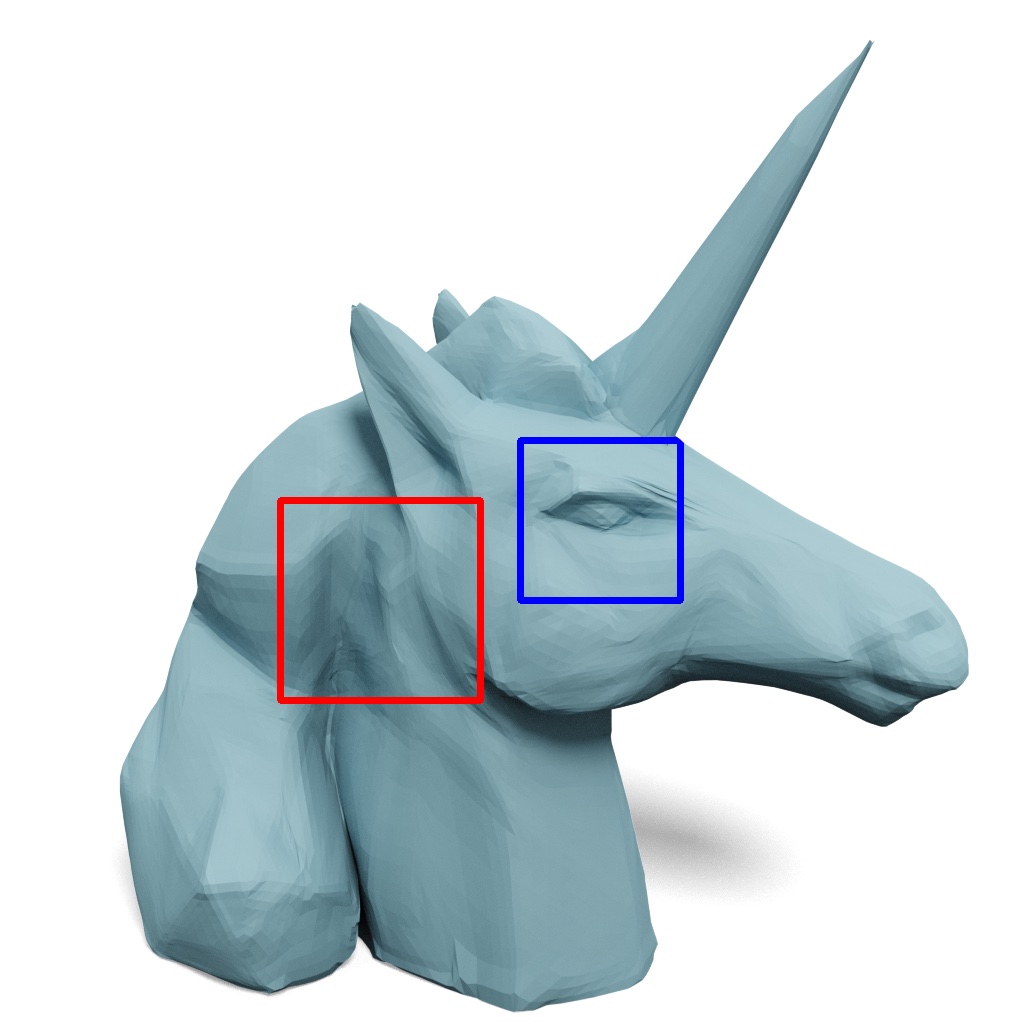}} \\
  \vspace{1.0mm}
  \mpage{0.235}{\includegraphics[width=0.475\linewidth]{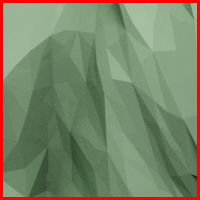} \hfill \includegraphics[width=0.475\linewidth]{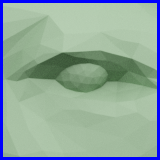}} \hfill
  \mpage{0.235}{\includegraphics[width=0.475\linewidth]{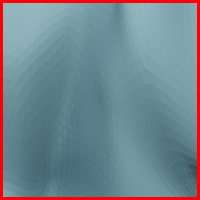} \hfill \includegraphics[width=0.475\linewidth]{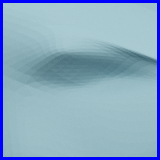}} \hfill
  \mpage{0.235}{\includegraphics[width=0.475\linewidth]{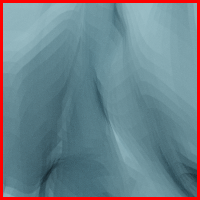} \hfill \includegraphics[width=0.475\linewidth]{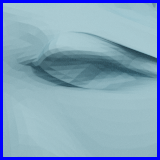}} \hfill
  \mpage{0.235}{\includegraphics[width=0.475\linewidth]{images/supp_gallery/55932_ours_2_crop1.png} \hfill \includegraphics[width=0.475\linewidth]{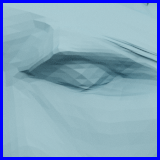}} \\
  \vspace{1.0mm}
  \mpage{0.235}{$CR$ / $d_\text{pm}$ ($\times 10^{-4}$) / $d_\text{normal}$} \hfill
  \mpage{0.235}{17.73 / 16.39 / 11.38$^\circ$} \hfill
  \mpage{0.235}{11.60 / 12.14 / 10.01$^\circ$} \hfill
  \mpage{0.235}{2.82 /  3.26 / 7.29$^\circ$} \\
  \vspace{1.0mm}
  \mpage{0.235}{\includegraphics[width=\linewidth]{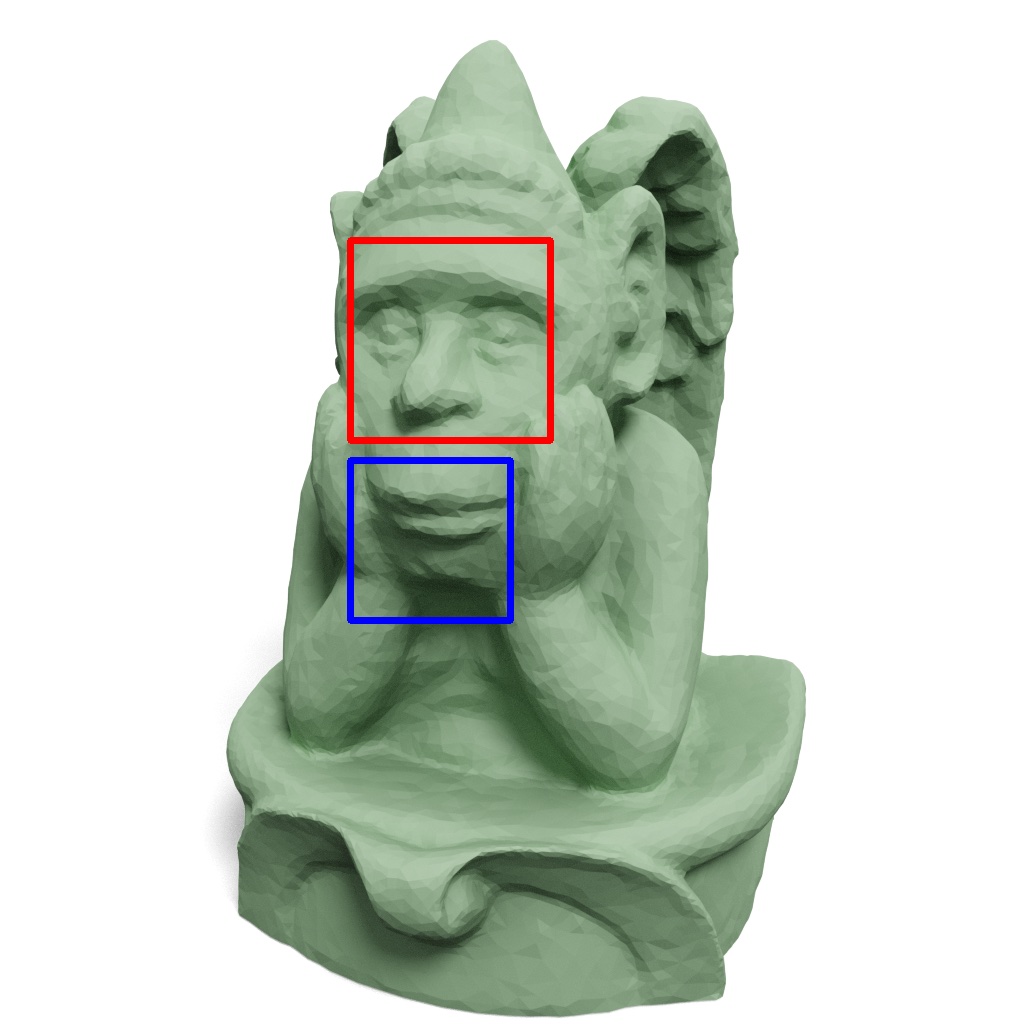}} \hfill
  \mpage{0.235}{\includegraphics[width=\linewidth]{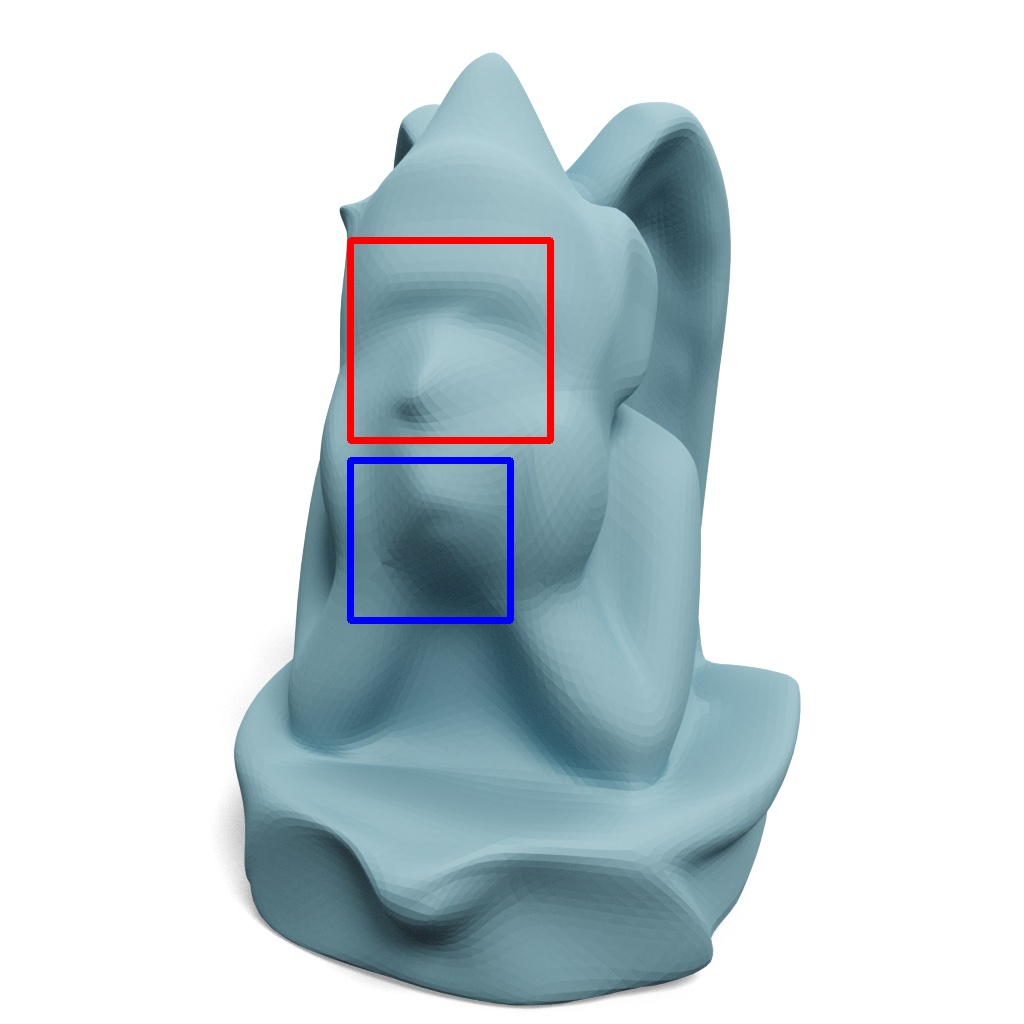}} \hfill
  \mpage{0.235}{\includegraphics[width=\linewidth]{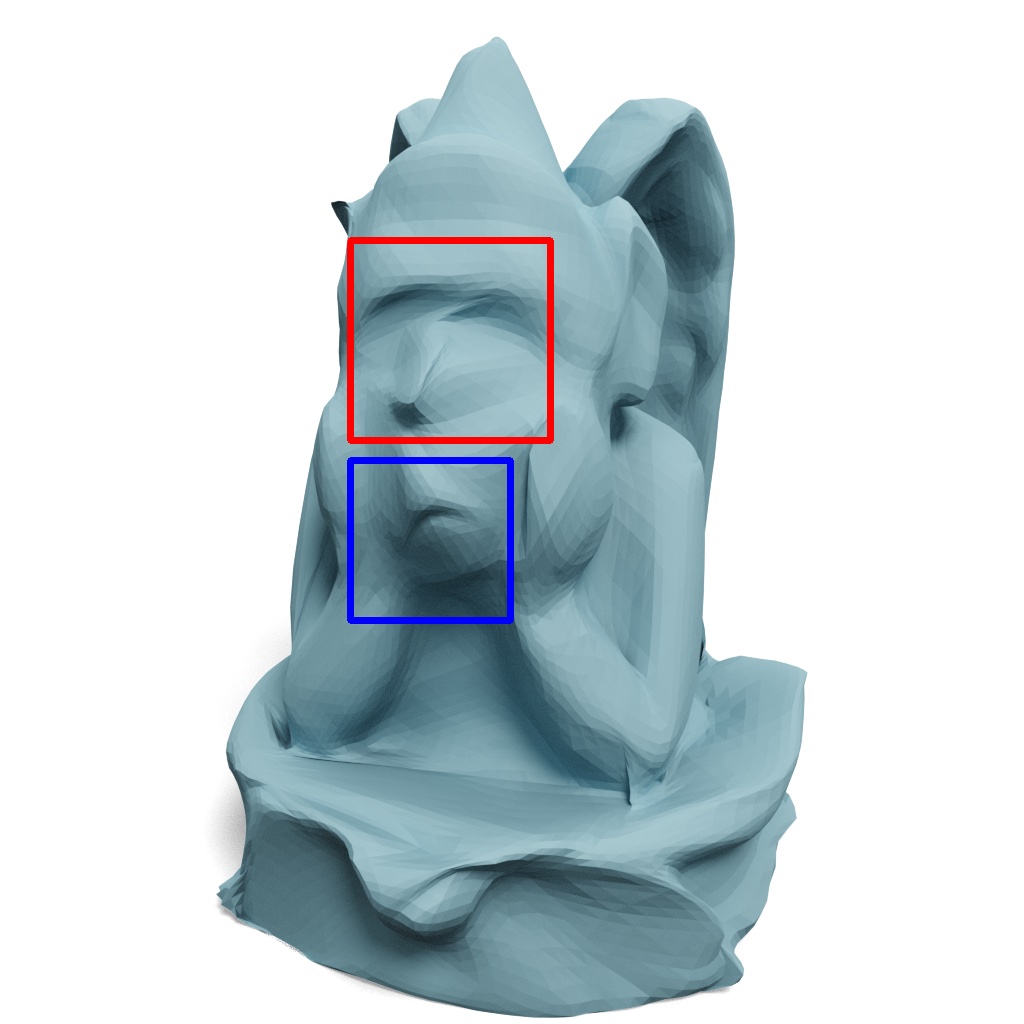}} \hfill
  \mpage{0.235}{\includegraphics[width=\linewidth]{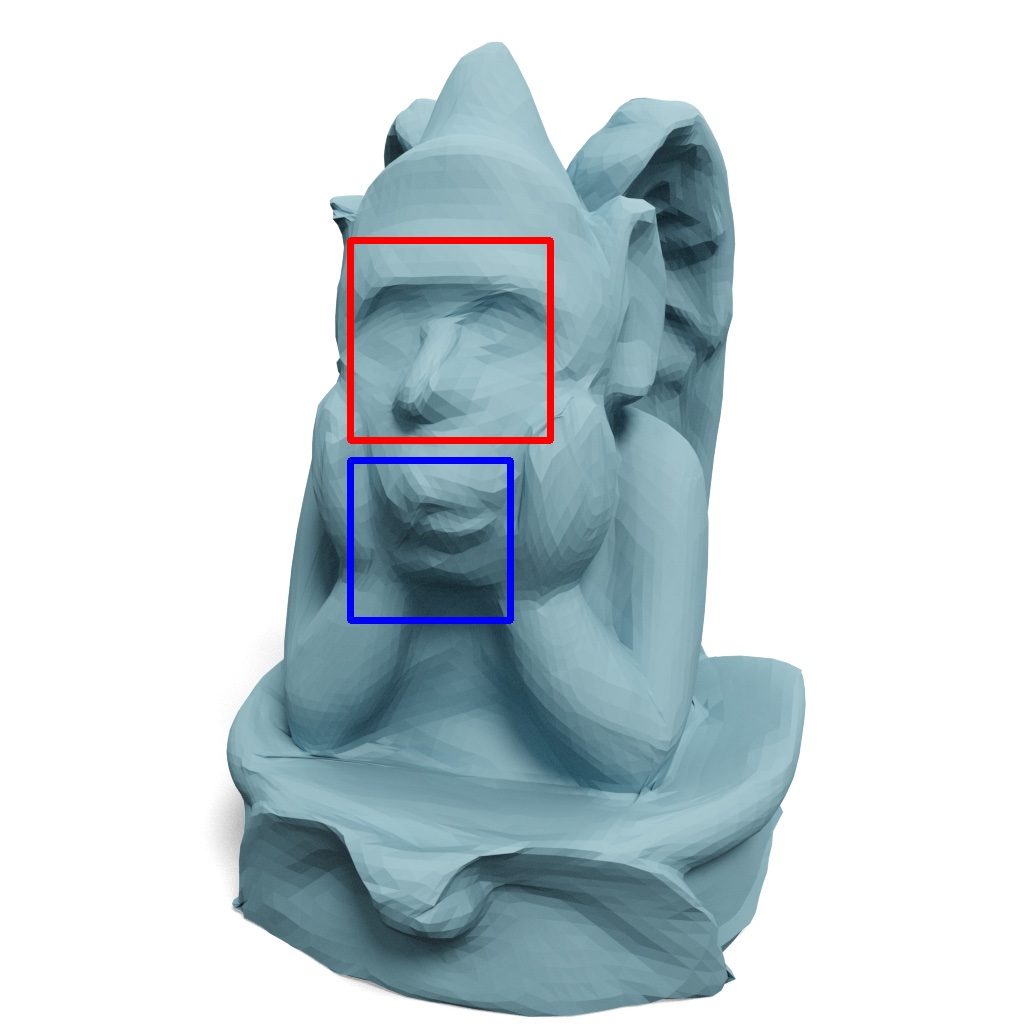}} \\
  \vspace{1.0mm}
  \mpage{0.235}{\includegraphics[width=0.475\linewidth]{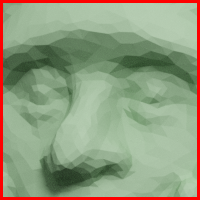} \hfill \includegraphics[width=0.475\linewidth]{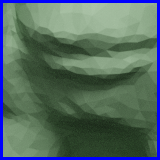}} \hfill
  \mpage{0.235}{\includegraphics[width=0.475\linewidth]{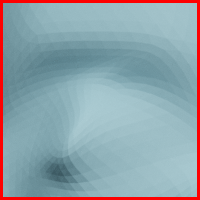} \hfill \includegraphics[width=0.475\linewidth]{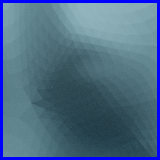}} \hfill
  \mpage{0.235}{\includegraphics[width=0.475\linewidth]{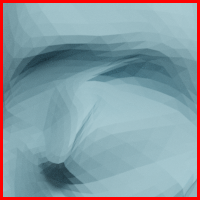} \hfill \includegraphics[width=0.475\linewidth]{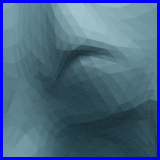}} \hfill
  \mpage{0.235}{\includegraphics[width=0.475\linewidth]{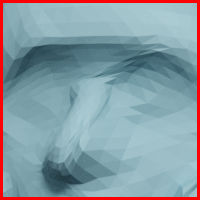} \hfill \includegraphics[width=0.475\linewidth]{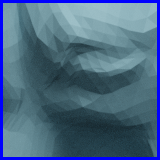}} \\
  \vspace{1.0mm}
  \mpage{0.235}{$CR$ / $d_\text{pm}$ ($\times 10^{-4}$) / $d_\text{normal}$} \hfill
  \mpage{0.235}{93.09 / 32.50 / 15.99$^\circ$} \hfill
  \mpage{0.235}{60.92 / 21.42 / 14.50$^\circ$} \hfill
  \mpage{0.235}{14.82 /  7.31 / 10.34$^\circ$} \\
  \vspace{1.0mm}
  \mpage{0.235}{Ground truth} \hfill
  \mpage{0.235}{Ours w/o features} \hfill
  \mpage{0.235}{Ours + 40 features} \hfill
  \mpage{0.235}{Ours + 400 features} \\
  \caption{
  \textbf{Progressive features.} 
  We show more examples where transmitting more features leads to better quantitative and qualitative results.
  }
  \label{supp-fig:prog-feat-1}
\end{figure*}

\begin{figure*}[t]
  \centering
  \mpage{0.235}{\includegraphics[width=\linewidth]{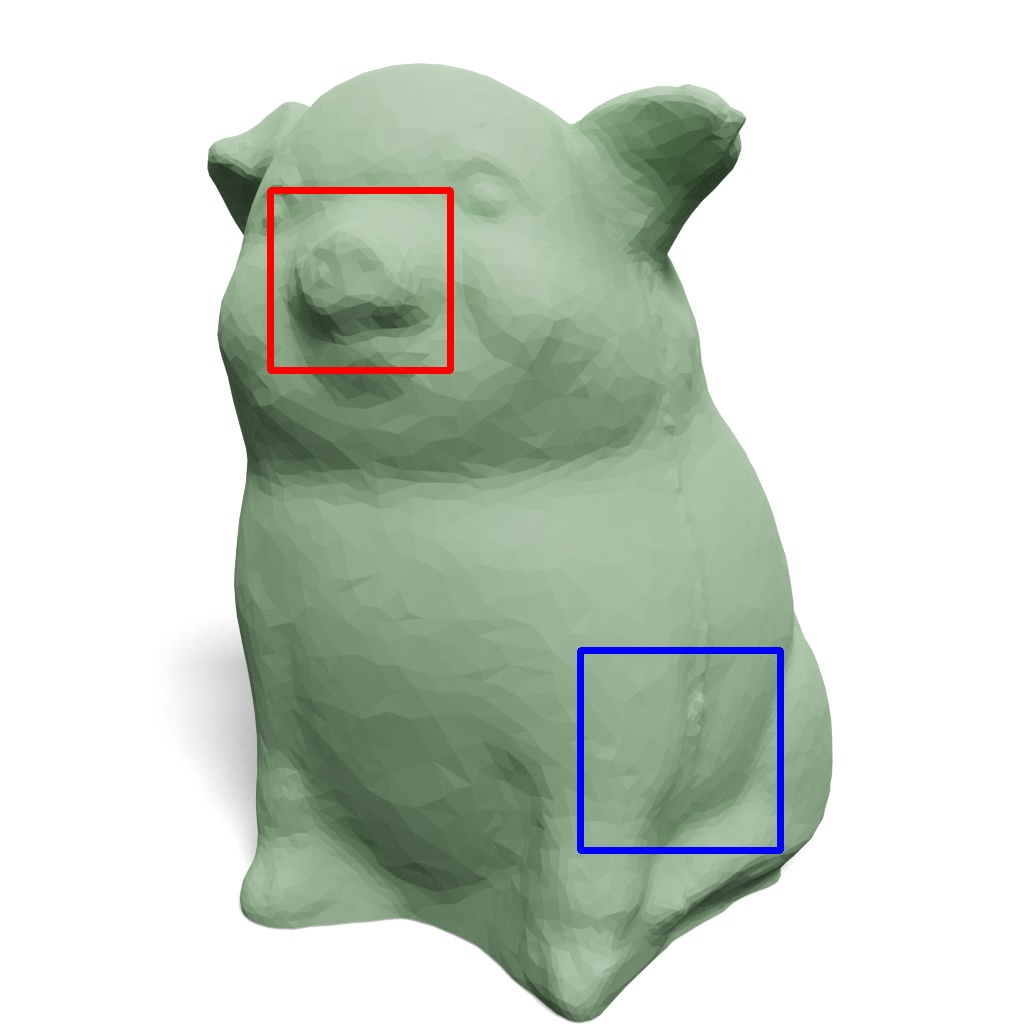}} \hfill
  \mpage{0.235}{\includegraphics[width=\linewidth]{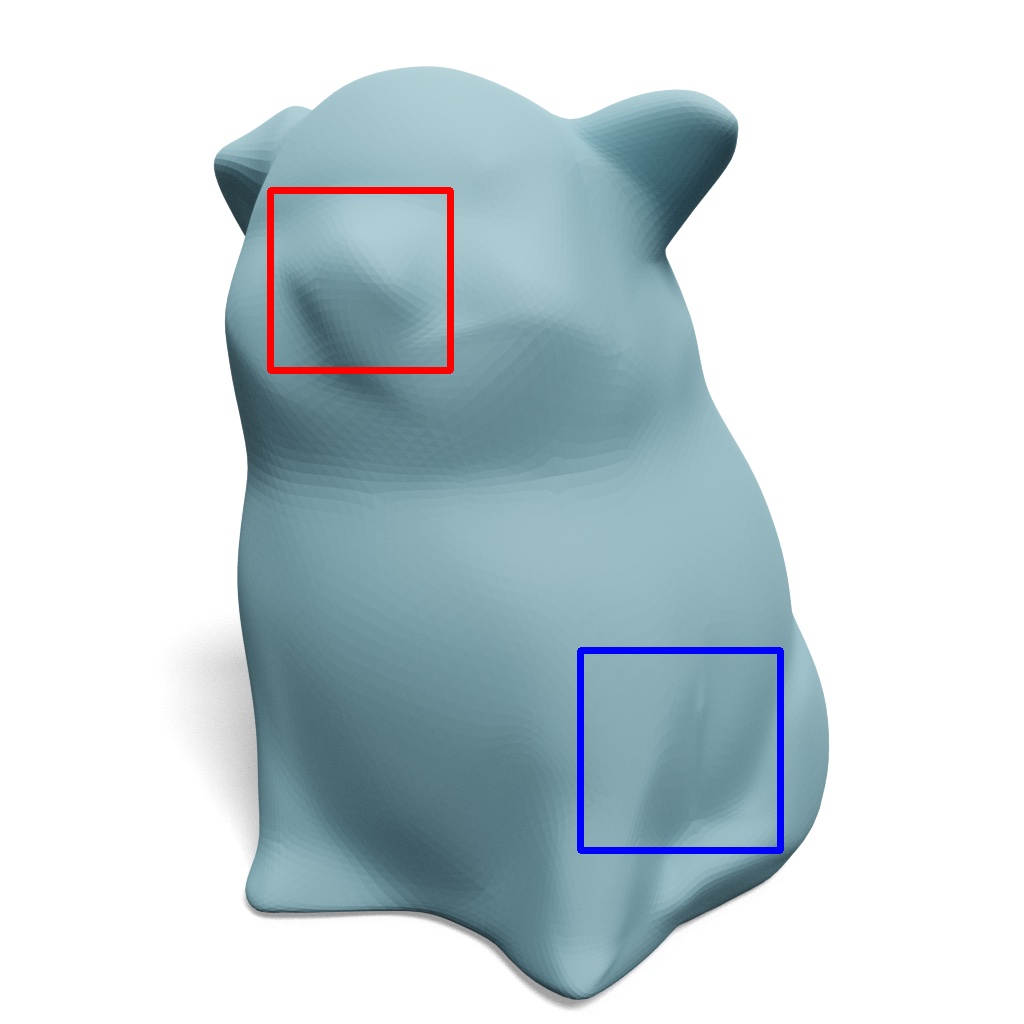}} \hfill
  \mpage{0.235}{\includegraphics[width=\linewidth]{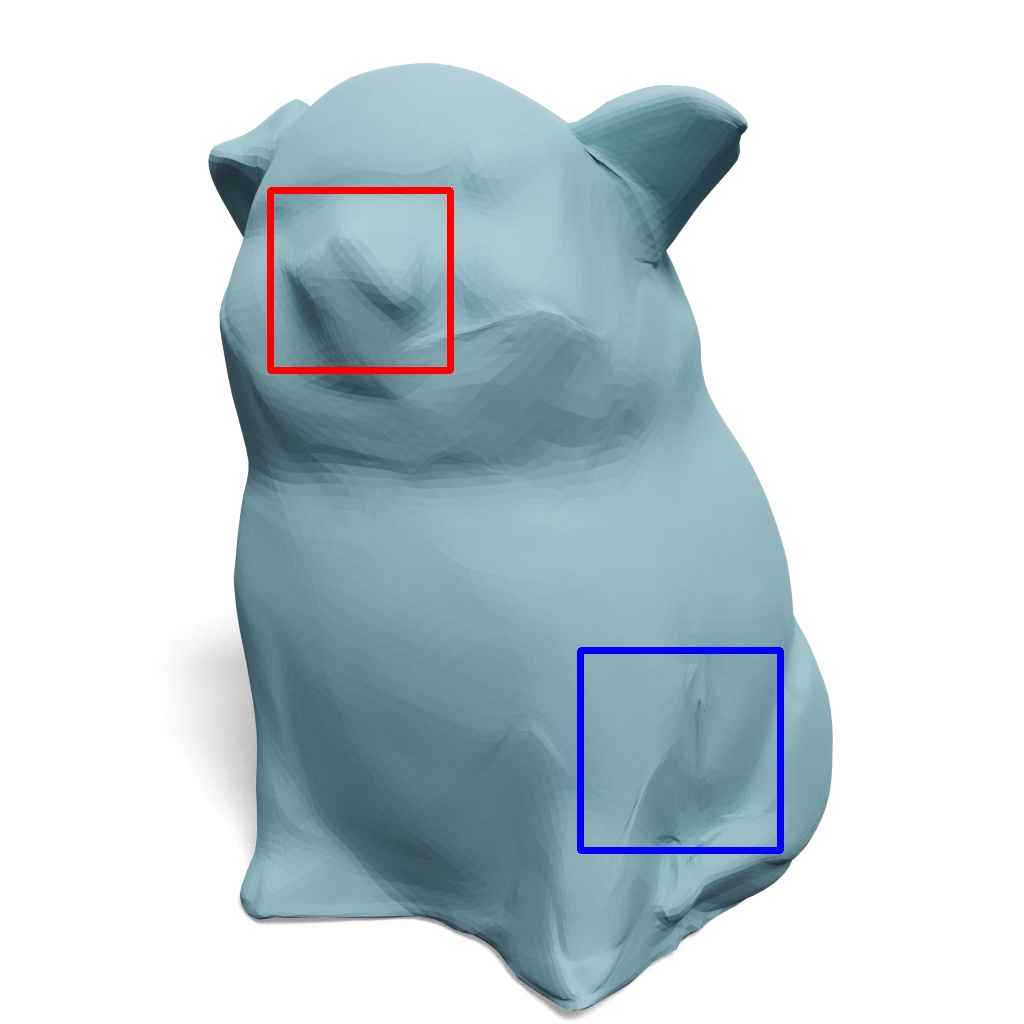}} \hfill
  \mpage{0.235}{\includegraphics[width=\linewidth]{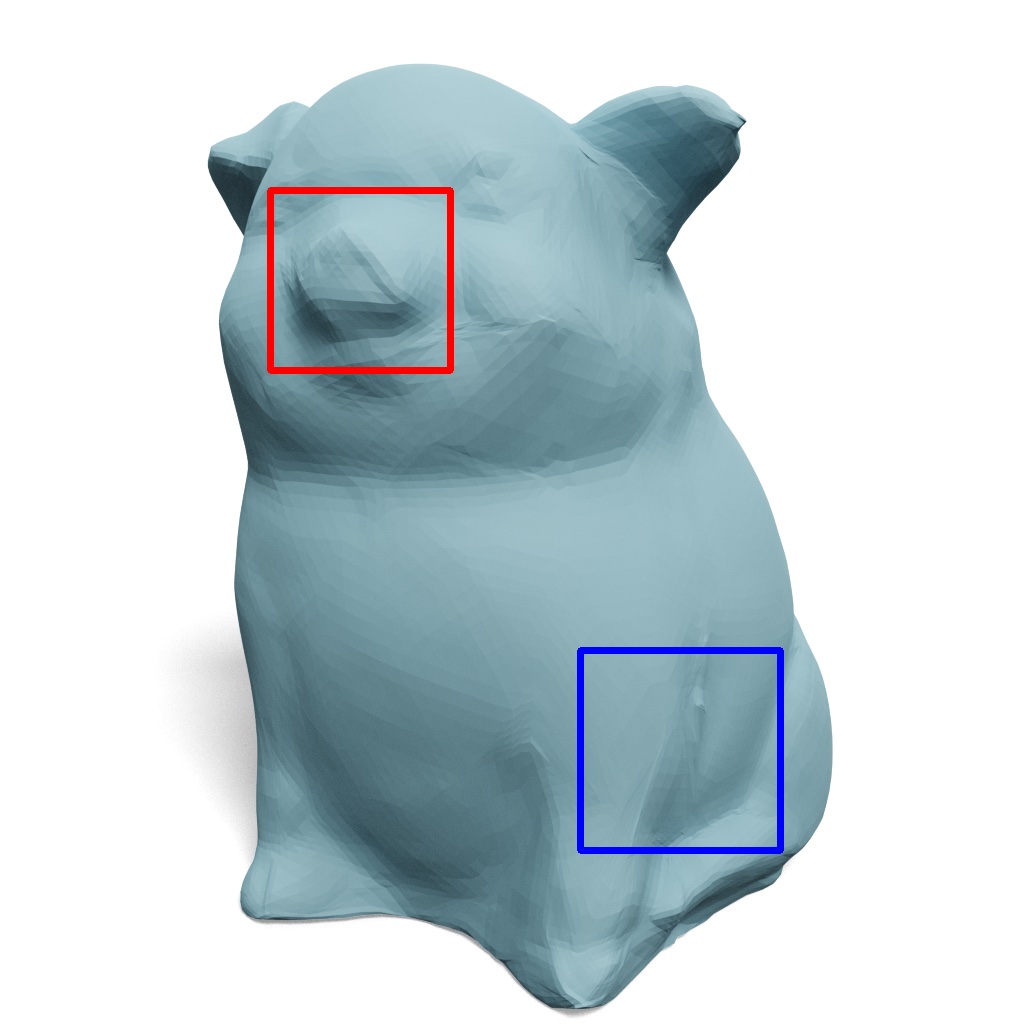}} \\
  \vspace{1.0mm}
  \mpage{0.235}{\includegraphics[width=0.475\linewidth]{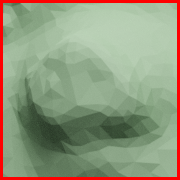} \hfill \includegraphics[width=0.475\linewidth]{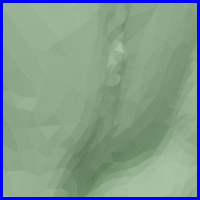}} \hfill
  \mpage{0.235}{\includegraphics[width=0.475\linewidth]{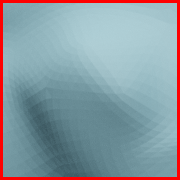} \hfill \includegraphics[width=0.475\linewidth]{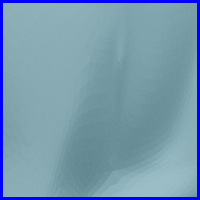}} \hfill
  \mpage{0.235}{\includegraphics[width=0.475\linewidth]{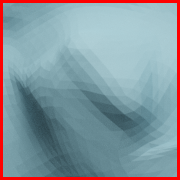} \hfill \includegraphics[width=0.475\linewidth]{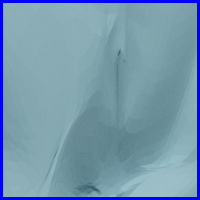}} \hfill
  \mpage{0.235}{\includegraphics[width=0.475\linewidth]{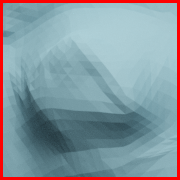} \hfill \includegraphics[width=0.475\linewidth]{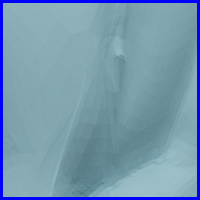}} \\
  \vspace{1.0mm}
  \mpage{0.235}{$CR$ / $d_\text{pm}$ ($\times 10^{-4}$) / $d_\text{normal}$} \hfill
  \mpage{0.235}{28.87 / 16.57 / 9.26$^\circ$} \hfill
  \mpage{0.235}{18.89 /  6.65 / 6.14$^\circ$} \hfill
  \mpage{0.235}{4.60 /  2.26 / 4.03$^\circ$} \\
  \vspace{1.0mm}
  \mpage{0.235}{\includegraphics[width=\linewidth]{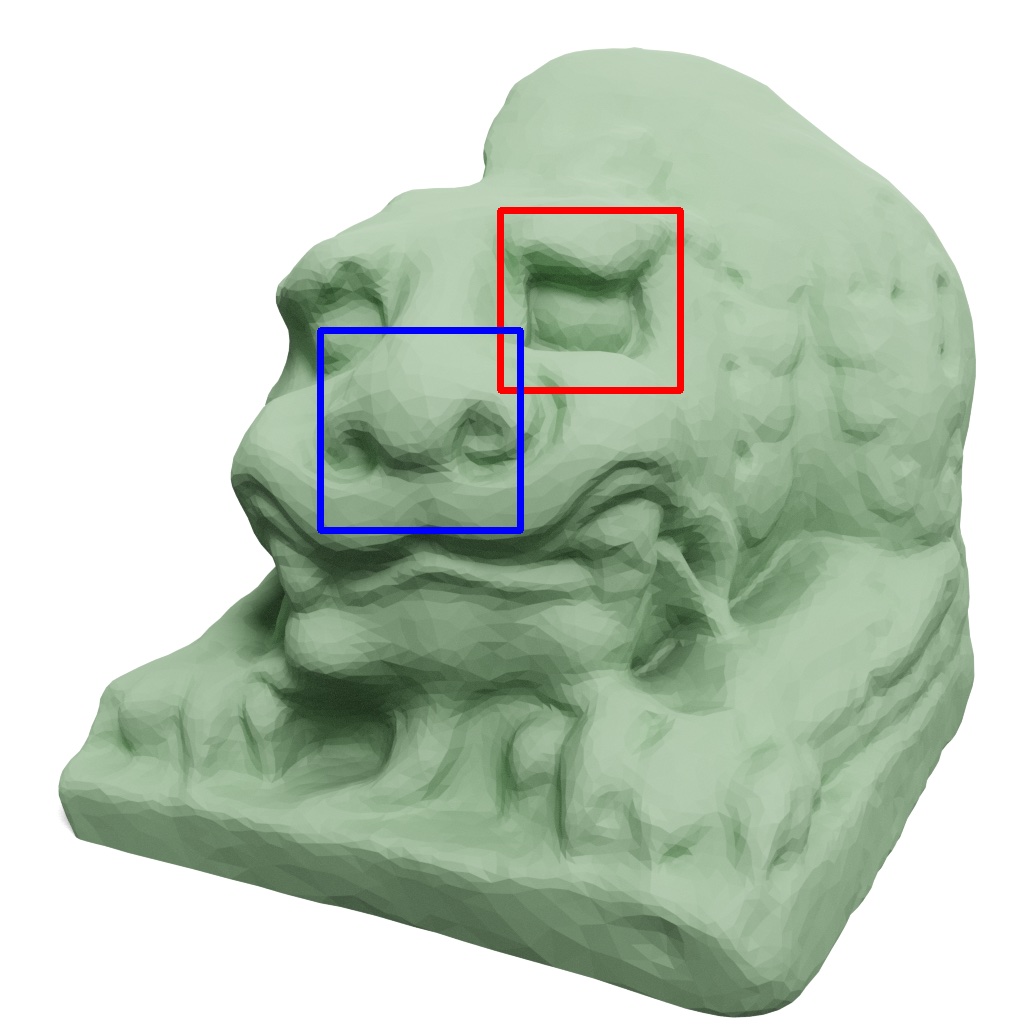}} \hfill
  \mpage{0.235}{\includegraphics[width=\linewidth]{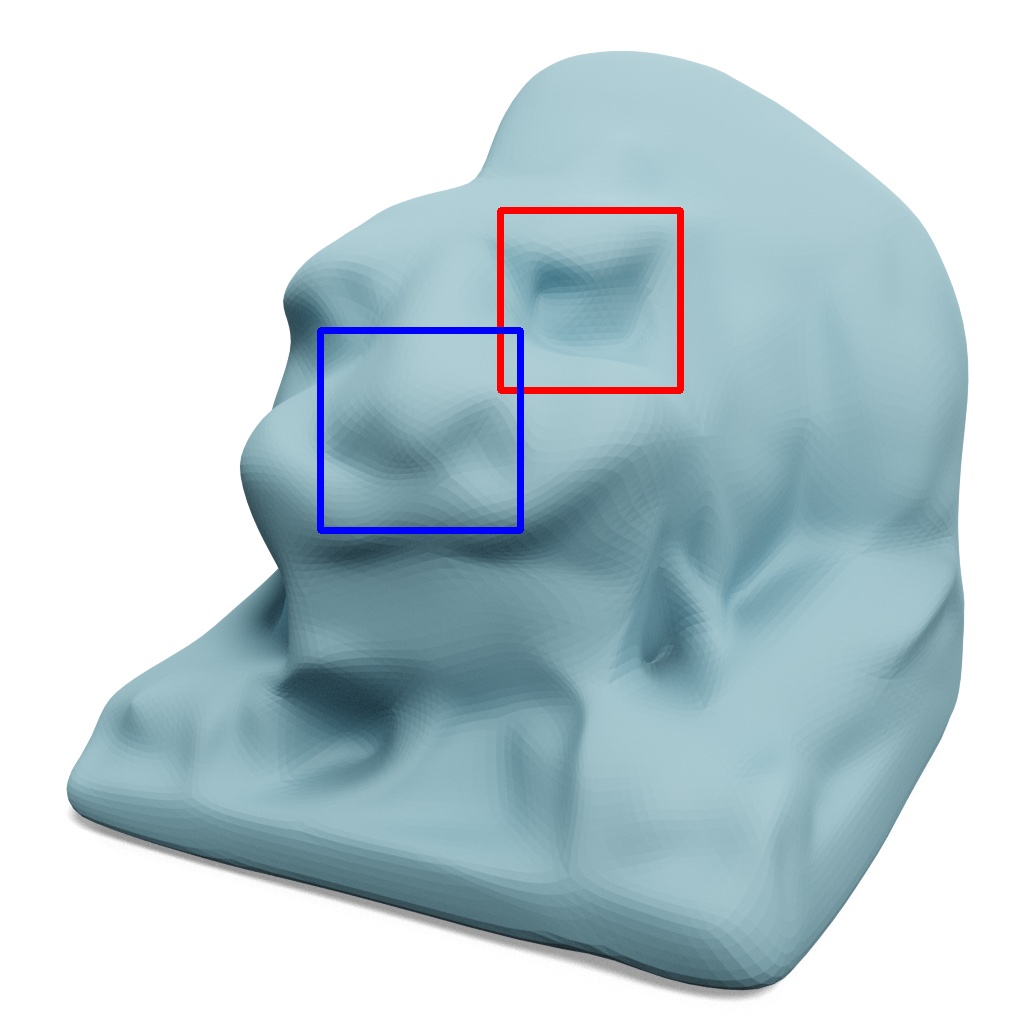}} \hfill
  \mpage{0.235}{\includegraphics[width=\linewidth]{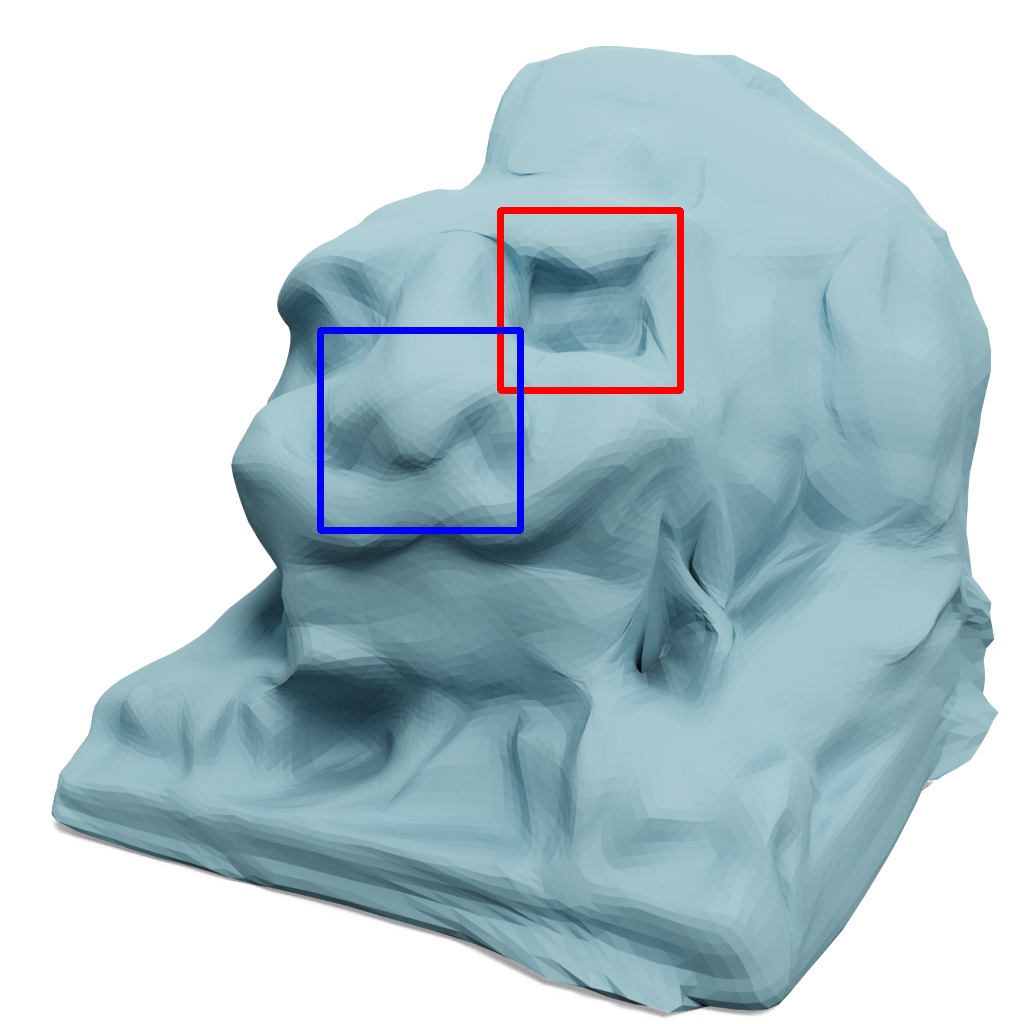}} \hfill
  \mpage{0.235}{\includegraphics[width=\linewidth]{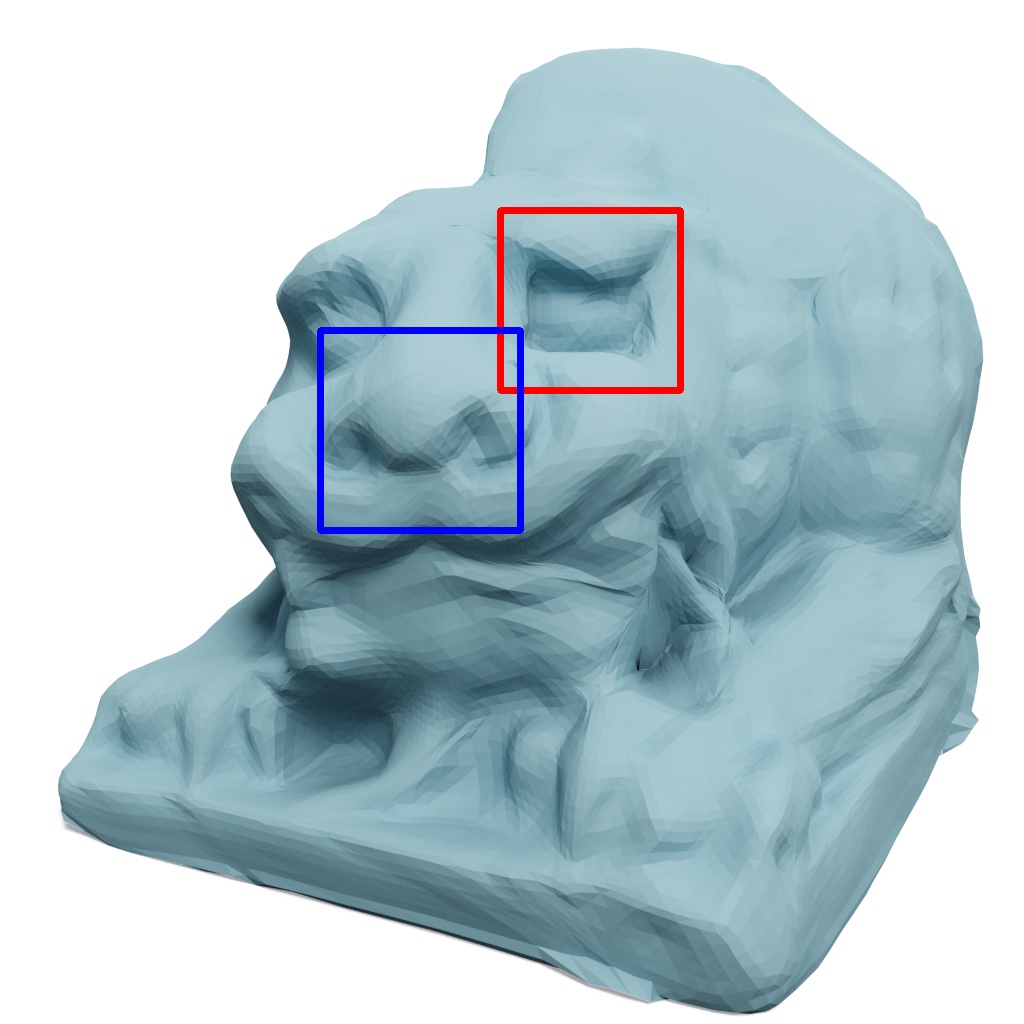}} \\
  \vspace{1.0mm}
  \mpage{0.235}{\includegraphics[width=0.475\linewidth]{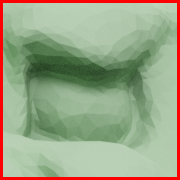} \hfill \includegraphics[width=0.475\linewidth]{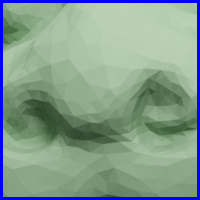}} \hfill
  \mpage{0.235}{\includegraphics[width=0.475\linewidth]{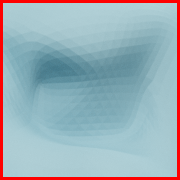} \hfill \includegraphics[width=0.475\linewidth]{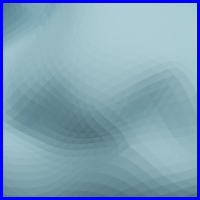}} \hfill
  \mpage{0.235}{\includegraphics[width=0.475\linewidth]{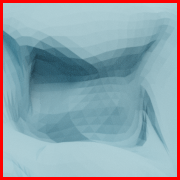} \hfill \includegraphics[width=0.475\linewidth]{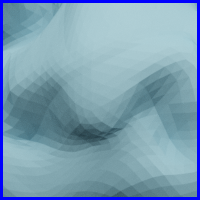}} \hfill
  \mpage{0.235}{\includegraphics[width=0.475\linewidth]{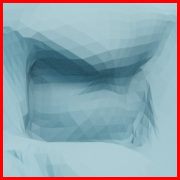} \hfill \includegraphics[width=0.475\linewidth]{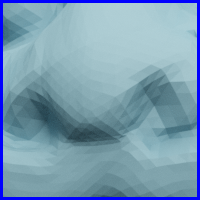}} \\
  \vspace{1.0mm}
  \mpage{0.235}{$CR$ / $d_\text{pm}$ ($\times 10^{-4}$) / $d_\text{normal}$} \hfill
  \mpage{0.235}{51.35 / 15.63 / 10.22$^\circ$} \hfill
  \mpage{0.235}{33.61 /  4.48 /  9.53$^\circ$} \hfill
  \mpage{0.235}{8.18  /  2.36 /  6.28$^\circ$} \\
  \vspace{1.0mm}
  \mpage{0.235}{\includegraphics[width=\linewidth]{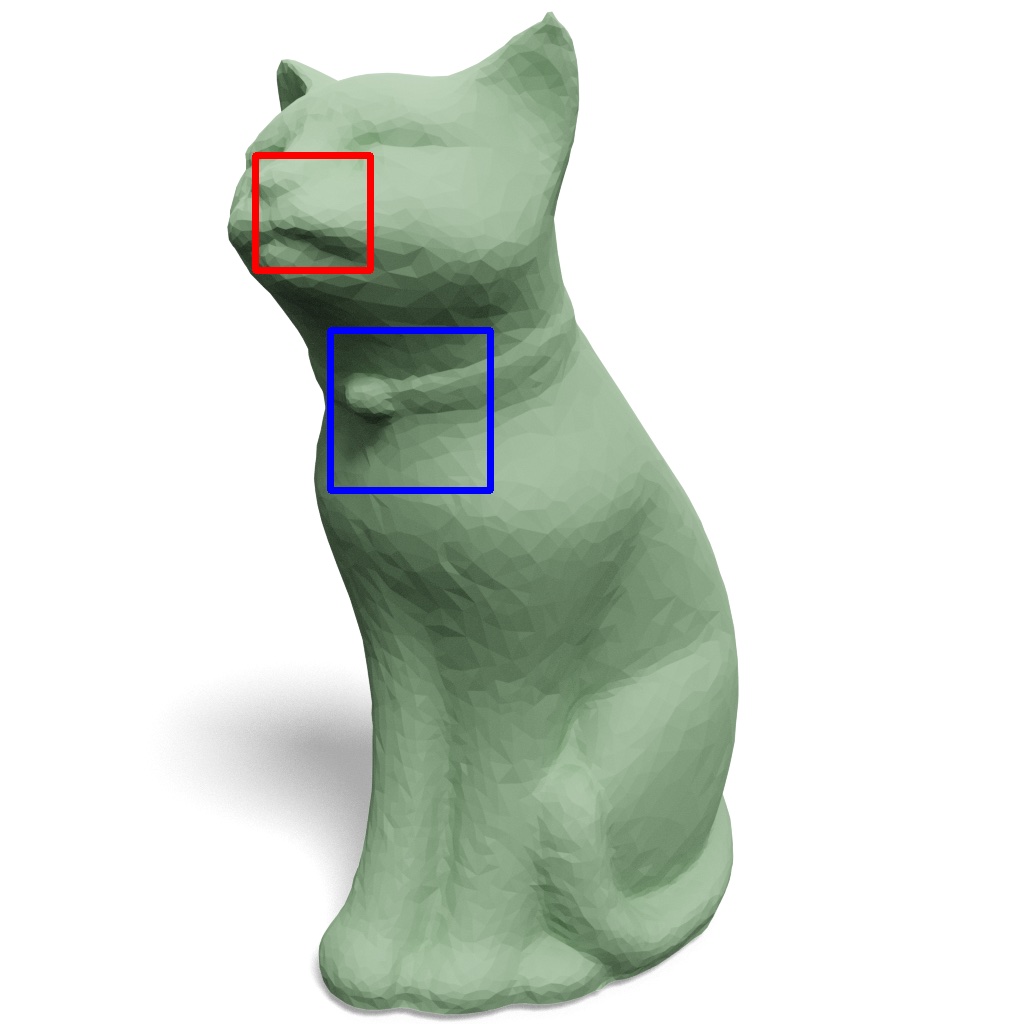}} \hfill
  \mpage{0.235}{\includegraphics[width=\linewidth]{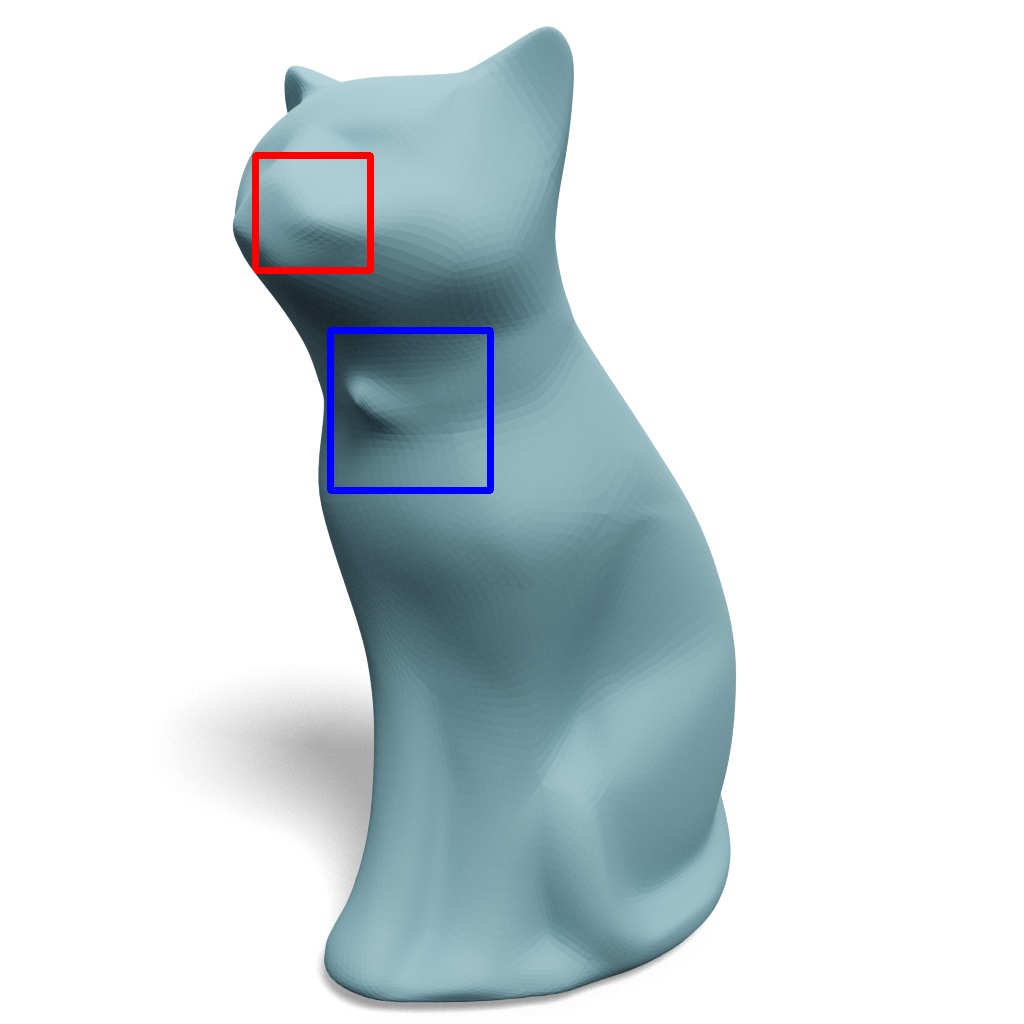}} \hfill
  \mpage{0.235}{\includegraphics[width=\linewidth]{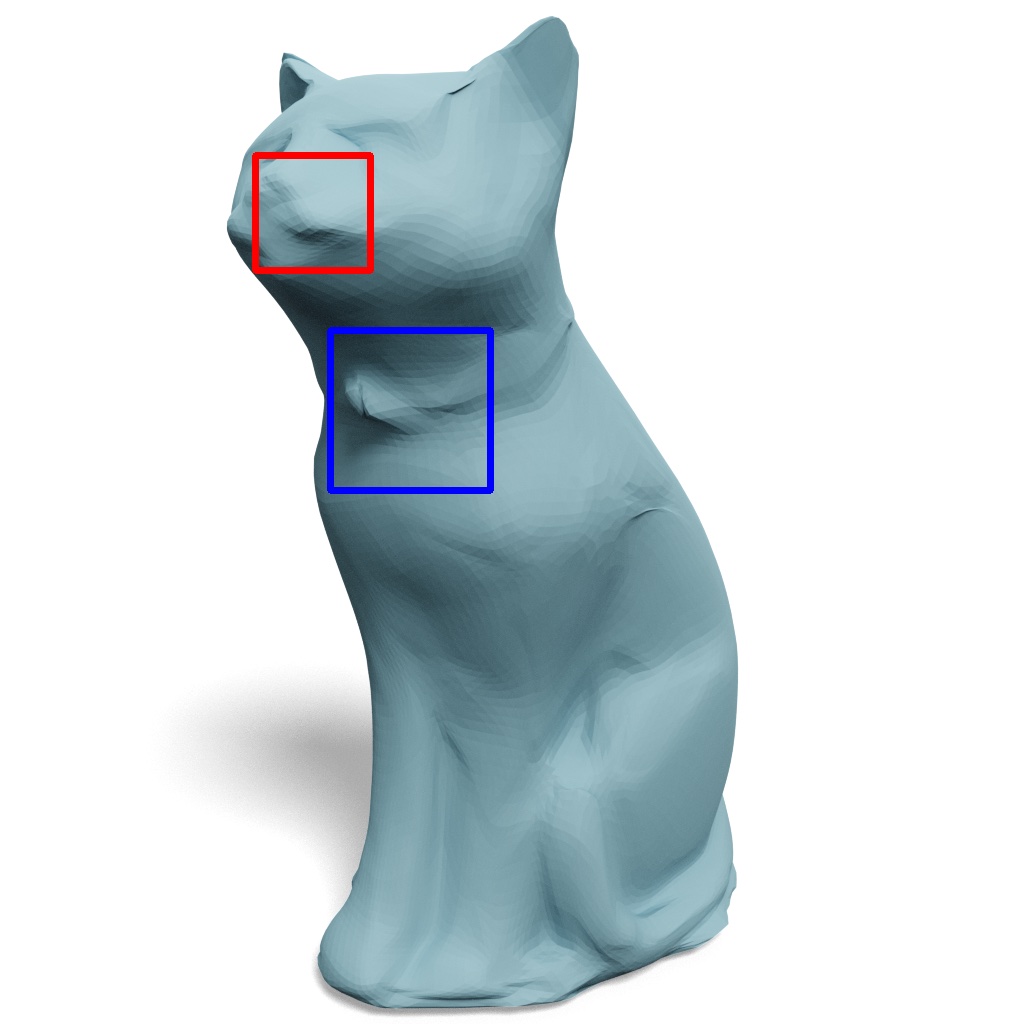}} \hfill
  \mpage{0.235}{\includegraphics[width=\linewidth]{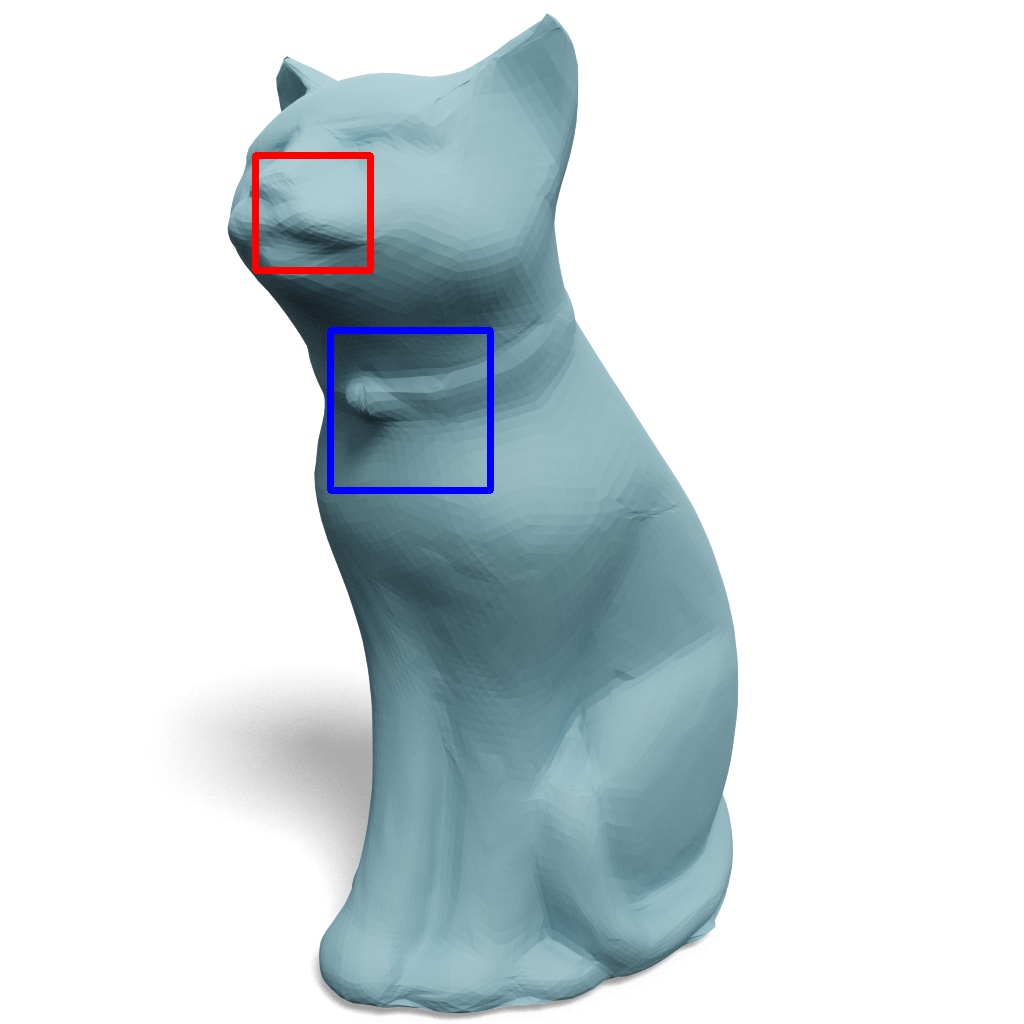}} \\
  \vspace{1.0mm}
  \mpage{0.235}{\includegraphics[width=0.475\linewidth]{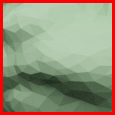} \hfill \includegraphics[width=0.475\linewidth]{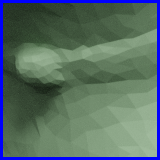}} \hfill
  \mpage{0.235}{\includegraphics[width=0.475\linewidth]{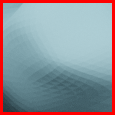} \hfill \includegraphics[width=0.475\linewidth]{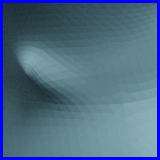}} \hfill
  \mpage{0.235}{\includegraphics[width=0.475\linewidth]{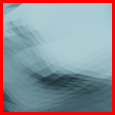} \hfill \includegraphics[width=0.475\linewidth]{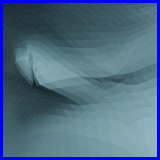}} \hfill
  \mpage{0.235}{\includegraphics[width=0.475\linewidth]{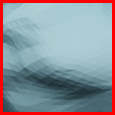} \hfill \includegraphics[width=0.475\linewidth]{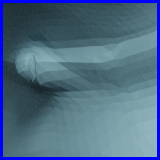}} \\
  \vspace{1.0mm}
  \mpage{0.235}{$CR$ / $d_\text{pm}$ ($\times 10^{-4}$) / $d_\text{normal}$} \hfill
  \mpage{0.235}{27.88 / 15.42 / 6.19$^\circ$} \hfill
  \mpage{0.235}{23.27 /  5.87 / 5.63$^\circ$} \hfill
  \mpage{0.235}{9.36  /  4.63 / 4.17$^\circ$} \\
  \vspace{1.0mm}
  \mpage{0.235}{Ground truth} \hfill
  \mpage{0.235}{Ours w/o features} \hfill
  \mpage{0.235}{Ours + 40 features} \hfill
  \mpage{0.235}{Ours + 400 features} \\
  \caption{
  \textbf{Progressive features.} 
  We show more examples where transmitting more features leads to better quantitative and qualitative results.
  }
  \label{supp-fig:prog-feat-2}
\end{figure*}

\begin{figure*}[t]
  \centering
  \mpage{0.235}{\includegraphics[width=\linewidth]{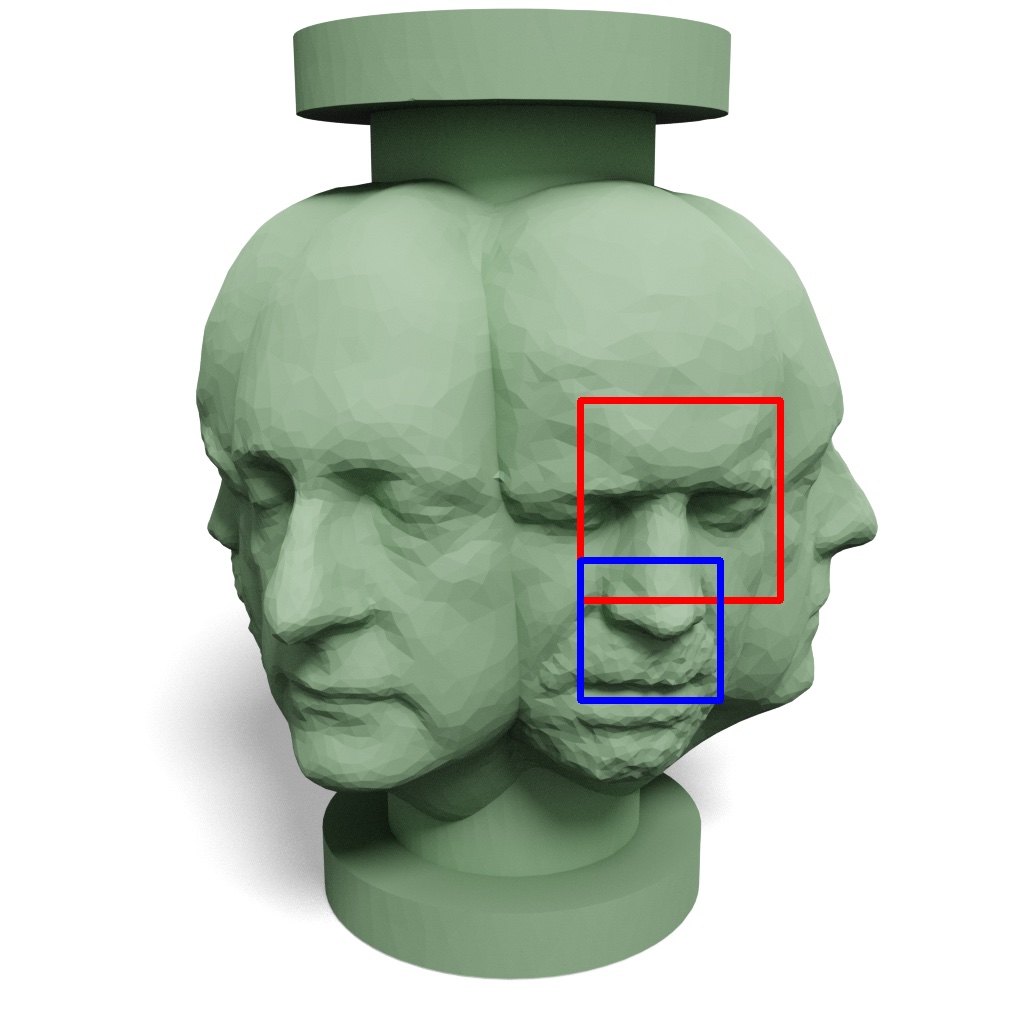}} \hfill
  \mpage{0.235}{\includegraphics[width=\linewidth]{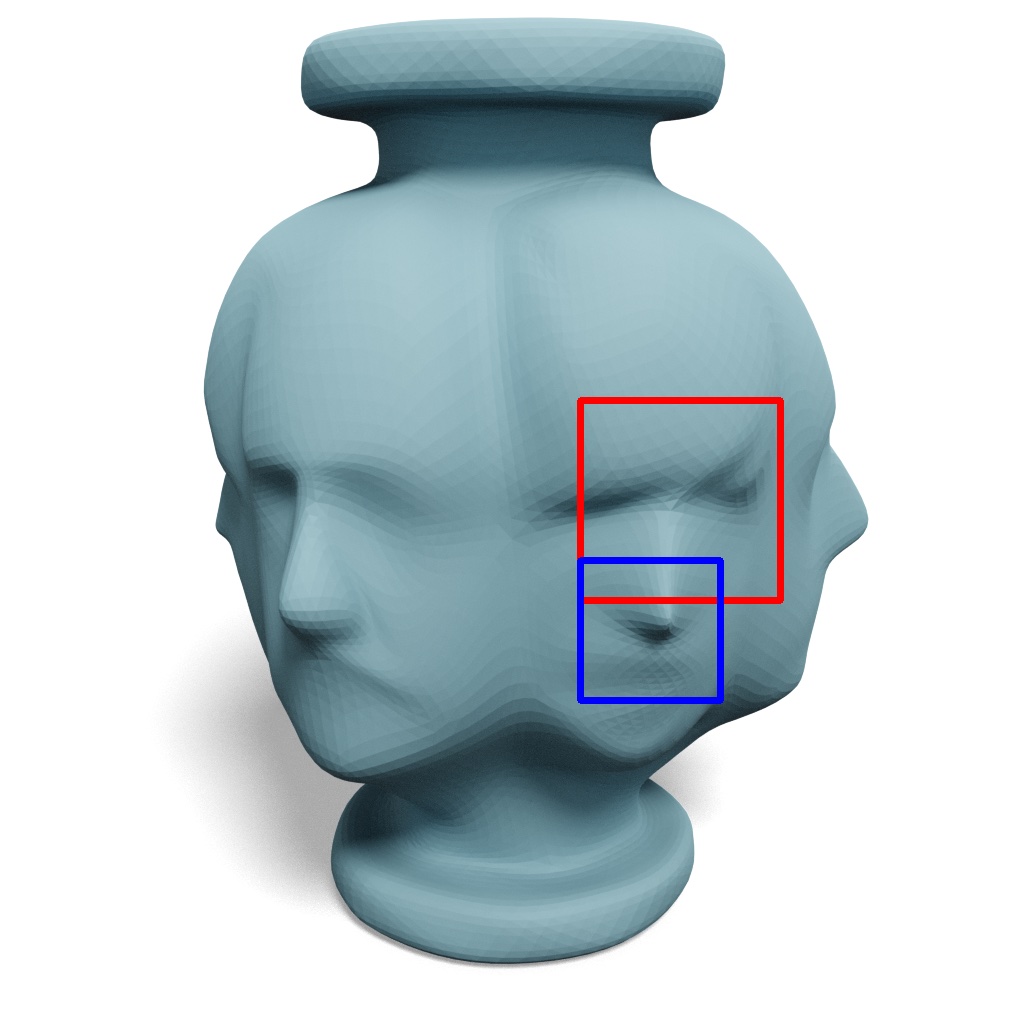}} \hfill
  \mpage{0.235}{\includegraphics[width=\linewidth]{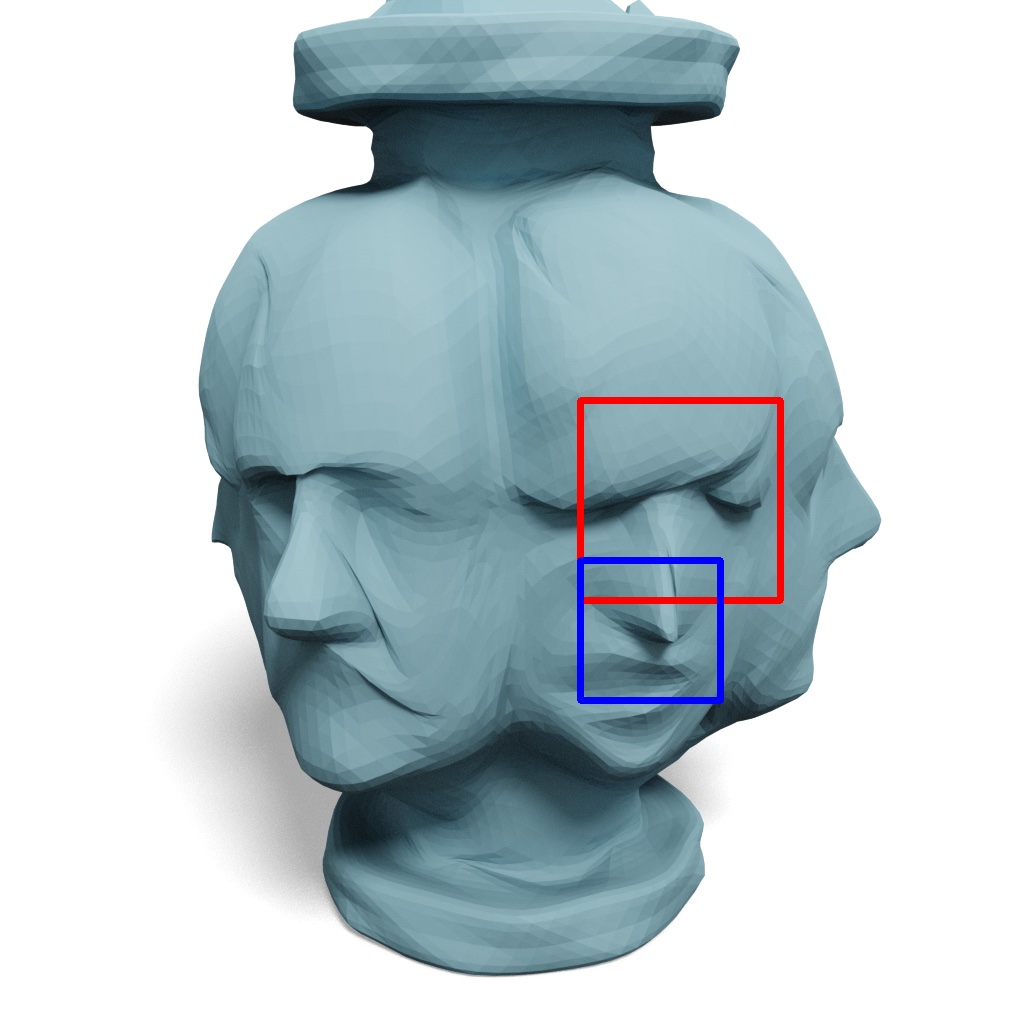}} \hfill
  \mpage{0.235}{\includegraphics[width=\linewidth]{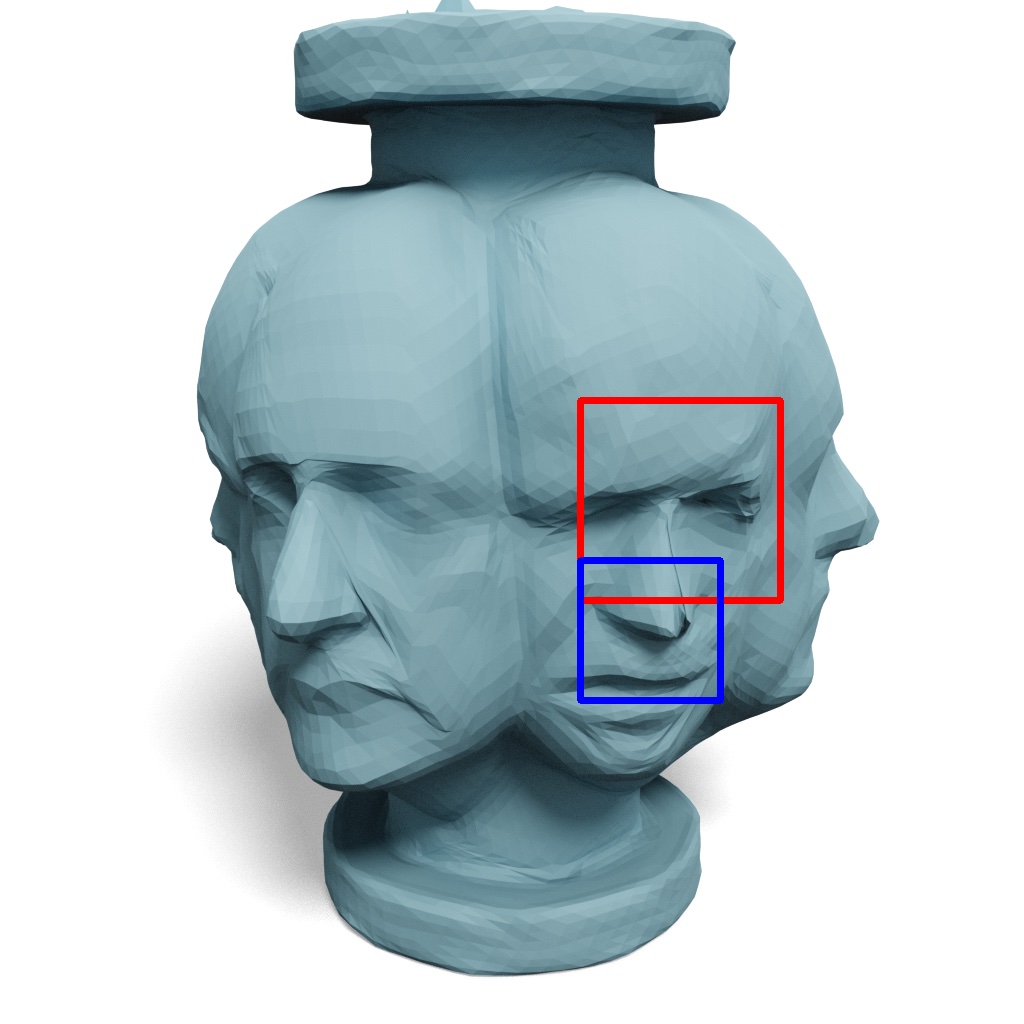}} \\
  \vspace{1.0mm}
  \mpage{0.235}{\includegraphics[width=0.475\linewidth]{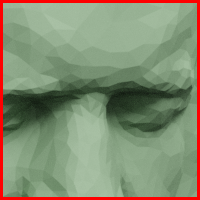} \hfill \includegraphics[width=0.475\linewidth]{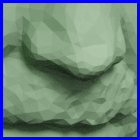}} \hfill
  \mpage{0.235}{\includegraphics[width=0.475\linewidth]{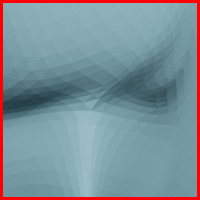} \hfill \includegraphics[width=0.475\linewidth]{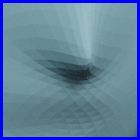}} \hfill
  \mpage{0.235}{\includegraphics[width=0.475\linewidth]{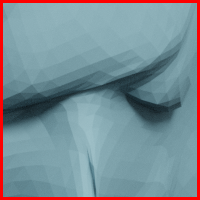} \hfill \includegraphics[width=0.475\linewidth]{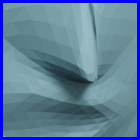}} \hfill
  \mpage{0.235}{\includegraphics[width=0.475\linewidth]{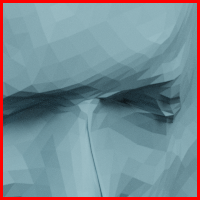} \hfill \includegraphics[width=0.475\linewidth]{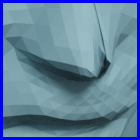}} \\
  \vspace{1.0mm}
  \mpage{0.235}{$CR$ / $d_\text{pm}$ ($\times 10^{-4}$) / $d_\text{normal}$} \hfill
  \mpage{0.235}{58.48 / 30.57 / 13.89$^\circ$} \hfill
  \mpage{0.235}{38.14 / 22.89 / 12.05$^\circ$} \hfill
  \mpage{0.235}{9.23 /  8.96  / 8.25$^\circ$} \\
  \vspace{1.0mm}
  \mpage{0.235}{\includegraphics[width=\linewidth]{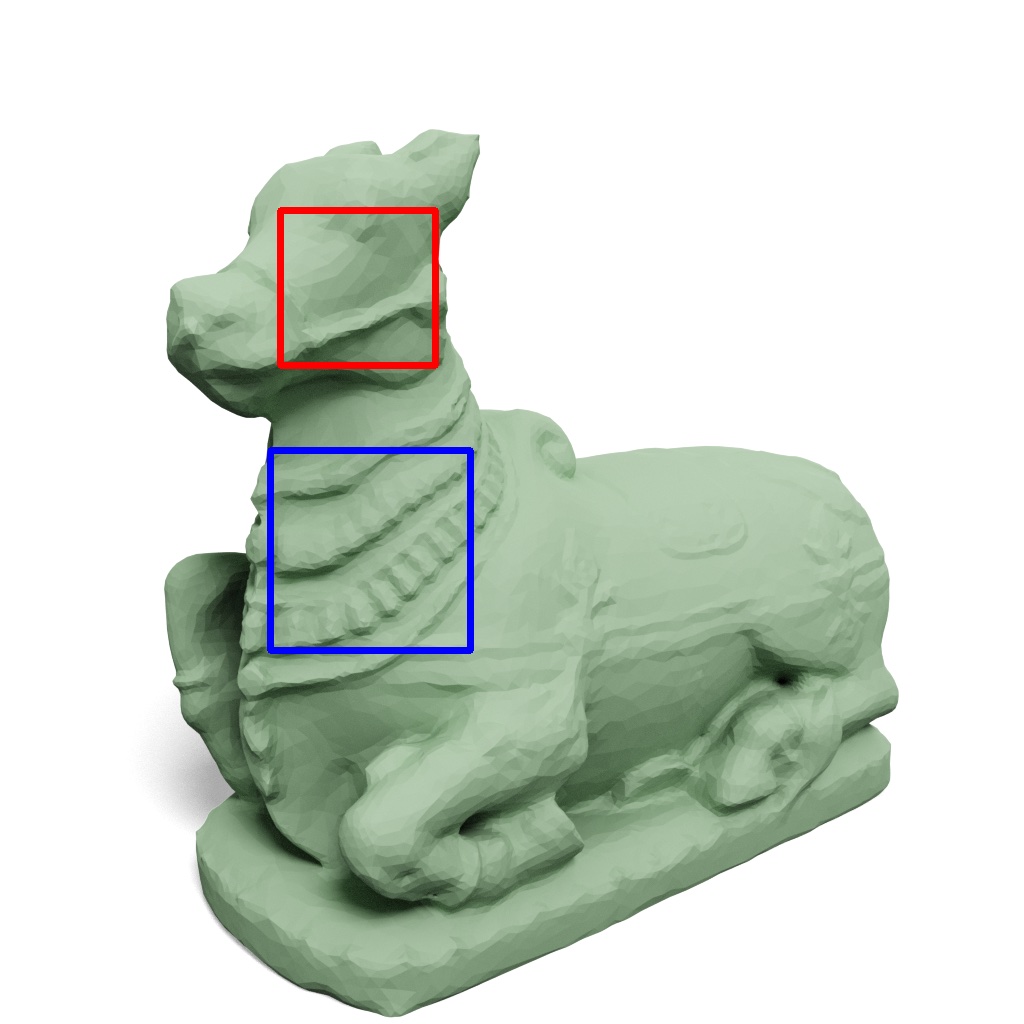}} \hfill
  \mpage{0.235}{\includegraphics[width=\linewidth]{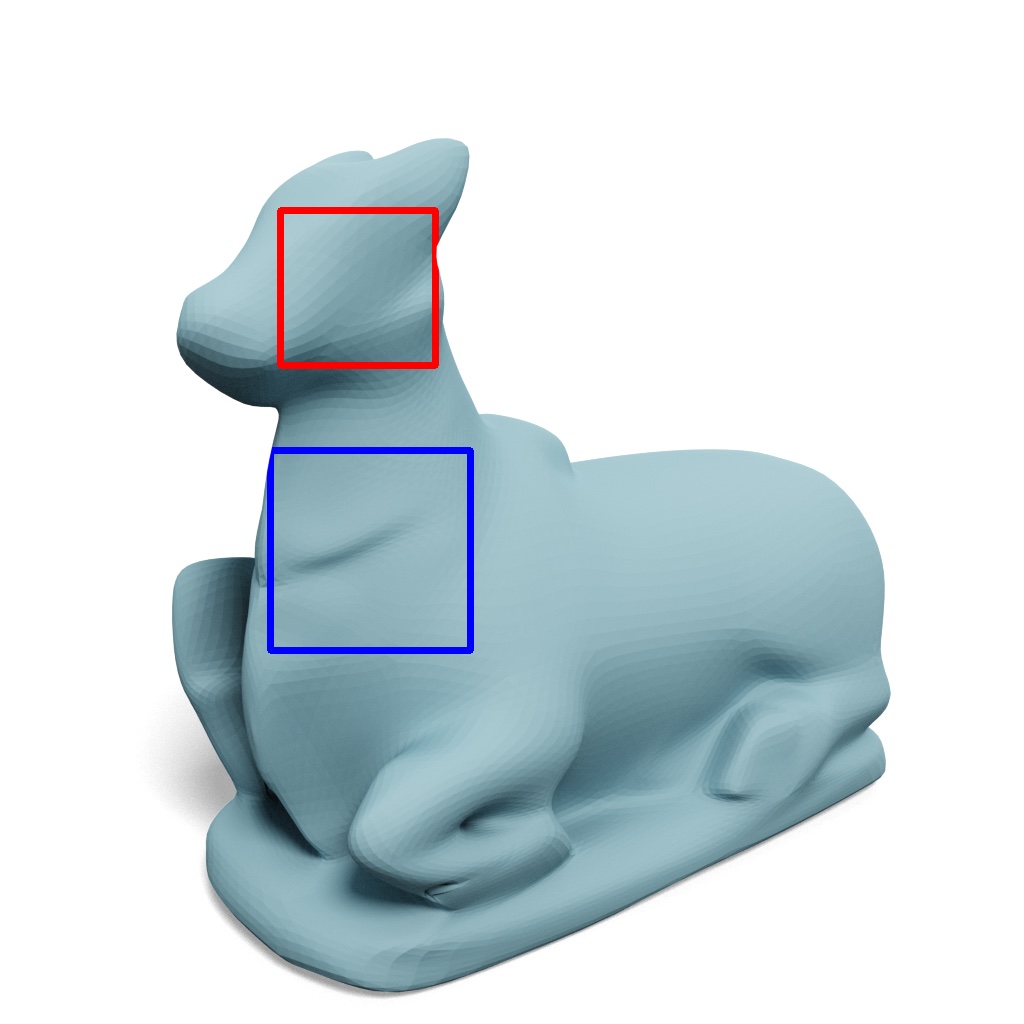}} \hfill
  \mpage{0.235}{\includegraphics[width=\linewidth]{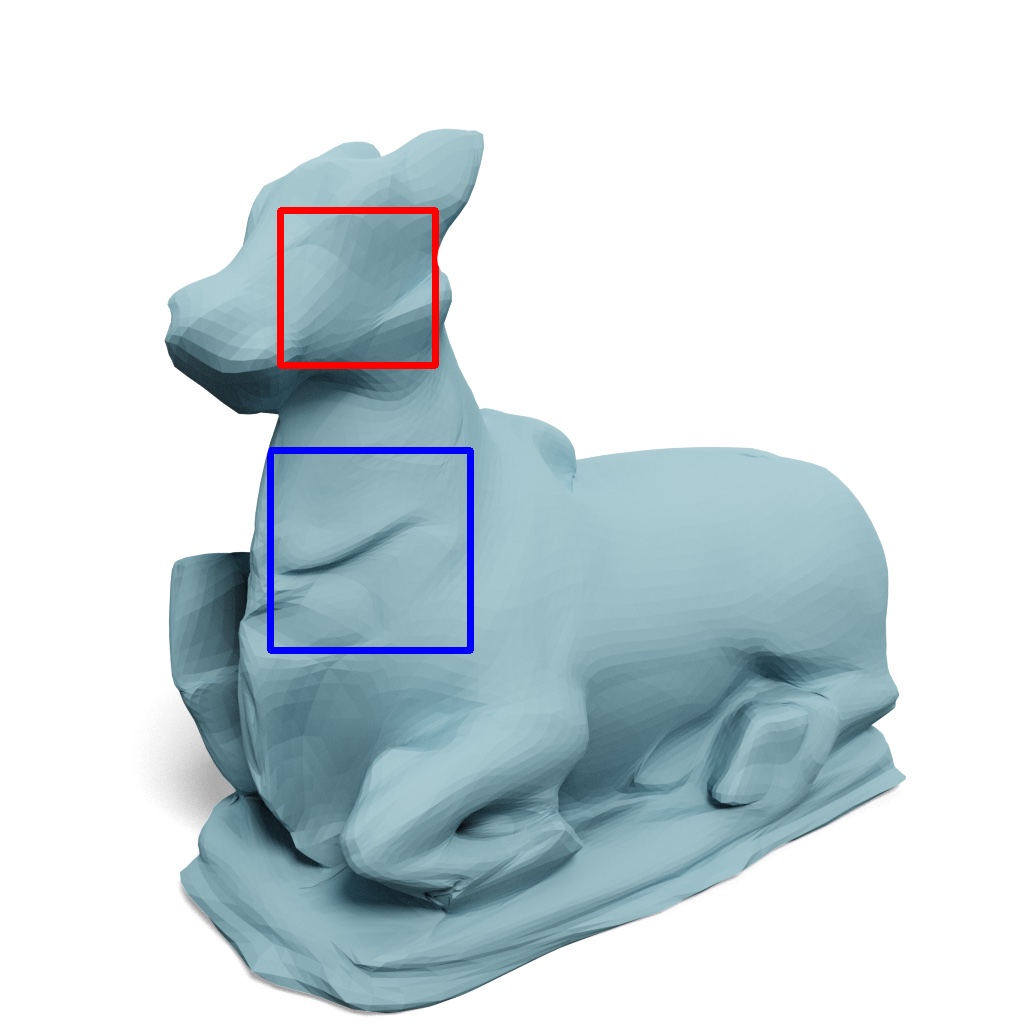}} \hfill
  \mpage{0.235}{\includegraphics[width=\linewidth]{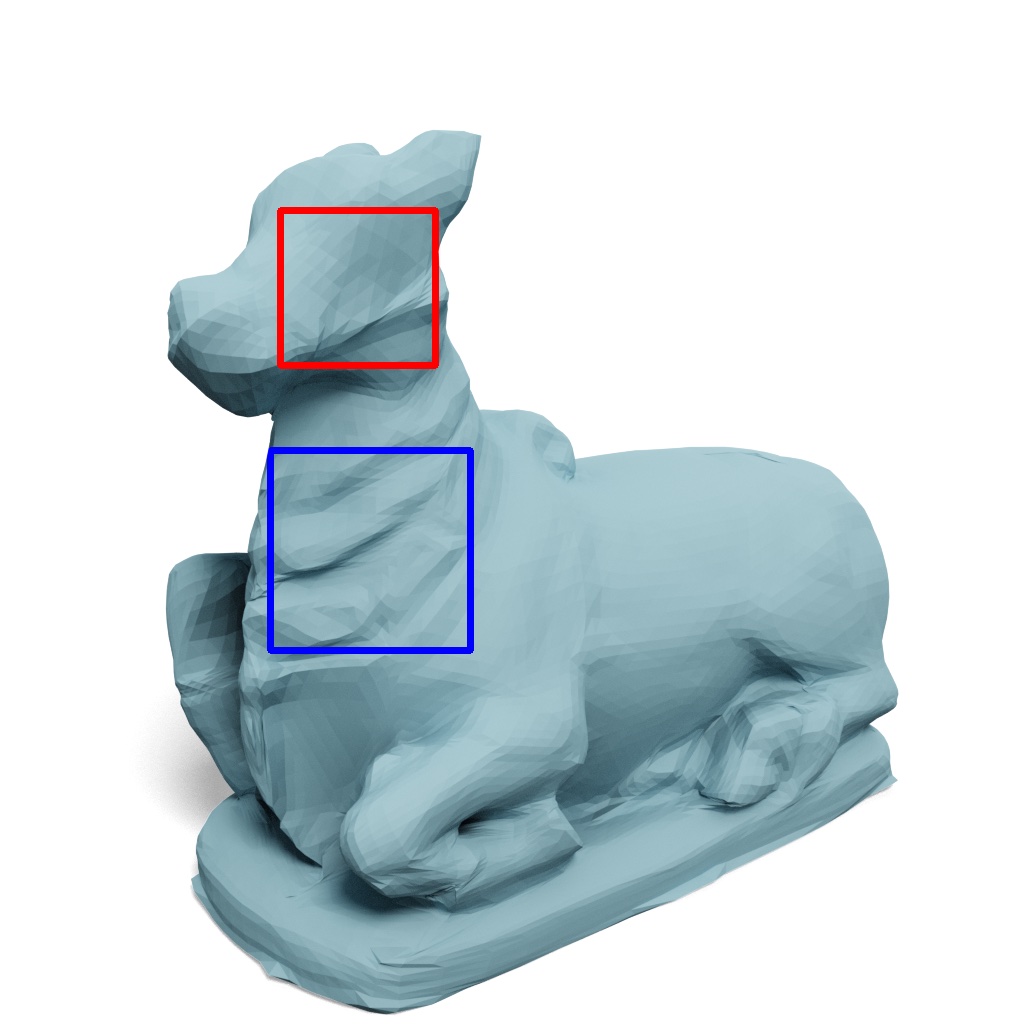}} \\
  \vspace{1.0mm}
  \mpage{0.235}{\includegraphics[width=0.475\linewidth]{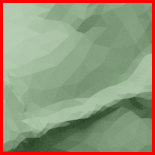} \hfill \includegraphics[width=0.475\linewidth]{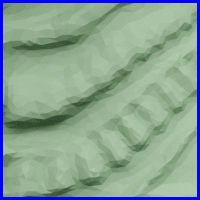}} \hfill
  \mpage{0.235}{\includegraphics[width=0.475\linewidth]{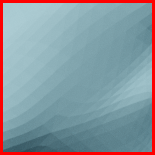} \hfill \includegraphics[width=0.475\linewidth]{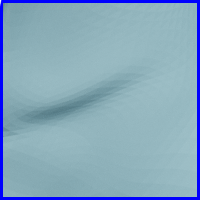}} \hfill
  \mpage{0.235}{\includegraphics[width=0.475\linewidth]{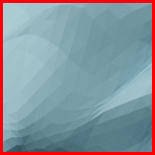} \hfill \includegraphics[width=0.475\linewidth]{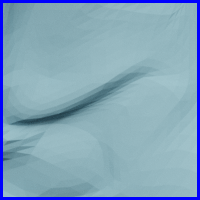}} \hfill
  \mpage{0.235}{\includegraphics[width=0.475\linewidth]{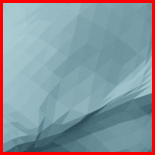} \hfill \includegraphics[width=0.475\linewidth]{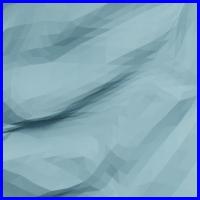}} \\
  \vspace{1.0mm}
  \mpage{0.235}{$CR$ / $d_\text{pm}$ ($\times 10^{-4}$) / $d_\text{normal}$} \hfill
  \mpage{0.235}{92.03 / 27.04 / 14.01$^\circ$} \hfill
  \mpage{0.235}{59.60 / 14.08 / 12.59$^\circ$} \hfill
  \mpage{0.235}{14.28 / 6.21 / 10.09$^\circ$} \\
  \vspace{1.0mm}
  \mpage{0.235}{\includegraphics[width=\linewidth]{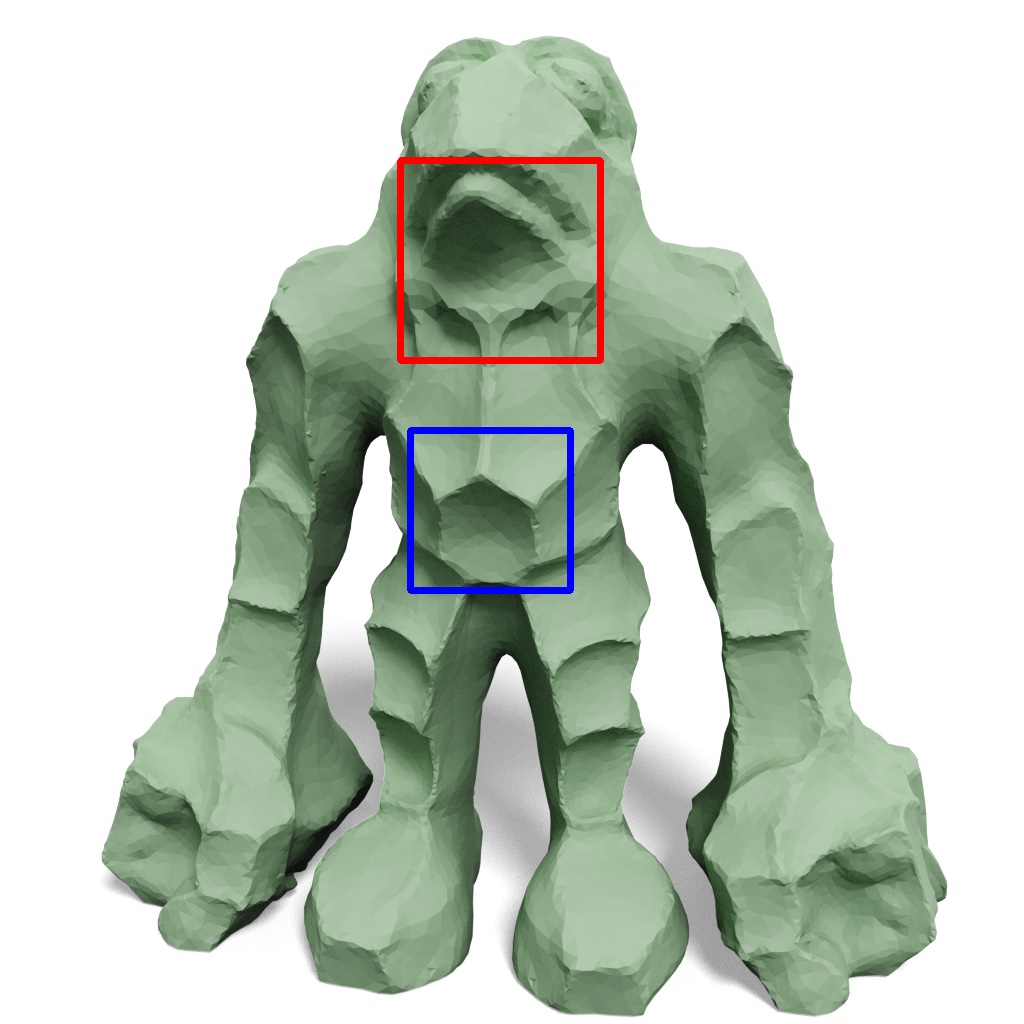}} \hfill
  \mpage{0.235}{\includegraphics[width=\linewidth]{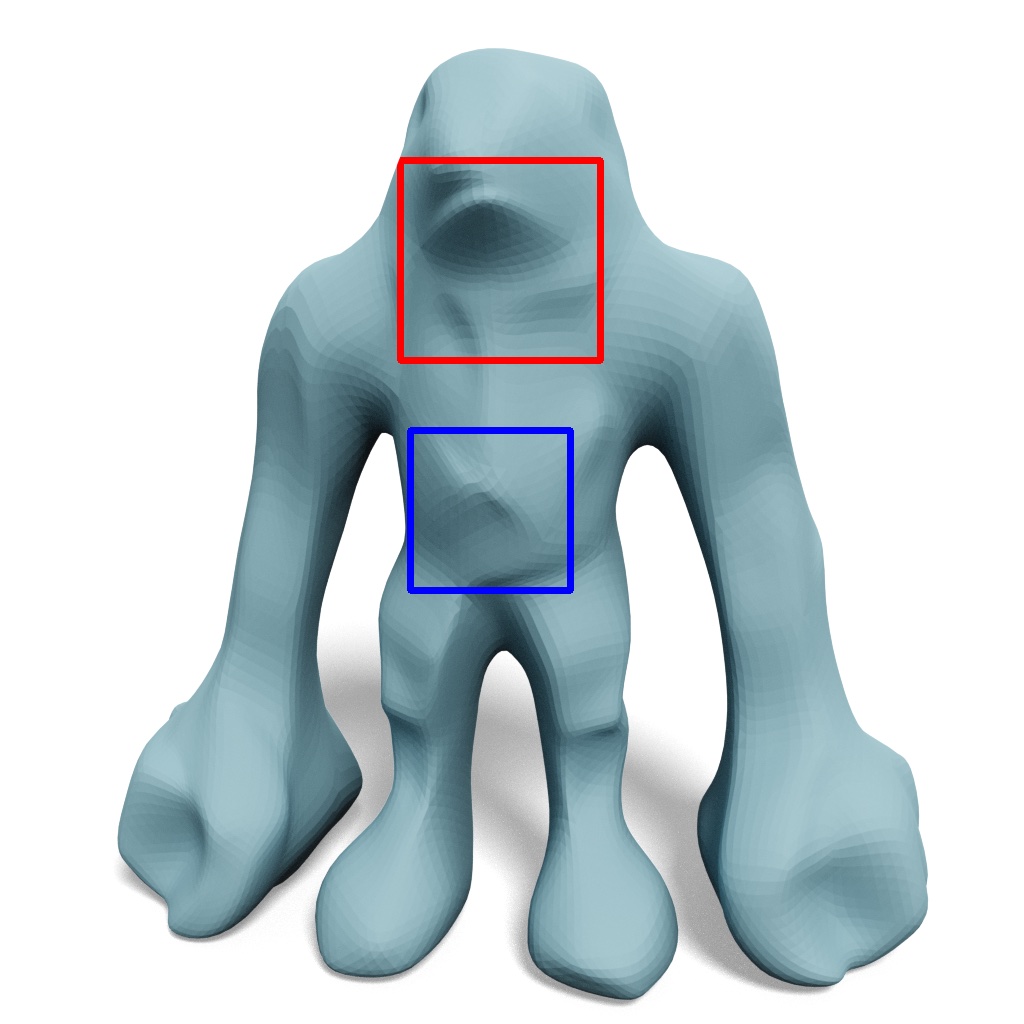}} \hfill
  \mpage{0.235}{\includegraphics[width=\linewidth]{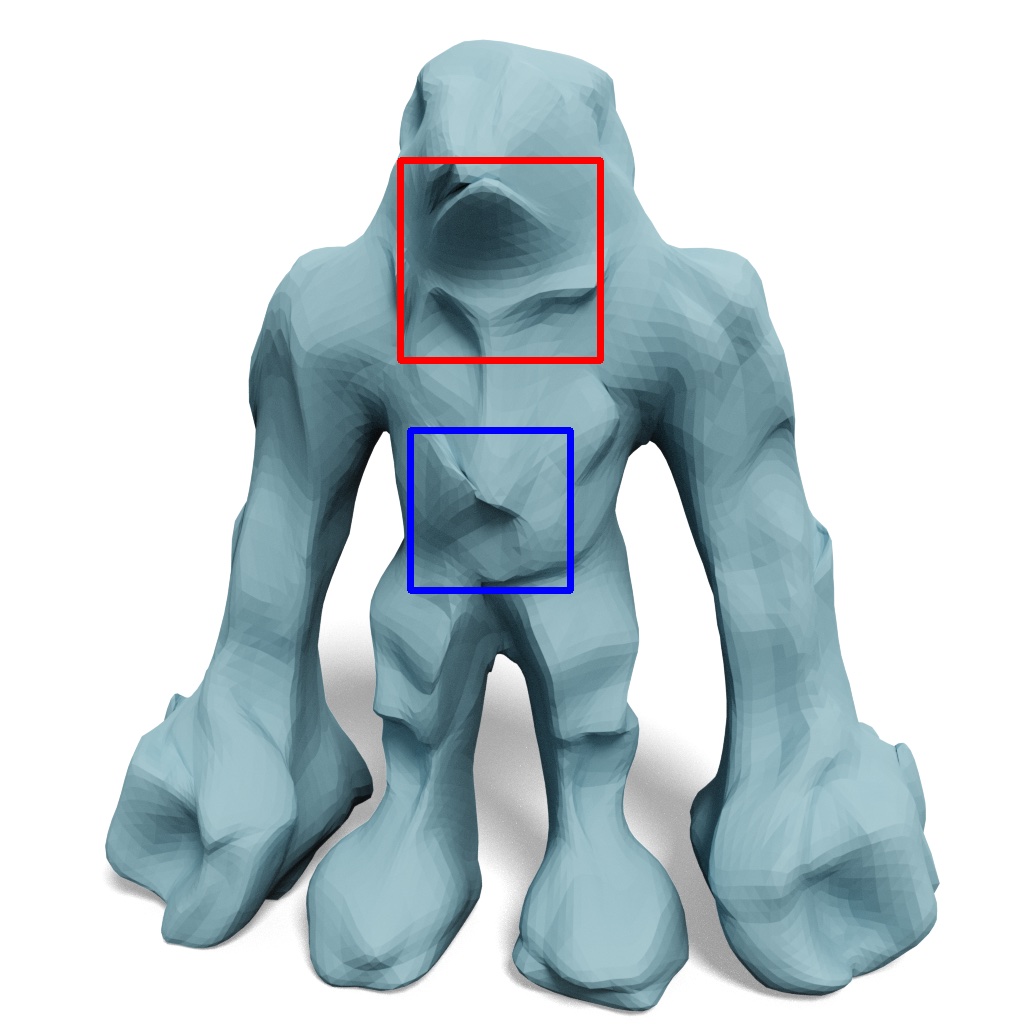}} \hfill
  \mpage{0.235}{\includegraphics[width=\linewidth]{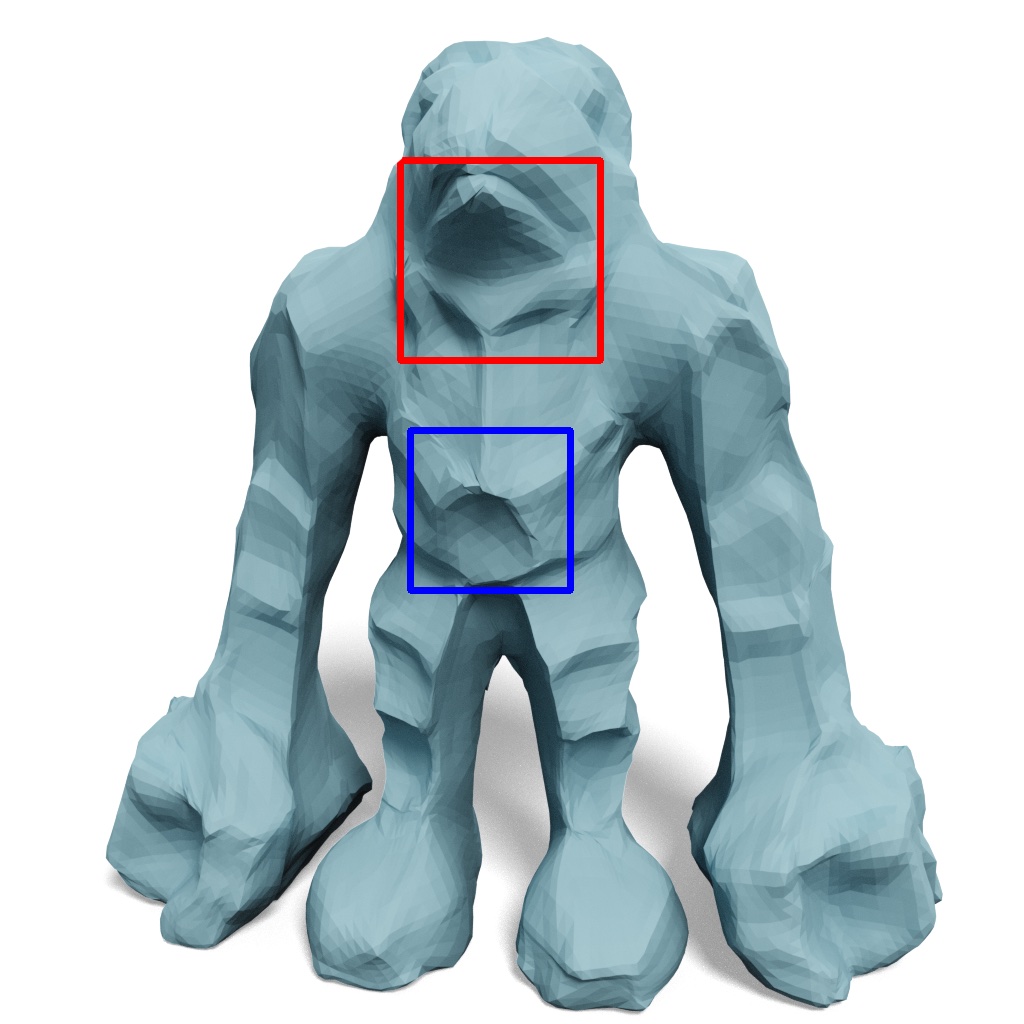}} \\
  \vspace{1.0mm}
  \mpage{0.235}{\includegraphics[width=0.475\linewidth]{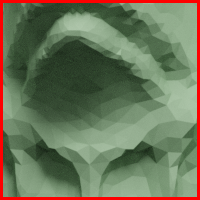} \hfill \includegraphics[width=0.475\linewidth]{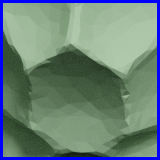}} \hfill
  \mpage{0.235}{\includegraphics[width=0.475\linewidth]{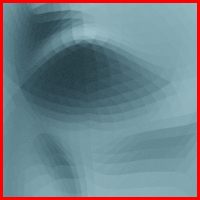} \hfill \includegraphics[width=0.475\linewidth]{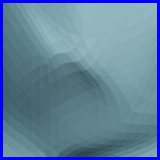}} \hfill
  \mpage{0.235}{\includegraphics[width=0.475\linewidth]{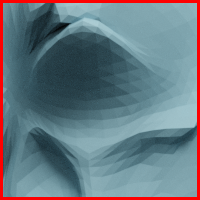} \hfill \includegraphics[width=0.475\linewidth]{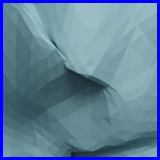}} \hfill
  \mpage{0.235}{\includegraphics[width=0.475\linewidth]{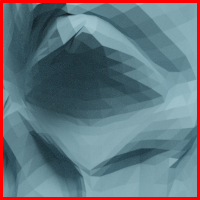} \hfill \includegraphics[width=0.475\linewidth]{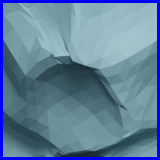}} \\
  \vspace{1.0mm}
  \mpage{0.235}{$CR$ / $d_\text{pm}$ ($\times 10^{-4}$) / $d_\text{normal}$} \hfill
  \mpage{0.235}{82.63 / 29.67 / 18.57$^\circ$} \hfill
  \mpage{0.235}{54.07 / 20.48 / 15.64$^\circ$} \hfill
  \mpage{0.235}{13.16 / 7.27 / 12.22$^\circ$} \\
  \vspace{1.0mm}
  \mpage{0.235}{Ground truth} \hfill
  \mpage{0.235}{Ours w/o features} \hfill
  \mpage{0.235}{Ours + 40 features} \hfill
  \mpage{0.235}{Ours + 400 features} \\
  \caption{
  \textbf{Progressive features.} 
  We show more examples where transmitting more features leads to better quantitative and qualitative results.
  }
  \vspace{-3.0mm}
  \label{supp-fig:prog-feat-3}
\end{figure*}

Figures~\ref{supp-fig:prog-feat-1}, \ref{supp-fig:prog-feat-2}, \ref{supp-fig:prog-feat-3} and~\ref{supp-fig:prog-feat-4} show ten visual results of our method.

\section{Failure Case}

\begin{figure*}[t]
  \centering
  \mpage{0.235}{\includegraphics[width=\linewidth]{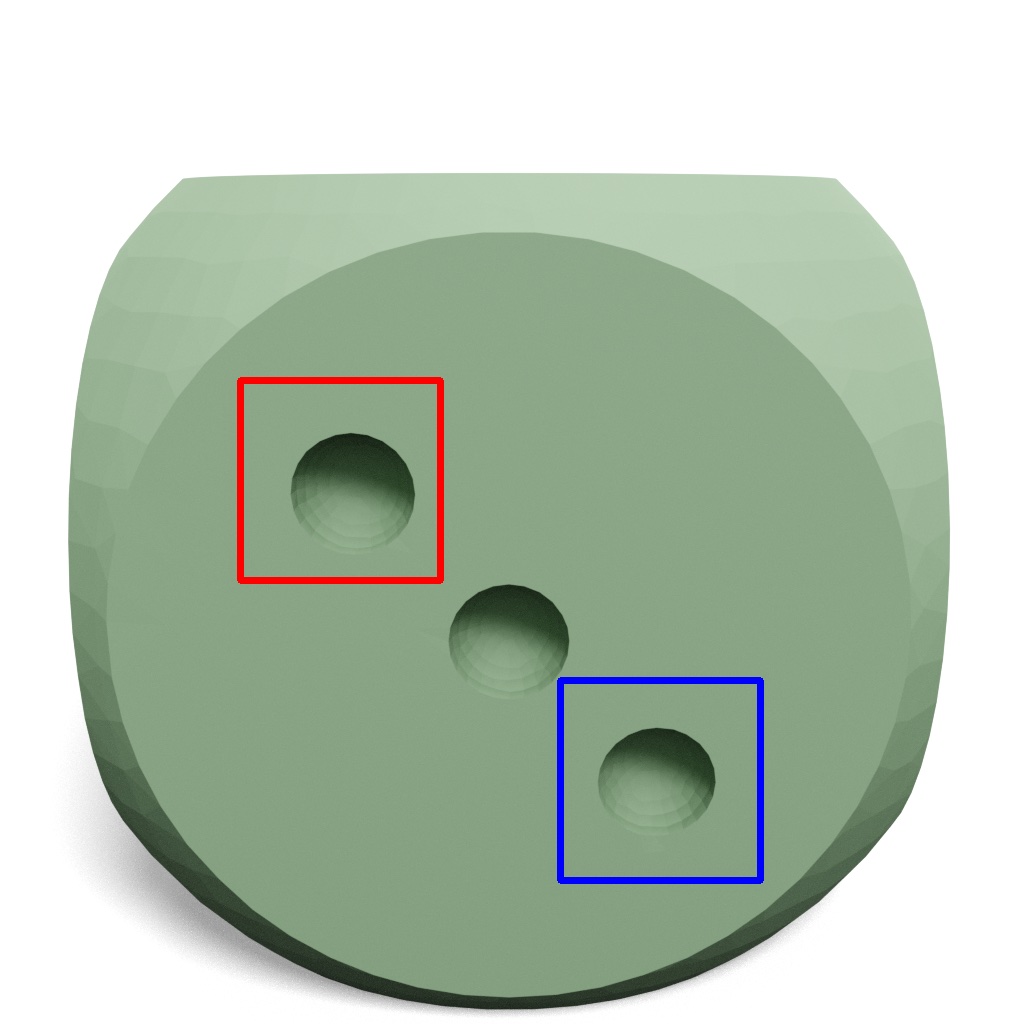}} \hfill
  \mpage{0.235}{\includegraphics[width=\linewidth]{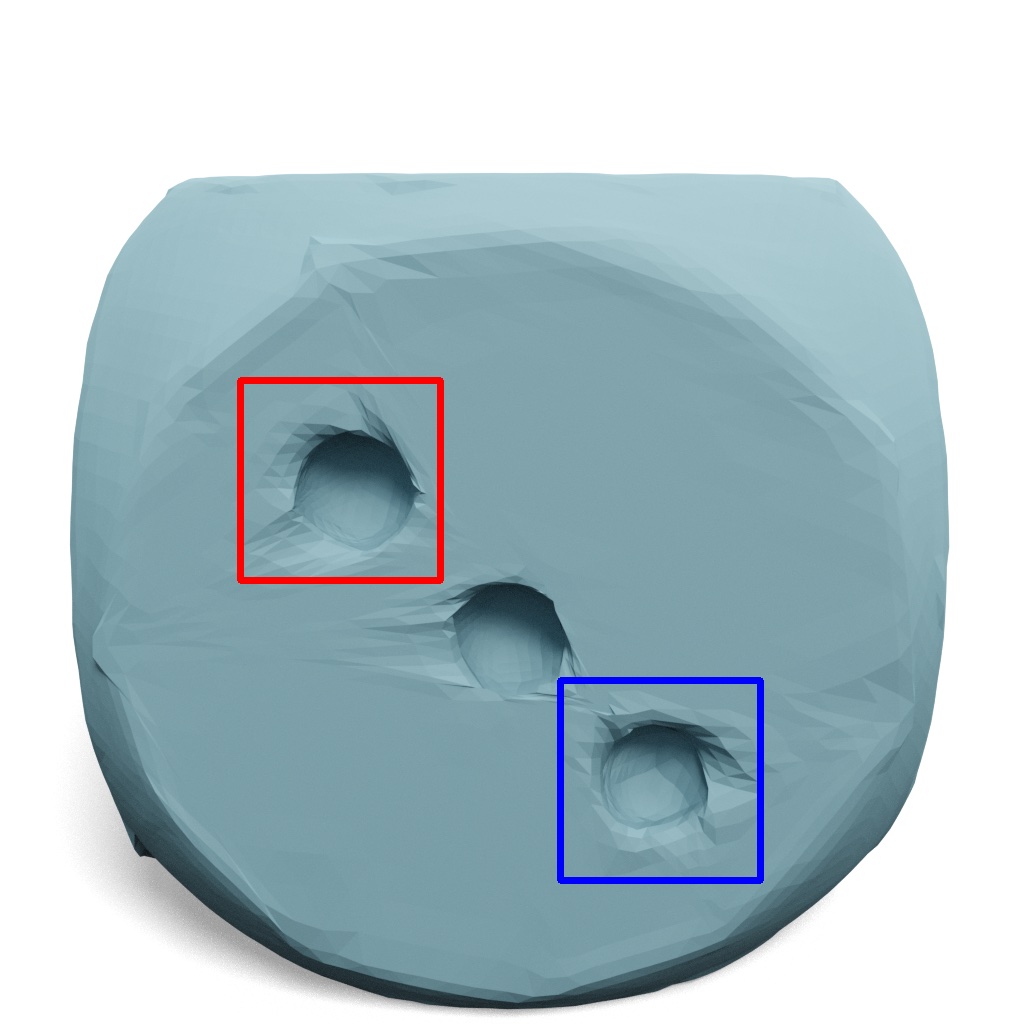}} \hfill
  \mpage{0.235}{\includegraphics[width=\linewidth]{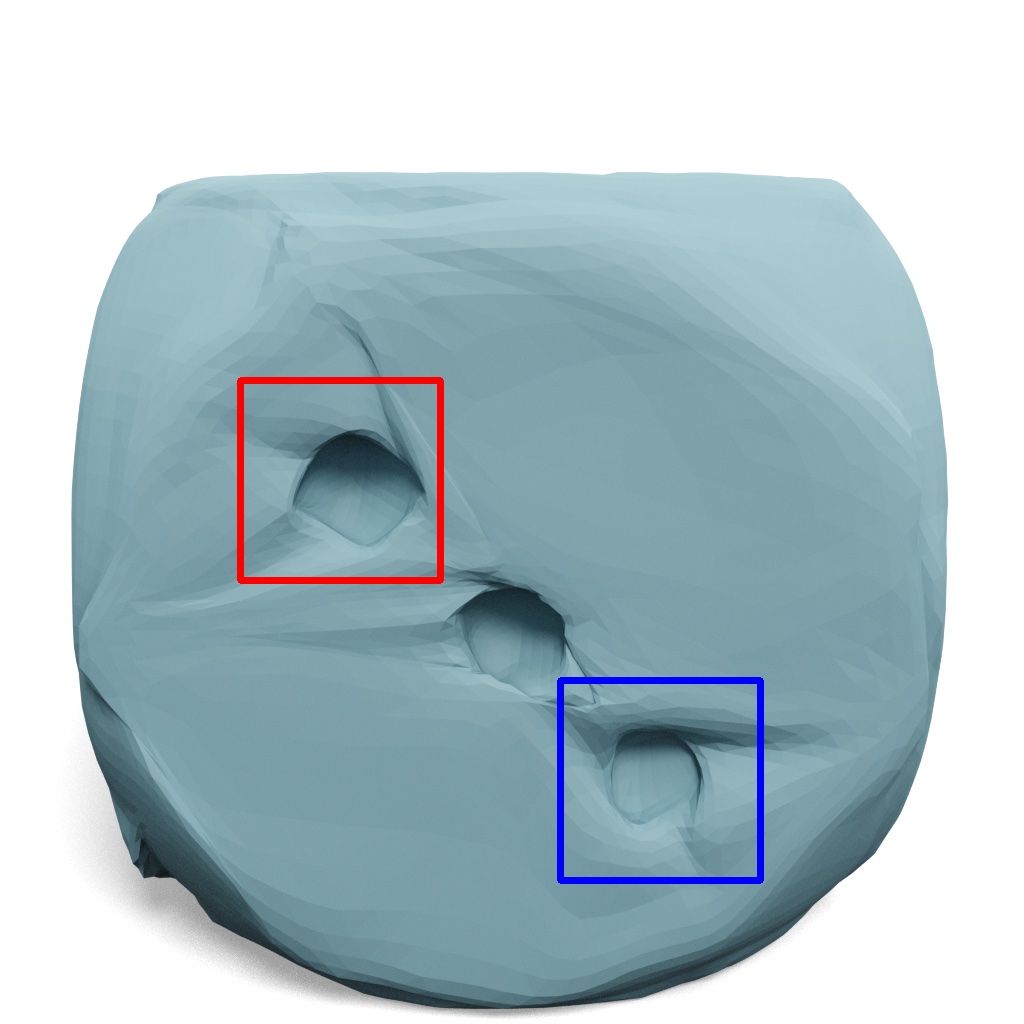}} \hfill
  \mpage{0.235}{\includegraphics[width=\linewidth]{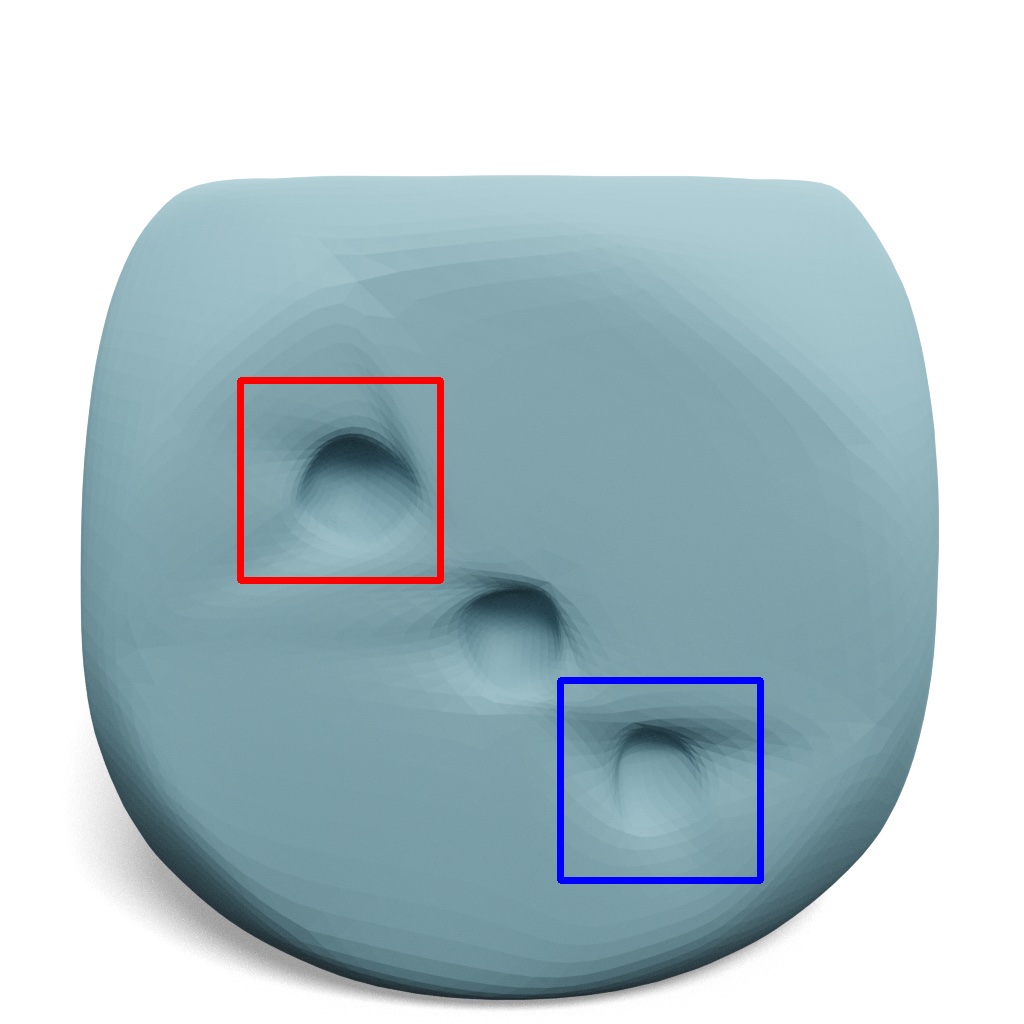}} \\
  \vspace{1.0mm}
  \mpage{0.235}{\includegraphics[width=0.475\linewidth]{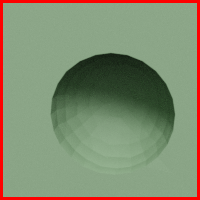} \hfill \includegraphics[width=0.475\linewidth]{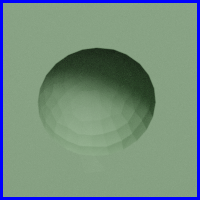}} \hfill
  \mpage{0.235}{\includegraphics[width=0.475\linewidth]{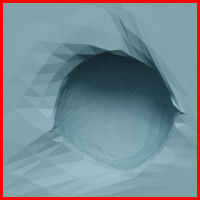} \hfill \includegraphics[width=0.475\linewidth]{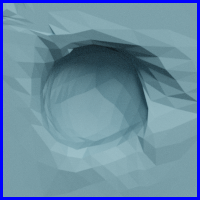}} \hfill
  \mpage{0.235}{\includegraphics[width=0.475\linewidth]{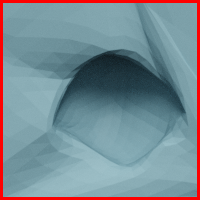} \hfill \includegraphics[width=0.475\linewidth]{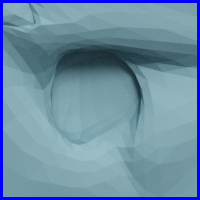}} \hfill
  \mpage{0.235}{\includegraphics[width=0.475\linewidth]{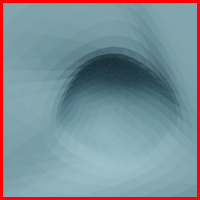} \hfill \includegraphics[width=0.475\linewidth]{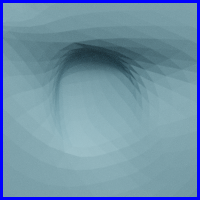}} \\
  \vspace{1.0mm}
  \mpage{0.235}{$CR$ / $d_\text{pm}$ ($\times 10^{-4}$) / $d_\text{normal}$} \hfill
  \mpage{0.235}{19.72 / 24.17 / 9.53$^\circ$} \hfill
  \mpage{0.235}{12.90 / 12.62 / 7.84$^\circ$} \hfill
  \mpage{0.235}{3.14 /  6.39 /  5.43$^\circ$} \\
  \vspace{1.0mm}
  \mpage{0.235}{Ground truth} \hfill
  \mpage{0.235}{Ours w/o features} \hfill
  \mpage{0.235}{Ours + 40 features} \hfill
  \mpage{0.235}{Ours + 400 features} \\
  \caption{
  \textbf{Progressive features.} 
  We show more examples where transmitting more features leads to better quantitative and qualitative results.
  }
  \label{supp-fig:prog-feat-4}
  \vspace{5.0mm}
  \mpage{0.31}{\includegraphics[width=\linewidth]{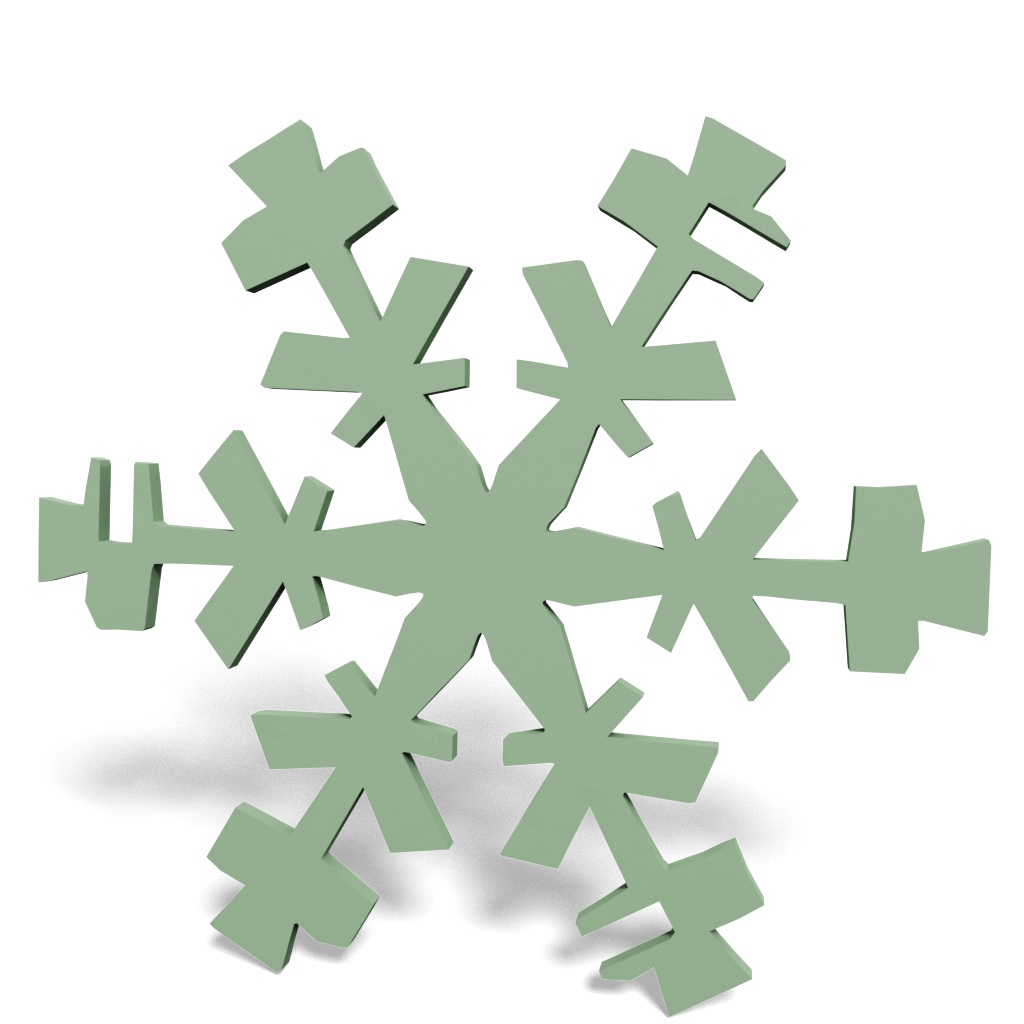}} \hfill
  \mpage{0.31}{\includegraphics[width=\linewidth]{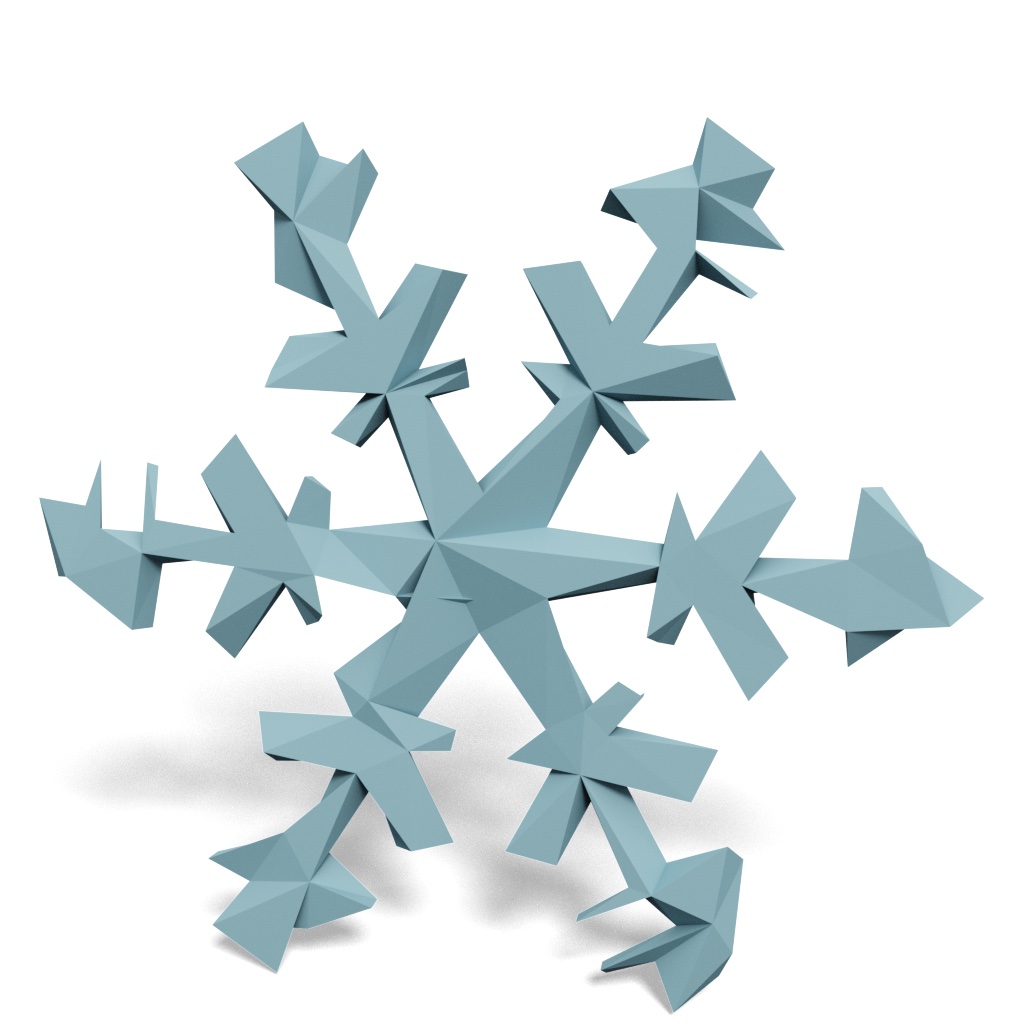}} \hfill
  \mpage{0.31}{\includegraphics[width=\linewidth]{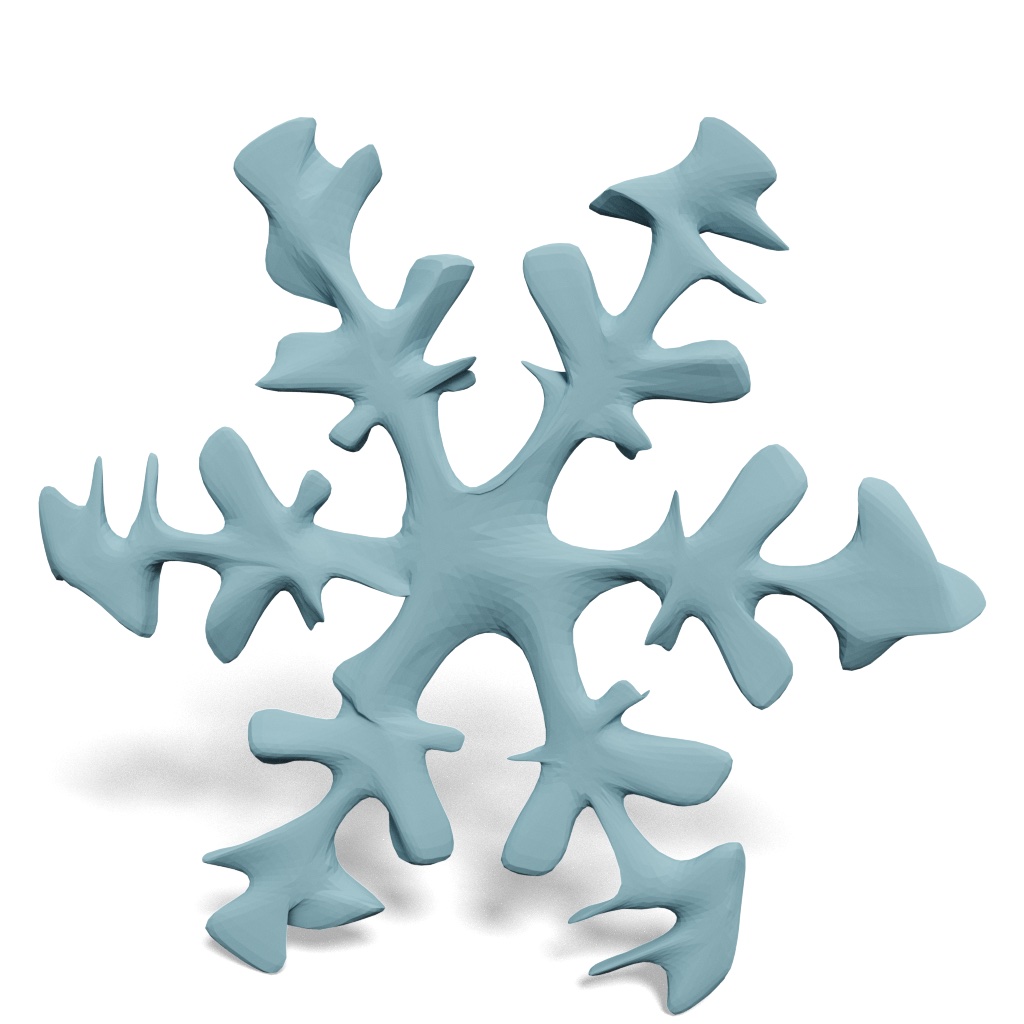}} \\
  \vspace{1.0mm}
  \mpage{0.31}{$CR$ / $d_\text{pm}$ ($\times 10^{-4}$) / $d_\text{normal}$} \hfill
  \mpage{0.31}{10.68 / 2.52 / 35.66$^\circ$} \hfill
  \mpage{0.31}{10.68 / 2.43 / 31.29$^\circ$} \\
  \vspace{1.0mm}
  \mpage{0.31}{Ground truth} \hfill
  \mpage{0.31}{QSlim} \hfill
  \mpage{0.31}{Ours} \\
  \caption{
  \textbf{Failure case.} 
  Our method fails when applied to a shape with complex topological details and intricate thin features.
  }
  \label{supp-fig:exp-failure}
\end{figure*}

Figure~\ref{supp-fig:exp-failure} shows a failure case of our method. 
Our method fails when applied to a shape with complex topological details and intricate thin features. 

\clearpage 

\bibliographystyle{ACM-Reference-Format}
\bibliography{references.bib}

\clearpage

\begin{figure*}[!htb]
  \centering
  \mpage{0.132}{\includegraphics[width=\linewidth]{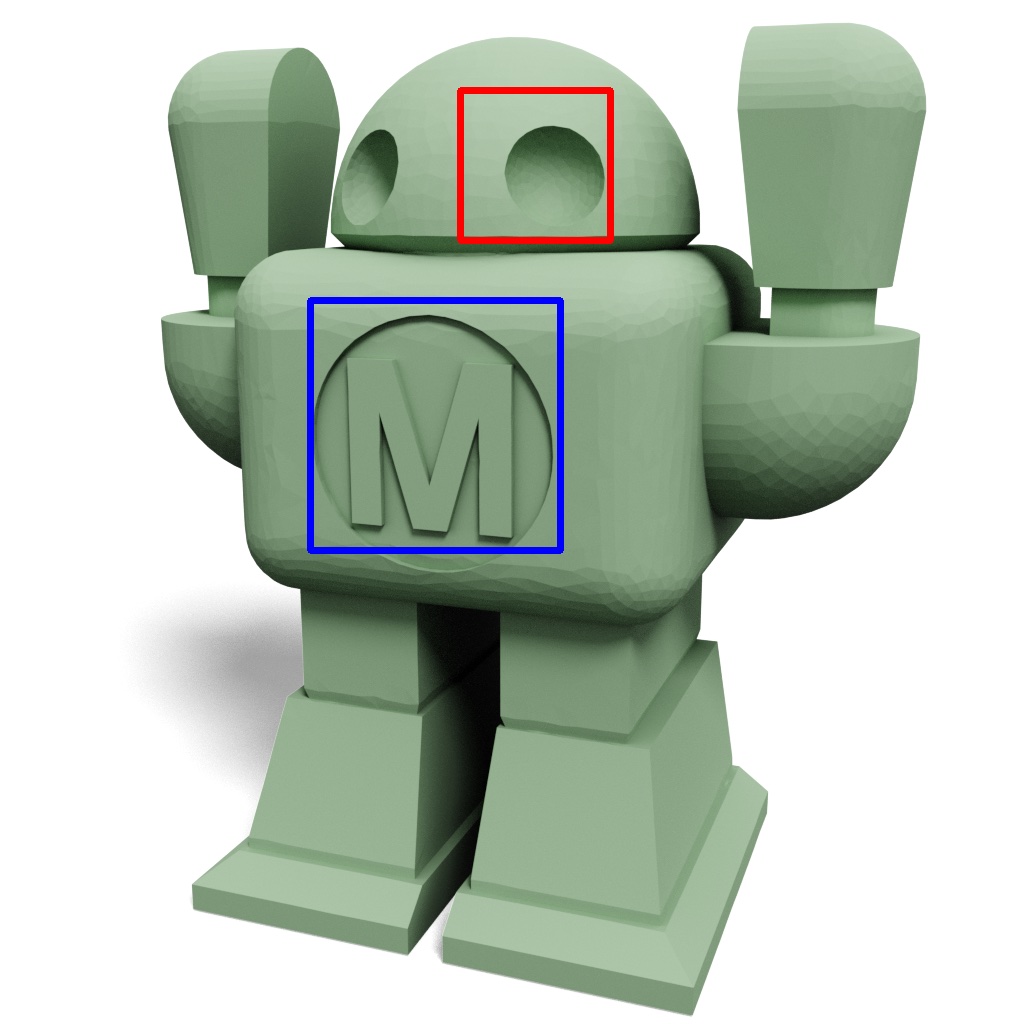}} \hfill
  \mpage{0.132}{\includegraphics[width=\linewidth]{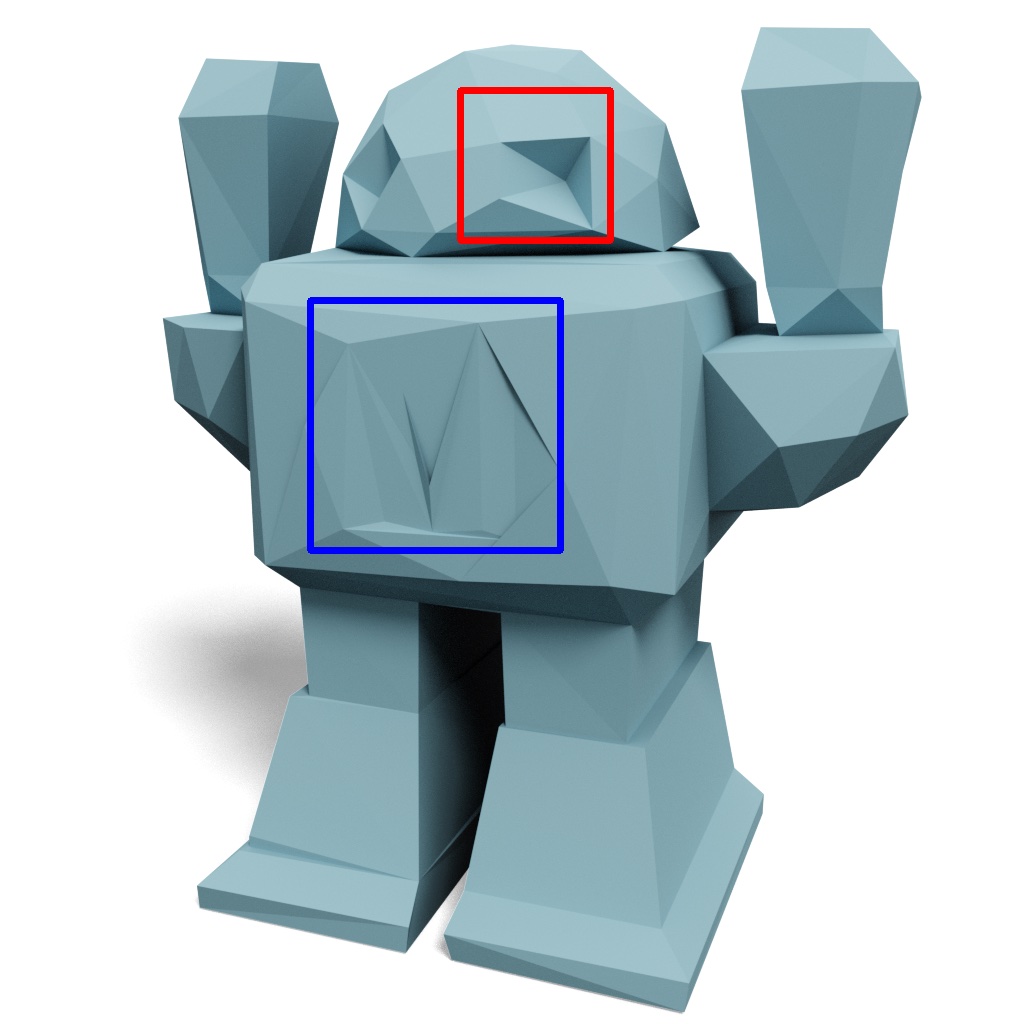}} \hfill
  \mpage{0.132}{\includegraphics[width=\linewidth]{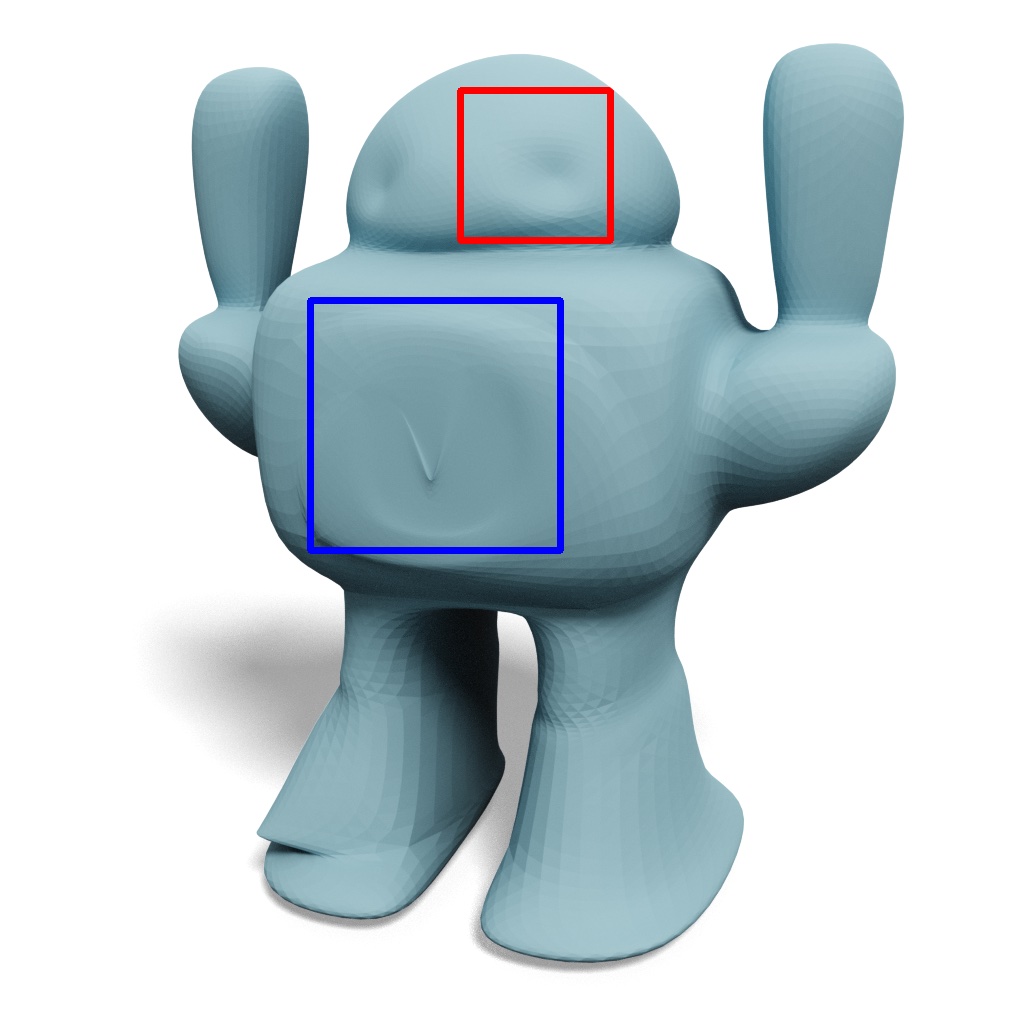}} \hfill
  \mpage{0.132}{\includegraphics[width=\linewidth]{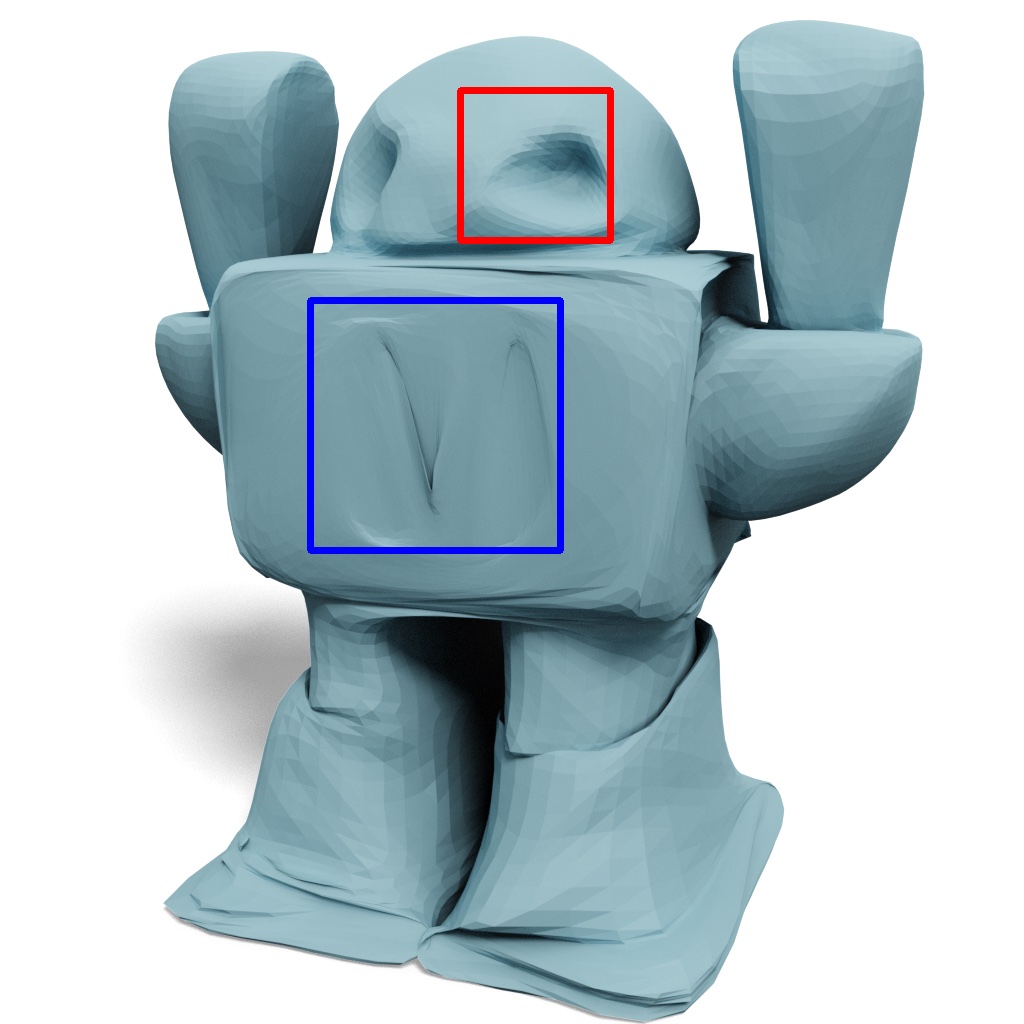}} \hfill
  \mpage{0.132}{\includegraphics[width=\linewidth]{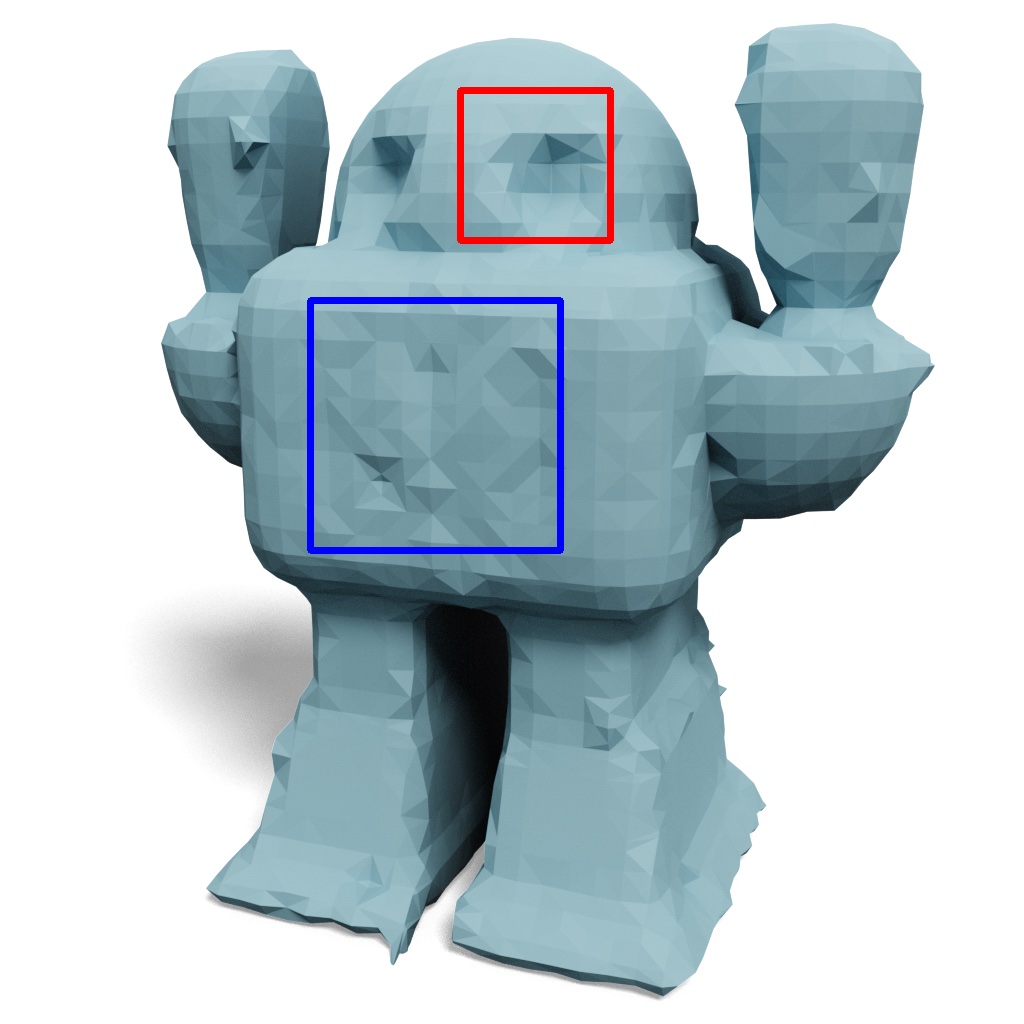}} \hfill
  \mpage{0.132}{\includegraphics[width=\linewidth]{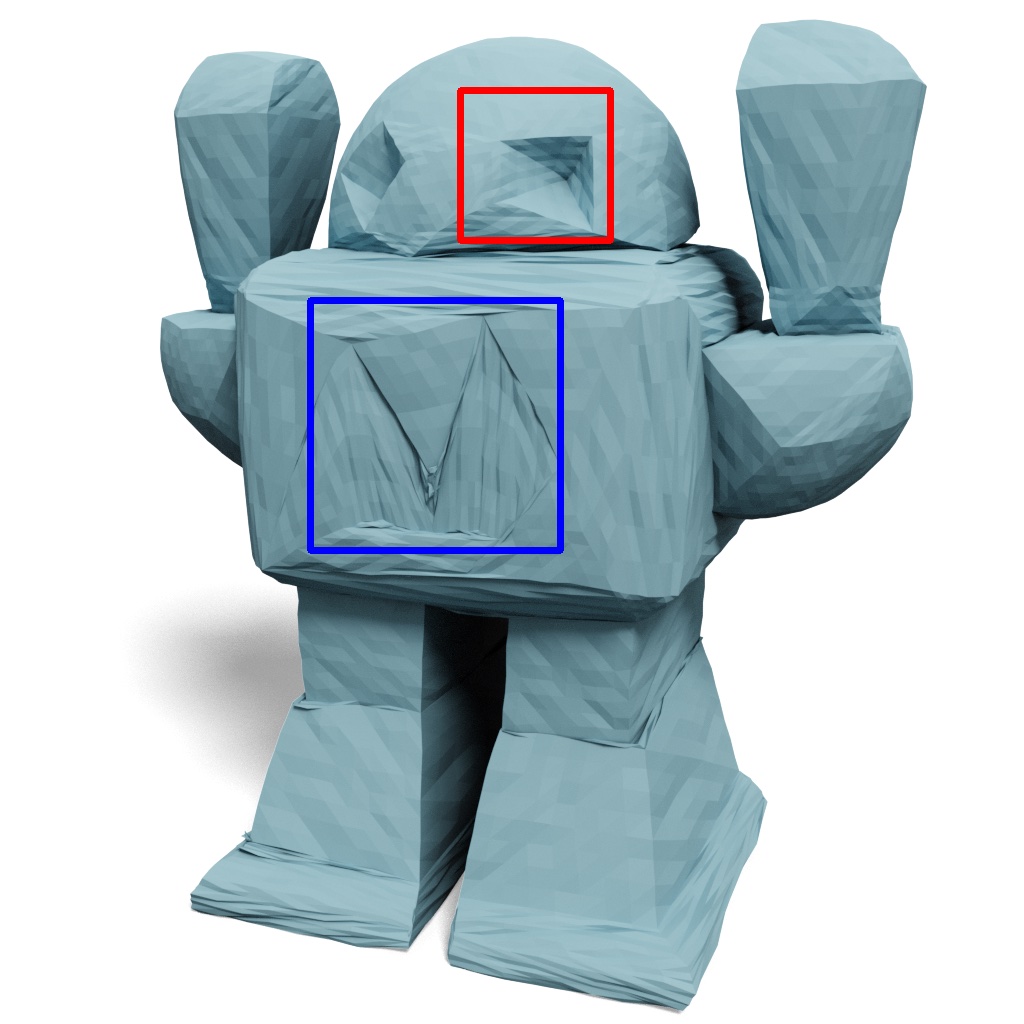}} \hfill
  \mpage{0.132}{\includegraphics[width=\linewidth]{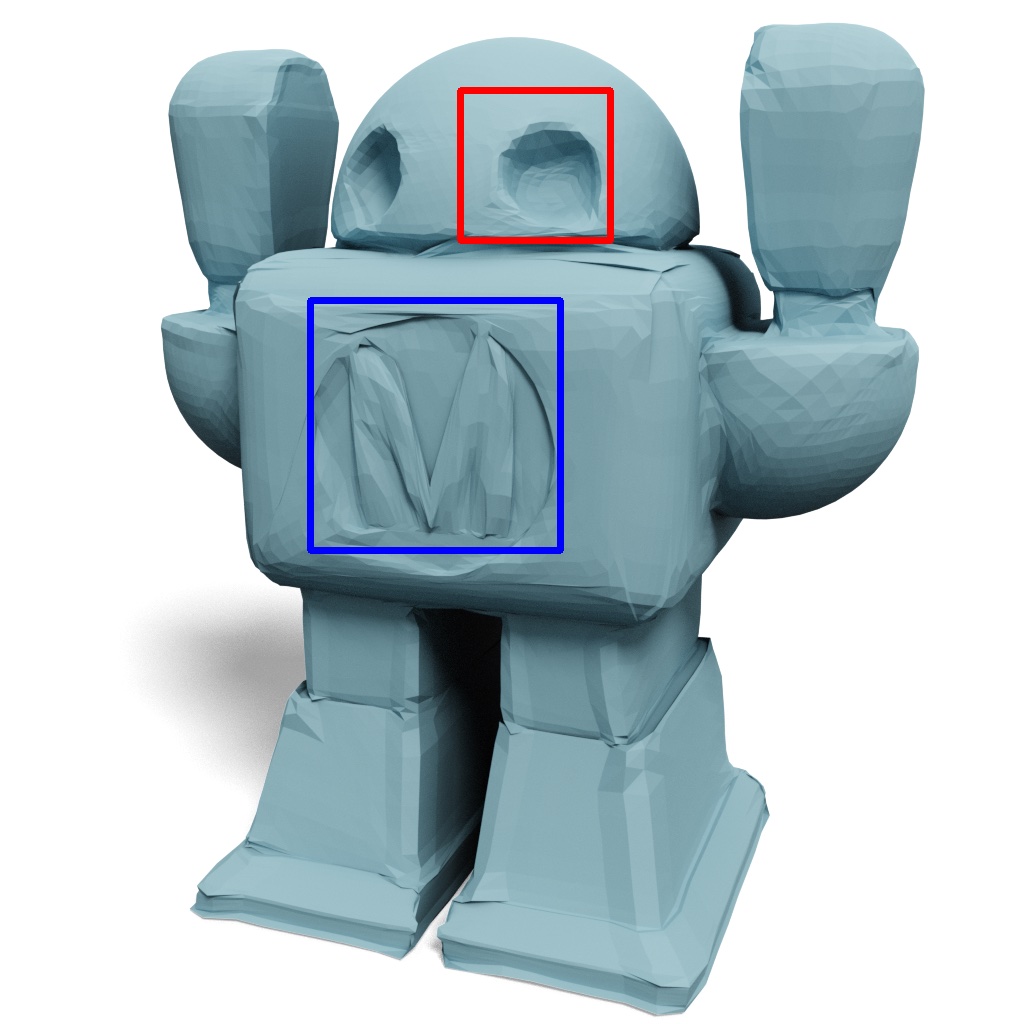}} \\
  \vspace{1.0mm}
  \mpage{0.132}{\includegraphics[width=0.475\linewidth]{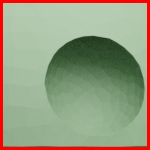} \hfill \includegraphics[width=0.475\linewidth]{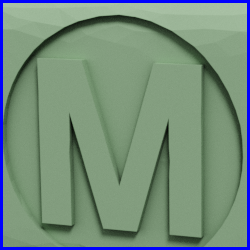}} \hfill
  \mpage{0.132}{\includegraphics[width=0.475\linewidth]{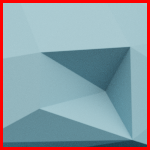} \hfill \includegraphics[width=0.475\linewidth]{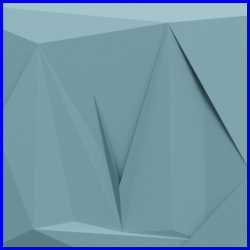}} \hfill
  \mpage{0.132}{\includegraphics[width=0.475\linewidth]{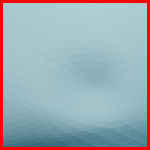} \hfill \includegraphics[width=0.475\linewidth]{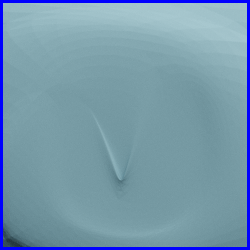}} \hfill
  \mpage{0.132}{\includegraphics[width=0.475\linewidth]{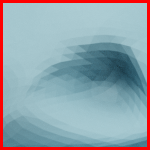} \hfill \includegraphics[width=0.475\linewidth]{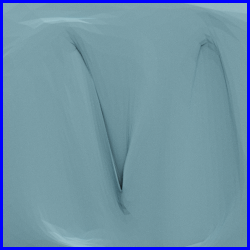}} \hfill 
  \mpage{0.132}{\includegraphics[width=0.475\linewidth]{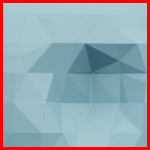} \hfill \includegraphics[width=0.475\linewidth]{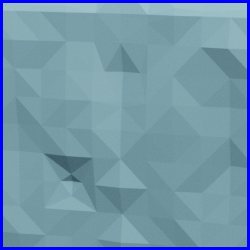}} \hfill
  \mpage{0.132}{\includegraphics[width=0.475\linewidth]{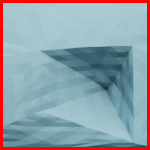} \hfill \includegraphics[width=0.475\linewidth]{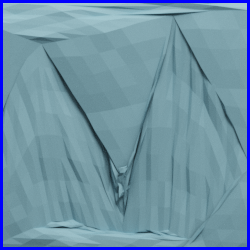}} \hfill
  \mpage{0.132}{\includegraphics[width=0.475\linewidth]{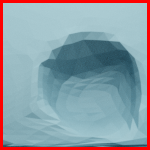} \hfill \includegraphics[width=0.475\linewidth]{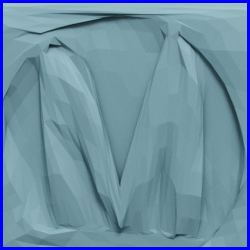}} \\
  \vspace{1.0mm}
  \mpage{0.132}{$CR = 19.90$} \hfill
  \mpage{0.132}{36.06 / 12.75$^\circ$} \hfill
  \mpage{0.132}{103.51 / 19.66$^\circ$} \hfill
  \mpage{0.132}{60.03 / 23.56$^\circ$} \hfill
  \mpage{0.132}{24.84 / 18.45$^\circ$} \hfill
  \mpage{0.132}{17.06 / 14.90$^\circ$} \hfill
  \mpage{0.132}{5.63 / 9.96$^\circ$} \\
  \vspace{1.0mm}
  \mpage{0.132}{\includegraphics[width=\linewidth]{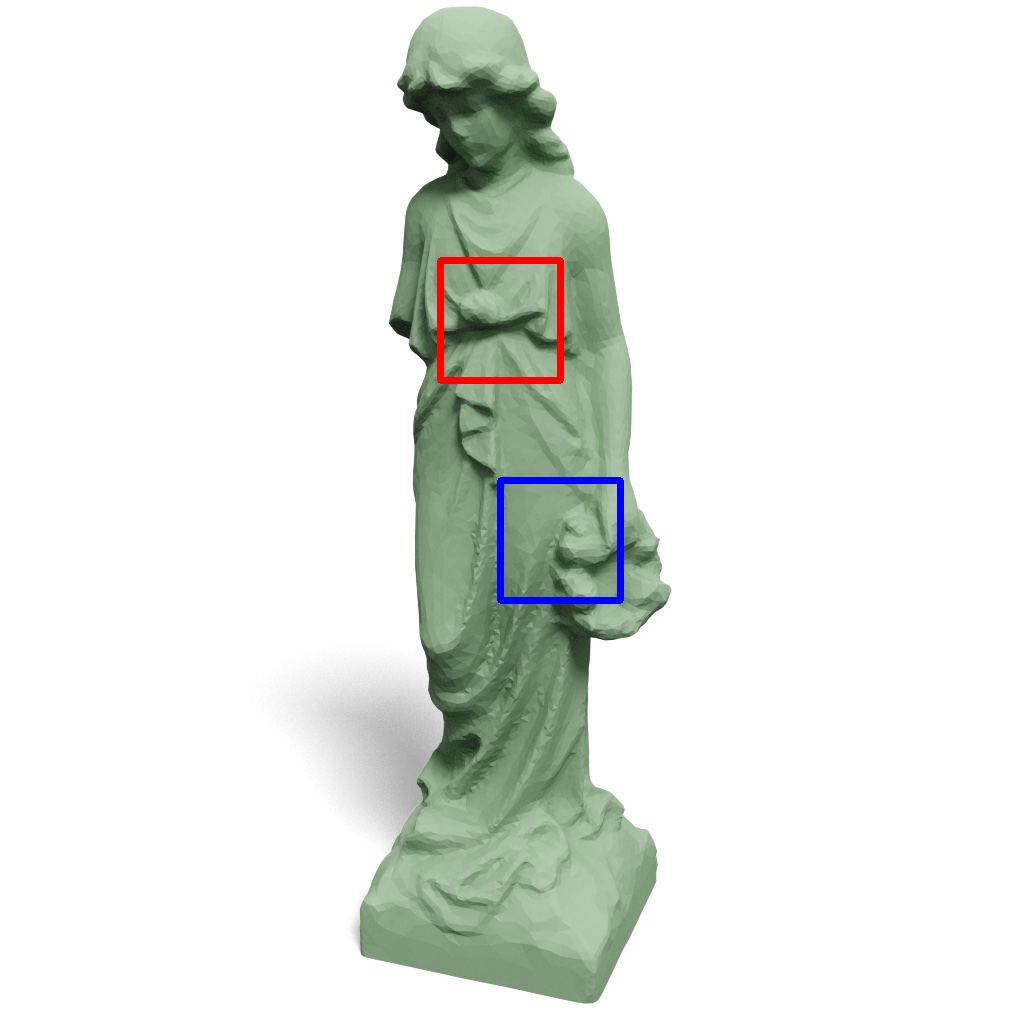}} \hfill
  \mpage{0.132}{\includegraphics[width=\linewidth]{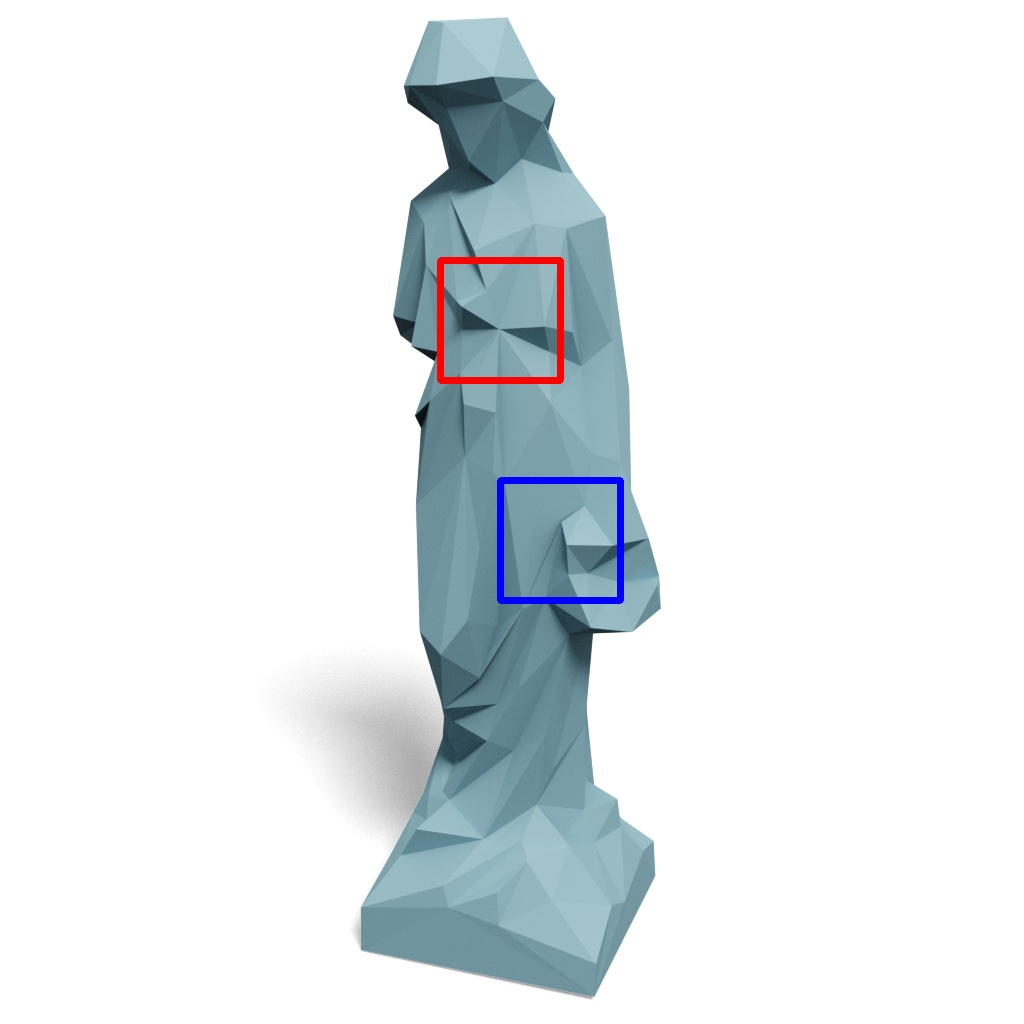}} \hfill
  \mpage{0.132}{\includegraphics[width=\linewidth]{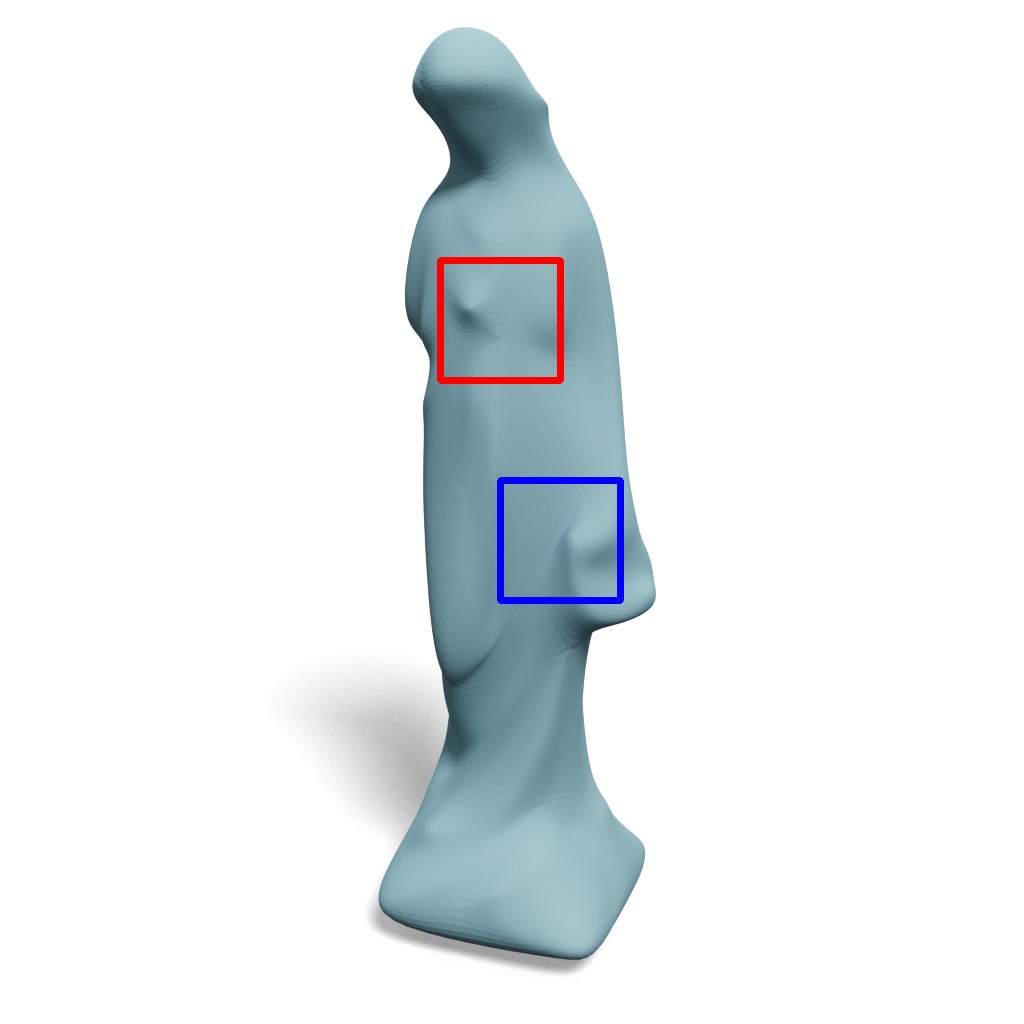}} \hfill
  \mpage{0.132}{\includegraphics[width=\linewidth]{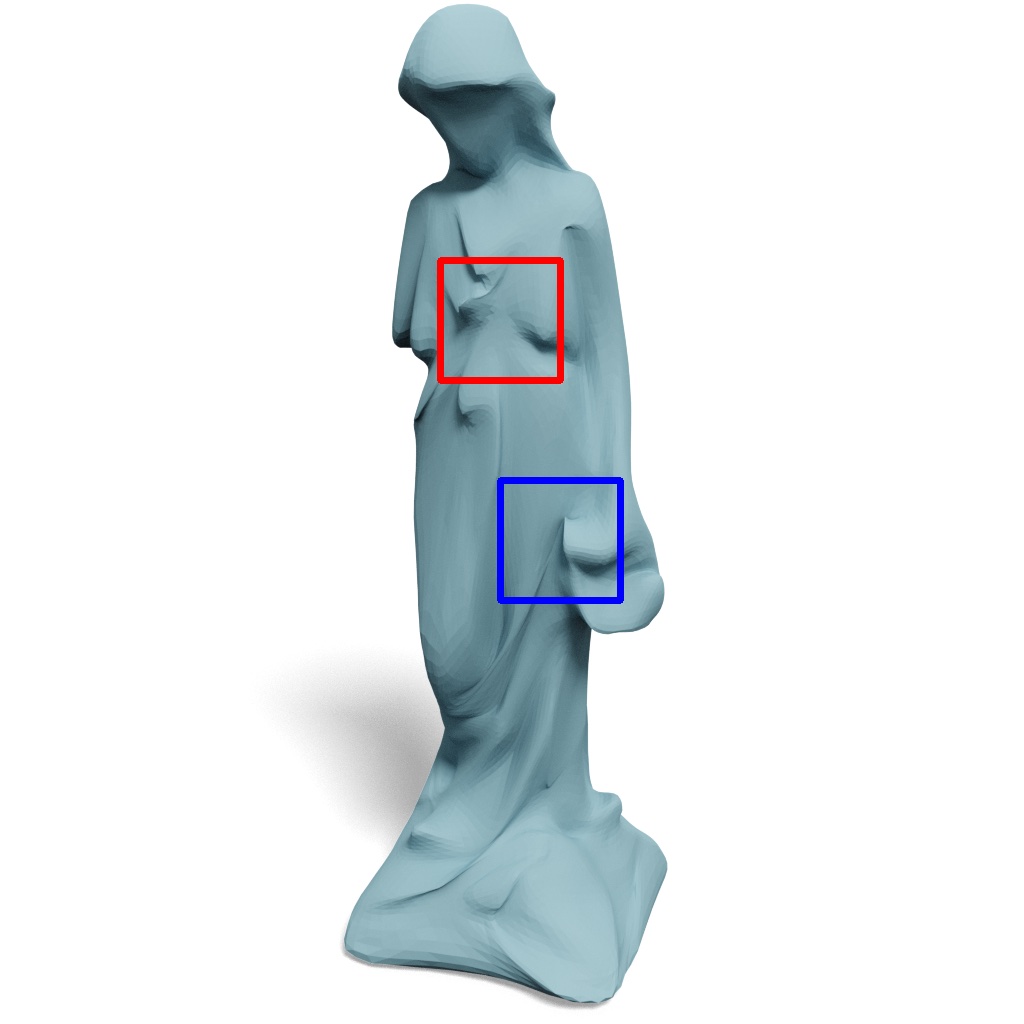}} \hfill
  \mpage{0.132}{\includegraphics[width=\linewidth]{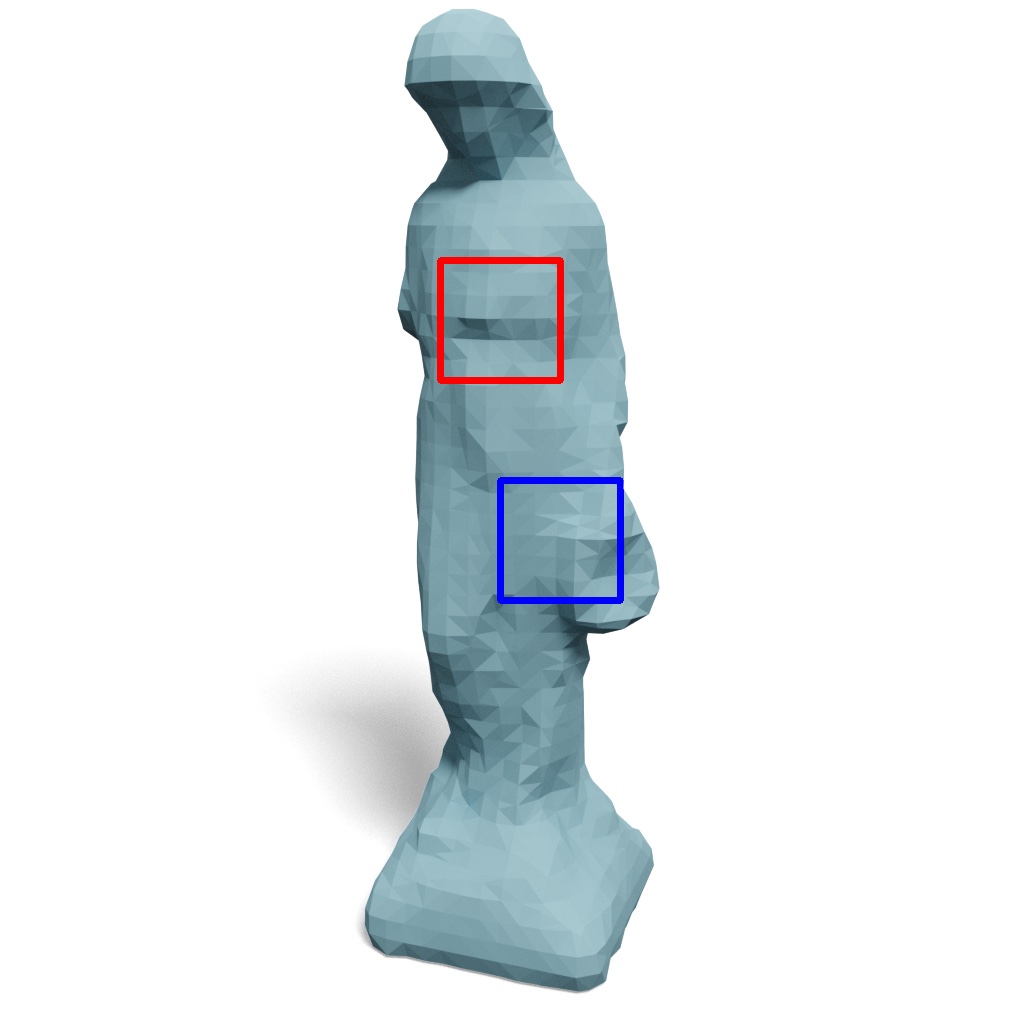}} \hfill
  \mpage{0.132}{\includegraphics[width=\linewidth]{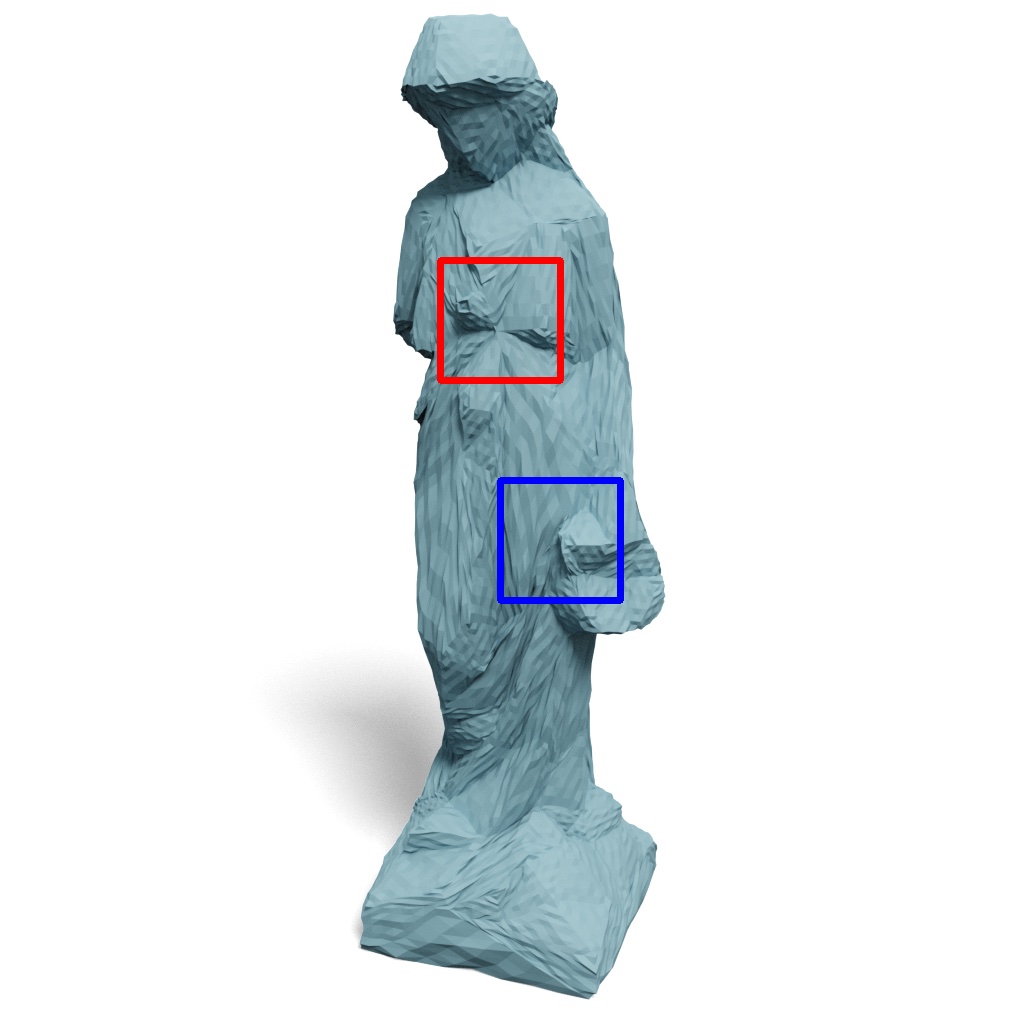}} \hfill
  \mpage{0.132}{\includegraphics[width=\linewidth]{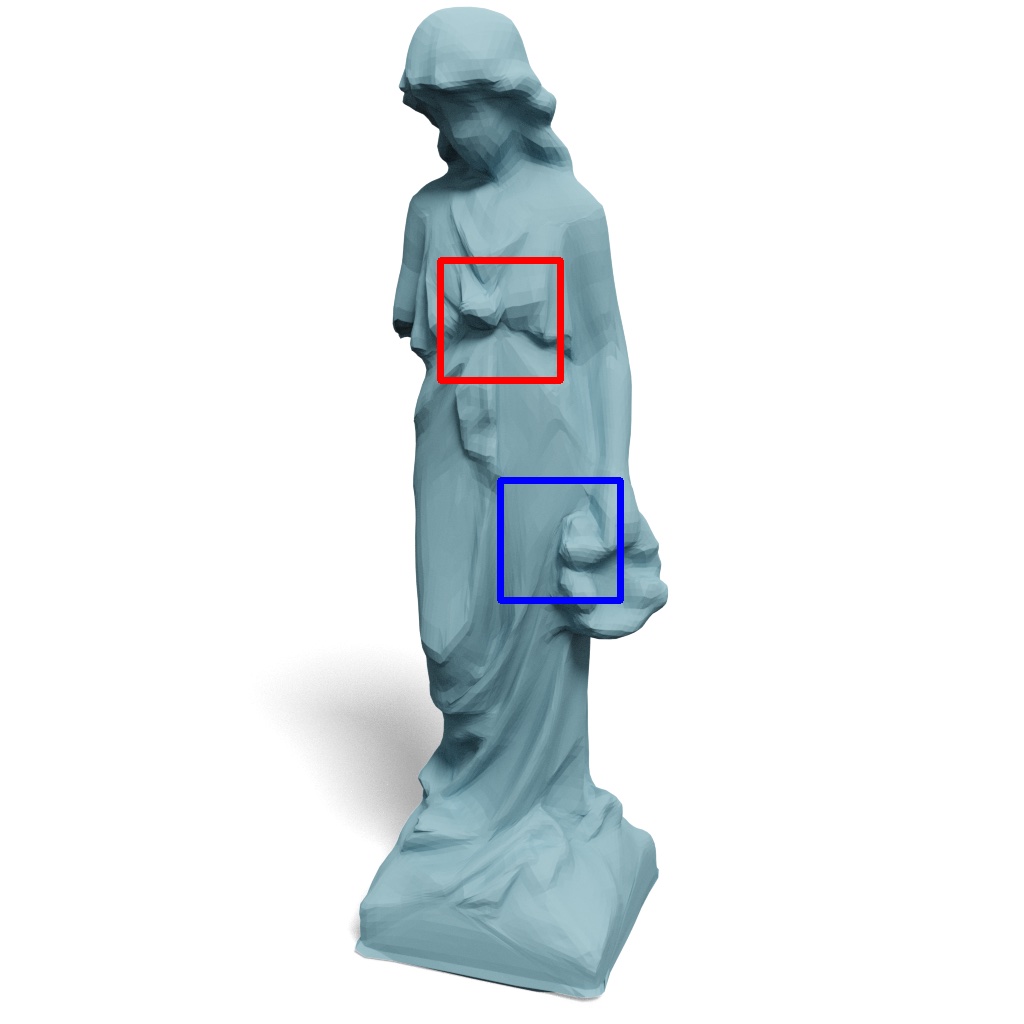}} \\
  \vspace{1.0mm}
  \mpage{0.132}{\includegraphics[width=0.475\linewidth]{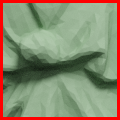} \hfill \includegraphics[width=0.475\linewidth]{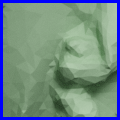}} \hfill
  \mpage{0.132}{\includegraphics[width=0.475\linewidth]{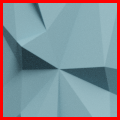} \hfill \includegraphics[width=0.475\linewidth]{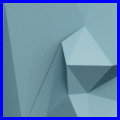}} \hfill
  \mpage{0.132}{\includegraphics[width=0.475\linewidth]{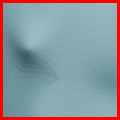} \hfill \includegraphics[width=0.475\linewidth]{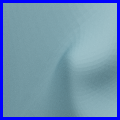}} \hfill
  \mpage{0.132}{\includegraphics[width=0.475\linewidth]{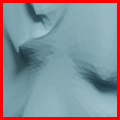} \hfill \includegraphics[width=0.475\linewidth]{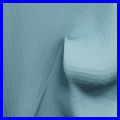}} \hfill 
  \mpage{0.132}{\includegraphics[width=0.475\linewidth]{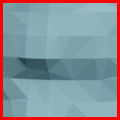} \hfill \includegraphics[width=0.475\linewidth]{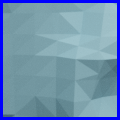}} \hfill
  \mpage{0.132}{\includegraphics[width=0.475\linewidth]{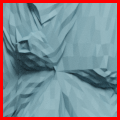} \hfill \includegraphics[width=0.475\linewidth]{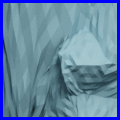}} \hfill
  \mpage{0.132}{\includegraphics[width=0.475\linewidth]{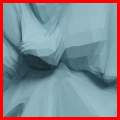} \hfill \includegraphics[width=0.475\linewidth]{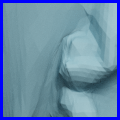}} \\
  \vspace{1.0mm}
  \mpage{0.132}{$CR = 33.75$} \hfill
  \mpage{0.132}{27.80 / 17.19$^\circ$} \hfill
  \mpage{0.132}{56.15 / 21.65$^\circ$} \hfill
  \mpage{0.132}{34.64 / 19.89$^\circ$} \hfill
  \mpage{0.132}{26.18 / 23.61$^\circ$} \hfill
  \mpage{0.132}{19.87 / 18.22$^\circ$} \hfill
  \mpage{0.132}{4.74 / 10.10$^\circ$} \\
  \vspace{1.0mm}
  \mpage{0.132}{Ground truth} \hfill
  \mpage{0.132}{QSlim} \hfill
  \mpage{0.132}{Loop} \hfill
  \mpage{0.132}{Butterfly} \hfill
  \mpage{0.132}{SubdivFit} \hfill
  \mpage{0.132}{Neural Subdiv} \hfill
  \mpage{0.132}{Ours} \\
  \caption{
  \textbf{Visual comparisons with decimation and subdivision methods on Thingi10K.} 
  We report the $d_\text{pm}$ ($\times 10^{-4}$) / $d_\text{normal}$ results under each method. The compression ratio ($CR$) is the same for all methods on the same shape and is reported under the ground-truth example.
  }
  \label{app-fig:exp-thingi10k-results}
  \mpage{0.22}{\includegraphics[width=\linewidth]{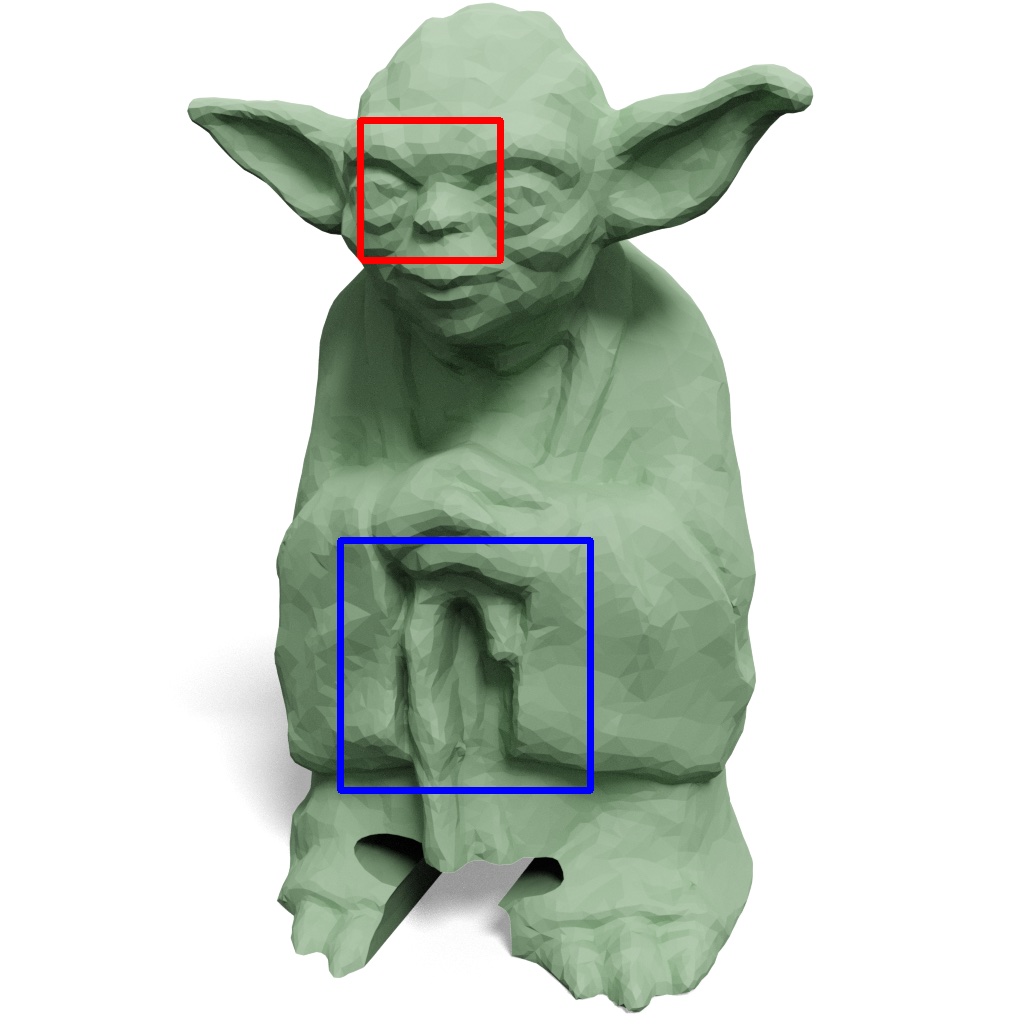}} \hfill
  \mpage{0.22}{\includegraphics[width=\linewidth]{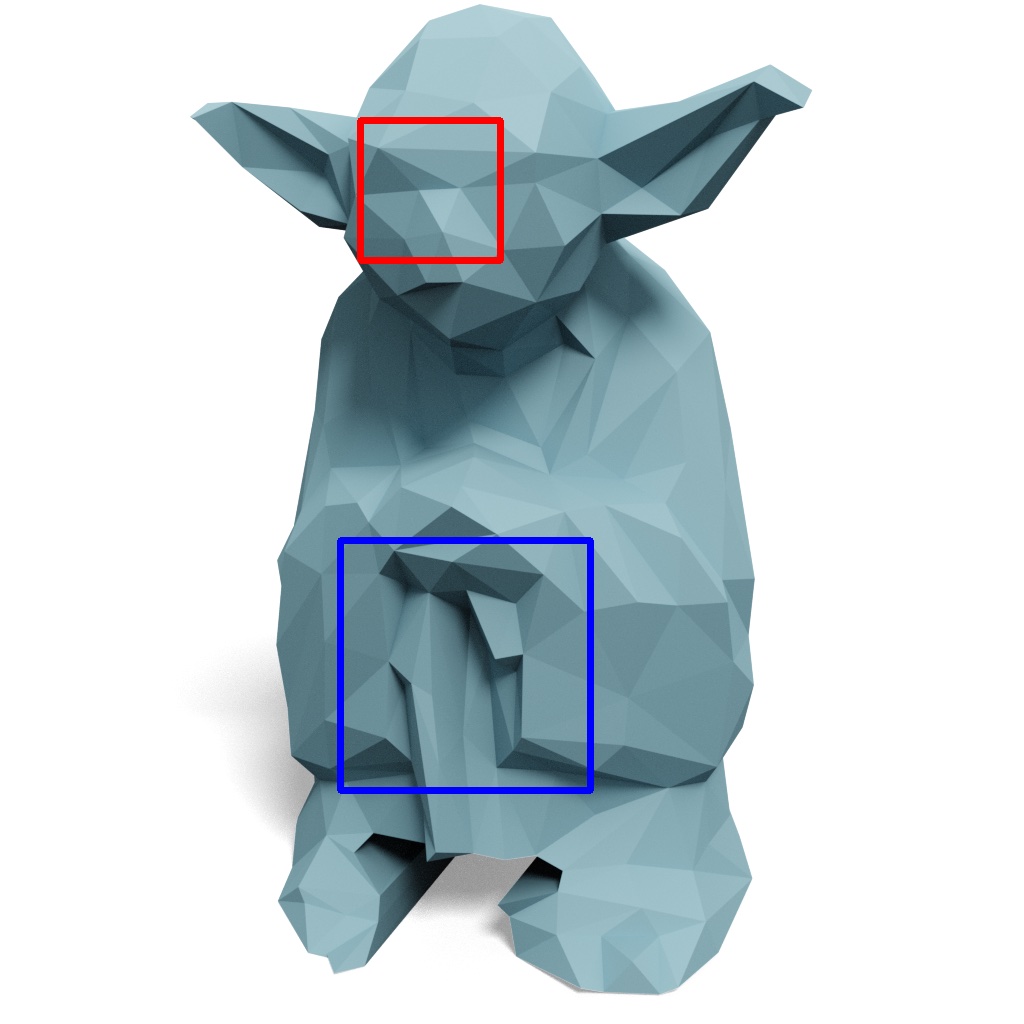}} \hfill
  \mpage{0.22}{\includegraphics[width=\linewidth]{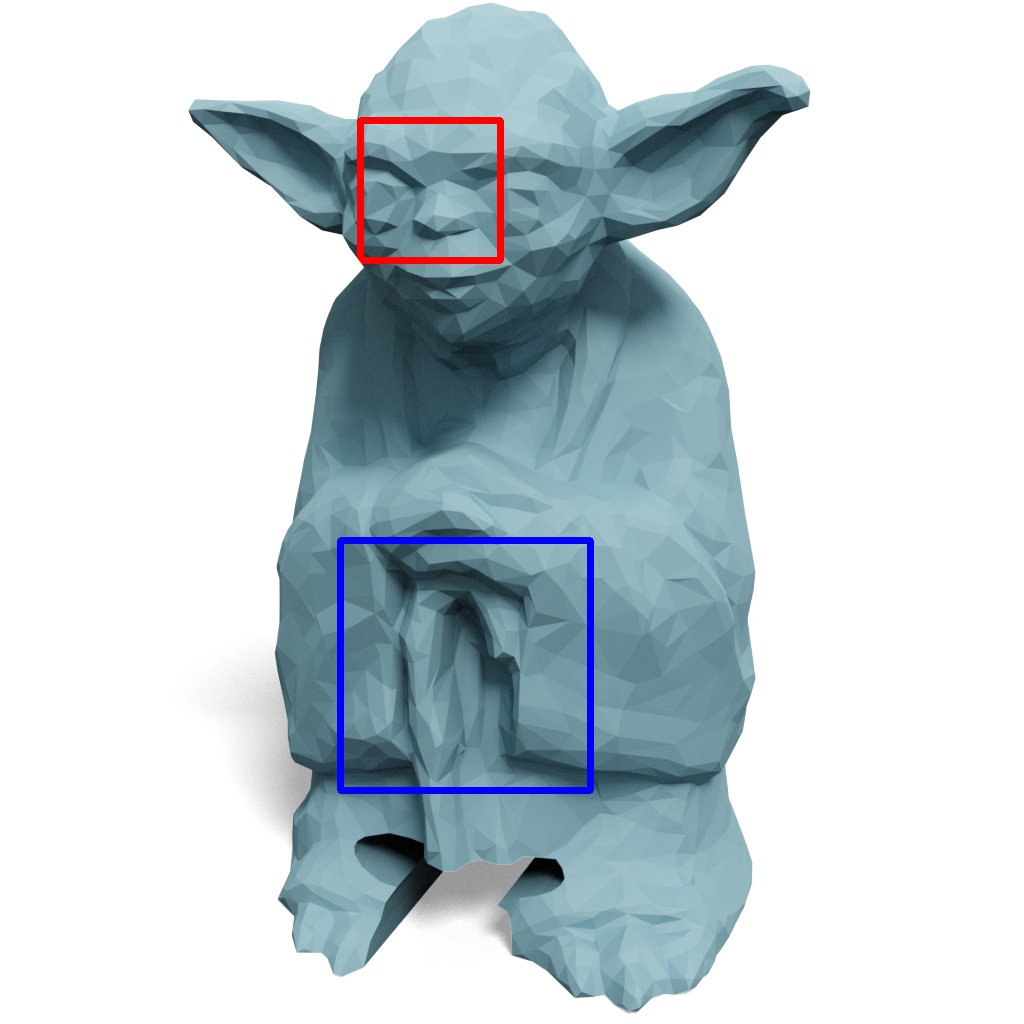}} \hfill
  \mpage{0.22}{\includegraphics[width=\linewidth]{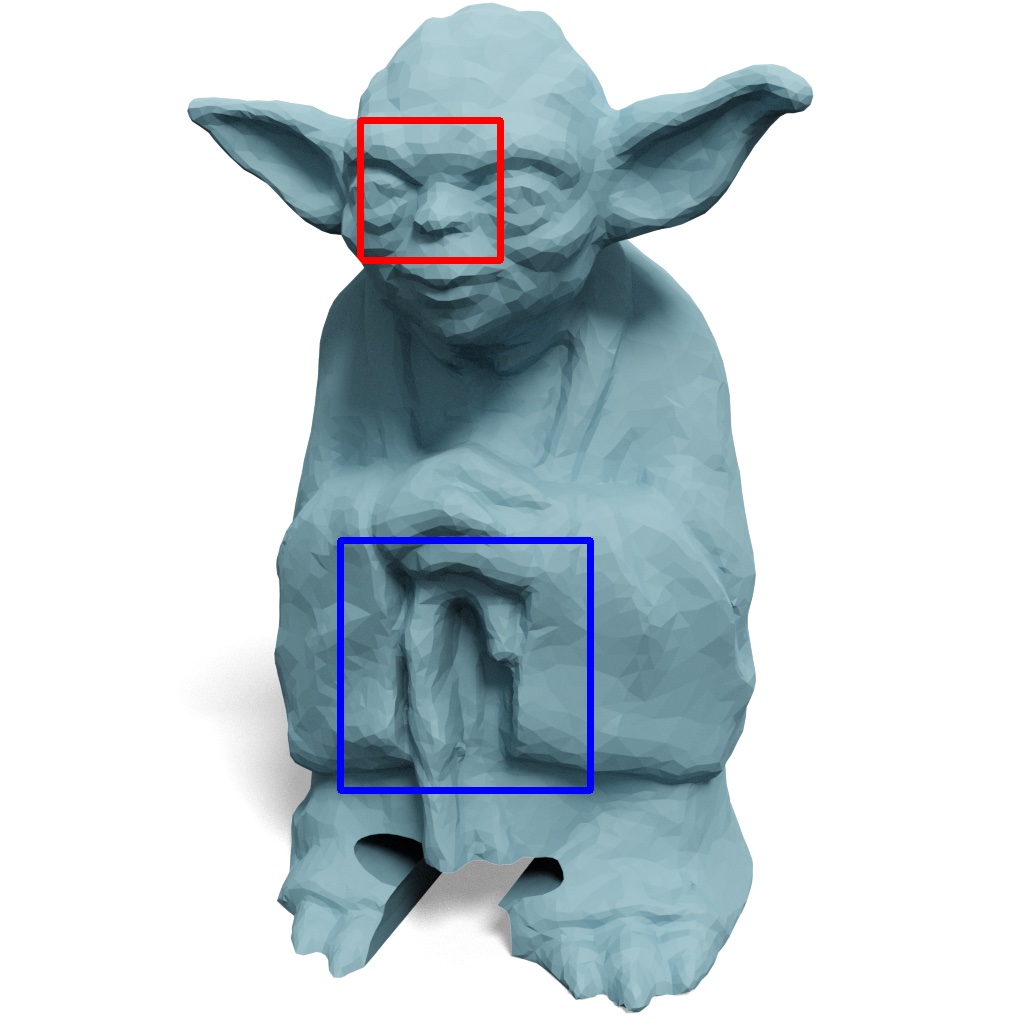}} \\
  \vspace{1.0mm}
  \mpage{0.22}{\includegraphics[width=0.475\linewidth]{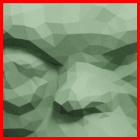} \hfill \includegraphics[width=0.475\linewidth]{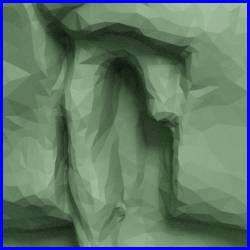}} \hfill
  \mpage{0.22}{\includegraphics[width=0.475\linewidth]{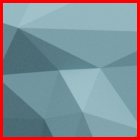} \hfill \includegraphics[width=0.475\linewidth]{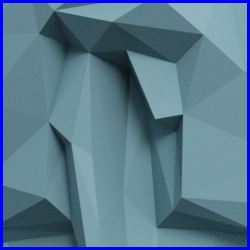}} \hfill
  \mpage{0.22}{\includegraphics[width=0.475\linewidth]{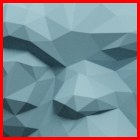} \hfill \includegraphics[width=0.475\linewidth]{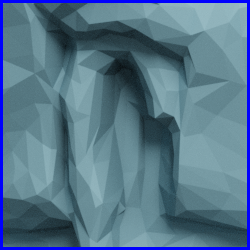}} \hfill
  \mpage{0.22}{\includegraphics[width=0.475\linewidth]{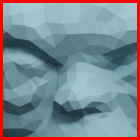} \hfill \includegraphics[width=0.475\linewidth]{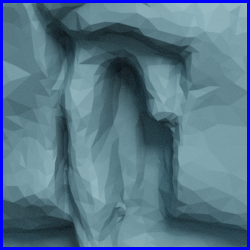}} \\
  \vspace{1.0mm}
  \mpage{0.22}{$CR$ / $d_\text{pm}$ ($\times 10^{-4}$) / $d_\text{normal}$} \hfill
  \mpage{0.22}{39.83 / 20.28 / 13.37$^\circ$} \hfill
  \mpage{0.22}{7.18 /  7.28 /  7.40$^\circ$} \hfill
  \mpage{0.22}{3.17 /  0.41 /  0.18$^\circ$} \\
  \vspace{1.0mm}
  \mpage{0.22}{Ground truth} \hfill
  \mpage{0.22}{Progressive Meshes} \hfill
  \mpage{0.22}{Progressive Meshes} \hfill
  \mpage{0.22}{Progressive Meshes} \\
  \mpage{0.22}{\includegraphics[width=\linewidth]{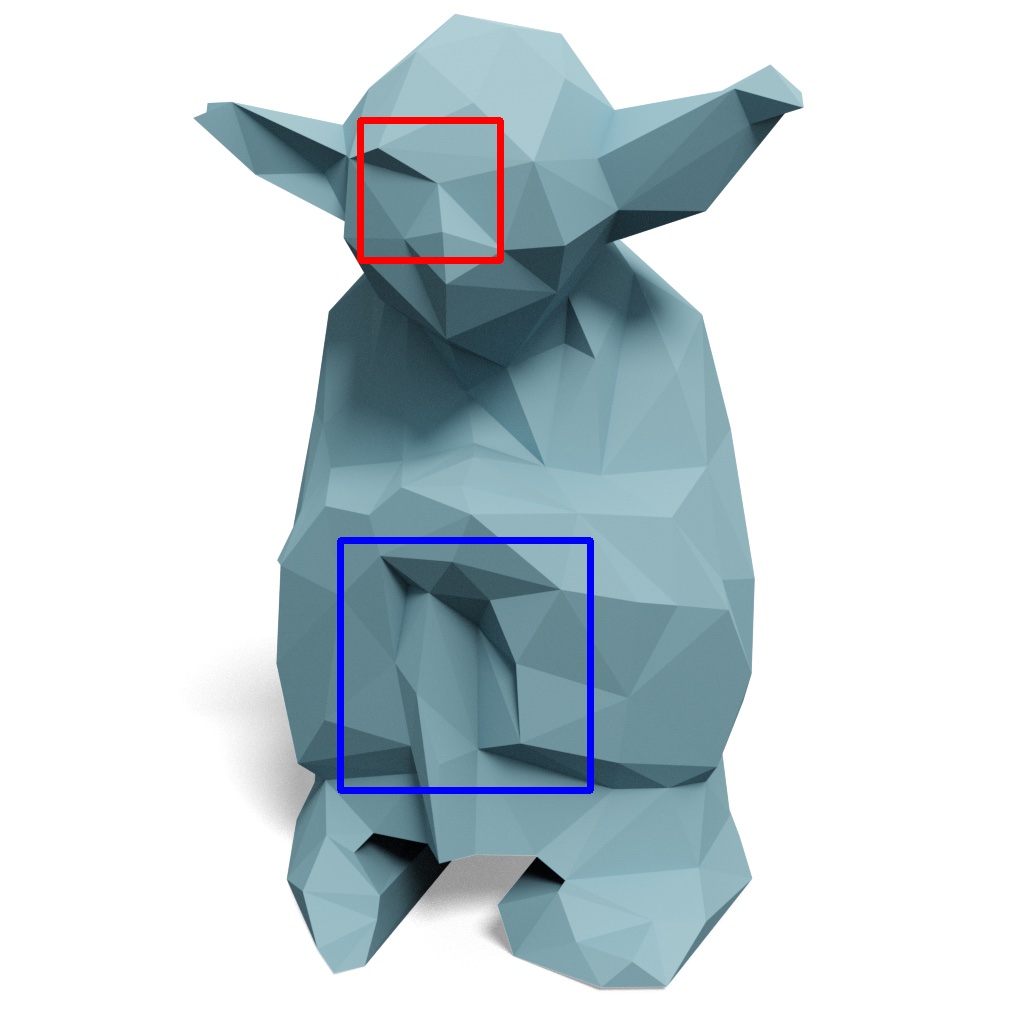}} \hfill
  \mpage{0.22}{\includegraphics[width=\linewidth]{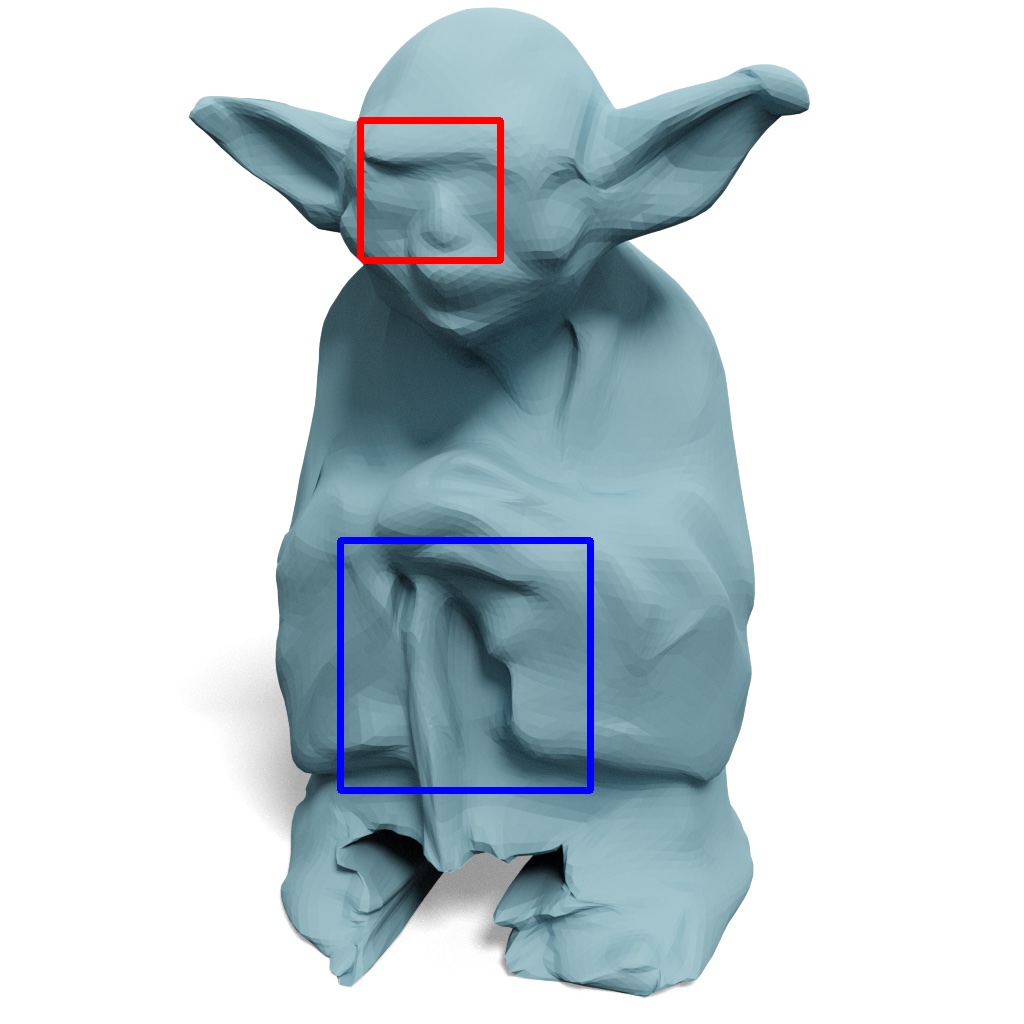}} \hfill
  \mpage{0.22}{\includegraphics[width=\linewidth]{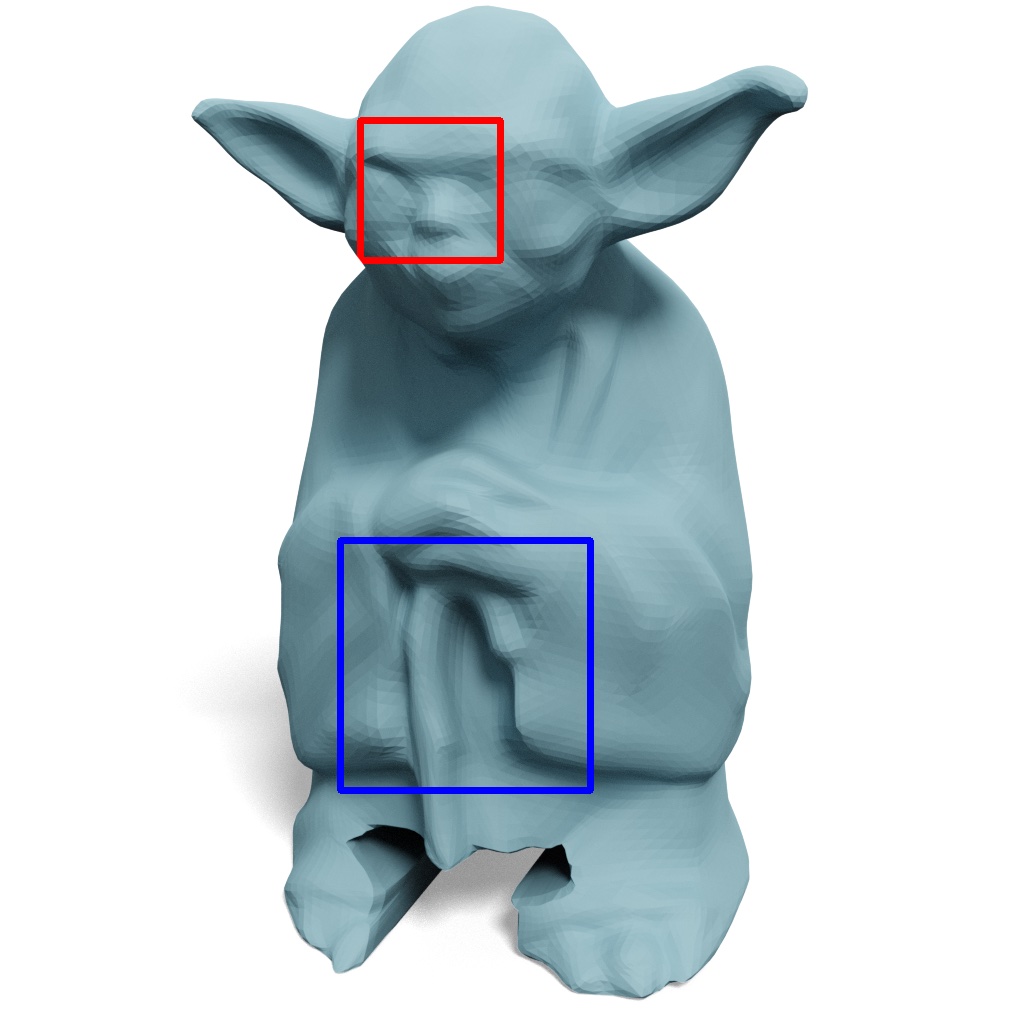}} \hfill
  \mpage{0.22}{\includegraphics[width=\linewidth]{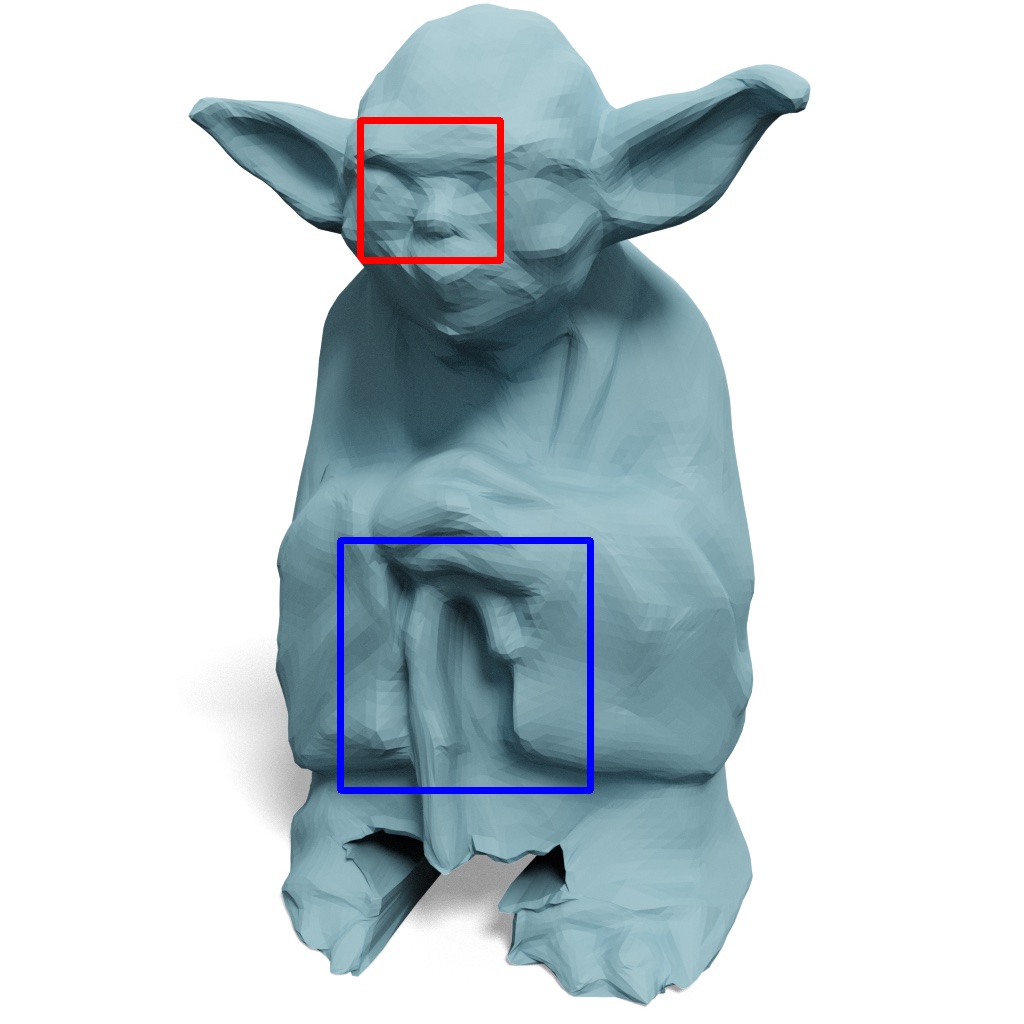}} \\
  \vspace{1.0mm}
  \mpage{0.22}{\includegraphics[width=0.475\linewidth]{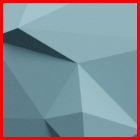} \hfill \includegraphics[width=0.475\linewidth]{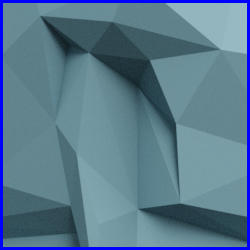}} \hfill
  \mpage{0.22}{\includegraphics[width=0.475\linewidth]{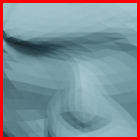} \hfill \includegraphics[width=0.475\linewidth]{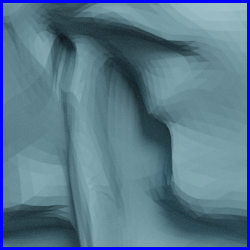}} \hfill
  \mpage{0.22}{\includegraphics[width=0.475\linewidth]{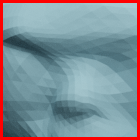} \hfill \includegraphics[width=0.475\linewidth]{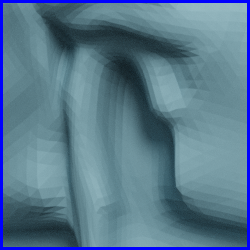}} \hfill
  \mpage{0.22}{\includegraphics[width=0.475\linewidth]{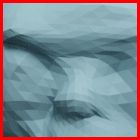} \hfill \includegraphics[width=0.475\linewidth]{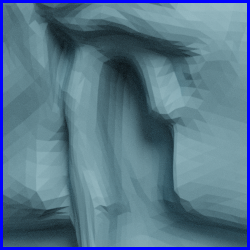}} \\
  \vspace{1.0mm}
  \mpage{0.22}{39.83 / 30.75 / 13.86$^\circ$} \hfill
  \mpage{0.22}{39.83 / 15.59 / 11.54$^\circ$} \hfill
  \mpage{0.22}{7.18 / 7.70 / 7.66$^\circ$} \hfill
  \mpage{0.22}{3.17 / 4.89 / 8.15$^\circ$} \\
  \vspace{1.0mm}
  \mpage{0.22}{QSlim} \hfill
  \mpage{0.22}{Ours w/o features} \hfill
  \mpage{0.22}{Ours + 40 features} \hfill
  \mpage{0.22}{Ours + 400 features} \\
  \caption{
  \textbf{Comparison to Progressive Meshes.} 
  }
  \label{app-fig:exp-prog-meshes}
\end{figure*}

\begin{figure*}[t]
  \centering
  \mpage{0.235}{\includegraphics[width=\linewidth]{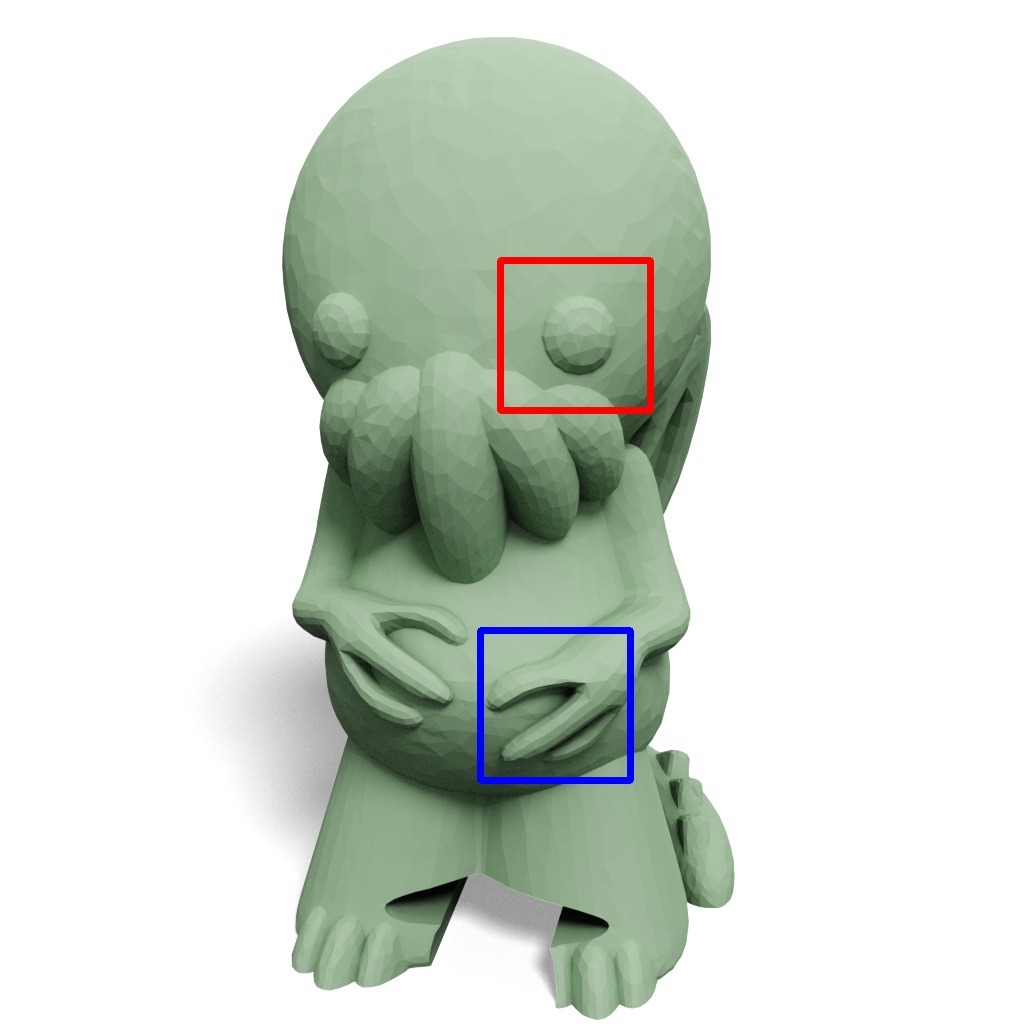}} \hfill
  \mpage{0.235}{\includegraphics[width=\linewidth]{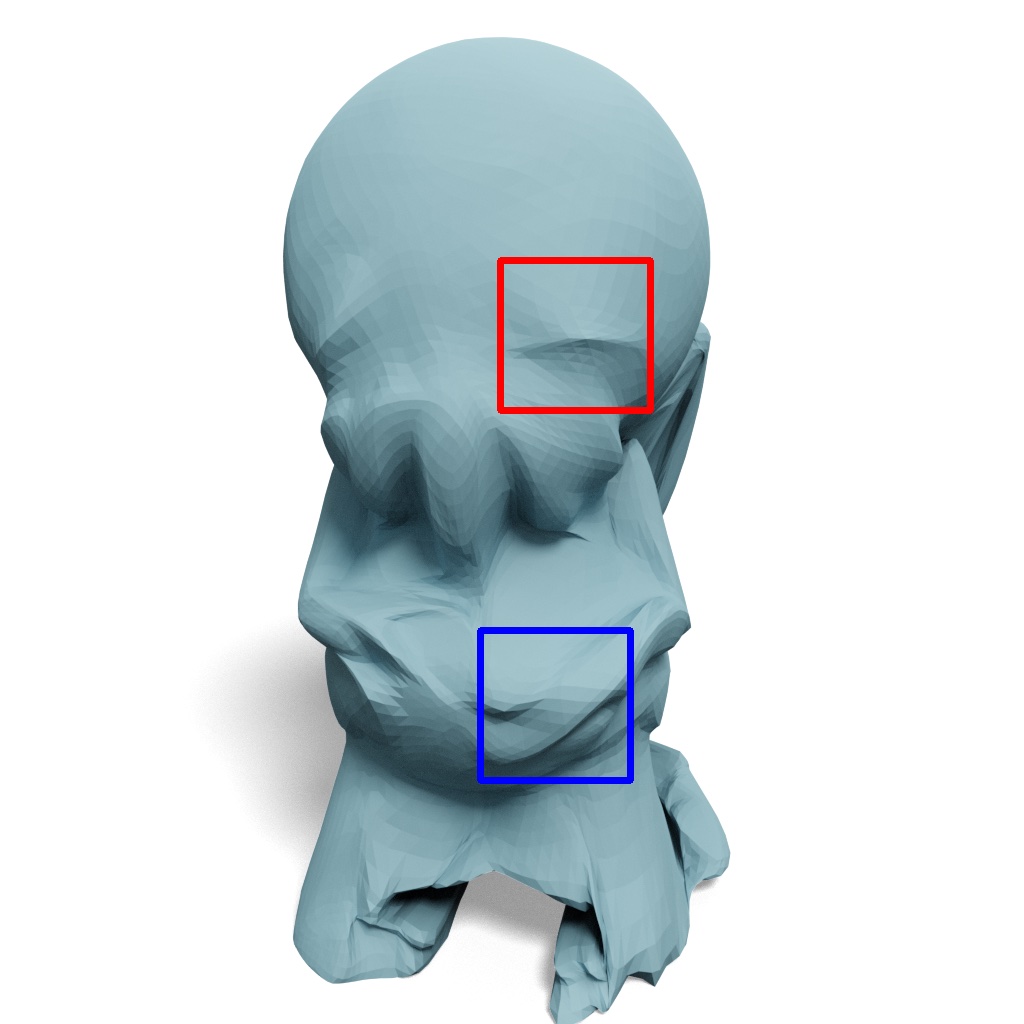}} \hfill
  \mpage{0.235}{\includegraphics[width=\linewidth]{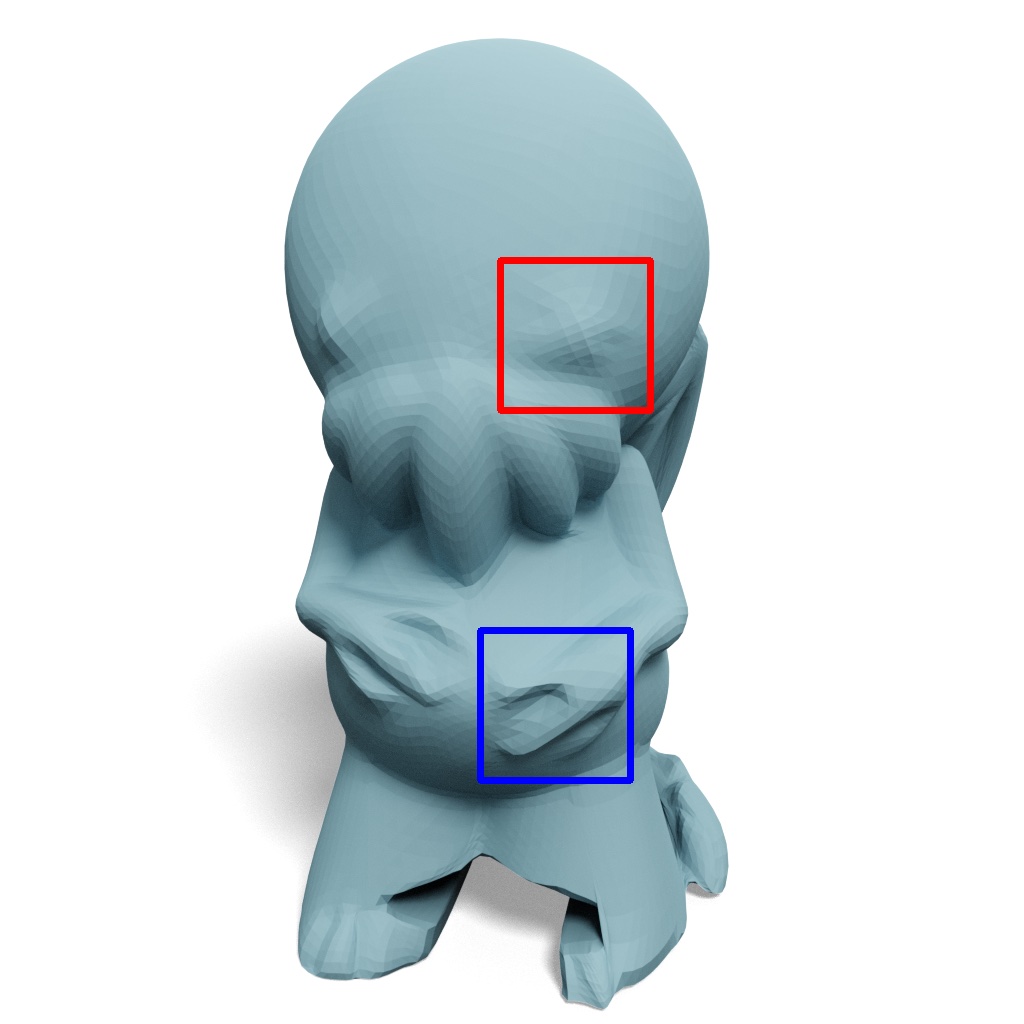}} \hfill
  \mpage{0.235}{\includegraphics[width=\linewidth]{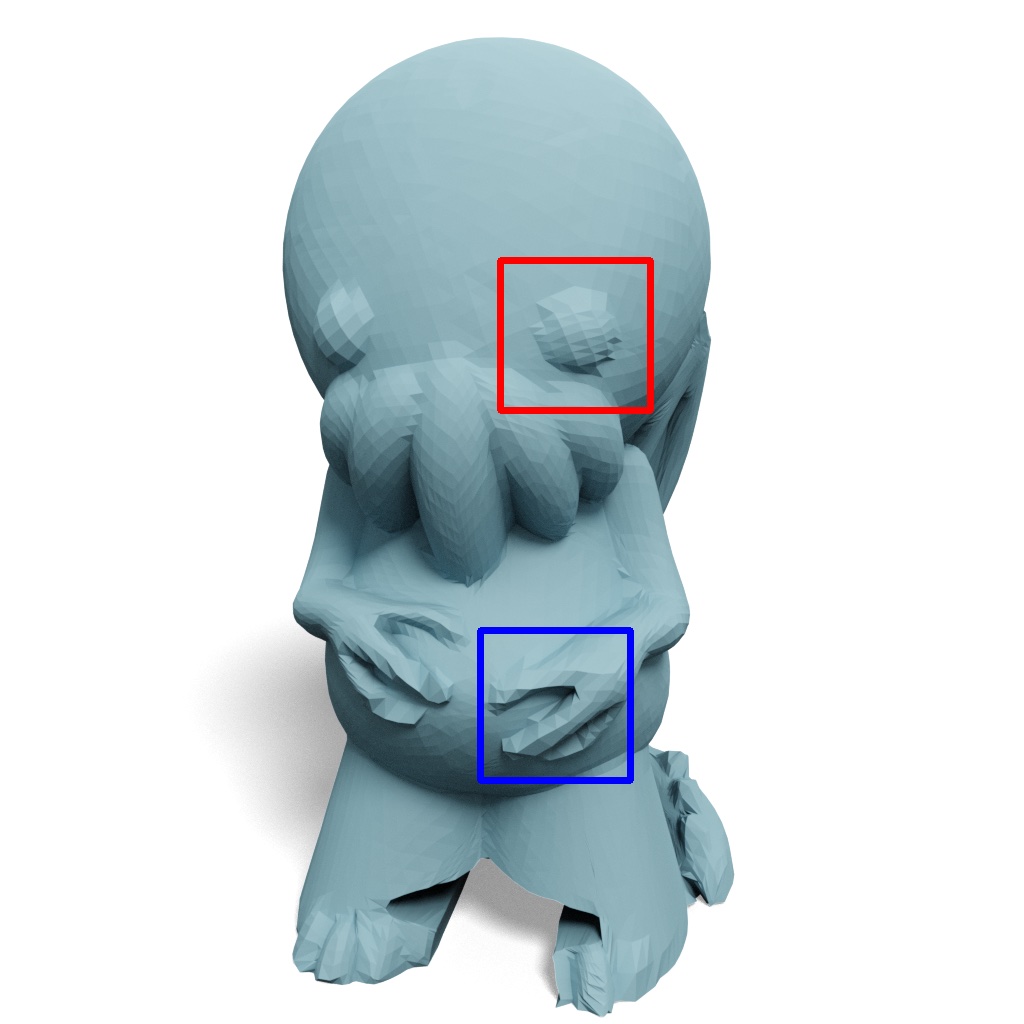}} \\
  \vspace{1.0mm}
  \mpage{0.235}{\includegraphics[width=0.475\linewidth]{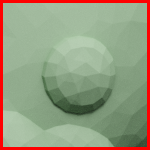} \hfill \includegraphics[width=0.475\linewidth]{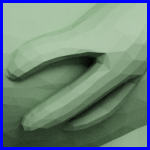}} \hfill
  \mpage{0.235}{\includegraphics[width=0.475\linewidth]{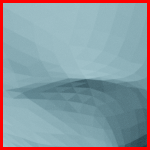} \hfill \includegraphics[width=0.475\linewidth]{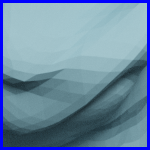}} \hfill
  \mpage{0.235}{\includegraphics[width=0.475\linewidth]{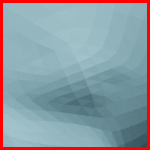} \hfill \includegraphics[width=0.475\linewidth]{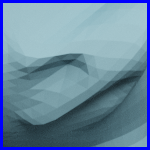}} \hfill
  \mpage{0.235}{\includegraphics[width=0.475\linewidth]{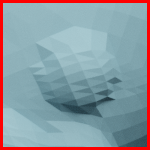} \hfill \includegraphics[width=0.475\linewidth]{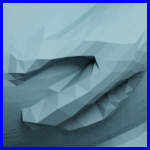}} \\
  \vspace{1.0mm}
  \mpage{0.235}{$CR$ / $d_\text{pm}$ ($\times 10^{-4}$) / $d_\text{normal}$} \hfill
  \mpage{0.235}{43.72 / 18.90 / 16.08$^\circ$} \hfill
  \mpage{0.235}{28.31 / 11.83 /  9.28$^\circ$} \hfill
  \mpage{0.235}{6.79 /  3.01 /  7.61$^\circ$} \\
  \vspace{1.0mm}
  \mpage{0.235}{\includegraphics[width=\linewidth]{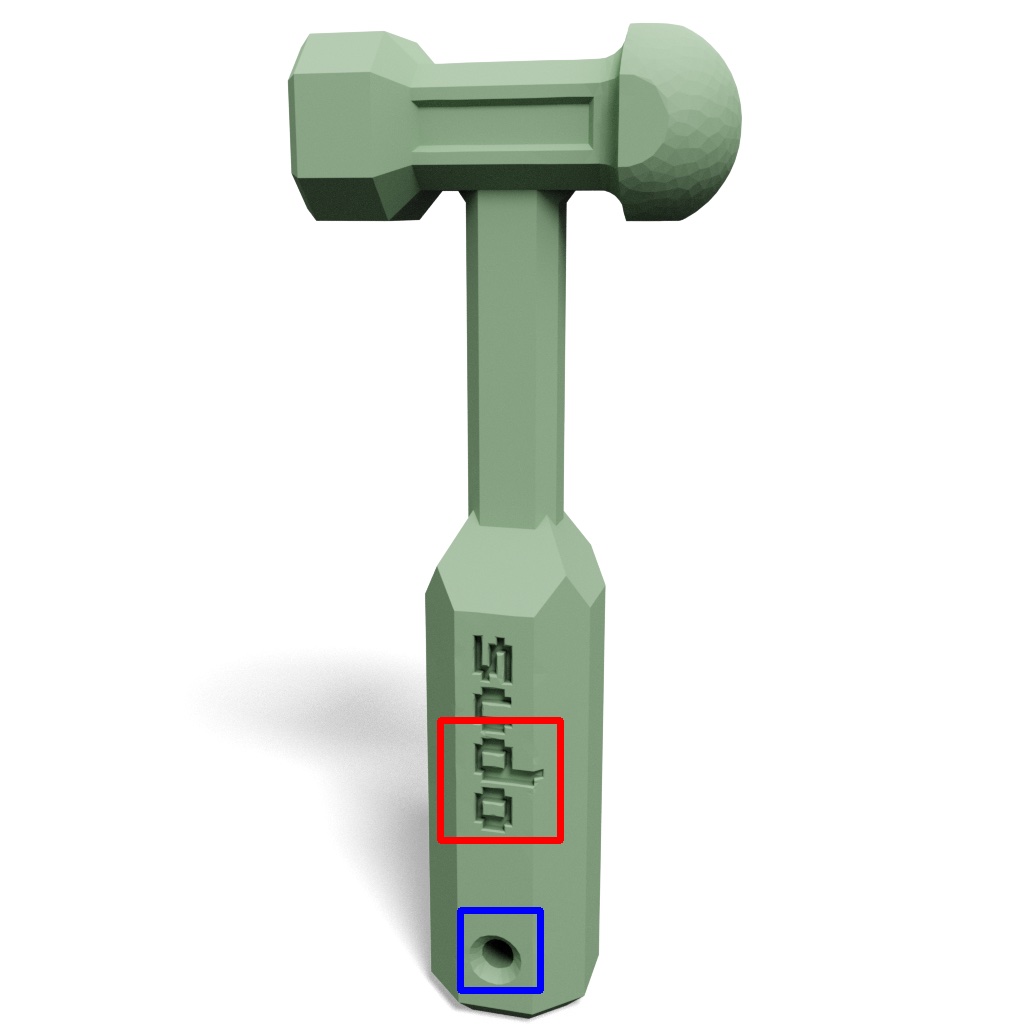}} \hfill
  \mpage{0.235}{\includegraphics[width=\linewidth]{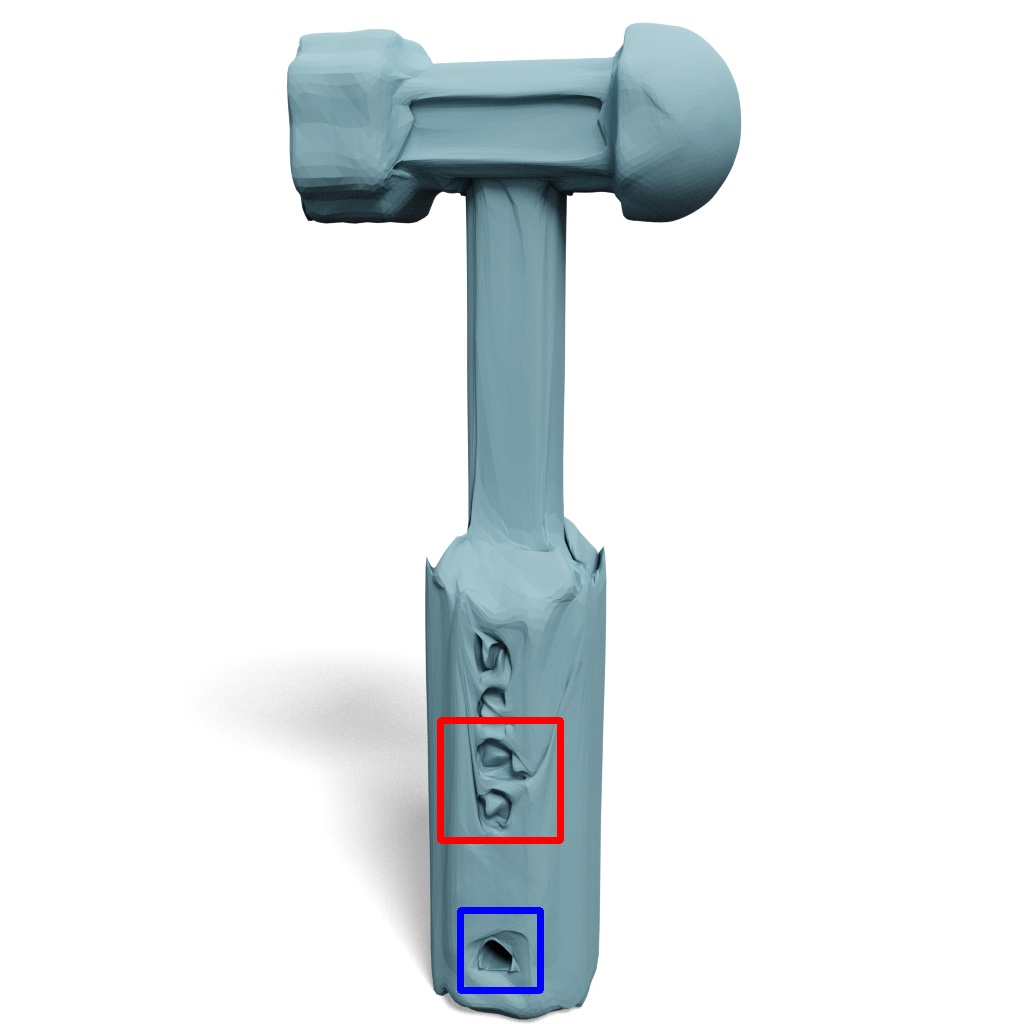}} \hfill
  \mpage{0.235}{\includegraphics[width=\linewidth]{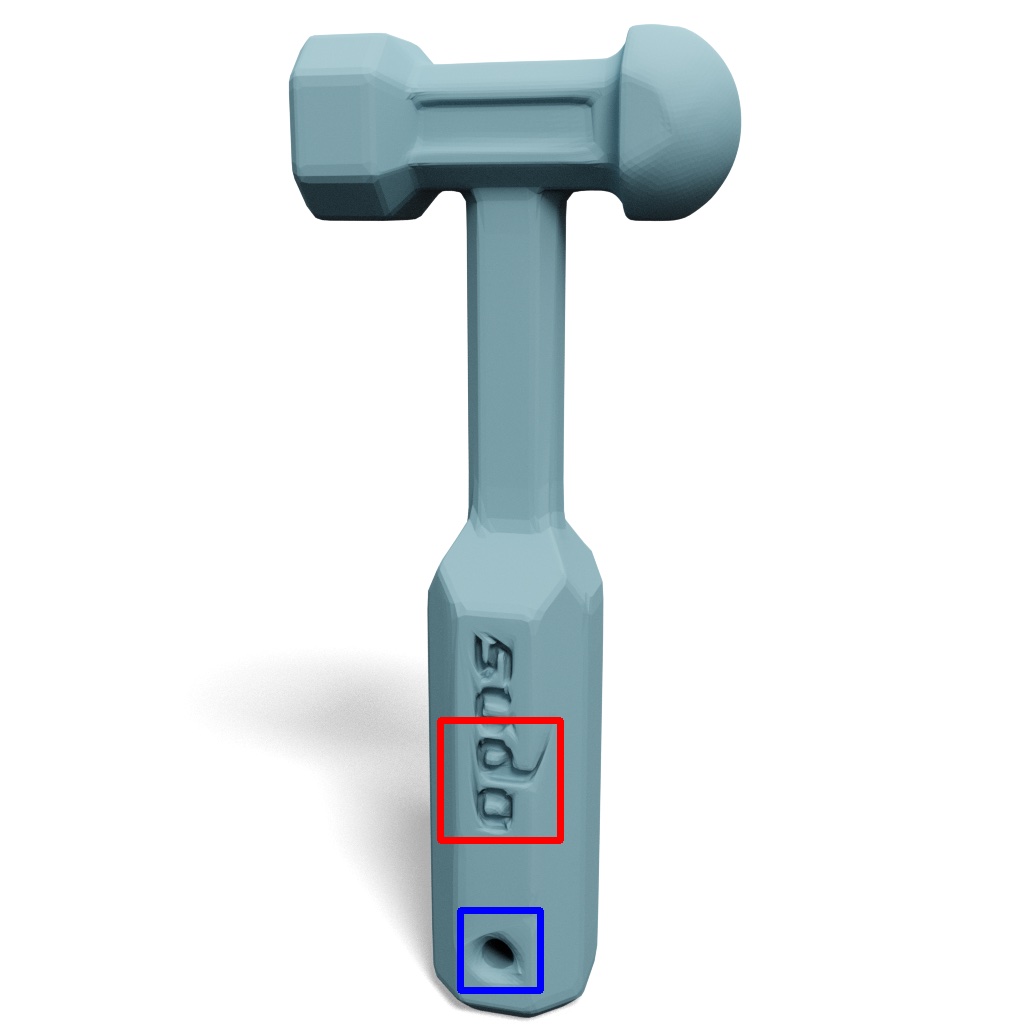}} \hfill
  \mpage{0.235}{\includegraphics[width=\linewidth]{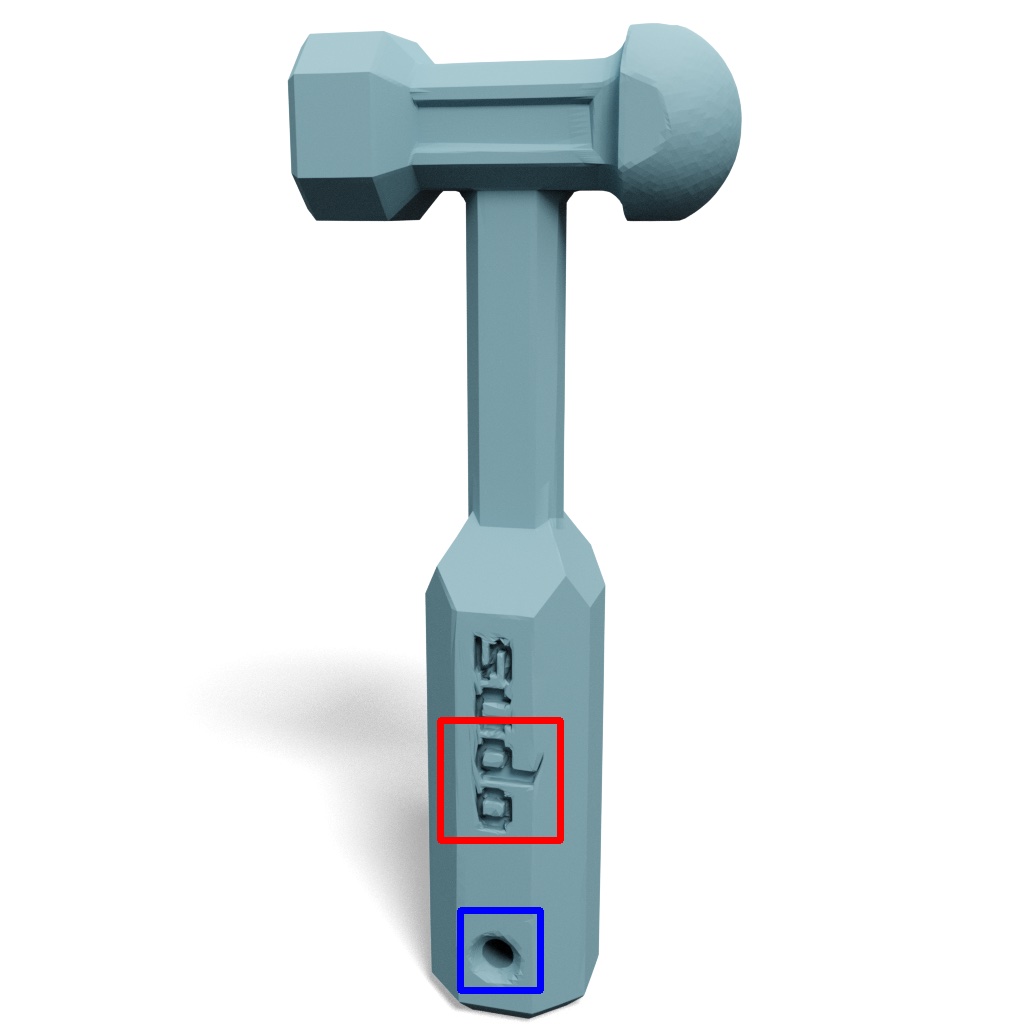}} \\
  \vspace{1.0mm}
  \mpage{0.235}{\includegraphics[width=0.475\linewidth]{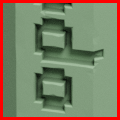} \hfill \includegraphics[width=0.475\linewidth]{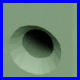}} \hfill
  \mpage{0.235}{\includegraphics[width=0.475\linewidth]{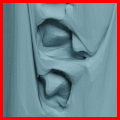} \hfill \includegraphics[width=0.475\linewidth]{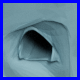}} \hfill
  \mpage{0.235}{\includegraphics[width=0.475\linewidth]{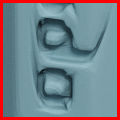} \hfill \includegraphics[width=0.475\linewidth]{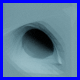}} \hfill
  \mpage{0.235}{\includegraphics[width=0.475\linewidth]{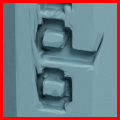} \hfill \includegraphics[width=0.475\linewidth]{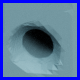}} \\
  \vspace{1.0mm}
  \mpage{0.235}{$CR$ / $d_\text{pm}$ ($\times 10^{-4}$) / $d_\text{normal}$} \hfill
  \mpage{0.235}{11.17 / 12.11 / 14.29$^\circ$} \hfill
  \mpage{0.235}{7.28 /  3.16 / 5.60$^\circ$} \hfill
  \mpage{0.235}{1.76 /  0.56 / 2.71$^\circ$} \\
  \vspace{1.0mm}
  \mpage{0.235}{\includegraphics[width=\linewidth]{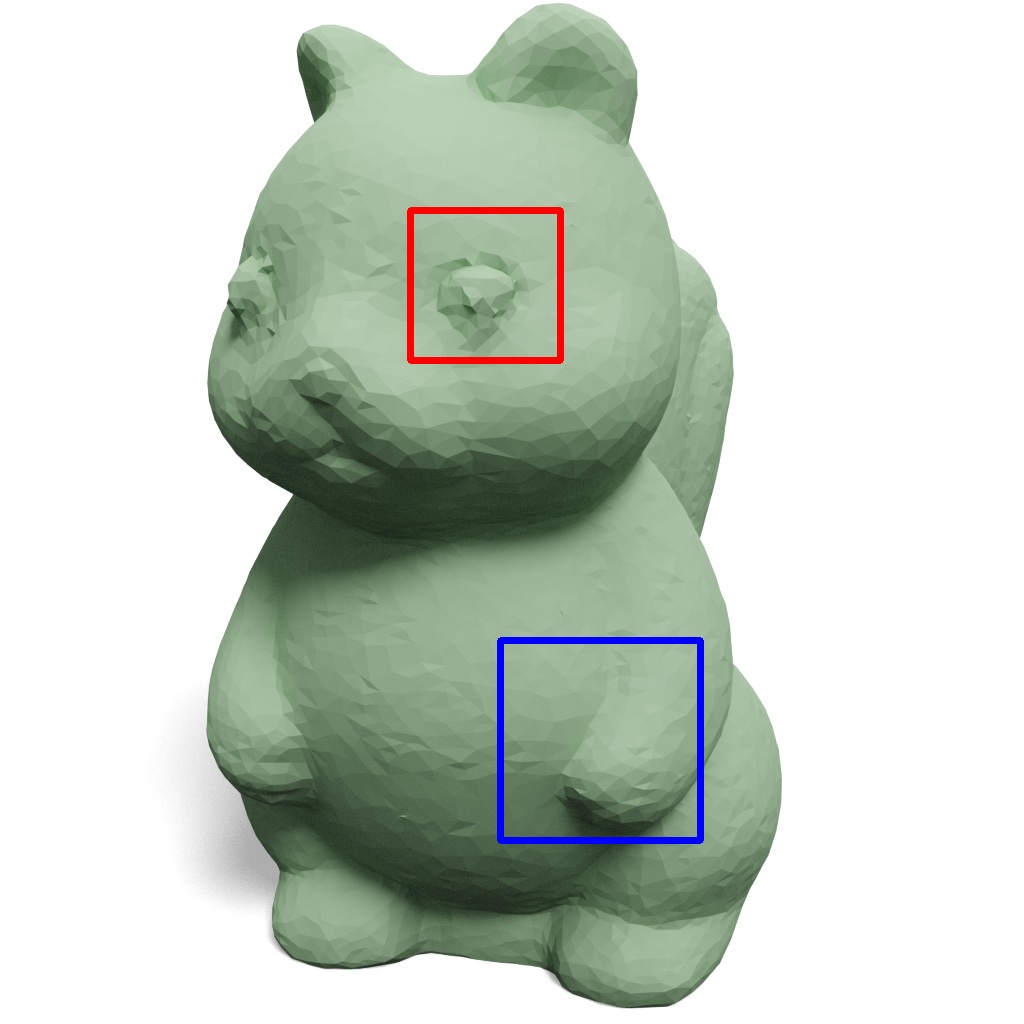}} \hfill
  \mpage{0.235}{\includegraphics[width=\linewidth]{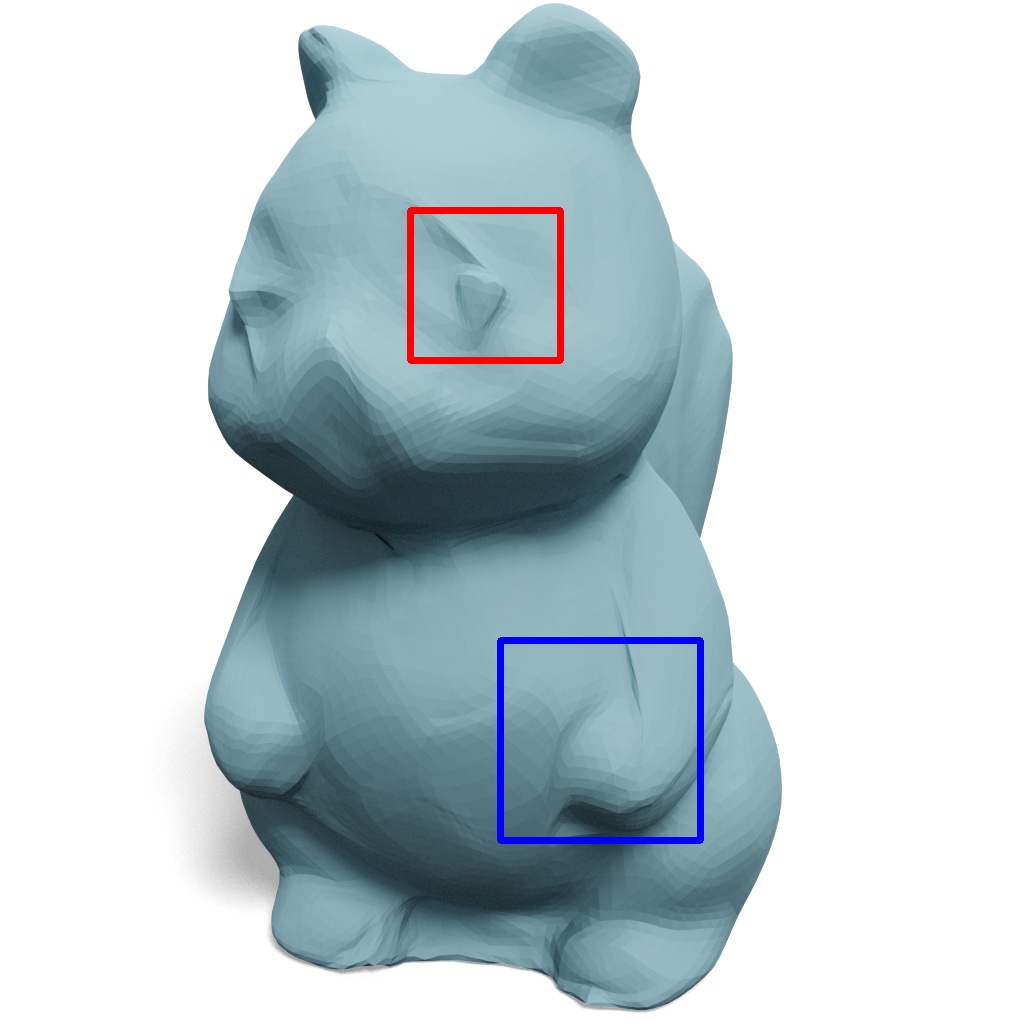}} \hfill
  \mpage{0.235}{\includegraphics[width=\linewidth]{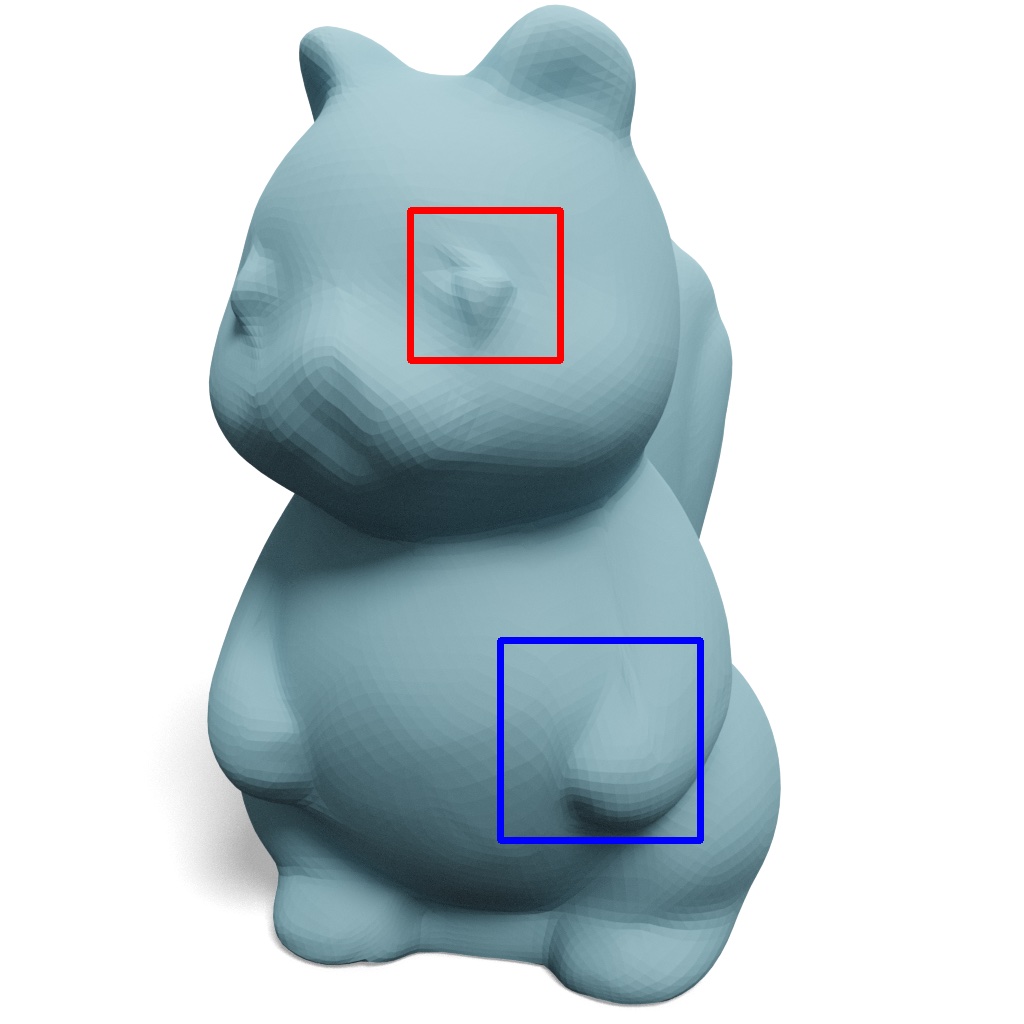}} \hfill
  \mpage{0.235}{\includegraphics[width=\linewidth]{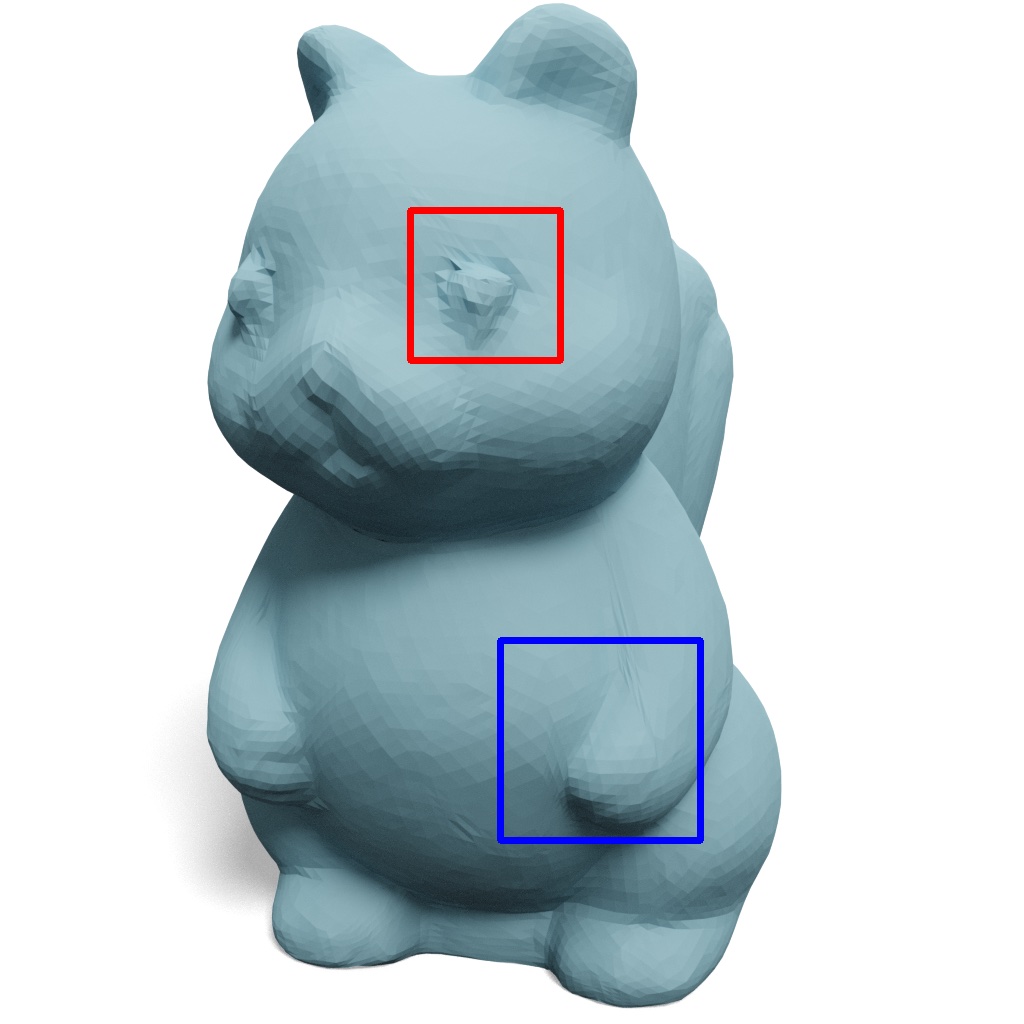}} \\
  \vspace{1.0mm}
  \mpage{0.235}{\includegraphics[width=0.475\linewidth]{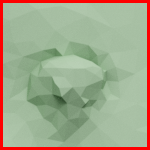} \hfill \includegraphics[width=0.475\linewidth]{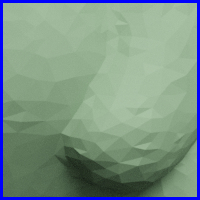}} \hfill
  \mpage{0.235}{\includegraphics[width=0.475\linewidth]{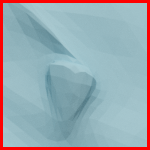} \hfill \includegraphics[width=0.475\linewidth]{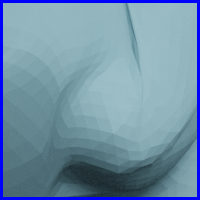}} \hfill
  \mpage{0.235}{\includegraphics[width=0.475\linewidth]{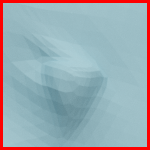} \hfill \includegraphics[width=0.475\linewidth]{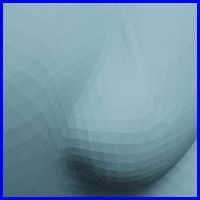}} \hfill
  \mpage{0.235}{\includegraphics[width=0.475\linewidth]{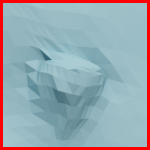} \hfill \includegraphics[width=0.475\linewidth]{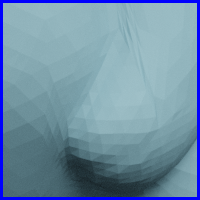}} \\
  \vspace{1.0mm}
  \mpage{0.235}{$CR$ / $d_\text{pm}$ ($\times 10^{-4}$) / $d_\text{normal}$} \hfill
  \mpage{0.235}{42.02 / 8.24 / 7.40$^\circ$} \hfill
  \mpage{0.235}{27.50 / 6.57 / 5.06$^\circ$} \hfill
  \mpage{0.235}{6.69 / 1.44 / 3.98$^\circ$} \\
  \vspace{1.0mm}
  \mpage{0.235}{Ground truth} \hfill
  \mpage{0.235}{Ours w/o features} \hfill
  \mpage{0.235}{Ours + 40 features} \hfill
  \mpage{0.235}{Ours + 400 features} \\
  \caption{
  \textbf{Progressive features.} 
  We show more examples where transmitting more features leads to better quantitative and qualitative results.
  }
  \label{app-fig:prog-feat}
\end{figure*}

\end{document}